\documentclass[10pt,twocolumn,letterpaper]{article}

\usepackage[pagenumbers]{cvpr} % To force page numbers, e.g. for an arXiv version

\definecolor{cvprblue}{rgb}{0.21,0.49,0.74}
\usepackage[pagebackref,breaklinks,colorlinks,allcolors=cvprblue]{hyperref}

\def\paperID{*****} % *** Enter the Paper ID here
\def\confName{CVPR}
\def\confYear{2026}

\usepackage{tabularx}
\usepackage{float}
\usepackage{multirow}
\usepackage{array}

\title{Probing Memes in LLMs: A Paradigm for the Entangled Evaluation World}

\author{%
  \textbf{Luzhou Peng}$^{1,3}$ \quad
  \textbf{Zhengxin Yang}$^{1,2}$\thanks{Zhengxin Yang is the corresponding author} \quad
  \textbf{Honglu Ji}$^{3}$ \quad
  \textbf{Yikang Yang}$^{1,3}$ \\
  \textbf{Fanda Fan}$^{1,2}$ \quad
  \textbf{Wanling Gao}$^{1,2}$ \quad
  \textbf{Jiayuan Ge}$^{1,2}$ \quad
  \textbf{Yilin Han}$^{4}$ \quad  \textbf{Jianfeng Zhan}$^{1,2}$ \\
  Institute of Computing Technology, Chinese Academy of Sciences$^{1}$\\
  BenchCouncil (International Open Benchmark Council)$^{2}$\\
  University of Chinese Academy of Sciences$^{3}$\\
  School of Artificial Intelligence and Data Science, Hebei University of Technology$^{4}$\\
  \texttt{\{pengluzhou24s, yangzhengxin\}@ict.ac.cn}
}

\begin{document}
\maketitle
\begin{abstract}
Current evaluation paradigms for large language models (LLMs) characterize models and datasets separately, yielding coarse descriptions: items in datasets are treated as pre-labeled entries, and models are summarized by overall scores such as accuracy, together ignoring the diversity of population-level model behaviors across items with varying properties.
To address this gap, this paper conceptualizes LLMs as composed of memes, a notion introduced by Dawkins as cultural genes that replicate knowledge and behavior. Building on this perspective, the Probing Memes paradigm reconceptualizes evaluation as an entangled world of models and data. It centers on a Perception Matrix that captures model-item interactions, enabling Probe Properties for characterizing items and Meme Scores for depicting model behavioral traits. 
Applied to 9 datasets and 4,507 LLMs, Probing Memes reveals hidden capability structures and quantifies phenomena invisible under traditional paradigms (e.g., elite models failing on problems that most models answer easily). It not only supports more informative and extensible benchmarks but also enables population-based evaluation of LLMs.
\end{abstract}    
\section{Introduction}
\label{sec:intro}

To advance the development and understanding of large language models (LLMs), researchers have devoted sustained efforts to improving benchmark design across a broad range of domains~\citep{hendrycks2020measuring,hendrycksmath2021,srivastava2023beyond}. To make evaluation more effective, recent work has pursued two complementary directions: introducing increasingly challenging or cost-efficient datasets~\citep{phan2025humanity,polo2024tinybenchmarks}, and expanding evaluation metrics beyond simple accuracy to capture richer performance dimensions~\citep{checklist:acl20,bommasani2023holistic,guo2025mathematical}. However, limitations remain: current approaches often treat models and datasets in isolation, leading to overly coarse descriptions. Consequently, evaluations may lack depth and fail to reveal phenomena that emerge when data and models are analyzed in a population context.

\begin{figure}[t]
  \centering
  \includegraphics[width=\linewidth]{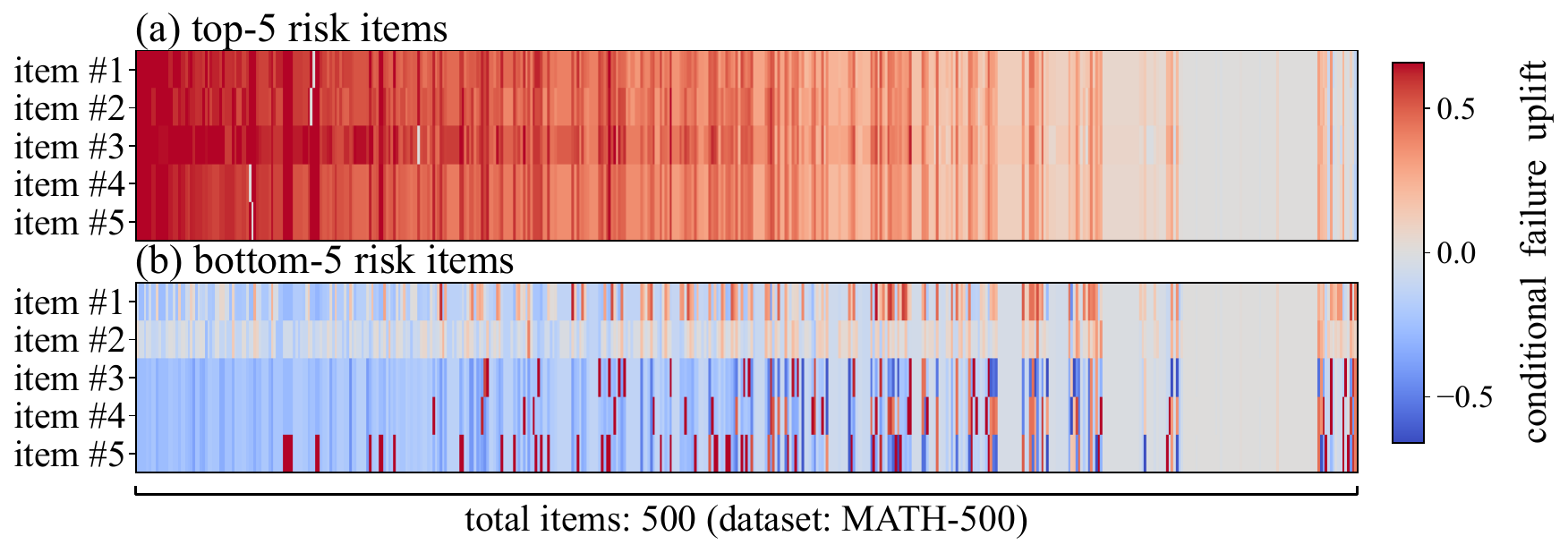}
  \caption{\textbf{High-risk items' failure correlates with wider errors across the dataset.}
  Rows are the 5 highest-risk items in (a) and the 5 lowest-risk items in (b); columns are all MATH-500 items. For a row item $i$ and a column item $k$, the color shows how much the failure probability of item $k$ rises when failing item $i$.}
  \label{fig:risk-case}
\end{figure}

\begin{figure}[b] 
  \centering
  \includegraphics[width=\linewidth]{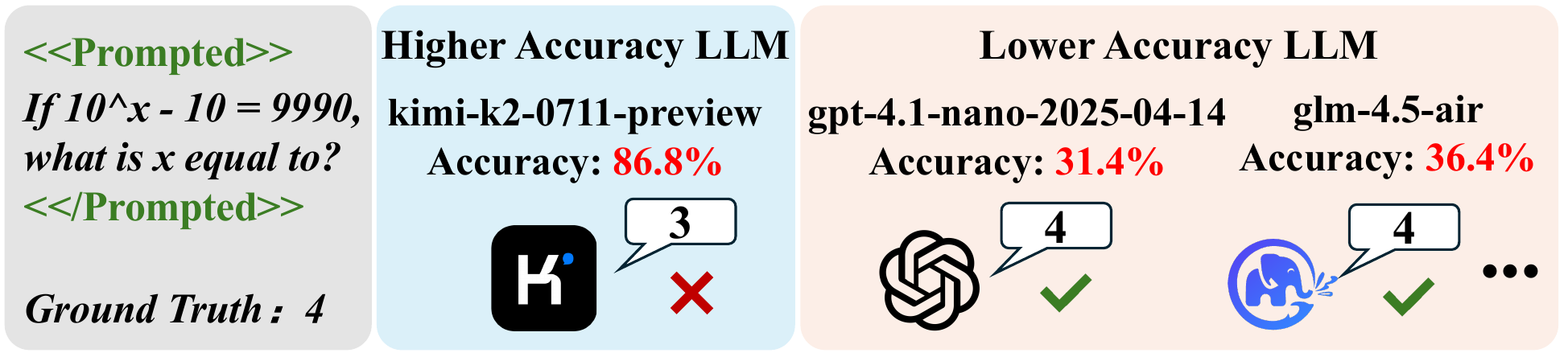}
\caption{\textbf{A surprising case across LLMs on MATH-500.} Kimi-k2, despite higher overall accuracy, fails on this item, whereas lower-accuracy LLMs (GPT-4.1-nano, GLM-4.5-air) succeed.}
  \label{fig:surprise-case-math}
\end{figure}

On the data side, individual items are usually defined only by pre-assigned labels, without further characterization of their latent properties or their ability to differentiate model capabilities. This limits the explanatory power of datasets. For example, some items exhibit riskiness, where failing them strongly correlates with broader error across the dataset (Figure~\ref{fig:risk-case} provides an intuitive illustration). 
On the model side, although many new evaluation metrics have been proposed, they largely broaden the range of overall evaluation scores rather than revealing the deeper structure of model behaviors. Fine-grained differences are often obscured within overall scores, yet such differences typically surface only through population-level comparisons. For instance, certain elite models that excel in overall metrics nevertheless display anomalous errors on questions that most other models solve with ease (Figure~\ref{fig:surprise-case-math} provides an intuitive illustration).

\begin{figure*}[t]
  \centering
  \includegraphics[width=0.8\linewidth]{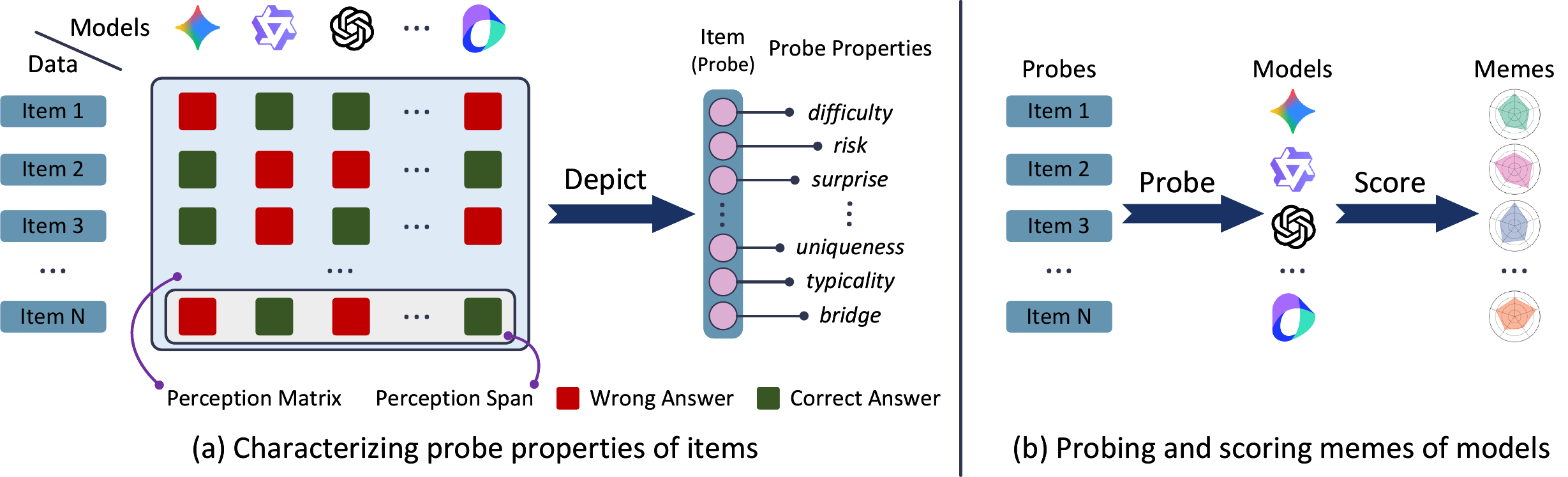}
\caption{\textbf{Overview of the Probing Memes Paradigm.} 
Starting from the Perception Matrix, the paradigm computes diverse item-level properties to construct probes, which are then used to detect models’ memes, providing an interpretable view of fine-grained behavioral structure and underlying capabilities.}
  \label{fig:paradigm}
\end{figure*}

These phenomena highlight the inadequacy of existing evaluation paradigms. To address this gap, this paper introduces the Probing Memes paradigm. As shown in Figure~\ref{fig:paradigm}, the paradigm situates evaluation within a world jointly shaped by interactions between data and models. Here, the notion of meme is borrowed\footnote{In \emph{The Selfish Gene}~\citep{dawkins1976selfish}, memes are described as ``tunes, ideas, catch-phrases, clothes fashions, ways of making pots or of building arches,'' highlighting cultural units replicated through imitation. For the conception of memes and their application to LLMs, refer to the Section~\ref{sec:related_works}.} and metaphorically extended to the context of LLM evaluation, denoting latent units of model behavior that can be revealed through probing. From this perspective, the behavioral traits of LLMs are conceptualized as composed of memes. At the same time, each data item is treated as a Meme Probe (MP) designed to elicit and expose particular aspects of these behavioral traits.

Interactions between probes and models yield the Perception Matrix. Analyzing this matrix enables two complementary abstractions. On the data side, the ability of an item to elicit specific memes is captured by its Meme Probe Properties (MPPs). These properties are derived by generalizing across models and data contexts, revealing deeper characteristics of data and enabling more principled dataset optimization. On the model side, latent memes can be organized into Meme Scores (MSs), making meaningful differences in behavioral traits across models explicit and interpretable. Crucially, MPPs and MSs are designed to be extensible, allowing researchers to define new properties or scores to meet diverse evaluation needs. Together, these dual abstraction moves beyond conventional reliance on overall metrics, enabling evaluation that is more flexible and fine-grained.

The paradigm is validated through applications to 9 datasets and 4,507 LLMs. First, analyses are conducted on 28 models from 11 institutions across MATH-500~\citep{lightman2023let}, MMLU-Redux~\citep{gema-etal-2025-done}, and SimpleQA~\citep{wei2024measuring}, focusing on probe- and model-level perspectives. At the probe level, the analysis illustrates how individual items in the entangled evaluation world can reveal fine-grained insights, such as the fact that datasets like SimpleQA contain a large number of hard items that are nevertheless answered correctly by some weaker models while missed by some stronger ones (quantified by \emph{surprise}, a MPP). At the model level, the analysis reveals differences invisible to conventional evaluations: models with similar overall accuracy may have distinct behavioral traits that favor different item types (e.g., \emph{Caution}, an MS, captures specialization on easy but risk items), enabling more principled model selection for diverse requirements. Second, applying the paradigm to the Open LLM Leaderboard~\citep{open-llm-leaderboard-v2} (6 datasets; 4,479 models) demonstrates scalability while preserving interpretability and flexibility at scale. Overall, the experiments validate the paradigm by making previously obscured, fine-grained behavioral phenomena explicit, underscoring the necessity of moving toward population-based, entangled evaluation.

In conclusion, the contributions of this work are threefold:
\begin{enumerate}
    \item[*] It introduces the \textbf{Probing Memes} paradigm, which places evaluation within an entangled world jointly shaped by data and model interactions;
    \item[*] It formalizes two abstractions, \textbf{Meme Probe Properties} and \textbf{Meme Scores}, enabling structured and extensible characterization across data and models;
    \item[*] It validates the paradigm via large-scale experiments on 9 datasets and 4,507 LLMs, revealing fine-grained phenomena hidden under conventional evaluations.
\end{enumerate}
\section{The Probing Memes Paradigm}
This section introduces the Probing Memes paradigm in three steps: first, formalizing the paradigm as an evaluation paradigm within the entangled world shaped by data-model interactions; second, characterizing MPPs that enable the detection of latent memes; and third, defining MSs as structured representations of model behavioral traits.

\subsection{Formalization of the Paradigm}
\label{subsec:formalization_of_paradigm}
The Probing Memes paradigm can be formalized by specifying data, models, and their interaction. Let $\mathcal{D}=\{(x_i,y_i)\}_{i=1}^n$ denote a dataset of paired data items, where each pair consists of an input $x_i$ and a reference output $y_i$. Let $\mathcal{M}=\{M_j\}_{j=1}^m$ be a collection of LLMs, each viewed as a mapping $M_j:\mathcal{X}\to\mathcal{O}$. For any $(x_i,y_i)$, model $M_j$ produces an output $o_{ij}=M_j(x_i)\in\mathcal{O}$.

A judging function $J:\mathcal{O}\times\mathcal{Y}\to\{0,1\}$ returns a binary correctness indicator for each paired output $(o_{ij},y_i)$, which termed a \emph{perception unit}:

\begin{equation}\label{eq:p_unit}
    P_{ij} = J\big(M_j(x_i), y_i\big).
\end{equation}

Collecting all results yields the \emph{Perception Matrix} $P \in \{0,1\}^{n \times m}$, where rows correspond to probes (items) and columns to models. Figure~\ref{fig:paradigm} provides an intuitive schematic. Each probe $i$ is associated with a perception span $P_i$ (the $i$-th row of $P$; a population-level success/failure pattern on that probe), summarizing how that probe is perceived across the model population and serving as the basis for MPPs.

In memetic terms, this paper posits an underlying \emph{meme space} $\mathcal{V} = \{\mu_1,\mu_2,\dots,\mu_R\}$ of elementary memes and associates each model $M_j$ with a subset $\mathcal{V}_j \subseteq \mathcal{V}$ of memes it carries. The Perception Matrix $P$ provides the empirical interface between these unobserved memes and their observable probe-level expressions on $\mathcal{D}$, preserving the data--model interaction structure and supporting probe properties that quantify how items reveal model behavioral traits.

\subsection{Meme Probe Properties}
\label{subsec:mpps}
The degree to which a probe elicits and exposes specific memes in a model population depends on its intrinsic properties. These properties, termed Meme Probe Properties (MPPs), offer a structured lens for characterizing how individual items reveal meme-related behavioral traits within the joint interaction of model and data populations.

Formally, let $\mathcal{A} \;=\; \{\alpha_{1},\alpha_{2},\dots,\alpha_{K}\}$ denote a \emph{property space}, where coordinate $a_{k}$ represents one property dimension. In this work, $K$ is instantiated as $6$, with dimensions corresponding to six well-designed properties: \emph{difficulty}, \emph{risk}, \emph{surprise}, \emph{uniqueness}, \emph{typicality}, and \emph{bridge}. The following outlines each MPP together with its intended role, definitions, and notation.

\noindent\textbf{Difficulty.} 
A probe should dynamically provide a difficulty baseline based on the performance of the model population. 
Formally, the \emph{difficulty} of the $i$-th data item can be quantified as
\begin{equation}\label{eq:mpp_d}
    d_i = 1 - \frac{1}{|\mathcal{M}|}\sum_{j=1}^{|\mathcal{M}|} P_{ij},
\end{equation}
where $P_{ij}$ denotes the perception unit of model $M_j$ on probe $i$ as defined in Equation~\ref{eq:p_unit}, and $\mathcal{M}$ denotes the model population. 
Intuitively, $d_i$ measures the proportion of models that fail on probe $i$, 
so a higher value indicates greater difficulty relative to the population baseline.

\noindent\textbf{Risk.} 
A probe should reveal high-risk failure modes: failing this probe should be associated with higher failure rates on many other probes. Formally, the \emph{risk} of probe $i$ is defined as
\begin{equation}
\label{eq:risk}
r_i \;=\; \text{scale}_i \cdot \frac{1}{n-1}\sum_{k\neq i}\mathrm{CF}_{i\to k},
\end{equation}
where $\text{scale}_i$ is a coverage factor that downweights probes failed by only a few models.
Here $\mathrm{CF}_{i\to k}$ is an association score in the spirit of the Certainty Factor~\citep{delgado2002association}, measuring the change in the probability of failing probe $k$ conditioned on failing probe $i$. A larger positive $\mathrm{CF}_{i\to k}$ indicates that models failing $i$ tend to fail $k$ more often than expected from $k$'s \emph{difficulty} alone. Complete definitions are provided in Appendix~\ref{appendix:mpps_risk}.

\noindent\textbf{Surprise.} 
A probe should expose anomalies in which stronger models fail on relatively easy probes, or conversely, weaker models succeed on difficult probes, especially when such failures or successes are rare across the population.
Formally, for the easy-side case, the surprise of probe $i$ is
\begin{equation}
s^{\text{easy}}_i \;=\; \bigl(-\ln d_i\bigr) \cdot \frac{1}{|F_i|}\sum_{j \in F_i} a_j,
\end{equation}
where $d_i$ is the difficulty of probe $i$ as defined in Equation~\ref{eq:mpp_d}, 
$F_i=\{j \mid P_{ij}=0\}$ is the set of model indices such that $M_j$ fails probe $i$, and $a_j$ represents the ability of model $M_j$ based on its accuracy over all probes.

Intuitively, $s^{\text{easy}}_i$ becomes large when a probe is solved by most models but disproportionately fails by stronger ones, while $s^{\text{hard}}_i$ highlights the reverse case. The formal definition of $s^{\text{hard}}_i$ is provided in Appendix~\ref{appendix:mpps_surprise}.
Finally, the overall surprise of probe $i$ is given by
\begin{equation}\label{eq:mpp_s}
    s_i = \tfrac{1}{2}\bigl(s^{\text{easy}}_i + s^{\text{hard}}_i\bigr).
\end{equation}

\begin{table*}[t]
\centering
\caption{\textbf{Definitions of Meme Scores with semantic interpretation.}}
\label{tab:memescores_definitions}
\setlength{\tabcolsep}{5pt}
\renewcommand{\arraystretch}{1.12}
\footnotesize

\begin{tabularx}{\linewidth}{@{}
p{3.5cm}
>{\raggedright\arraybackslash}X
>{\centering\arraybackslash}p{2.95cm}
@{}}
\toprule
\textbf{Meme score} & \textbf{Interpretation} & \textbf{Construction} \\
\midrule

\multicolumn{3}{@{}l}{\textbf{Property-derived (1D) meme scores}}\\[-1pt]
Difficulty   & Performs well on difficult probes. & \(f(d)\) \\
Uniqueness   & Performs well on probes with rare behavioral patterns. & \(f(u)\) \\
Risk         & Resists probes that tend to fail alongside many other probes. & \(f(r)\) \\
Surprise     & Handles probes with anomalous behavioral patterns. & \(f(s)\) \\
Typicality   & Proficiency on prototypical probes that represent major behavior clusters. & \(f(t)\) \\
Bridge       & Proficiency on probes that connect multiple behavioral clusters. & \(f(b)\) \\

\addlinespace[4pt]
\multicolumn{3}{@{}l}{\textbf{Predefined (2D) meme scores}}\\[-1pt]
Mastery      & Performs well on difficult, prototypical probes & \(f(t,d)\) \\
Ingenuity    & Flexibility on probes with rare and anomalous behavior patterns. & \(f(u,s)\) \\
Robustness   & Remains correct on high-risk probes at cross-cluster intersections. & \(f(b,r)\) \\

\addlinespace[4pt]
\multicolumn{3}{@{}l}{\textbf{Predefined (3D) meme score}}\\[-1pt]
Caution      & Avoids errors on easy, prototypical, yet high-risk probes. & \(f(t,1-d,r)\) \\
\bottomrule
\end{tabularx}
\end{table*}

\noindent\textbf{Uniqueness.}
A probe is unique if its perception span is dissimilar to those of other probes.
Let $S_{ik}=\mathrm{sim}(P_i,P_k)\in[0,1]$ denote the Hamming similarity (see Appendix~\ref{appendix:hamming_similarity}) between probes $i$ and $k$ computed from their perception spans.
The \emph{uniqueness} of probe $i$ is defined as
\begin{equation}\label{eq:mpp_u}
u_i \;=\; 1 - \frac{1}{\,n-1\,}\sum_{\substack{k=1\\k\ne i}}^{n} S_{ik}\;\in[0,1].
\end{equation}
Intuitively, $u_i$ is high when probe $i$ has a low average similarity to other probes, indicating a more distinctive failure/success pattern across the model population.

\noindent\emph{Cluster Construction.}
Given two probes $i$ and $k$, their similarity $\mathrm{sim}(P_i,P_k)$ is measured by the Hamming similarity, which compares the perception spans element-wise and quantifies the extent to which the model population exhibits similar success/failure patterns on the two probes. Based on this similarity, an undirected weighted graph $G=(V,E)$ is defined, where each node corresponds to a probe, and an edge $(i,k)$ is included if $\mathrm{sim}(P_i,P_k)\ge\tau$, with the edge weight set to the similarity value. Here, $\tau$ controls the sparsity of the graph. A hierarchical clustering procedure~\citep{müllner2011modernhierarchicalagglomerativeclustering} on this graph produces a partition $\mathcal{C}=\{C_1,C_2,\ldots,C_L\}$ of probes into clusters. These clusters are subsequently used to define the \emph{typicality} and \emph{bridge} properties (see Appendix~\ref{appendix:clustering} for detailed cluster process).

\noindent\textbf{Typicality.}
A probe should be considered a prototype if its perception span shows similarity to other probes in the cluster.
Given the cluster partition $\mathcal{C}$ and the similarity matrix $S$, the \emph{typicality} value of probe $i\in C_\ell$ is defined as
\begin{equation}\label{eq:mpp_t}
t_i =
\begin{cases}
\displaystyle
\tfrac{1}{2}+\tfrac{1}{2}\,h(|C_\ell|)\,\mathrm{Intra}(C_\ell), & \text{if } i=p_{C_\ell},\\[8pt]
\displaystyle
g(|C_\ell|)\,\mathrm{Cen}(i;C_\ell), & \text{otherwise},
\end{cases}
\end{equation}
where $p_{C_\ell}$ denotes the prototype probe of cluster $C_\ell$ (i.e., the member with the largest aggregate similarity to other probes in the cluster), $\mathrm{Intra}(C_\ell)$ is the average similarity among probes within $C_\ell$, $\mathrm{Cen}(i;C_\ell)$ quantifies the similarity between probe $i$ and the rest of its cluster, and $h(\cdot)$ and $g(\cdot)$ are cluster-size-dependent scaling functions (with detailed formal definitions provided in Appendix~\ref{appendix:typicality}).

\noindent\textbf{Bridge.}
A probe is bridge-like if its similarities span multiple clusters rather than concentrating in a single cluster.
Inspired by the Participation Coefficient~\citep{guimera2005functional}, given the cluster partition $\mathcal{C}$ and similarity matrix $S$, the bridge value of probe $i$ is defined as
\begin{equation}\label{eq:mpp_b}
b_i \;=\;
1 \;-\; \sum_{\ell=1}^{L}
\left(
\frac{\sum_{k\in C_\ell,\;k\neq i} S_{ik}}{\sum_{k\neq i} S_{ik}}
\right)^{\!2}\ \in[0,1).
\end{equation}
Intuitively, $b_i$ is small when most of $i$'s similarity concentrates within a single cluster, and increases as its similarity mass is more evenly distributed across clusters.

\subsection{Meme Scores of LLMs}
\label{sec2:meme_scores_of_LLMs}
Given the \emph{meme space} $\mathcal{V}$ and the \emph{property space} $\mathcal{A}$, any nonempty subset $U\subseteq \mathcal{A}$ is mapped by a \textit{construction operator} $f$ to an induced meme $v=f(U)\in\mathcal{V}$. Here, $v$ captures the behavioral trait defined by combining the property dimensions in $U$. The operator $f$ further induces a probe weight $w(U)$, specifying how probes contribute to the MS of $v$. Thus, the MS of model $M_j$ under meme $v$ is defined as a generic weight-aggregated score over the $j$-th column of the Perception Matrix:
\begin{equation}
\label{eq:meme_score_function}
\mathrm{MemeScore}(v;\,M_j)
= \mathrm{Score}\!\bigl(w(U);\;P_{\cdot j}\bigr).
\end{equation}
where $\mathrm{Score}(\cdot)$ is the generic aggregation operator.

\begin{figure*}[t]
  \centering
  \includegraphics[width=\linewidth]{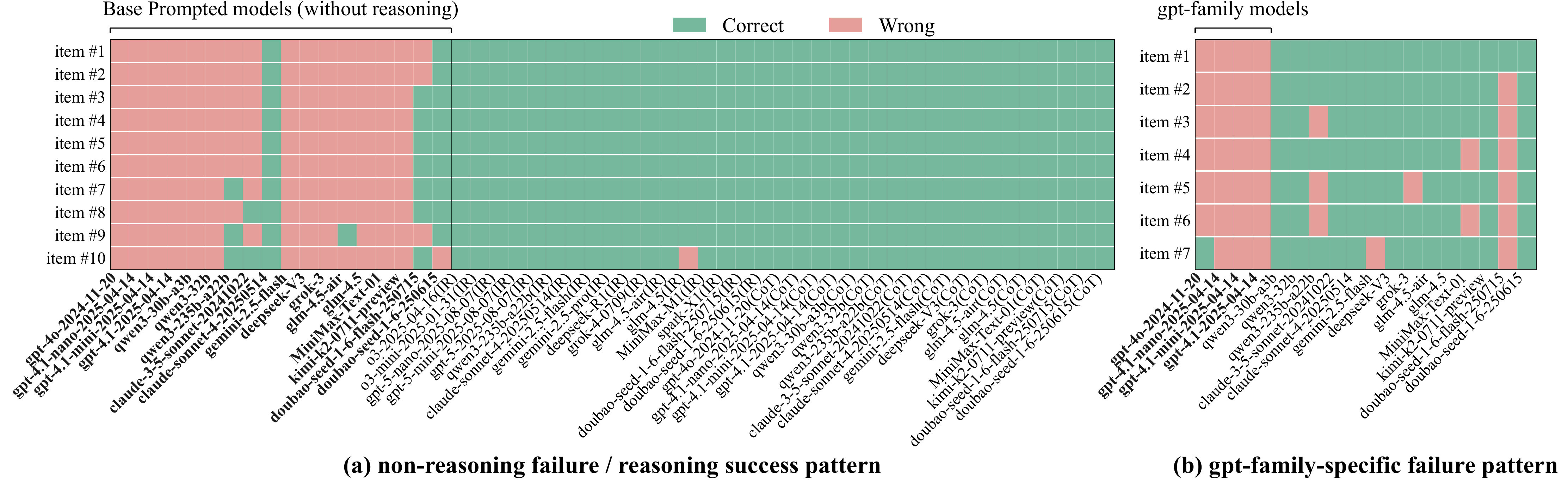}
  \caption{\textbf{Probe clusters reveal distinct population behavioral patterns (Curated Population).} Within each cluster, rows represent items and their corresponding perception spans, which record the error of every model. Two representative clusters are shown: (a) A cluster where all base-prompted variants fail while stronger reasoning modes succeed, suggesting that explicit reasoning improves reliability on this type of item. (b) A cluster where gpt-family models fail consistently despite high accuracy for many other models.
 }
\label{fig:probe_behavioral_pattern}
\end{figure*}

This paper instantiates a set of meme scores for analysis. In particular, property-derived (1D) meme scores are mapped from individual MPPs. Moreover, since \emph{uniqueness}, \emph{typicality}, and \emph{bridge} more directly capture behavioral information, they are further combined with other properties to form a set of predefined 2D meme scores. The score set further includes a predefined 3D meme score, \emph{Caution}. Table~\ref{tab:memescores_definitions} summarizes the property-derived (1D) meme scores and predefined (2D--3D) meme scores, together with their semantic interpretations, where \(d,u,r,s,t,b\) denote the probe-side properties \emph{difficulty}, \emph{uniqueness}, \emph{risk}, \emph{surprise}, \emph{typicality}, and \emph{bridge}, respectively. Implementation details are provided in Appendix~\ref{appendix:meme_scores}.

\section{Experiments and Analyses}
\label{sec:experiments_analyses}
The experiments are based on a \textbf{Curated Population} with standardized evaluation configurations, and an \textbf{Open LLM Population} based on publicly available leaderboard results. This section first specifies the experimental setup for the two population-based experiments, and then presents data-side and model-side results along with analyses for each.

\subsection{Experimental Setups}
\label{subsec:experimental_setups}
\paragraph{Curated Population.} The curated population includes 28 widely used, publicly available LLMs from 11 providers, spanning a broad range of model sizes and versions from October 22, 2024 (claude-3-5-sonnet-20241022) to August 7, 2025 (gpt-5-2025-08-07). Three widely used datasets across distinct tasks (mathematics, general knowledge, and question answering) are selected: MATH-500~\citep{lightman2023let}, MMLU-Redux~\citep{gema-etal-2025-done}, and SimpleQA~\citep{wei2024measuring}. The full model list and dataset information are provided in Appendix~\ref{app:models-curated} and~\ref{app:datasets-curated}.

The study analyzes three reasoning modes: \textbf{default prompting (Base)}, \textbf{chain-of-thought prompting (CoT)}, and \textbf{internal reasoning (IR)}. \textbf{Base} and \textbf{CoT} differ only in the prompting template: \textbf{CoT} uses an explicit chain-of-thought instruction to elicit step-by-step reasoning, whereas \textbf{Base} uses a default instruction without any reasoning cue. 
\textbf{IR} refers to models that execute multi-step reasoning intrinsically without requiring an explicit chain-of-thought prompt, as in so-called reasoning models such as DeepSeek-R1~\citep{guo2025deepseek}. 

\begin{figure*}[t]
  \centering
  \includegraphics[width=\linewidth]{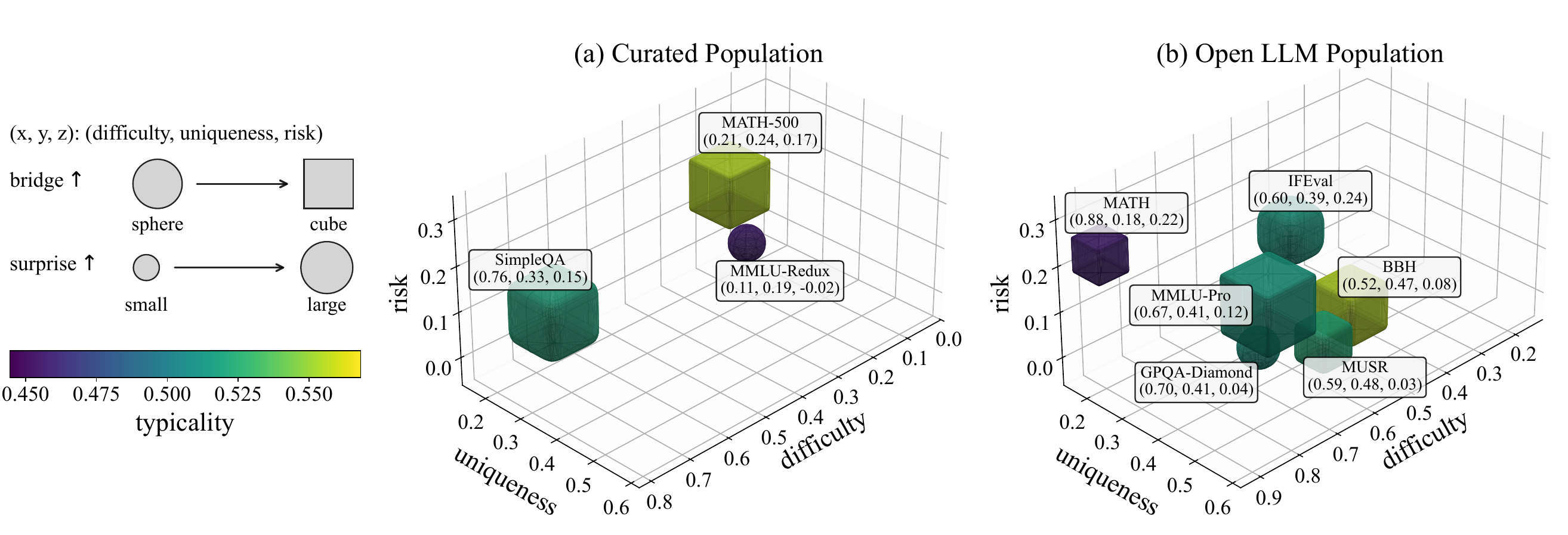}
  \caption{\textbf{A 3D landscape of datasets in the lens of Probe Properties.} Six Probe Properties are visualized via the 3D position (axes), color, marker size, and shape, using dataset-level averages over all probes.}
\label{fig:datasets_landscape_3d}
\end{figure*}

Accordingly, two prompting templates are used: a default template and a chain-of-thought template. \textbf{Base} and \textbf{IR} use the default template, whereas \textbf{CoT} uses the chain-of-thought template. Each model is evaluated under a subset of the three modes: reasoning models that support internal reasoning are evaluated under \textbf{IR}, while models without internal reasoning support are evaluated under \textbf{Base} and \textbf{CoT}. For each dataset, under these reasoning modes and corresponding hyperparameter settings, the 28 models produce 53 distinct evaluation results, which constitute the Perception Matrix; therefore, the subsequent analysis considers $|\mathcal{M}|=53$. Hyperparameter settings and prompt templates are provided in the Appendix~\ref{app:hyperparameters} and~\ref{app:prompt_templates}.

\begin{figure}[!b]
  \centering
  \includegraphics[width=0.8\linewidth]{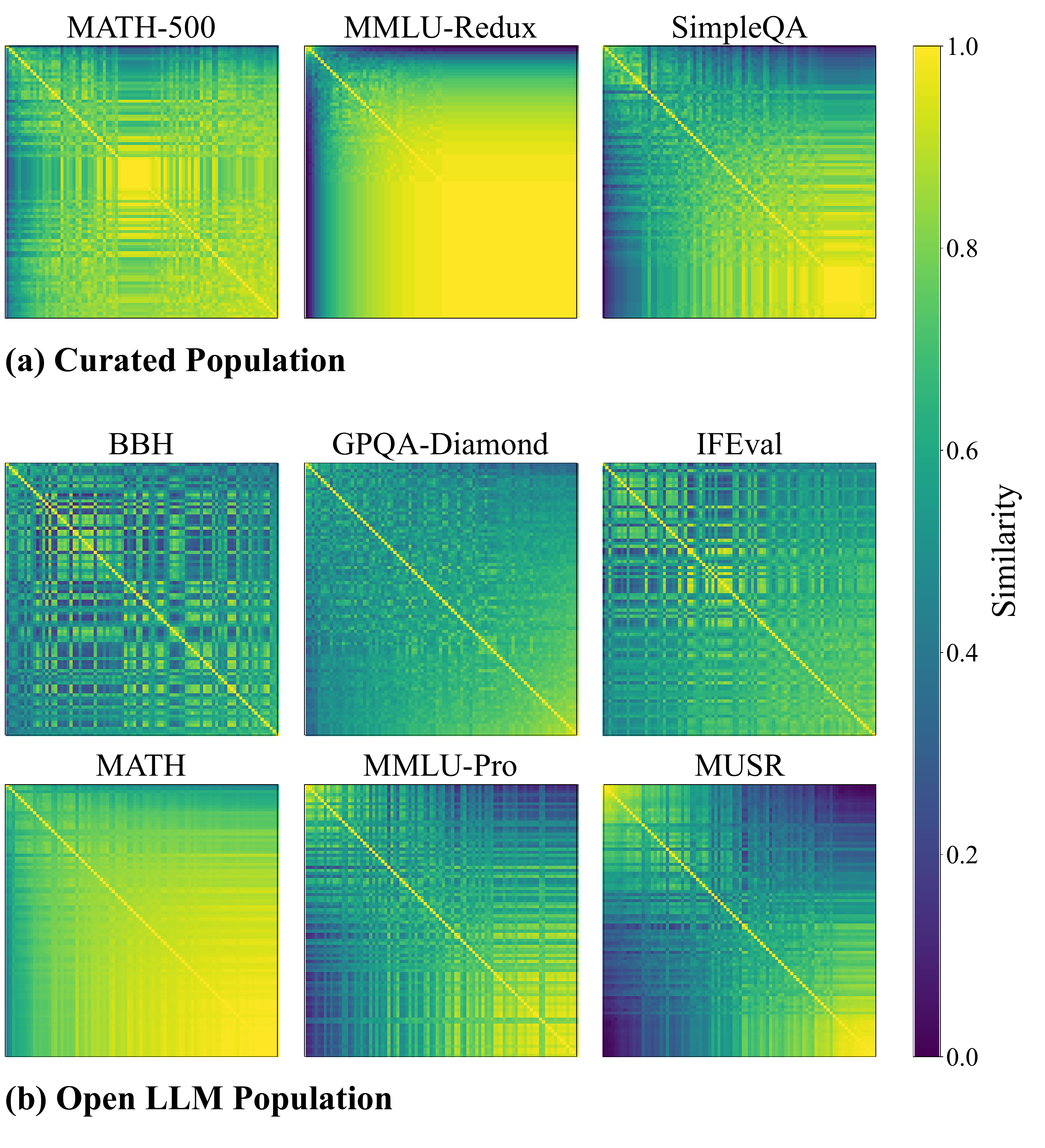}
  \caption{\textbf{Probe similarity heatmaps across datasets.}}
  \label{fig:heat_map}
\end{figure}

\paragraph{Open LLM Population (Large-Scale).}
Results for 4{,}479 models across six datasets are collected from the Open LLM Leaderboard~\citep{open-llm-leaderboard-v2}. These leaderboard-reported results are used to construct a Perception Matrix, enabling an instantiation of the Probing Memes paradigm at scale.
Additional details on the included models and datasets are provided in Appendix~\ref{app:OpenLLM_models_and_datasets}.

\subsection{Data-side Characterization}
\label{subsec:probe_side_characterization}

\paragraph{Probe-revealed Population Behavioral Structure.}
Following the formalization in Section~\ref{subsec:formalization_of_paradigm}, each item $i$ (probe $i$) is associated with a perception span $P_i$, corresponding to a row of the Perception Matrix $P$, which captures the model population's behavioral pattern on that probe.

Building on the similarity and clustering procedure in Section~\ref{subsec:mpps}, items can then be grouped into interpretable clusters using perception-span similarity, revealing distinct population-level behaviors.
For example, Figure~\ref{fig:probe_behavioral_pattern} shows a cluster of seven items on which the gpt-family models consistently fail while most other models succeed. Such patterns, exposed by the paradigm, reveal family-specific failure modes that are obscured under aggregate metrics, enabling model providers to perform targeted diagnosis and optimization.

Across datasets, probe--probe similarity patterns differ systematically, indicating distinct population-level behavioral organizations across model populations. This contrast is particularly evident in MMLU-Redux, where only a small fraction of probes exhibit rare population behaviors (high \emph{uniqueness}), while the majority share highly similar response patterns. Such long-tailed behavioral structure suggests a practical direction for item selection and dataset design: prioritizing behaviorally distinctive probes with informative MPPs. Figure~\ref{fig:heat_map} visualizes these probe--probe similarities as heatmaps for all datasets, with probes sorted by \emph{uniqueness}. Additional heatmaps sorted by other MPPs are provided in Appendix~\ref{app:more_heatmaps}.

\paragraph{Dataset Landscape induced from Probe Properties.} Each dataset can be represented by the average values of the six MPPs computed over its probes, enabling a compact characterization of dataset-level behavioral profiles. Figure~\ref{fig:datasets_landscape_3d} visualizes these profiles in a shared landscape, highlighting clear variation across datasets along multiple interpretable MPP dimensions. 

For example, under the Curated Population, SimpleQA exhibits notably high \emph{difficulty} and \emph{surprise}, indicating many probes on which weaker models succeed despite the overall hardness of the items. In the Open LLM Population, a similarly interesting phenomenon emerges: IFEval, though easier overall than GPQA-Diamond (\emph{difficulty} 0.60 vs.\ 0.70), shows substantially higher \emph{risk} (0.24 vs.\ 0.04), suggesting that some probes can be relatively easy yet highly risky, and these probes deserve attention in dataset design, rather than focusing solely on making items harder.

Together, this representation enables direct cross-dataset behavioral comparison along interpretable dimensions, supporting informed dataset selection and benchmark refinement for both benchmark designers and model providers. More fine-grained per-probe property distributions are provided in Appendix~\ref{app:distribution_of_MPPs}.

\subsection{Model-side Meme Profiling}
\label{subsec:model_side_meme_profiling}

\begin{table*}[!t]
\centering
\caption{\textbf{Meme Scores (Curated Population; Averaged over MATH-500, MMLU-Redux, and SimpleQA).} The table reports meme scores across models, including property-derived 1D meme scores, predefined 2D meme scores (Mastery, Ingenuity, and Robustness), and a predefined 3D meme score (Caution). Models are sorted by Accuracy. {\color{blue}Blue} indicates rank improvement compared with Accuracy rank, and {\color{red}red} indicates rank degradation. For brevity, only the top five, the five around the median, and the bottom five models under the Accuracy ranking are shown. The complete ranking results are provided in Appendix~\ref{app:MemeScore_Leaderboard_curated_full}.}
\resizebox{\textwidth}{!}{
\begin{tabular}{cccccccccccc}
\toprule
\multirow{2}{*}{\textbf{Model}} & \multirow{2}{*}{\textbf{Accuracy}} &
\multicolumn{6}{c}{\textbf{Property-derived 1D meme scores}} &
\multicolumn{4}{c}{\textbf{Predefined 2D/3D meme scores}} \\
\cmidrule(lr){3-8}\cmidrule(lr){9-12}
 &  & \textbf{Difficulty} & \textbf{Uniqueness} & \textbf{Risk} & \textbf{Surprise} & \textbf{Typicality} & \textbf{Bridge} & \textbf{Mastery} & \textbf{Ingenuity} & \textbf{Robustness} & \textbf{Caution} \\
\midrule
gemini-2.5-pro(IR) & 82.57 & 68.51\, \textcolor{red}{(-2)} & 76.69\, \textcolor{red}{(-3)} & 90.07 & 71.23\, \textcolor{red}{(-2)} & 84.37 & 79.08 & 72.75 & 62.51\, \textcolor{red}{(-3)} & 86.91 & 95.34 \\
grok-4-0709(IR) & 81.92 & 68.70\, \textcolor{blue}{(+1)} & 79.48\, \textcolor{blue}{(+1)} & 89.29 & 71.62 & 83.39 & 77.55 & 72.54 & 66.89\, \textcolor{blue}{(+1)} & 84.89 & 95.20\, \textcolor{red}{(-1)} \\
gpt-5-2025-08-07(IR) & 81.51 & 68.52\, \textcolor{blue}{(+1)} & 79.19\, \textcolor{blue}{(+1)} & 89.05 & 71.79\, \textcolor{blue}{(+2)} & 83.22 & 77.04\, \textcolor{red}{(-1)} & 72.46 & 66.28\, \textcolor{blue}{(+1)} & 84.49\, \textcolor{red}{(-1)} & 95.32\, \textcolor{blue}{(+1)} \\
o3-2025-04-16(IR) & 81.14 & 67.70 & 78.40\, \textcolor{blue}{(+1)} & 88.81 & 70.58 & 83.04 & 77.06\, \textcolor{blue}{(+1)} & 72.02 & 64.80\, \textcolor{blue}{(+1)} & 84.79\, \textcolor{blue}{(+1)} & 95.16 \\
grok-3(CoT) & 76.69 & 59.63 & 70.05\, \textcolor{red}{(-2)} & 83.78 & 65.51 & 78.00 & 73.58\, \textcolor{red}{(-1)} & 64.31 & 59.25 & 79.61\, \textcolor{red}{(-1)} & 91.65\, \textcolor{red}{(-1)} \\
{\Large$\boldsymbol{\vdots}$} & {\Large$\boldsymbol{\vdots}$} & {\Large$\boldsymbol{\vdots}$} & {\Large$\boldsymbol{\vdots}$} & {\Large$\boldsymbol{\vdots}$} & {\Large$\boldsymbol{\vdots}$} & {\Large$\boldsymbol{\vdots}$} & {\Large$\boldsymbol{\vdots}$} & {\Large$\boldsymbol{\vdots}$} & {\Large$\boldsymbol{\vdots}$} & {\Large$\boldsymbol{\vdots}$} & {\Large$\boldsymbol{\vdots}$} \\
o3-mini-2025-01-31(IR) & 67.39 & 49.95\, \textcolor{blue}{(+4)} & 56.67\, \textcolor{blue}{(+1)} & 69.83\, \textcolor{red}{(-1)} & 60.86\, \textcolor{blue}{(+7)} & 65.62\, \textcolor{red}{(-1)} & 67.00\, \textcolor{blue}{(+1)} & 52.33\, \textcolor{blue}{(+3)} & 51.41\, \textcolor{blue}{(+7)} & 67.91\, \textcolor{red}{(-1)} & 73.96\, \textcolor{red}{(-3)} \\
gpt-5-nano-2025-08-07(IR) & 66.81 & 51.42\, \textcolor{blue}{(+9)} & 58.43\, \textcolor{blue}{(+5)} & 68.60\, \textcolor{red}{(-1)} & 61.61\, \textcolor{blue}{(+12)} & 64.59\, \textcolor{red}{(-1)} & 66.50\, \textcolor{red}{(-1)} & 52.59\, \textcolor{blue}{(+5)} & 55.37\, \textcolor{blue}{(+15)} & 67.17\, \textcolor{red}{(-3)} & 71.80\, \textcolor{red}{(-4)} \\
claude-3-5-sonnet-20241022(CoT) & 66.08 & 43.66\, \textcolor{red}{(-3)} & 53.21 & 71.73\, \textcolor{blue}{(+4)} & 54.42\, \textcolor{red}{(-5)} & 65.86\, \textcolor{blue}{(+2)} & 66.67\, \textcolor{blue}{(+1)} & 47.88\, \textcolor{red}{(-2)} & 42.70\, \textcolor{red}{(-8)} & 70.54\, \textcolor{blue}{(+10)} & 80.82\, \textcolor{blue}{(+7)} \\
doubao-seed-1-6-250615 & 65.01 & 44.42\, \textcolor{red}{(-1)} & 50.47\, \textcolor{red}{(-3)} & 68.17 & 55.86\, \textcolor{red}{(-2)} & 63.69 & 66.17 & 46.83\, \textcolor{red}{(-2)} & 42.80\, \textcolor{red}{(-6)} & 67.72\, \textcolor{blue}{(+1)} & 75.19\, \textcolor{blue}{(+2)} \\
doubao-seed-1-6-flash-250715(IR) & 64.45 & 48.32\, \textcolor{blue}{(+5)} & 51.62\, \textcolor{blue}{(+1)} & 65.43\, \textcolor{red}{(-4)} & 58.13\, \textcolor{blue}{(+4)} & 62.09 & 65.77\, \textcolor{red}{(-2)} & 50.34\, \textcolor{blue}{(+3)} & 47.76\, \textcolor{blue}{(+5)} & 65.98\, \textcolor{red}{(-3)} & 67.09\, \textcolor{red}{(-6)} \\
{\Large$\boldsymbol{\vdots}$} & {\Large$\boldsymbol{\vdots}$} & {\Large$\boldsymbol{\vdots}$} & {\Large$\boldsymbol{\vdots}$} & {\Large$\boldsymbol{\vdots}$} & {\Large$\boldsymbol{\vdots}$} & {\Large$\boldsymbol{\vdots}$} & {\Large$\boldsymbol{\vdots}$} & {\Large$\boldsymbol{\vdots}$} & {\Large$\boldsymbol{\vdots}$} & {\Large$\boldsymbol{\vdots}$} & {\Large$\boldsymbol{\vdots}$} \\
MiniMax-Text-01 & 48.85 & 25.98\, \textcolor{blue}{(+1)} & 41.34\, \textcolor{blue}{(+6)} & 49.40 & 47.23 & 42.09 & 45.72\, \textcolor{red}{(-1)} & 26.11\, \textcolor{blue}{(+1)} & 40.77\, \textcolor{blue}{(+8)} & 44.52\, \textcolor{red}{(-1)} & 51.87\, \textcolor{blue}{(+1)} \\
qwen3-30b-a3b & 46.62 & 23.44 & 29.14\, \textcolor{red}{(-1)} & 45.40 & 45.40 & 38.53 & 48.58\, \textcolor{blue}{(+1)} & 23.42 & 35.19 & 46.76\, \textcolor{blue}{(+1)} & 42.06 \\
qwen3-32b & 43.80 & 20.61 & 28.78\, \textcolor{red}{(-1)} & 41.59\, \textcolor{red}{(-1)} & 42.97\, \textcolor{red}{(-1)} & 35.80 & 44.66 & 20.32 & 33.34\, \textcolor{red}{(-2)} & 41.66 & 38.85\, \textcolor{red}{(-1)} \\
glm-4.5-air & 43.64 & 18.30 & 31.57\, \textcolor{blue}{(+2)} & 42.37\, \textcolor{blue}{(+1)} & 43.60\, \textcolor{blue}{(+1)} & 34.25 & 42.77 & 17.82 & 36.61\, \textcolor{blue}{(+3)} & 40.36 & 40.94\, \textcolor{blue}{(+1)} \\
gpt-4.1-nano-2025-04-14 & 38.79 & 17.25 & 26.77 & 36.67 & 39.48 & 29.95 & 39.21 & 17.57 & 33.67\, \textcolor{blue}{(+1)} & 36.62 & 32.45 \\
\bottomrule
\end{tabular}}
\label{tab:meme_scores_summary}
\end{table*}

\paragraph{Meme score results.}
In this paradigm, models are described not only by overall accuracy but also by a set of Meme Scores that correspond to different behavioral traits, with detailed interpretations provided in Table~\ref{tab:memescores_definitions}. As shown in Table~\ref{tab:meme_scores_summary}, the Curated Population is ranked under different Meme Scores and compared against the accuracy ranking. Substantial rank shifts across multiple Meme Scores indicate that models with similar accuracy can exhibit different behavioral traits. 

For example, \textit{gpt-5-nano-2025-08-07(IR)} and \textit{claude-3-5-sonnet-20241022(CoT)} achieve comparable accuracy (66.81 vs.\ 66.08) and are adjacent in the accuracy ranking, yet they show clearly different Meme Scores (e.g., \emph{Difficulty}: 51.42 vs.\ 43.66; \emph{Caution}: 71.80 vs.\ 80.82), with corresponding rank shifts relative to the accuracy ranking (\textit{gpt-5-nano-2025-08-07(IR)}: \emph{Difficulty} + 9, \emph{Caution} - 4; \textit{claude-3-5-sonnet-20241022(CoT)}: \emph{Difficulty} - 3, \emph{Caution} + 7). These differences suggest behavioral traits that are not captured by accuracy alone. In particular, \textit{gpt-5-nano-2025-08-07(IR)} appears more capable on harder items (as indicated by the \emph{Difficulty} MS), whereas \textit{claude-3-5-sonnet-20241022(CoT)} appears more reliable on easy but high-risk typical items (as indicated by the \emph{Caution} MS).

Similar discrepancies are observed in the Open LLM Population. For instance, \textit{Galactic-Qwen-14B-Exp2} drops by 419 positions in \emph{Robustness} relative to its accuracy ranking. The ranking results for the Open LLM Population are provided in Appendix~\ref{app:MemeScore_Leaderboard_openllm}.

\begin{figure}[!b]
  \centering
\includegraphics[width=0.7\linewidth]{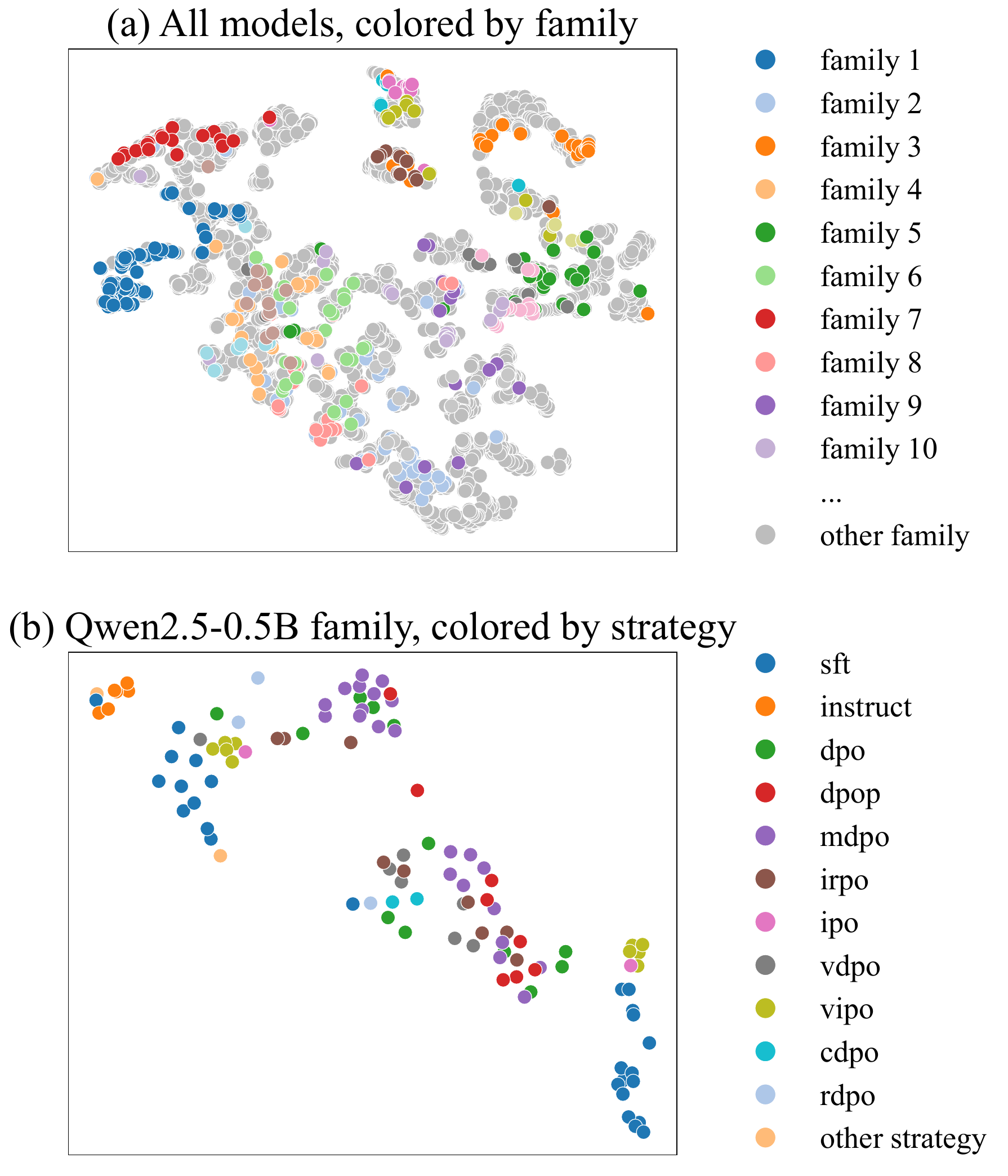} 
\caption{\textbf{Commonality and divergence among models revealed by Meme Scores on MMLU-Pro (Open LLM population).}
(a) Models are colored by family based on their reported shared base model (i.e., the same pre-trained checkpoint or architecture) on Hugging Face.
The 20 largest families are color-coded, and the remaining models are shown in gray.
(b) 
Models in the Qwen2.5-0.5B family are colored by training strategy.}
 \label{fig:hf_models_tsne_panel_two}
\end{figure}

\paragraph{Commonality and divergence among models revealed by Meme Scores.}  
Figure~\ref{fig:hf_models_tsne_panel_two} shows a t-SNE visualization of the Open LLM population, where each model is represented by its 1D Meme Scores. Using each model's reported base model on Hugging Face, models are labeled into families or training strategies. Notably, models from the same family tend to lie closer together in the visualization; for instance, the blue (family~1), green (family~5), and red (family~7) models each form clearly separated groups. Furthermore, focusing on the Qwen2.5-0.5B family and marking different training strategies reveals a similar effect: for example, models with \emph{sft} (blue), \emph{instruct} (orange), and \emph{mdpo} (purple) exhibit clear separation. Results on other datasets are provided in Appendix~\ref{app:commondity_characterize_Meme_scores}.

In the Open LLM population (4,479 open-source models), Meme Scores provide a behavioral characterization that can distinguish models not only across families but also within the same family. The same MS-based approach can be applied to closed-source model populations, helping identify commonalities and divergences that may relate to differences in training data, base models, and training strategies.

\section{Implications: What the Probing Memes Paradigm Enables}
\label{sec:implications}
\paragraph{Decision Support via Meme-Guided Model Routing}
To demonstrate that Meme Scores are not only interpretable for characterizing models' behavioral traits but also applicable in tasks such as model selection, this study conducts a meme-guided routing experiment. Based on Meme Scores computed on MATH-500, two models with similar overall accuracy are selected: one with a higher \emph{Difficulty} Meme Score (derived from the Probe Property \emph{difficulty}), which indicates that the model is better at harder items, and the other with a lower \emph{Difficulty} MS, which indicates a better fit for easier items.

The routing is evaluated on an unseen split of the MATH dataset~\citep{hendrycksmath2021} (excluding MATH-500), where each item is pre-annotated with a difficulty level. Harder items and easier items are then routed to the model that is more suitable for them based on the MS. Compared with baselines, the routing yields consistent accuracy improvements, reaching an accuracy improvement of up to 3.15 percentage points. Overall, these results highlight that the Probing Memes paradigm and its Meme Scores can support effective model selection in settings such as task-aware routing and multi-agent pipelines. Full experimental details are provided in Appendix~\ref{app:meme-guided-routing}.

\paragraph{Behavioral Diagnosis via Probe-Level Surprise Analysis}
The Probing Memes Paradigm supports fine-grained analysis at the probe level, enabling targeted investigations of individual items within a dataset. As a diagnostic case study, high-surprise items are examined, which capture rare but informative phenomena such as stronger models failing on easy items while weaker models succeeding on harder ones.

Repeated evaluations under the Curated Population are used to analyze the underlying factors contributing to high surprise. The analysis shows that these items can be systematically distinguished into cases driven by genuine model capability versus those dominated by stochastic guessing behavior. Interestingly, a counterintuitive effect is observed: adding an explicit ``don't guess'' instruction to the prompt can, in some cases, improve model accuracy on these high-surprise items.

These findings illustrate how probe-level MPPs enable concrete behavioral diagnosis beyond aggregate performance metrics, providing actionable insight into how and why models succeed or fail on specific items. Additional analyses and details are provided in Appendix~\ref{app:high-surprise-case}.

\paragraph{Robustness of Probe Properties and Meme Scores via Population Subsampling}
To assess the stability of Probe Properties and Meme Scores estimated from model populations, random subsampling experiments are conducted over the Curated Population, with subsample sizes ranging from 5 to 50. The results show that most Probe Properties and Meme Score rankings stabilize rapidly as population size increases, and become highly stable once the subsample size reaches approximately $30$--$40$ models (i.e., more than half of $|\mathcal{M}|$).

The remaining variability is primarily concentrated in MPPs such as \emph{uniqueness} and \emph{bridge}, along with their corresponding \emph{bridge-based} Meme Scores. This sensitivity is expected, as these measures depend on global probe--probe relations and clustering structure, which are inherently more sensitive to population size. 

Overall, these findings indicate that the Probing Memes Paradigm can yield stable Probe Properties and Meme Scores with a moderate-sized model population, supporting its practical deployment on real-world and diverse model collections. Additional analyses and details are provided in Appendix~\ref{app:stability_population_size}.
\section{Related Work}
\label{sec:related_works}
\paragraph{Memetics and its applications to LLMs.} Dawkins first proposed the concept of \textit{memes} ~\citep{dawkins1976selfish}, drawing an analogy to genes in cultural transmission. Building on this analogy, memetics introduced memotype and phemotype, paralleling genotype and phenotype \citep{grant1990memetic, blackmore2000meme, fomin2019memes}. In memetics, the \textit{memotype} refers to the actual information content of a meme, while the \textit{phemotype} denotes its concrete manifestation as produced by the memotype under specific conditions. Within LLM research, memetics has been applied to model ideological propagation \citep{10894714} and to explain reasoning behaviors \citep{BirchallManuscript-BIRPAA-2}.

Compared with the memetics definition of a meme as a basic unit of cultural transmission (e.g., a tune, idea, catch-phrase, or way of acting), the Probing Memes Paradigm views a meme as a latent behavioral factor shared across models that can be probed through items with different, purposefully designed properties. To better measure the extent to which different models carry such memes, this work introduces Meme Scores, analogous to phemotypes in memetics, to quantify the memes expressed on probes.

\paragraph{Comparative LLM Behavior Analysis.}  
Recent work studies LLMs comparatively, showing that model outputs exhibit structured similarity and divergence rather than independent failures. A central line analyzes correlated errors: \citet{bradley2024llms} finds high agreement on incorrect options and uses clustering method to reveal model families and shared failure modes; \citet{Goel2025GreatMT} further finds that as the capabilities of LLMs increase, errors across models become more correlated, leading to a higher tendency to select the same incorrect options; and \citet{kim2025correlated} provides large-scale evidence across hundreds of models, linking higher agreement rates to shared developer, architecture, and size. A related direction profiles models beyond outcome correlations:~\citet{pei2025behavioral} constructs behavioral fingerprints by querying models with diagnostic prompts and scoring responses with LLM-based judging, yielding capability profiles (e.g., world-model reasoning and metacognition), highlighting systematic cross-model differences and linking these behavioral abilities to developer-specific alignment strategies.

Compared with these works, the Probing Memes paradigm goes a step further: it first derives probe properties that reflect model behaviors from a model population’s actual performance on items, and then induces models’ behavioral traits (Memes), moving beyond merely showing that models err on different items. Moreover, it can be applied directly to mainstream datasets without specially designed prompts to elicit behaviors~\citep{pei2025behavioral}.

\paragraph{Population-based IRT analysis in LLMs.} 
Item Response Theory (IRT)~\citep{BakerKim2004} estimates a respondent's latent ability together with item difficulty and discrimination from observed responses. \citet{magis2013note} provides the general form of the four-parameter logistic (4PL) model:
\begin{equation}
P_j(\theta)
= c_j + (d_j - c_j)\,\frac{\exp\!\left[a_j(\theta - b_j)\right]}{1+\exp\!\left[a_j(\theta - b_j)\right]}.
\end{equation}
It specifies the probability that item $j$ is answered correctly given an individual's ability $\theta$ and item parameters $(a_j,b_j,c_j,d_j)$.
Common variants such as 1PL, 2PL, and 3PL can be viewed as special cases obtained by imposing constraints on the 4PL form. Under the standard IRT assumption, $P_j(\theta)$ increases with $\theta$, i.e., higher latent ability implies a higher probability of answering item $j$ correctly~\citep{wallmark2024introducing}.

By migrating IRT from psychology and education, \citet{kipnis2024metabench}, \citet{polo2024tinybenchmarks}, and \citet{schilling2025lifting} all fit latent model ability jointly with item parameters such as discrimination and difficulty ($a_j$ and $b_j$); \citet{kipnis2024metabench} and \citet{polo2024tinybenchmarks} further leverage item information for subset selection and adaptive testing to reduce evaluation cost, while \citet{schilling2025lifting} uses a Bayesian 2PL formulation to provide calibrated ability estimates with uncertainty and to examine how leaderboard rankings can shift under an IRT lens.

IRT relies on a prespecified parametric probabilistic model, so data-side descriptions typically reduce to model parameters such as difficulty and discrimination, making it difficult to directly construct diverse and heterogeneous item properties and their explicit mapping to multi-dimensional model capabilities as in the Probing Memes Paradigm. On the model side, the IRT ability $\theta$ of an LLM is largely determined by the tested dataset: when items are mainly mathematics or physics, $\theta$ is often interpreted as overall domain ability. In contrast, the Probing Memes Paradigm uses property-induced characterizations to more directly distinguish behavioral traits such as proficiency on hard items, caution under high-risk but easy items.
\section{Conclusion}
This paper takes the view that evaluating large language models requires considering models and data jointly, rather than in isolation. Based on this view, the Probing Memes paradigm is proposed to analyze model behavior through systematic interactions between model populations and data populations. By assigning interpretable Meme Probe Properties to data items and characterizing model responses as Meme Scores, the paradigm provides diagnostic insight into dataset items and enables fine-grained comparison of models beyond overall accuracy. Experiments on both the Curated Population and the large-scale Open LLM Population show that this approach is scalable, stable under population subsampling, and useful for diagnosing behavioral differences and failure patterns in practice.
{
    \small
    \bibliographystyle{ieeenat_fullname}
    \bibliography{main}
}
\clearpage
\setcounter{page}{1}
\onecolumn
\appendix

\section{Meme, Property, and Meme Score (Extended Discussion)}
\label{appendix:meme_spaces}

\subsection{Meme space.}
Assume a meme space
\begin{equation}
\mathcal{V} = \{\mu_1,\mu_2,\dots,\mu_R\},
\end{equation}
where each $\mu_r$ denotes a meme, viewed as a latent capability factor that underlies models’ behavioral traits.
For each model $M_j$,
\begin{equation}
\mathcal{V}_j \subseteq \mathcal{V}
\end{equation}
denotes the set of memes carried by $M_j$.
As discussed in subsection~\ref{subsec:formalization_of_paradigm}, the Perception Matrix provides an empirical view of how these latent memes are expressed on the finite probe set in $\mathcal{D}$.

\subsection{Property space.}
Probe Properties are defined via a property space
\begin{equation}
\mathcal{A} \;=\; \{\alpha_{1},\alpha_{2},\dots,\alpha_{K}\},
\end{equation}
where $\alpha_{k}$ denotes the $k$-th probe-side property dimension.
In this work, $K$ is instantiated as $6$, with dimensions corresponding to six well-designed properties: \emph{difficulty}, \emph{risk}, \emph{surprise}, \emph{uniqueness}, \emph{typicality}, and \emph{bridge}. These property dimensions can be extended or refined to support richer probing objectives.

\subsection{Memes induced by property subsets.}
\label{appendix:memes_induced_by_properties}
In the Probing Memes paradigm, evaluation is operationalized by inducing memes from observable probe properties: combinations of probe-side property dimensions highlight different behavioral traits manifested by models on the probe set.

Given the collection of property dimensions $\mathcal{A}$ and a non-empty subset $U\subseteq \mathcal{A}$, a construction operator $f$ maps the properties in $U$ to a detected meme
\begin{equation}
v \;=\; f(U)\in\mathcal{V}.
\end{equation}
Let $D = |U|$ be the number of property dimensions in the subset.
For a particular construction operator $f$, the set of all memes built from $D$-property subsets is
\begin{equation}
\mathcal{V}_D
  \;=\;
  \{\, f(U) \;\mid\; U\subseteq \,\mathcal{A},\ |U|=D \,\}
  \;\subseteq\;
  \mathcal{V},
  \qquad D = 1,2,\dots,K-1,
\end{equation}
where $|\mathcal{V}_D| = \binom{K}{D}$ under the convention that different subsets $U$ correspond to different induced memes (for a fixed $f$).
This work primarily considers the induced meme family $\bigcup_{D=1}^{K-1}\mathcal{V}_D$, which is inherently extensible via alternative choices of $U$ (and potentially different designs of $f$).

\section{Probe Properties}

\subsection{Additional Discussion on MPPs}

\subsubsection{Risk}\label{appendix:mpps_risk}
A probe is risky if models that fail it tend to fail many other probes beyond their baselines.

\medskip
\paragraph{Failure sets (from Perception Matrix \(P\)).}
\begin{equation}
F_i = \big\{\, j\in\{1,\ldots,m\}\ \big|\ P_{ij}=0 \,\big\},\qquad |F_i|=\sum_{j=1}^{m}(1-P_{ij}).
\end{equation}

\medskip
\paragraph{Conditional co-wrong rate (from probe $i$ to $k$).}
\begin{equation}
\widehat{\Pr}_{i\to k}
=
\begin{cases}
\dfrac{|\,F_i\cap F_k\,|}{|F_i|}, & |F_i|>0,\\[8pt]
0, & |F_i|=0.
\end{cases}
\end{equation}

\medskip
\paragraph{Baseline for probe \(k\).} Using the \emph{difficulty} of probe \(k\) as its baseline failure rate helps quantify the degree of failure uplift relative to \(k\)'s own baseline: 
\begin{equation}
d_k \;=\; 1-\dfrac{1}{m}\sum_{j=1}^{m} P_{kj}.
\end{equation}

\medskip
\paragraph{Two-sided certainty factor.} Inspired by~\citep{delgado2002association}, the increase in conditional failure rate is two-sidedly normalized according to the relative magnitude of $\widehat{\Pr}_{i\to k}$ and the baseline difficulty $d_k$, so that the strength of the uplift becomes fairly comparable across probes with different difficulties.
\begin{equation}
\mathrm{CF}_{i\to k}
=
\begin{cases}
\dfrac{\widehat{\Pr}_{i\to k}-d_k}{1-d_k}, & \widehat{\Pr}_{i\to k}\ge d_k,\\[10pt]
\dfrac{\widehat{\Pr}_{i\to k}-d_k}{d_k},   & \widehat{\Pr}_{i\to k}< d_k,
\end{cases}
\qquad
\mathrm{CF}_{i\to i}=0.
\end{equation}

\medskip
\paragraph{Coverage factor for probe \(i\).}
To downweight probes failed by only a few models, a coverage factor is introduced based on the failure-set size:
\begin{equation}
\text{scale}_i = \frac{\log\!\big(1+|F_i|\big)}{\log(1+m)} \in [0,1].
\end{equation}

\medskip
\paragraph{Aggregated risk.}
\begin{equation}
r_i = \text{scale}_i\cdot \frac{1}{n-1}\sum_{\substack{k=1\\k\ne i}}^{n} \mathrm{CF}_{i\to k}.
\end{equation}

\subsubsection{Surprise}\label{appendix:mpps_surprise}
A probe is surprising if it triggers rare cross-population outcomes, such as elite models failing on an otherwise easy probe or weaker models succeeding on an otherwise hard probe.

\medskip
\paragraph{Failure/Success sets (from Perception Matrix \(P\)).}
\begin{equation}
F_i = \{\, j\in\{1,\ldots,m\}\mid P_{ij}=0 \,\},\qquad
R_i = \{\, j\in\{1,\ldots,m\}\mid P_{ij}=1 \,\},\qquad |R_i|=m-|F_i|.
\end{equation}

\medskip
\paragraph{Model-side weights for \(M_j\).}
\begin{equation}
\mathrm{acc}_j = \frac{1}{n}\sum_{i=1}^{n} P_{ij},\qquad
\overline{\mathrm{acc}} = \frac{1}{m}\sum_{j=1}^{m}\mathrm{acc}_j,
\qquad
a_j^{\mathrm{res}} = \mathrm{acc}_j - \overline{\mathrm{acc}}.
\end{equation}
Nonnegative, symmetric residual weights are defined by
\begin{equation}
a_j \;=\; \max\!\bigl(a_j^{\mathrm{res}},\,0\bigr),\qquad
e_j \;=\; \max\!\bigl(-a_j^{\mathrm{res}},\,0\bigr).
\end{equation}
Here, $a_j$ assigns higher weight to generally stronger models (above-average accuracy), so their failures contribute more to surprise, while $e_j$ assigns higher weight to generally weaker models (below-average accuracy), so their successes contribute more to surprise. 

\medskip
\paragraph{Per-side contributions and aggregate surprise.}  Inspired by the TF-IDF~\citep{Rajaraman_Ullman_2011} method, per-side surprise is defined by
\begin{equation}
s_i^{(\mathrm{easy})}
= (-\ln d_i)\,\frac{1}{|F_i|}\sum_{j\in F_i} a_j,
\qquad
s_i^{(\mathrm{hard})}
= \bigl(-\ln(1-d_i)\bigr)\,\frac{1}{|R_i|}\sum_{j\in R_i} e_j.
\end{equation}
Here, $s_i^{(\mathrm{easy})}$ measures the surprise of elite models failing on an otherwise easy probe, while $s_i^{(\mathrm{hard})}$ measures the surprise of weaker models succeeding on an otherwise hard probe.
\begin{equation}
s_i = \tfrac{1}{2}\Big(s_i^{(\mathrm{easy})}+s_i^{(\mathrm{hard})}\Big).
\end{equation}

\subsubsection{Hamming Similarity}\label{appendix:hamming_similarity}
Given two probes \(i\) and \(k\), their similarity \(\mathrm{sim}(P_i, P_k)\) is measured by a Hamming-style similarity derived from the idea of Hamming Distance~\citep{6772729}.
Here, the perception span of probe \(i\) is the row vector
\begin{equation}
P_i = (P_{i1},\ldots,P_{im}) \in \{0,1\}^m .
\end{equation}
The similarity is defined as:
\begin{equation}
\mathrm{sim}(P_i,P_k)
= \frac{1}{m}\sum_{j=1}^{m} \mathbf{1}[\,P_{ij}=P_{kj}\,]
\;\in[0,1].
\end{equation}
Let \(S\in[0,1]^{n\times n}\) collect all pairwise similarities, i.e., \(S_{ik}=\mathrm{sim}(P_i,P_k)\) with \(S_{ii}=1\).

\subsubsection{Clustering}\label{appendix:clustering}

\paragraph{Reduction of identical rows.}
During clustering, any probes with identical perception spans (identical rows in \(P\)) are treated as one probe.
This accelerates computation and guarantees identical rows are assigned to the same cluster.

\paragraph{Hierarchical clustering (threshold \(\tau\)).}
Fix a similarity threshold \(\tau\in[0,1]\).
\begin{enumerate}
\item Build a sparse graph by keeping only pairs with \(S_{ik}\ge \tau\) (and ignoring self-edges).
\item For each connected component of this graph, run hierarchical agglomerative clustering with \emph{complete linkage} on the dissimilarity \(D_{ik}=1-S_{ik}\).
\item Cut the dendrogram at height \(1-\tau\) to obtain the final clusters.
\end{enumerate}
This procedure mirrors the implementation: threshold first to form components, then apply complete-linkage HAC within each component and cut at \(1-\tau\).

\subsubsection{Typicality}\label{appendix:typicality}
A probe is typical if it well represents its cluster in the similarity space.

\paragraph{Cluster partition.}
Assume all probes are grouped into \(L\) clusters
\begin{equation}
\mathcal{C}=\{\,C_1, C_2, \ldots, C_L\,\}, \qquad C_\ell \subseteq \{1,\ldots,n\}.
\end{equation}
Each cluster \(C_\ell\) contains probes with similar perception spans.

\paragraph{Cluster-level similarity summaries.}
For a cluster \(C_\ell\), define the average intra-cluster similarity
\begin{equation}
\mathrm{Intra}(C_\ell)
\;=\;
\frac{1}{|C_\ell|(|C_\ell|-1)}
\sum_{i\in C_\ell}\ \sum_{\substack{k\in C_\ell\\k\neq i}} S_{ik},
\end{equation}
and for each probe \(i\in C_\ell\), define its average similarity to the rest of the cluster
\begin{equation}
\mathrm{Cen}(i;C_\ell)
\;=\;
\frac{1}{|C_\ell|-1}
\sum_{\substack{k\in C_\ell\\k\neq i}} S_{ik}.
\end{equation}

\paragraph{Prototype of a cluster.}
Within each cluster \(C_\ell\), the prototype \(p_{C_\ell}\) is the probe most similar to other members:
\begin{equation}
p_{C_\ell} \;=\; \arg\max_{i\in C_\ell}\; \sum_{\substack{k\in C_\ell\\k\neq i}} S_{ik}.
\end{equation}

\paragraph{Size-dependent scaling.}
To account for varying cluster sizes, a size-dependent scaling is introduced so that prototypes in larger clusters are treated as more typical, while non-prototype probes contribute less. Accordingly, define
\begin{equation}
h(n)=\sqrt{\frac{\ln n}{1+\ln n}},\qquad
g(n)=\frac{1}{\sqrt{1+\ln n}}.
\end{equation}

\paragraph{Typicality of probe \(i\).}
For each probe \(i\in C_\ell\),
\begin{equation}
t_i =
\begin{cases}
\displaystyle
\tfrac{1}{2}+\tfrac{1}{2}\,h(|C_\ell|)\,\mathrm{Intra}(C_\ell), & \text{if } i=p_{C_\ell}\ \text{(prototype)},\\[8pt]
\displaystyle
g(|C_\ell|)\,\mathrm{Cen}(i;C_\ell), & \text{otherwise (non-prototype)}.
\end{cases}
\end{equation}
The constant 1/2 is added to initialize the typicality of the prototype probe in a cluster with a single element.

\section{Meme Scores}
\label{appendix:meme_scores}
\subsection{Notation}
As defined in subsection~\ref{subsec:formalization_of_paradigm}, $P\in\{0,1\}^{n\times m}$ is the Perception Matrix with entries $P_{ij}$ (probe $i$, model $j$).
The meme space $\mathcal{V}$, the collection of probe-side property dimensions $\mathcal{A} \;=\; \{\alpha_{1},\alpha_{2},\dots,\alpha_{K}\}$, and the induced meme $v=f(U)$ for any non-empty subset $U\subseteq \mathcal{A}$ follow the definitions in Appendix~\ref{appendix:memes_induced_by_properties}.

For a property dimension $\alpha_k\in \mathcal{A}$, let $\rho_i(\alpha_k)\in\mathbb{R}$ denote its MPP value on probe $i$. Collecting these values over all probes yields the vector $\rho(\alpha_k)\in\mathbb{R}^n$. In this work, the construction operator $f$ is operationally defined as a weight-construction rule that maps each property subset $U$ to a probe-weight vector $w(U)$, thereby specifying how probes contribute to the Meme Score associated with the induced meme $v=f(U)$. The subsequent subsections specify this construction via normalization (Appendix~\ref{appendix:meme_norm}) and property-subset composition (Appendix~\ref{appendix:meme_weights}).

\subsection{Normalization}
\label{appendix:meme_norm}
Given $\rho(\alpha_k)$, define the $z$-score normalized values
\begin{equation}\label{eq:meme_zscore}
z_i(\alpha_k) \;=\; \frac{\rho_i(\alpha_k)-\mu(\rho(\alpha_k))}{\sigma(\rho(\alpha_k))},
\end{equation}
The normalized values are then shifted to a nonnegative scale:
\begin{equation}\label{eq:meme_shift}
\tilde{\rho}_i(\alpha_k) \;=\; z_i(\alpha_k) - \min_{i'\in\{1,\ldots,n\}} z_{i'}(\alpha_k).
\end{equation}

\subsection{Property-subset weights}
\label{appendix:meme_weights}
For a non-empty subset $U\subseteq \mathcal{A}$ (thus an induced meme $v=f(U)$), the unnormalized weight of probe $i$ is
\begin{equation}\label{eq:meme_weight_raw}
\phi_i(U) \;=\; \prod_{\alpha_k\in U}\tilde{\rho}_i(\alpha_k),
\end{equation}
and the normalized weight vector $w(U)\in\mathbb{R}_{\ge 0}^n$ is
\begin{equation}\label{eq:meme_weight_norm}
w_i(U) \;=\; \frac{\phi_i(U)}{\sum_{l=1}^{n}\phi_l(U)},
\qquad \sum_{i=1}^{n} w_i(U) = 1.
\end{equation}

\subsection{Meme score}
\label{appendix:meme_score_def}
The meme score of model $M_j$ under the induced meme $v=f(U)$ is defined as
\begin{equation}\label{eq:meme_score}
\mathrm{MemeScore}(v;\,M_j)
\;=\;
\sum_{i=1}^{n} w_i(U)\,P_{ij}
\;\in\;[0,1],
\qquad v=f(U).
\end{equation}

\section{Experimental Details}
\label{app:experimental_details} 

\subsection{Models}
\label{app:models-curated} 
The following 28 models are included in this work. OpenAI: gpt-4.1-2025-04-14, gpt-4.1-mini-2025-04-14, gpt-4.1-nano-2025-04-14,
gpt-4o-2024-11-20, o3-2025-04-16, o3-mini-2025-01-31, gpt-5-2025-08-07, gpt-5-mini-2025-08-07,
gpt-5-nano-2025-08-07~\citep{achiam2023gpt,openai2025gpto3o4mini,openai2025gpt5};
Anthropic: claude-3-5-sonnet-20241022, claude-sonnet-4-20250514~\citep{anthropic2024claude35sonnet,anthropic2024claude4};
Google: gemini-2.5-flash, gemini-2.5-pro~\citep{comanici2025gemini};
DeepSeek: deepseek-V3, deepseek-R1~\citep{liu2024deepseek,guo2025deepseek};
Alibaba: qwen3-235b-a22b, qwen3-30b-a3b, qwen3-32b~\citep{yang2025qwen3};
xAI: grok-3, grok-4-0709\citep{xai2024grok3,xai2025grok4};
MiniMax: MiniMax-Text-01, MiniMax-M1~\citep{li2025minimax};
ByteDance: doubao-seed-1-6-250615, doubao-seed-1-6-flash-250715\citep{bytedance2025doubaoseed16,bytedance2025doubaoseed16flash};
Zhipu AI: glm-4.5, glm-4.5-air~\citep{zeng2025glm};
Moonshot AI: kimi-k2-0711-preview~\citep{team2025kimi};
iFlytek: spark-X1~\citep{iflytek2025sparkx1}.

\subsection{Datasets}
\label{app:datasets-curated}
There are three datasets involved in the experiment part of this work, including MATH-500~\citep{lightman2023let}, MMLU-Redux~\citep{gema-etal-2025-done}, and SimpleQA~\citep{wei2024measuring}. 

\textbf{MATH-500} is a 500-problem subset of MATH~\citep{hendrycksmath2021} selected by the OpenAI team. Its items come from high-school mathematics competitions, and many are challenging for humans. Overall, MATH-500 offers (i) hard items with strong discriminative power across models, and (ii) broad coverage of items that admit step-by-step solutions, making it well-suited for evaluating mathematical and multi-step reasoning performance.

\textbf{MMLU-Redux} is a revised version of MMLU~\citep{hendrycks2020measuring} dataset. It comprises 5,399 choice questions spanning 57 subject areas, including fields such as mathematics, physics, chemistry, political science, economics, law, and philosophy. Its broad subject coverage enables a comprehensive assessment of general knowledge across disciplines. In addition, MMLU-Redux is composed entirely of multiple-choice questions. This format allows for straightforward and highly accurate evaluation of model answers.

\textbf{SimpleQA} is a challenging dataset comprising 4,326 commonsense question-answer pairs. A distinctive feature of this dataset is its ternary answer evaluation scheme (``Correct'', ``Incorrect'', or ``Not Attempted''); for consistency, this study treated ``Not Attempted'' answers as ``Incorrect''. Due to its difficulty, few models performed well in our experiments. The dataset's broad topical coverage enables a comprehensive evaluation of model capabilities across diverse domains. Moreover, all questions are phrased precisely and unambiguously. Their high reliability is ensured through a rigorous validation process involving multiple annotators who independently provided and cross-verified answers. This procedure guarantees a unique gold answer for each question.

\subsection{Hyperparameters}
\label{app:hyperparameters}
For models without internal reasoning, temperature is set to 0, top-p to 1, and max tokens to 8192. Internal reasoning models often output a large amount of reasoning content. For models with internal reasoning, max tokens is set to 28672, and the remaining parameters follow the providers' defaults because these models often do not support very low temperature settings, and defaults are recommended. For internal reasoning models from the Qwen family, the max tokens parameter does not include the number of reasoning tokens, so set max tokens to 8192 and thinking budget to 20480 (for limiting max reasoning output).

\subsection{Prompt Templates}
\label{app:prompt_templates}
The boxes below present the prompts used for each dataset and reasoning mode, where ``\textless question text\textgreater'' denotes the text of the question. These prompts follow certain conventions. For instance, all prompts specify the required answer format, which varies across datasets. Furthermore, when testing under the Chain-of-Thought (CoT) setting, the phrase ``Please reason step by step'' is included to enable CoT reasoning, and models are instructed to output their reasoning process separately.

In terms of formatting, models are required to provide their final answers in the form ``Answer:'' followed by the answer, with no further explanation permitted. In addition, the MATH-500 prompt requires answers to be enclosed in \textbackslash boxed\{\} to facilitate the extraction of mathematical expressions; the MMLU-Redux prompt requires the answer to be a single letter corresponding to the selected option; and the SimpleQA prompt imposes no additional requirements beyond the ``Answer:'' format.

\newcommand{\PromptBox}[3]{
\noindent\textbf{#1}\hspace{0.8em}\emph{#2}\par
  \vspace{0.25em}
  \fbox{\parbox{0.97\linewidth}{\ttfamily\small #3}}
  \vspace{0.9em}
}

\PromptBox{MATH-500}{CoT Prompting (CoT).}{
Answer the following question.\\
Question: ``<question text>''\\
Please reason step by step.\\
Your response must strictly follow the format below:\\
Reasoning Process: \{Explain your reasoning step by step\}\\
Answer: \textbackslash boxed\{Your final result without any explanation\}
}

\PromptBox{MATH-500}{Default Prompting (Base) and Internal Reasoning (IR).}{
Answer the following question.\\
Question: ``<question text>''\\
Your response must strictly follow the format below:\\
Answer: \textbackslash boxed\{Your final result without any explanation\}
}

\PromptBox{MMLU-Redux}{CoT Prompting (CoT).}{
Answer the following question.\\
Question: ``<question text>''\\
Please reason step by step.\\
Your response must strictly follow the format below:\\
Reasoning Process: \{Explain your reasoning step by step\}\\
Answer: \{Your final choice letter without any explanation\}
}

\PromptBox{MMLU-Redux}{Default Prompting (Base) and Internal Reasoning (IR).}{
Answer the following question.\\
Question: ``<question text>''\\
Your response must strictly follow the format below:\\
Answer: \{Your final choice letter without any explanation\}
}

\PromptBox{SimpleQA}{CoT Prompting (CoT).}{
Answer the following question.\\
Question: ``<question text>''\\
Please reason step by step.\\
Your response must strictly follow the format below:\\
Reasoning Process: \{Explain your reasoning step by step\}\\
Answer: \{Your final answer without any explanation\}
}

\PromptBox{SimpleQA}{Default Prompting (Base) and Internal Reasoning (IR).}{
Answer the following question.\\
Question: ``<question text>''\\
Your response must strictly follow the format below:\\
Answer: \{Your final answer without any explanation\}
}

\subsection{Answer Verification}
\label{app:answer_verification}

Several issues arose during the extraction and evaluation of model responses. These included non-compliant output formats (despite explicit instructions), responses exceeding token limits, and model refusals to answer sensitive questions. Tables~\ref{tab:refuse_to_answer}, \ref{tab:over_token_limit}, and \ref{tab:error_format} present the statistics for these respective issues. The token limits are specified in Appendix~\ref{app:hyperparameters}.

\begin{table}[H]
\centering
\small
\caption{\textbf{Numbers of refusals to answer}}
\label{tab:refuse_to_answer}
\begin{tabular}{lrr}
\toprule
Model & MMLU-Redux & SimpleQA \\
\midrule
qwen3-235b-a22b(IR) & 6 & 34 \\
spark-x1(IR) & 12 & 7 \\
MiniMax-M1(IR) & 10 & 8 \\
qwen3-30b-a3b(CoT) & 5 & 8 \\
qwen3-235b-a22b(CoT) & 5 & 7 \\
glm-4.5(CoT) & 7 & 5 \\
qwen3-30b-a3b & 5 & 5 \\
qwen3-235b-a22b & 4 & 6 \\
qwen3-32b(CoT) & 5 & 5 \\
qwen3-32b & 2 & 5 \\
glm-4.5-air(CoT) & 3 & 3 \\
glm-4.5(IR) & 5 & 0 \\
glm-4.5-air(IR) & 3 & 1 \\
glm-4.5 & 1 & 0 \\
glm-4.5-air & 1 & 0 \\
others & 0 & 0 \\
\bottomrule
\end{tabular}
\vspace{2pt}
\begin{flushleft}
\end{flushleft}
\end{table}
\begin{table}[H]
\centering
\small
\caption{\textbf{Numbers of responses exceeding token limit}}
\label{tab:over_token_limit}
\begin{tabular}{lrrr}
\toprule
Model & MATH-500 & MMLU-Redux & SimpleQA \\
\midrule
glm-4.5-air(IR) & 31 & 506 & 452 \\
glm-4.5(IR) & 15 & 209 & 447 \\
gemini-2.5-flash(CoT) & 23 & 13 & 12 \\
doubao-seed-1-6-flash-250715(CoT) & 5 & 2 & 31 \\
glm-4.5-air(CoT) & 10 & 15 & 12 \\
glm-4.5(CoT) & 8 & 4 & 10 \\
MiniMax-M1(IR) & 0 & 2 & 18 \\
gpt-4.1-mini-2025-04-14(CoT) & 0 & 0 & 14 \\
gpt-4.1-2025-04-14(CoT) & 6 & 2 & 2 \\
kimi-k2-0711-preview(CoT) & 7 & 1 & 2 \\
kimi-k2-0711-preview & 6 & 0 & 0 \\
grok-4-0709(IR) & 2 & 1 & 3 \\
doubao-seed-1-6-250615(CoT) & 2 & 2 & 2 \\
doubao-seed-1-6-flash-250715(IR) & 2 & 0 & 3 \\
gpt-4.1-nano-2025-04-14(CoT) & 0 & 0 & 5 \\
gemini-2.5-flash & 2 & 0 & 2 \\
doubao-seed-1-6-flash-250715 & 1 & 0 & 2 \\
qwen3-30b-a3b(CoT) & 0 & 0 & 3 \\
deepseek-reasoner(IR) & 2 & 0 & 0 \\
gpt-4.1-nano-2025-04-14 & 0 & 0 & 2 \\
deepseek-chat(CoT) & 1 & 0 & 0 \\
o3-2025-04-16(IR) & 1 & 0 & 0 \\
gpt-5-2025-08-07(IR) & 1 & 0 & 0 \\
claude-sonnet-4-20250514(CoT) & 1 & 0 & 0 \\
doubao-seed-1-6-250615 & 0 & 1 & 0 \\
qwen3-30b-a3b & 0 & 0 & 1 \\
doubao-seed-1-6-250615(IR) & 0 & 0 & 1 \\
others & 0 & 0 & 0 \\
\bottomrule
\end{tabular}
\vspace{2pt}
\begin{flushleft}
\end{flushleft}
\end{table}
\begin{table}[H]
\centering
\small
\caption{\textbf{Numbers of malformed answers}}
\label{tab:error_format}
\begin{tabular}{lrrr}
\toprule
Model & MATH-500 & MMLU-Redux & SimpleQA \\
\midrule
doubao-seed-1-6-flash-250715(IR) & 100 & 956 & 838 \\
doubao-seed-1-6-flash-250715 & 19 & 707 & 912 \\
MiniMax-M1(IR) & 497 & 1119 & 21 \\
doubao-seed-1-6-250615 & 5 & 21 & 693 \\
spark-x1(IR) & 88 & 399 & 16 \\
gemini-2.5-flash(CoT) & 219 & 220 & 0 \\
gemini-2.5-flash(IR) & 249 & 122 & 4 \\
doubao-seed-1-6-flash-250715(CoT) & 39 & 26 & 292 \\
grok-3(CoT) & 282 & 24 & 9 \\
deepseek-reasoner(IR) & 299 & 4 & 0 \\
glm-4.5-air(IR) & 168 & 84 & 22 \\
grok-4-0709(IR) & 234 & 1 & 0 \\
qwen3-235b-a22b(IR) & 191 & 6 & 2 \\
glm-4.5(IR) & 163 & 3 & 3 \\
gemini-2.5-flash & 152 & 3 & 0 \\
qwen3-235b-a22b & 18 & 97 & 0 \\
doubao-seed-1-6-250615(CoT) & 3 & 12 & 100 \\
doubao-seed-1-6-250615(IR) & 2 & 12 & 101 \\
gpt-4o-2024-11-20(CoT) & 12 & 77 & 2 \\
gemini-2.5-pro(IR) & 61 & 9 & 0 \\
qwen3-235b-a22b(CoT) & 14 & 42 & 0 \\
kimi-k2-0711-preview(CoT) & 46 & 4 & 4 \\
gpt-4.1-2025-04-14(CoT) & 0 & 42 & 0 \\
MiniMax-Text-01(CoT) & 4 & 24 & 8 \\
gpt-4.1-nano-2025-04-14(CoT) & 5 & 31 & 0 \\
qwen3-32b(CoT) & 15 & 20 & 0 \\
o3-mini-2025-01-31(IR) & 1 & 11 & 21 \\
qwen3-30b-a3b(CoT) & 9 & 22 & 0 \\
grok-3 & 2 & 24 & 2 \\
qwen3-30b-a3b & 19 & 2 & 0 \\
deepseek-chat & 12 & 7 & 0 \\
kimi-k2-0711-preview & 18 & 1 & 0 \\
glm-4.5(CoT) & 10 & 8 & 0 \\
MiniMax-Text-01 & 0 & 14 & 4 \\
gpt-5-mini-2025-08-07(IR) & 0 & 11 & 5 \\
gpt-4.1-nano-2025-04-14 & 9 & 5 & 0 \\
qwen3-32b & 10 & 2 & 0 \\
gpt-4.1-mini-2025-04-14(CoT) & 4 & 7 & 0 \\
deepseek-chat(CoT) & 4 & 6 & 0 \\
claude-3-5-sonnet-20241022(CoT) & 1 & 8 & 0 \\
glm-4.5-air(CoT) & 1 & 7 & 0 \\
gpt-5-2025-08-07(IR) & 0 & 5 & 1 \\
gpt-4.1-2025-04-14 & 0 & 6 & 0 \\
claude-sonnet-4-20250514(IR) & 0 & 1 & 5 \\
o3-2025-04-16(IR) & 0 & 5 & 0 \\
claude-sonnet-4-20250514(CoT) & 0 & 4 & 0 \\
gpt-4.1-mini-2025-04-14 & 1 & 3 & 0 \\
gpt-5-nano-2025-08-07(IR) & 0 & 2 & 1 \\
glm-4.5 & 1 & 1 & 0 \\
glm-4.5-air & 0 & 2 & 0 \\
gpt-4o-2024-11-20 & 0 & 0 & 1 \\
others & 0 & 0 & 0 \\
\bottomrule
\end{tabular}
\vspace{2pt}
\begin{flushleft}
\end{flushleft}
\end{table}

The experiment applied two rounds of verification. The first round enforced the prompt's formatting requirements strictly: any response that failed to comply with the required format was treated as incorrect. The second round attempted to match and extract answers using a variety of possible formats, which did not conform to the prompt. Therefore, a purely formatting error was always regarded as incorrect in the first round verification, but could be viewed as correct in the second round verification if the model's output contained the correct answer. Responses that exceeded the token number limits and responses in which the model refused to answer were treated as incorrect in both verification rounds. The data presented in the experiments and analyses were obtained from the second round of verification.

Each round of answer verification comprises two steps: answer extraction and answer evaluation. The answer extractor extracts the model's answer (without any explanation) from the model's response, while the answer evaluator compares the extracted answer with the golden answer. Different datasets used different methods to extract and evaluate answers, and the methods are presented below.

\noindent\textbf{MATH-500.} MATH-500 dataset uses Math-Verify~\citep{Kydlicek_Math-Verify_Math_Verification} library to extract and evaluate answers. The extractor first attempts to extract the content enclosed by \textbackslash boxed\{\} from the model response using regular expressions. If the attempt fails, the response will be sent directly to Math-Verify. Math-Verify is capable of extracting answers in LaTeX format as well as numeric/expression formats from the model response. It uses the following formats to extract answers in descending priority:
\begin{itemize}
    \item Explicit final answer (e.g., ``Final answer is 3. I hope'');
    \item General final answer (e.g., ``final answer is 3'') and boxed expressions (e.g., \textbackslash boxed\{3\}) at the same priority;
    \item Answer with a colon (e.g., ``answer: 3'');
    \item Answer without a colon (e.g., ``answer is 3'');
    \item Unanchored matches (e.g., ``3'').
\end{itemize}
Unanchored matches carry some risk of extracting numbers/expressions that appear in the response but are not the model's perceived answer; however, manual per-item inspection found no such errors. After extraction, Math-Verify normalizes the answer format and then parses it with SymPy. The golden answer is likewise converted to SymPy, and Math-Verify judges correctness by comparing the two SymPy expressions.

\noindent\textbf{MMLU-Redux.} Regular expressions are used to extract the one-letter answer in MMLU-Redux. There are three modes in the answer extraction as follows:
\begin{itemize}
    \item Searching answer using ``Answer''/``answer'' anchor. If multiple matches occur, then take the last match.
    \item Searching answer with other anchors, like ``\{\}'' and ``**''. These anchors do not mean the letter beside them is definitely the answer. Therefore, the extractor accepts a match as an answer only if exactly one match is found.
    \item Full-string match. Sometimes models give one-letter responses, with no anchors existing in these responses. However, it is risky to extract non-anchor answers in responses. To address this issue, the extractor applies full-string matches, matching responses like ``A'', ``A.'' and so forth.
\end{itemize}
After the extraction, the evaluation step only requires a simple string comparison between the extracted answer and the golden answer.

\noindent\textbf{SimpleQA.} In SimpleQA, the extractor only extracts the content following ``Answer:'' as the answer, without any other format requirements. GPT-4.1 is employed as an LLM evaluator to evaluate the answers of the models under test by comparing their answers with the golden answer. The prompt for the LLM evaluator is the same as that in SimpleQA's official publication paper~\citep{wei2024measuring}. Under this prompt, the LLM evaluator classifies the answer into three categories: Correct, Incorrect, and Not Attempted. A response will be classified into ``Not Attempted'' if the model recognizes its inability to solve the problem and refrains from providing an answer. As long as it gives an answer, it will be classified into ``Correct'' or ``Incorrect''. In this work, only answers classified into the ``Correct'' category were regarded as correct answers, and other answers were all deemed incorrect.

\subsection{Open LLM Population Instantiation}
\subsubsection{Models and Datsets}
\label{app:OpenLLM_models_and_datasets}
Results from the Open LLM Leaderboard v2~\citep{open-llm-leaderboard-v2} on Hugging Face were used to assess the validity of the probing-memes paradigm. This work collects the publicly available response data of models from the Open LLM Leaderboard on Hugging Face and removes models with missing records as well as items with incomplete information, ensuring that all models have complete results on the same set of items. Six datasets were contained in the leaderboard: IFEval~\citep{zhou2023instruction}, MATH~\citep{hendrycksmath2021}, MUSR~\citep{sprague2023musr}, MMLU-Pro~\citep{wang2024mmlu}, BBH~\citep{suzgun-etal-2023-challenging} and GPQA~\citep{rein2311gpqa}. Among the three GPQA subsets (Diamond, Main, and Extended), this study chooses the GPQA-Diamond subset for analysis, as it provides higher data quality and is more representative.

Valid items from Open LLM Leaderboard are as follows: BBH contains 5,759 items, GPQA-Diamond contains 198 items, IFEval contains 541 items, MATH contains 1,324 items, MMLU-Pro contains 12,032 items, and MUSR contains 756 items.

\section{More Experimental Results}
\label{app:more_experimental_results}
\subsection{Supplementary Heatmap Visualizations}
\label{app:more_heatmaps}
This subsection provides additional probe--probe similarity heatmaps using alternative orderings by different MPPs (beyond the \emph{Uniqueness}-sorted visualization in the main text).

\begin{figure}[H]
  \centering
  \includegraphics[width=0.5\linewidth]{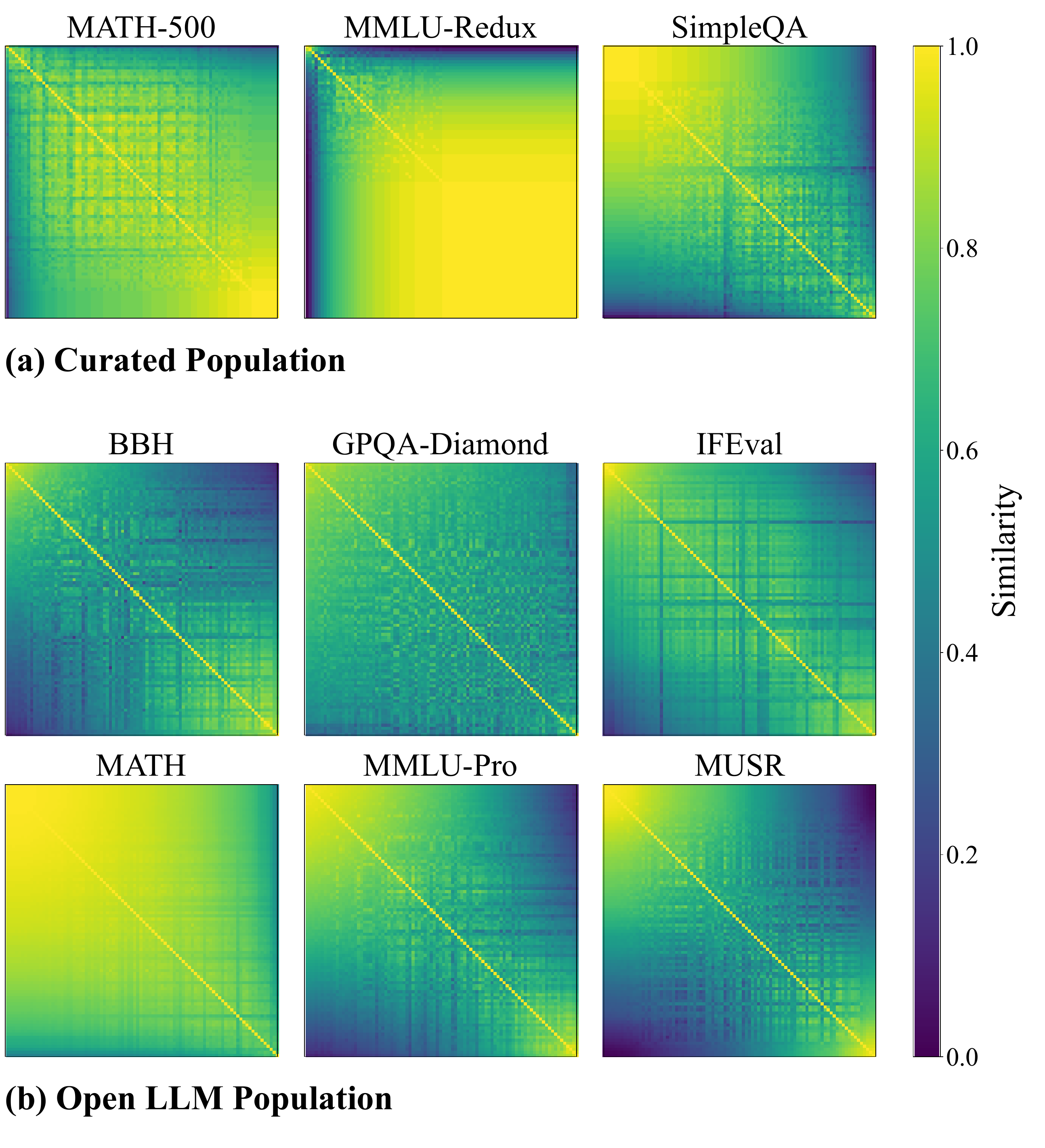}
  \caption{\textbf{Probe similarity heatmaps across datasets.}
  Probes are sorted by the value of \emph{difficulty}.}
  \label{fig:heat_map_dif}
\end{figure}

\begin{figure}[H]
  \centering
  \includegraphics[width=0.5\linewidth]{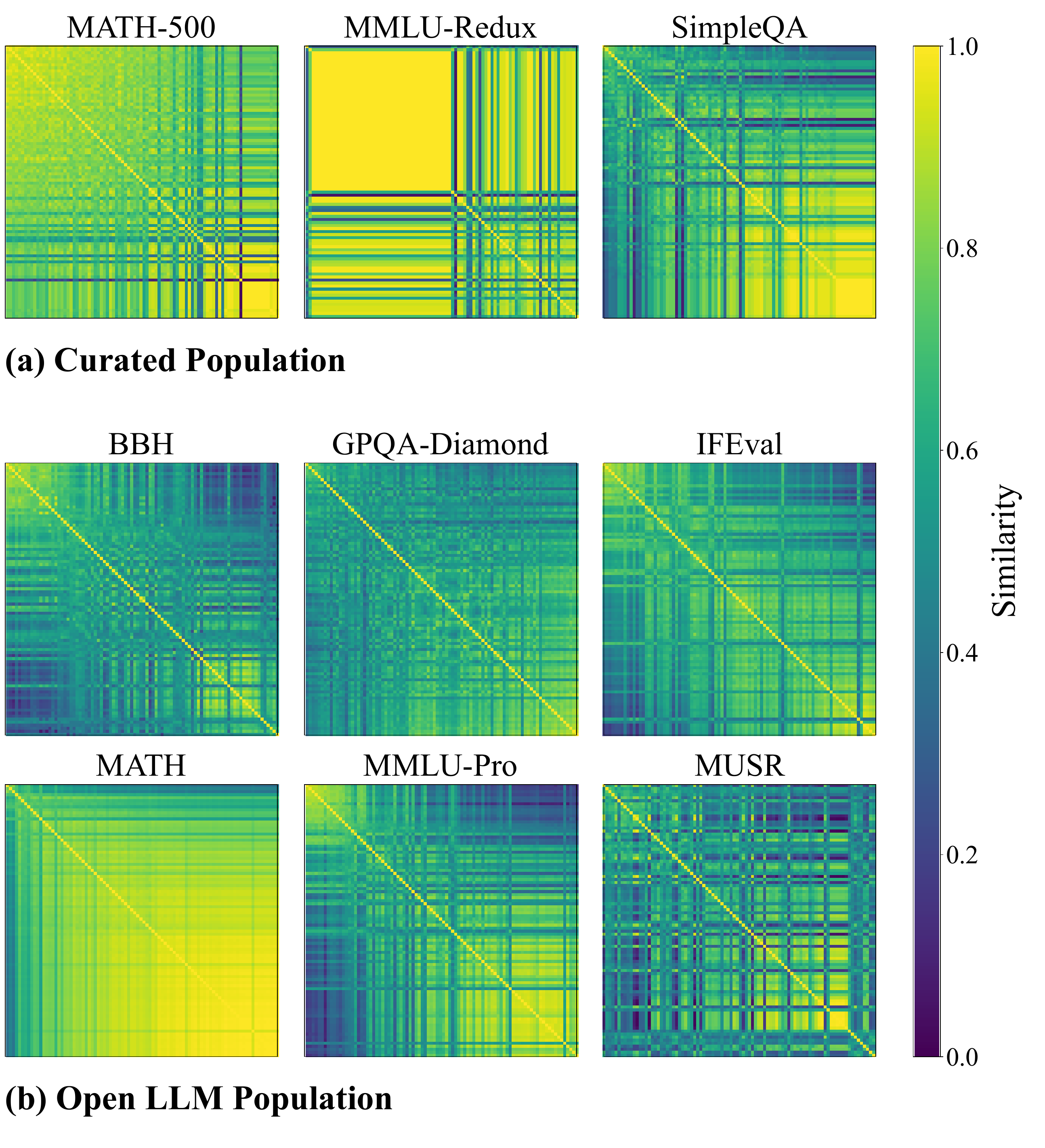}
  \caption{\textbf{Probe similarity heatmaps across datasets.}
  Probes are sorted by the value of \emph{risk}.}
  \label{fig:heat_map_risk}
\end{figure}

\begin{figure}[H]
  \centering
  \includegraphics[width=0.5\linewidth]{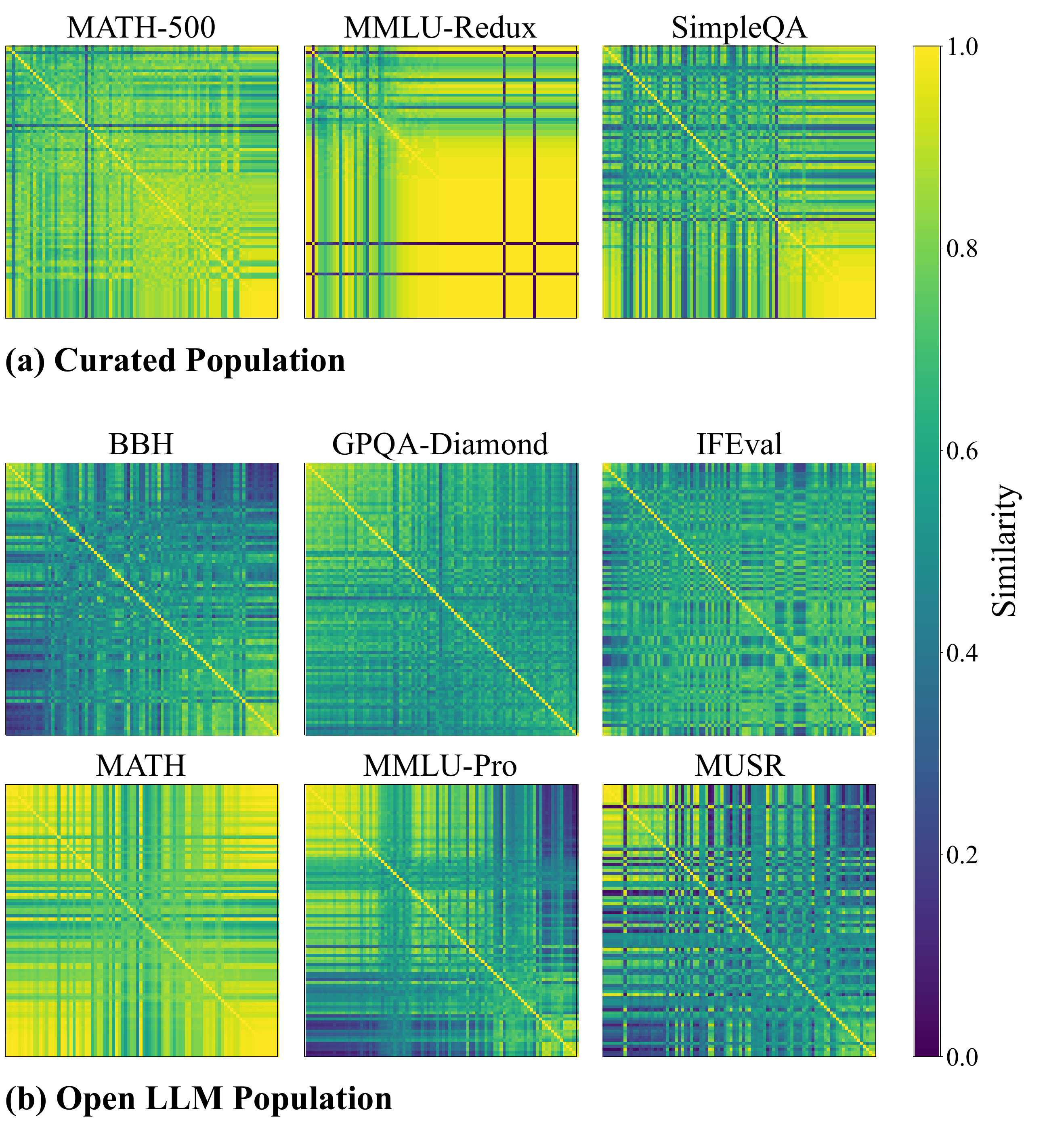}
  \caption{\textbf{Probe similarity heatmaps across datasets.}
  Probes are sorted by the value of \emph{surprise}.}
  \label{fig:heat_map_surprise}
\end{figure}

\begin{figure}[H]
  \centering
  \includegraphics[width=0.5\linewidth]{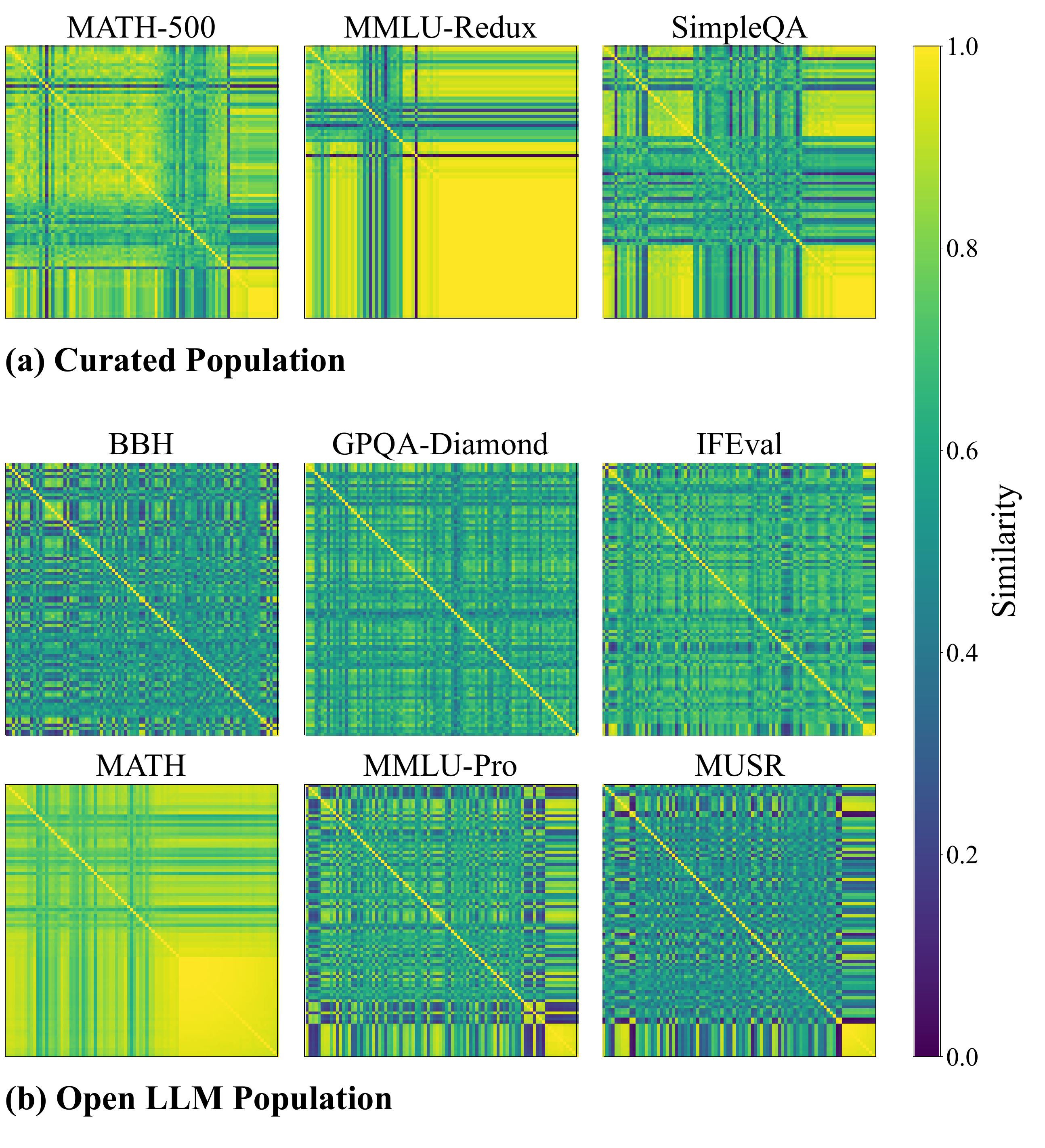}
  \caption{\textbf{Probe similarity heatmaps across datasets.}
  Probes are sorted by the value of \emph{typicality}.}
  \label{fig:heat_map_typicality}
\end{figure}

\begin{figure}[H]
  \centering
  \includegraphics[width=0.5\linewidth]{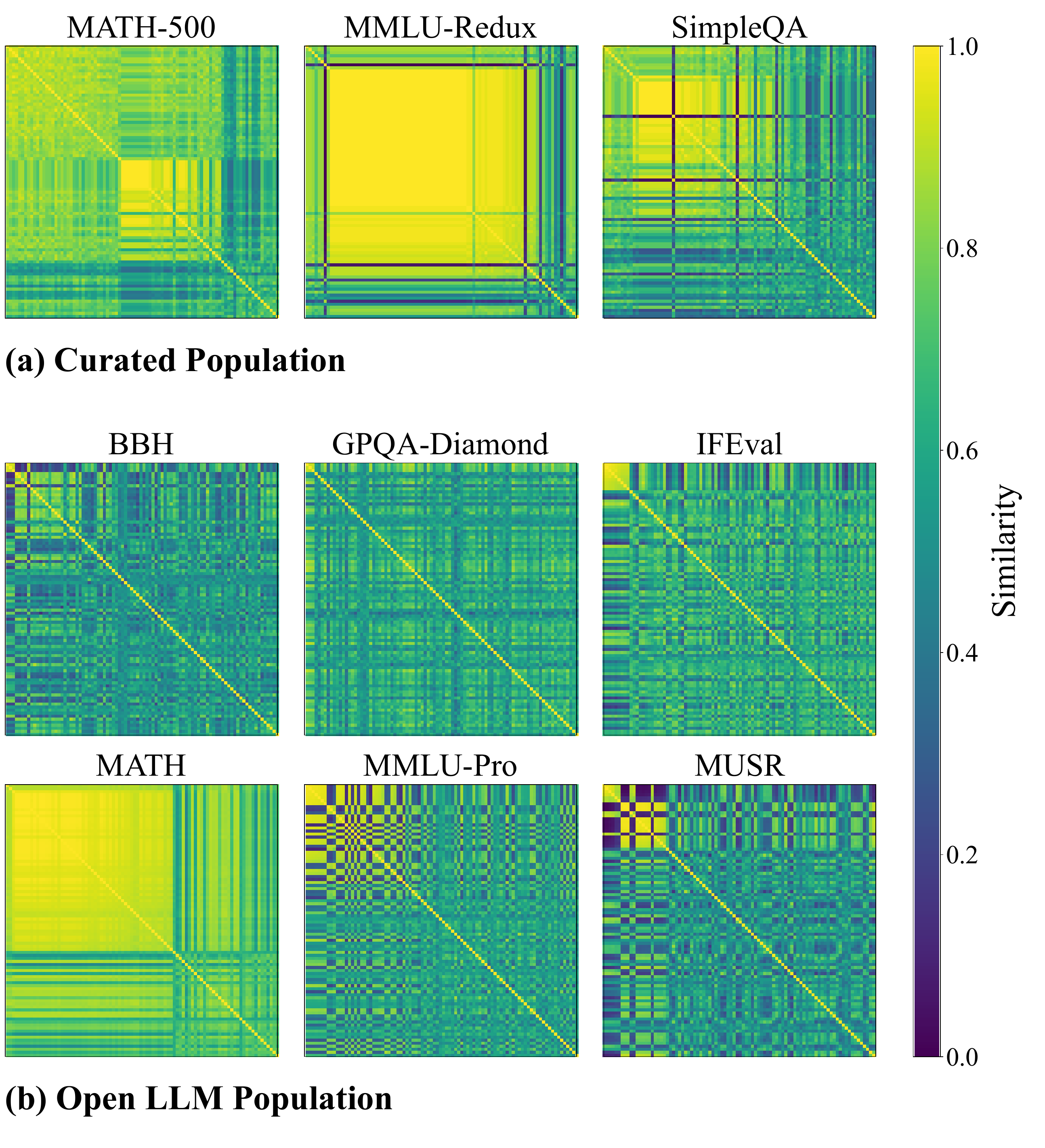}
  \caption{\textbf{Probe similarity heatmaps across datasets.}
  Probes are sorted by the value of \emph{bridge}.}
  \label{fig:heat_map_bridge}
\end{figure}

\subsection{Distribution of MPPs}
\label{app:distribution_of_MPPs}

\begin{figure}[H]
  \centering
  \includegraphics[width=0.6\linewidth]{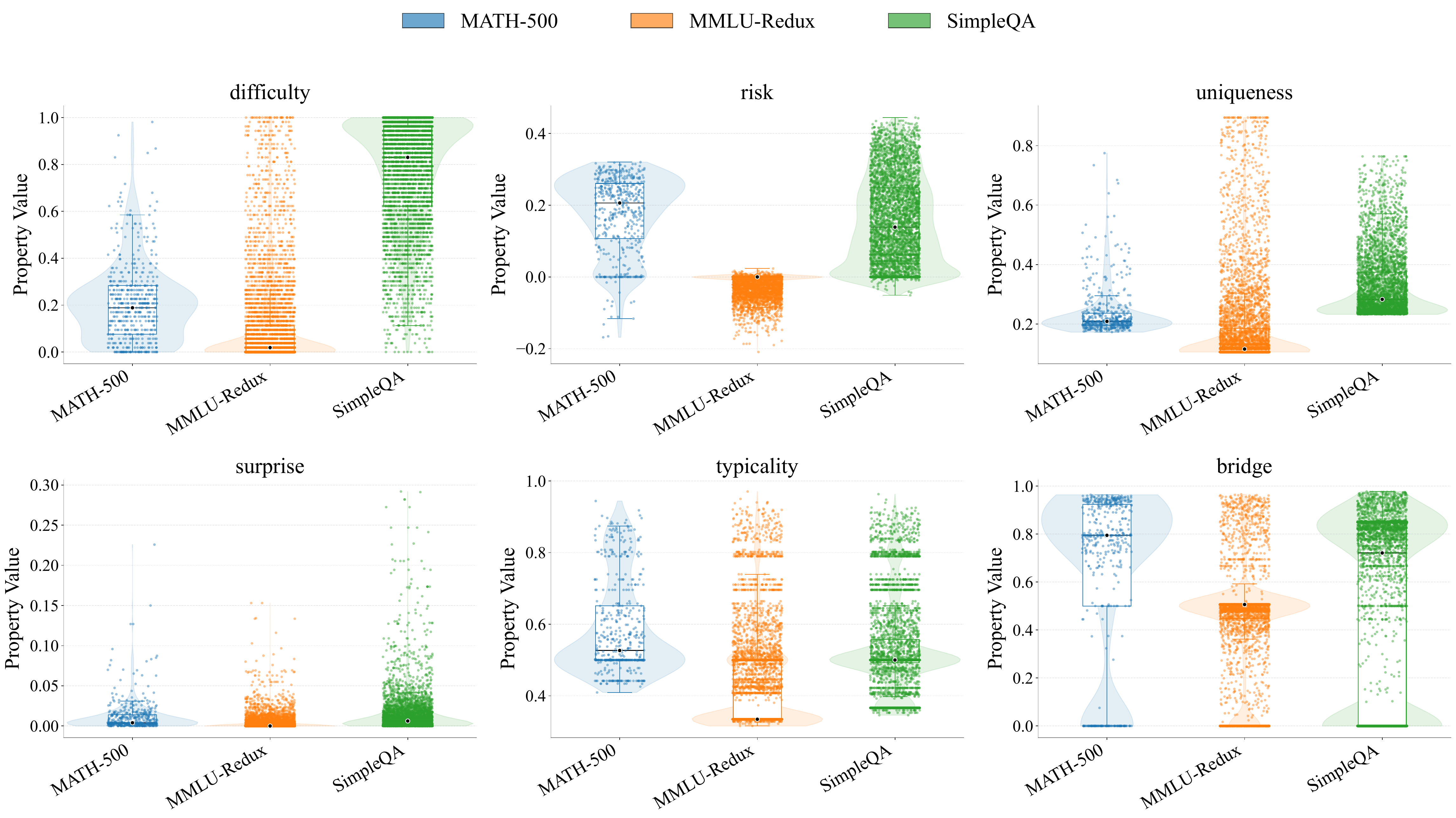}
  \caption{\textbf{Distribution of Probe Properties on each dataset (Curated Population).}}
  \label{fig:datasets_properties_boxgrid_raw}
\end{figure}

\begin{figure}[H]
  \centering
  \includegraphics[width=0.8\linewidth]{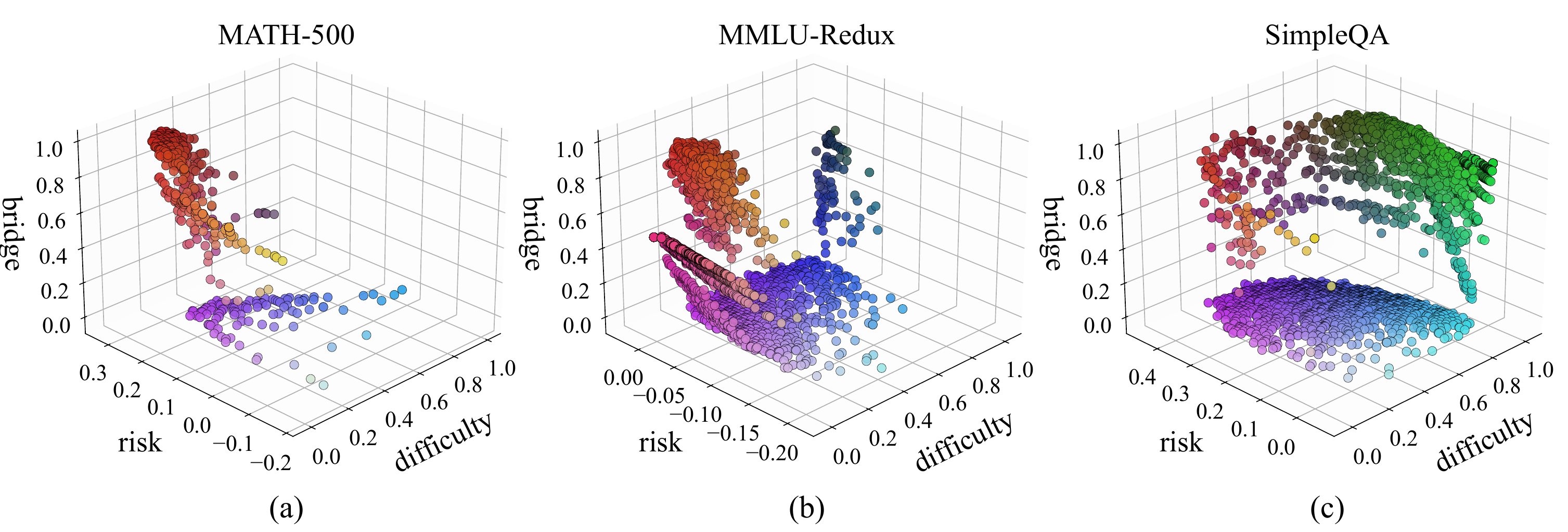}
  \caption{\textbf{Probe properties distributions across datasets (Curated Population).} Axes depict difficulty, risk, and bridge.}
  \label{fig:3d_dif_risk_br}
\end{figure}

\begin{figure}[H]
  \centering
  \includegraphics[width=0.8\linewidth]{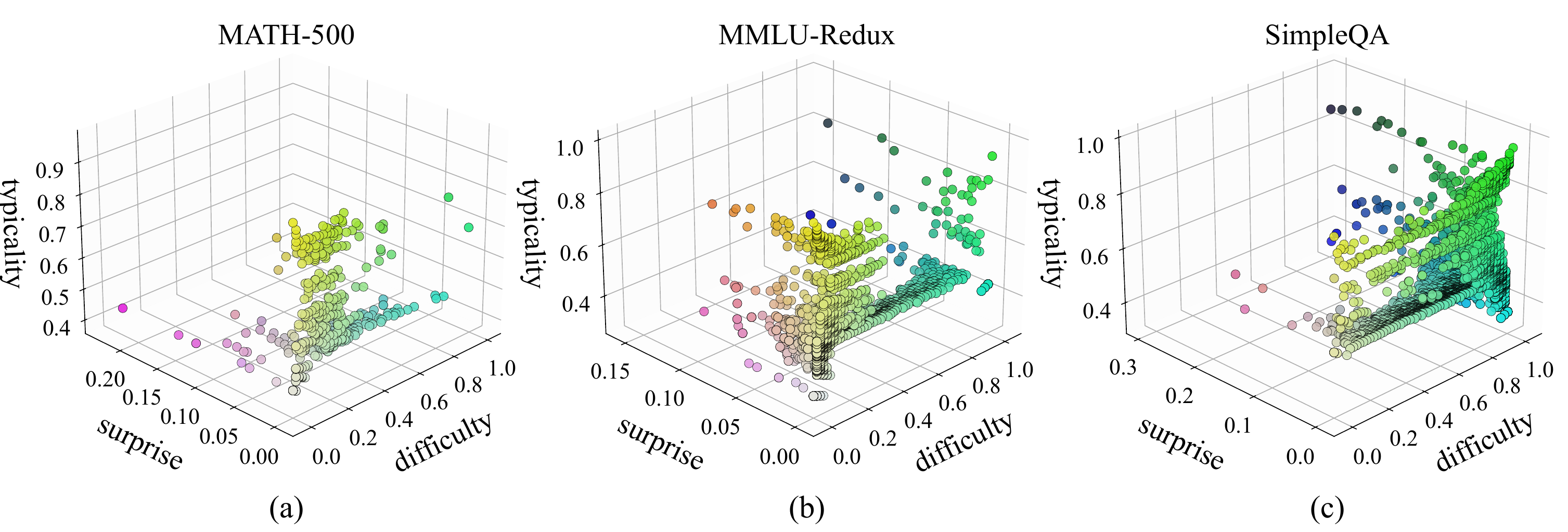}
  \caption{\textbf{Probe properties distributions across datasets (Curated Population).} Axes depict difficulty, surprise, and typicality.}
  \label{fig:3d_dif_sp_ty}
\end{figure}

\begin{figure}[H]
  \centering
  \includegraphics[width=0.8\linewidth]{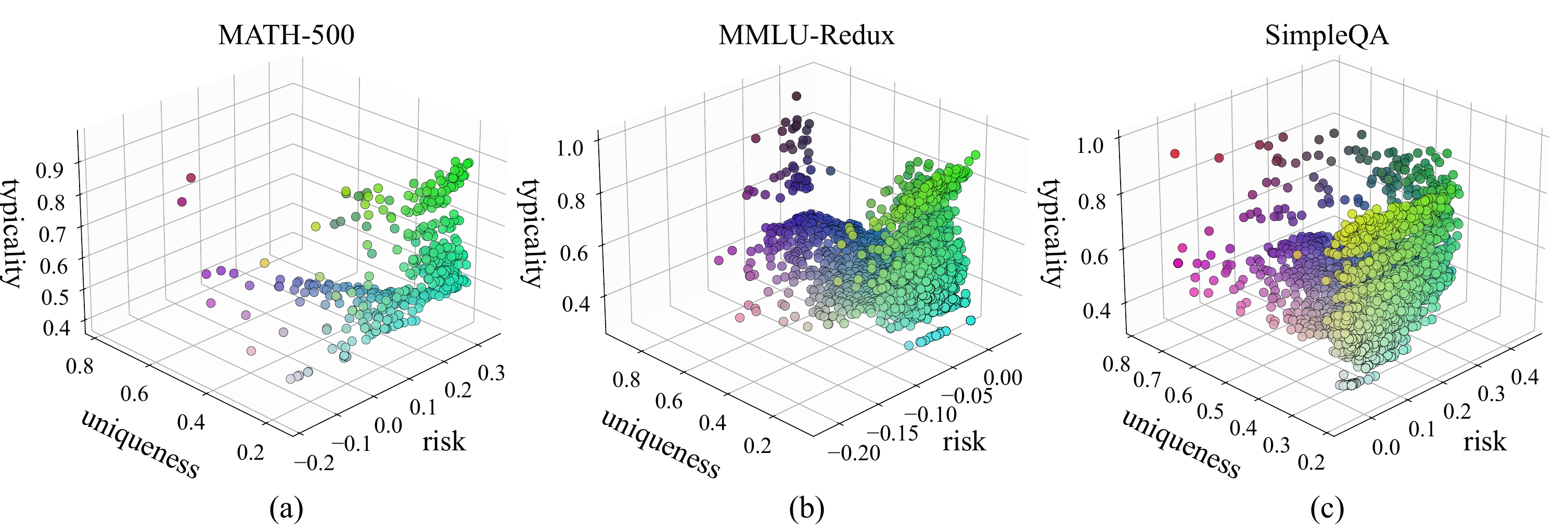}
  \caption{\textbf{Probe properties distributions across datasets (Curated Population).} Axes depict risk, uniqueness, and typicality.}
  \label{fig:3d_rk_uq_ty}
\end{figure}

\begin{figure}[H]
  \centering
  \includegraphics[width=0.9\linewidth]{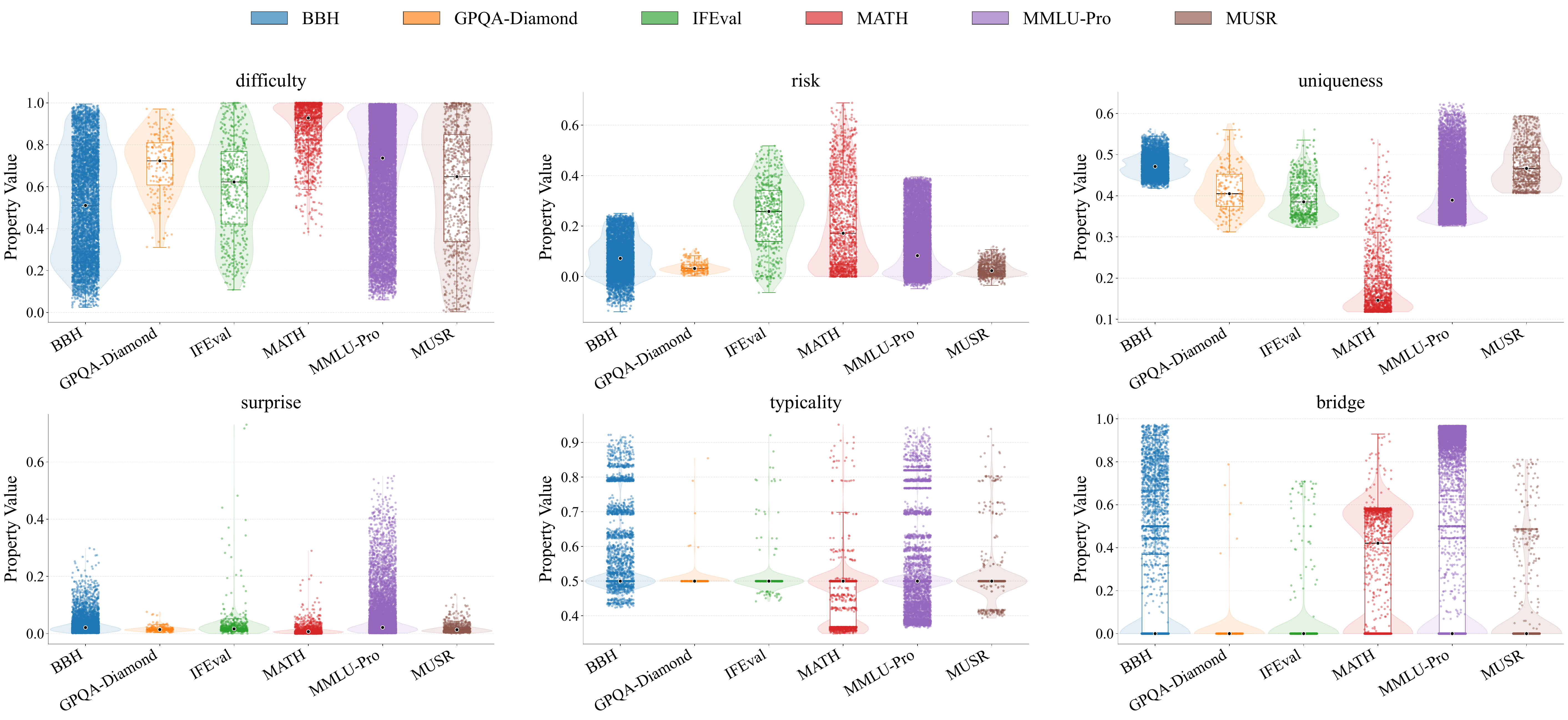}
  \caption{\textbf{Distribution of Probe Properties on each dataset (Open LLM Population).}}
  \label{fig:datasets_properties_boxgrid_raw—openllm}
\end{figure}

\begin{figure}[H]
  \centering
  \includegraphics[width=0.6\linewidth]{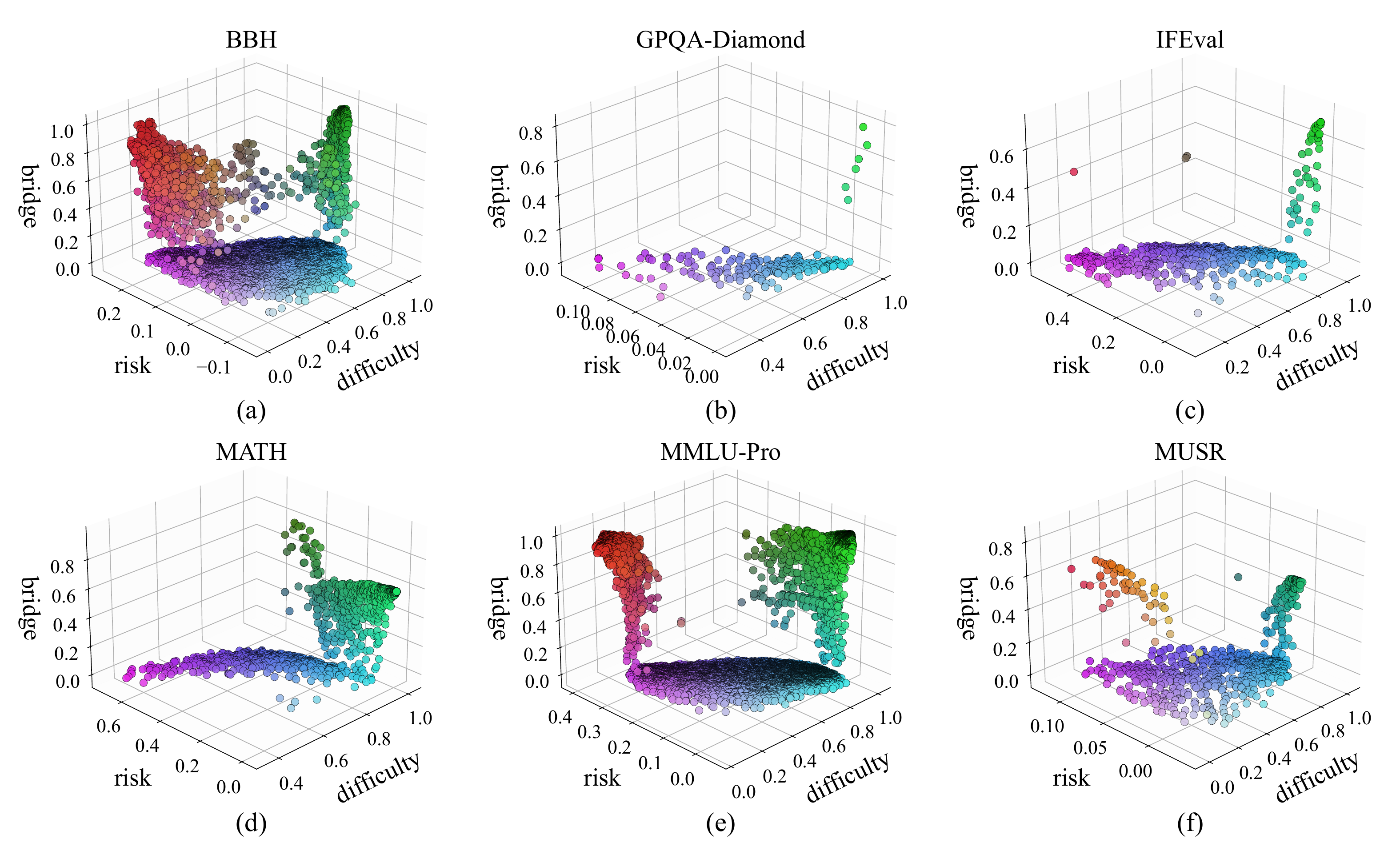}
  \caption{\textbf{Probe properties distributions across datasets (Open LLM Population).} Axes depict difficulty, risk, and bridge.}
  \label{fig:3d_dif_risk_br_openllm}
\end{figure}

\begin{figure}[H]
  \centering
  \includegraphics[width=0.8\linewidth]{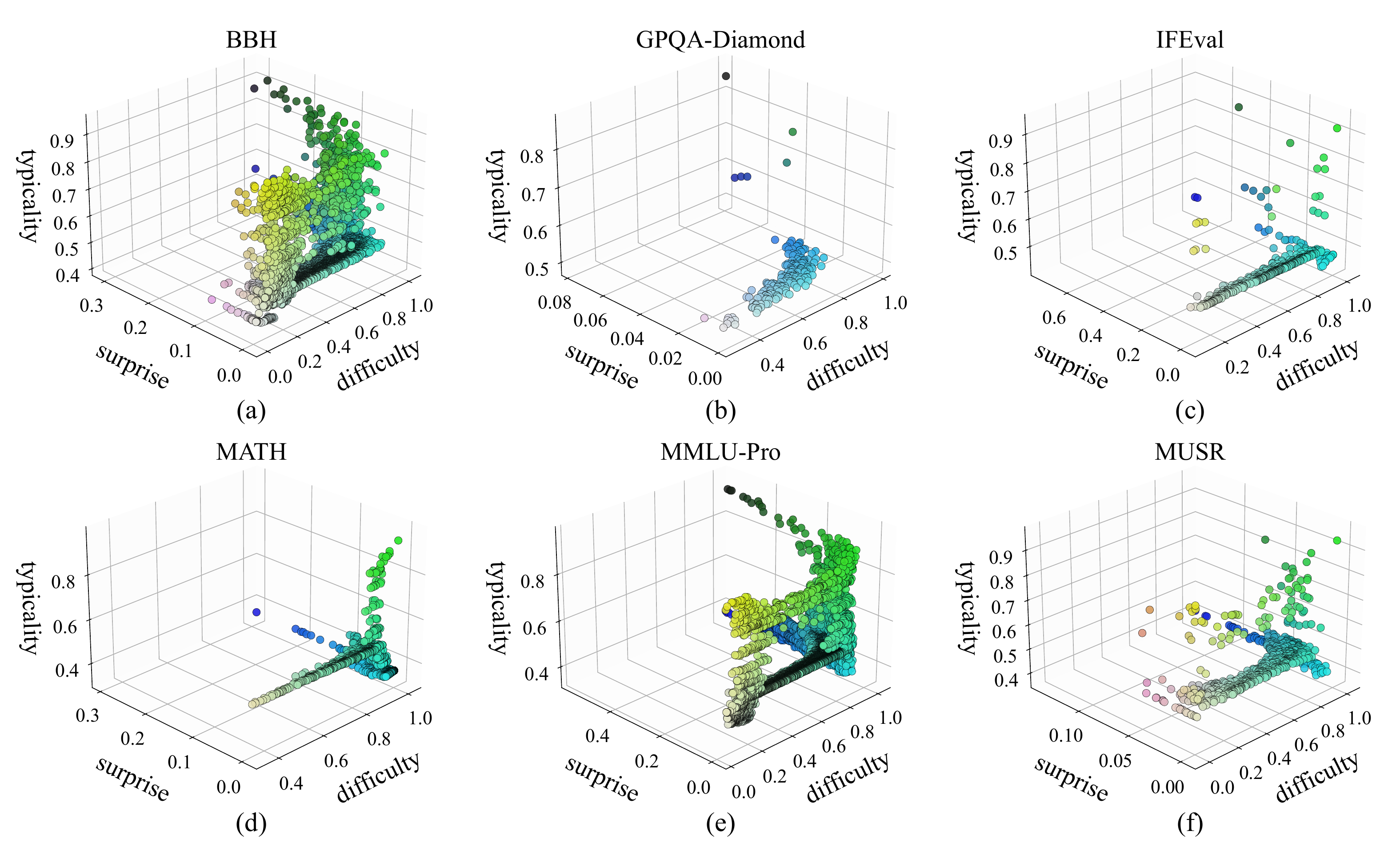}
  \caption{\textbf{Probe properties distributions across datasets (Open LLM Population).} Axes depict difficulty, surprise, and typicality.}
  \label{fig:3d_dif_sp_ty_openllm}
\end{figure}

\begin{figure}[H]
  \centering
  \includegraphics[width=0.8\linewidth]{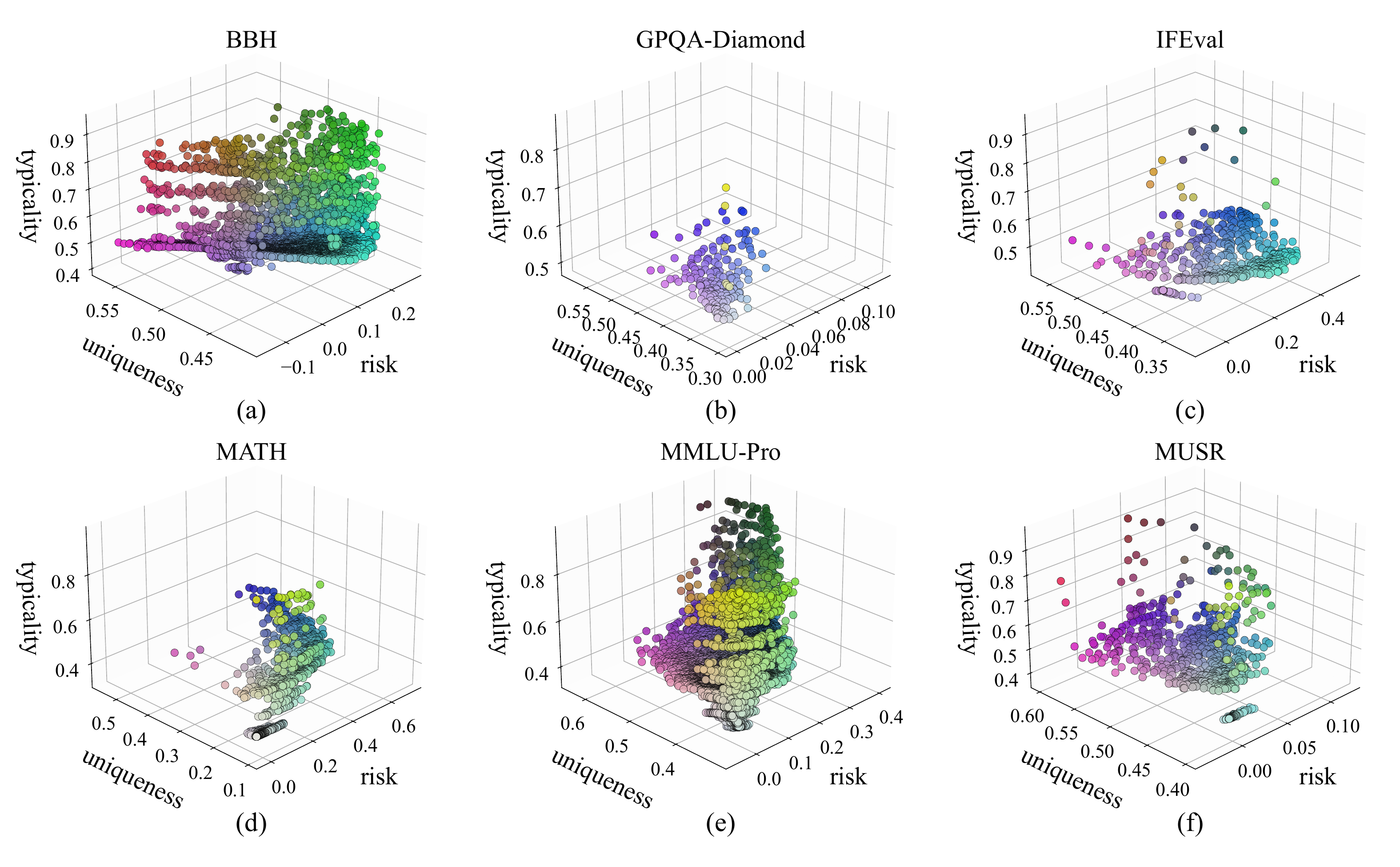}
  \caption{\textbf{Probe properties distributions across datasets (Open LLM Population).} Axes depict risk, uniqueness, and typicality.}
  \label{fig:3d_rk_uq_ty_openllm}
\end{figure}

\subsection{Meme Score Results on the Curated Population}
\label{app:MemeScore_Leaderboard_curated_full}

Table~\ref{tab:meme_scores_summary_full_curated} reports Meme Scores for all models in the Curated Population, ranked by overall accuracy, averaged over MATH-500, MMLU-Redux, and SimpleQA; Tables~\ref{tab:meme_scores_summary_full_curated-MATH-500}, \ref{tab:meme_scores_summary_full_curated-MMLU-Redux}, and \ref{tab:meme_scores_summary_full_curated-SimpleQA} report the corresponding results computed on MATH-500, MMLU-Redux, and SimpleQA, respectively.

\begin{table}[H]
\centering
\caption{\textbf{Meme Scores Curated Population; Averaged over MATH-500, MMLU-Redux, and SimpleQA).} The table reports meme scores across models, including property-derived 1D meme scores, predefined 2D meme scores (Mastery, Ingenuity, and Robustness), and a predefined 3D meme score (Caution). Models are sorted by Accuracy. {\color{blue}Blue} indicates rank improvement compared with Accuracy rank, and {\color{red}red} indicates rank degradation.}
\resizebox{\textwidth}{!}{
\begin{tabular}{cccccccccccc}
\toprule
\multirow{2}{*}{\textbf{Model}} & \multirow{2}{*}{\textbf{Accuracy}} &
\multicolumn{6}{c}{\textbf{Property-derived 1D meme scores}} &
\multicolumn{4}{c}{\textbf{Predefined 2D/3D meme scores}} \\
\cmidrule(lr){3-8}\cmidrule(lr){9-12}
 &  & \textbf{Difficulty} & \textbf{Uniqueness} & \textbf{Risk} & \textbf{Surprise} & \textbf{Typicality} & \textbf{Bridge} & \textbf{Mastery} & \textbf{Ingenuity} & \textbf{Robustness} & \textbf{Caution} \\
\midrule
gemini-2.5-pro(IR) & 82.57 & 68.51\, \textcolor{red}{(-2)} & 76.69\, \textcolor{red}{(-3)} & 90.07 & 71.23\, \textcolor{red}{(-2)} & 84.37 & 79.08 & 72.75 & 62.51\, \textcolor{red}{(-3)} & 86.91 & 95.34 \\
grok-4-0709(IR) & 81.92 & 68.70\, \textcolor{blue}{(+1)} & 79.48\, \textcolor{blue}{(+1)} & 89.29 & 71.62 & 83.39 & 77.55 & 72.54 & 66.89\, \textcolor{blue}{(+1)} & 84.89 & 95.20\, \textcolor{red}{(-1)} \\
gpt-5-2025-08-07(IR) & 81.51 & 68.52\, \textcolor{blue}{(+1)} & 79.19\, \textcolor{blue}{(+1)} & 89.05 & 71.79\, \textcolor{blue}{(+2)} & 83.22 & 77.04\, \textcolor{red}{(-1)} & 72.46 & 66.28\, \textcolor{blue}{(+1)} & 84.49\, \textcolor{red}{(-1)} & 95.32\, \textcolor{blue}{(+1)} \\
o3-2025-04-16(IR) & 81.14 & 67.70 & 78.40\, \textcolor{blue}{(+1)} & 88.81 & 70.58 & 83.04 & 77.06\, \textcolor{blue}{(+1)} & 72.02 & 64.80\, \textcolor{blue}{(+1)} & 84.79\, \textcolor{blue}{(+1)} & 95.16 \\
grok-3(CoT) & 76.69 & 59.63 & 70.05\, \textcolor{red}{(-2)} & 83.78 & 65.51 & 78.00 & 73.58\, \textcolor{red}{(-1)} & 64.31 & 59.25 & 79.61\, \textcolor{red}{(-1)} & 91.65\, \textcolor{red}{(-1)} \\
gpt-4.1-2025-04-14(CoT) & 75.93 & 56.15\, \textcolor{red}{(-3)} & 64.57\, \textcolor{red}{(-7)} & 83.57 & 62.89\, \textcolor{red}{(-6)} & 77.74 & 74.68\, \textcolor{blue}{(+1)} & 62.08 & 49.84\, \textcolor{red}{(-15)} & 81.62\, \textcolor{blue}{(+1)} & 91.86\, \textcolor{blue}{(+1)} \\
kimi-k2-0711-preview(CoT) & 74.47 & 56.30\, \textcolor{red}{(-1)} & 66.61\, \textcolor{red}{(-3)} & 80.80 & 64.17 & 75.15 & 71.30 & 60.29\, \textcolor{red}{(-1)} & 54.23\, \textcolor{red}{(-7)} & 75.78 & 89.19 \\
gemini-2.5-flash(IR) & 74.10 & 58.94\, \textcolor{blue}{(+2)} & 70.68\, \textcolor{blue}{(+3)} & 79.84 & 64.13 & 74.13 & 70.06 & 61.81\, \textcolor{blue}{(+1)} & 58.19\, \textcolor{blue}{(+1)} & 73.97\, \textcolor{red}{(-1)} & 87.36 \\
deepseek-R1(IR) & 73.61 & 58.07\, \textcolor{blue}{(+2)} & 70.20\, \textcolor{blue}{(+3)} & 79.22 & 64.17\, \textcolor{blue}{(+3)} & 72.95 & 69.10\, \textcolor{red}{(-1)} & 59.77 & 58.33\, \textcolor{blue}{(+3)} & 72.42\, \textcolor{red}{(-2)} & 86.99 \\
gemini-2.5-flash(CoT) & 72.07 & 53.64\, \textcolor{red}{(-3)} & 63.65\, \textcolor{red}{(-4)} & 77.74 & 61.53\, \textcolor{red}{(-5)} & 72.21 & 69.65\, \textcolor{blue}{(+1)} & 57.45\, \textcolor{red}{(-1)} & 51.76\, \textcolor{red}{(-7)} & 73.33 & 85.86 \\
claude-sonnet-4-20250514(IR) & 71.73 & 56.12\, \textcolor{blue}{(+1)} & 66.94\, \textcolor{blue}{(+3)} & 76.62 & 63.56\, \textcolor{blue}{(+1)} & 71.33 & 68.48\, \textcolor{red}{(-1)} & 57.92\, \textcolor{blue}{(+1)} & 56.50\, \textcolor{blue}{(+2)} & 71.21\, \textcolor{red}{(-1)} & 84.89 \\
glm-4.5(IR) & 71.14 & 52.63\, \textcolor{red}{(-4)} & 64.61 & 76.50 & 60.96\, \textcolor{red}{(-5)} & 69.64\, \textcolor{red}{(-2)} & 68.08\, \textcolor{red}{(-2)} & 54.84\, \textcolor{red}{(-3)} & 53.72\, \textcolor{red}{(-3)} & 71.12\, \textcolor{red}{(-1)} & 83.59\, \textcolor{red}{(-2)} \\
doubao-seed-1-6-250615(IR) & 71.05 & 55.99\, \textcolor{blue}{(+2)} & 66.88\, \textcolor{blue}{(+4)} & 74.84\, \textcolor{red}{(-3)} & 63.86\, \textcolor{blue}{(+4)} & 69.86 & 67.91\, \textcolor{red}{(-4)} & 57.11 & 57.49\, \textcolor{blue}{(+5)} & 69.94\, \textcolor{red}{(-7)} & 81.62\, \textcolor{red}{(-6)} \\
gpt-5-mini-2025-08-07(IR) & 70.95 & 55.58\, \textcolor{blue}{(+2)} & 65.31\, \textcolor{blue}{(+3)} & 74.68\, \textcolor{red}{(-3)} & 62.95\, \textcolor{blue}{(+3)} & 70.15\, \textcolor{blue}{(+2)} & 68.12\, \textcolor{blue}{(+1)} & 57.43\, \textcolor{blue}{(+2)} & 54.75\, \textcolor{blue}{(+2)} & 69.86\, \textcolor{red}{(-7)} & 80.78\, \textcolor{red}{(-7)} \\
kimi-k2-0711-preview & 69.78 & 49.94\, \textcolor{red}{(-7)} & 61.95\, \textcolor{red}{(-1)} & 75.24\, \textcolor{blue}{(+2)} & 60.11\, \textcolor{red}{(-5)} & 68.82\, \textcolor{red}{(-3)} & 66.72\, \textcolor{red}{(-10)} & 53.70\, \textcolor{red}{(-3)} & 53.40\, \textcolor{red}{(-1)} & 70.66\, \textcolor{red}{(-1)} & 82.26\, \textcolor{red}{(-2)} \\
deepseek-V3(CoT) & 69.76 & 48.39\, \textcolor{red}{(-7)} & 58.30\, \textcolor{red}{(-6)} & 75.16\, \textcolor{blue}{(+2)} & 59.51\, \textcolor{red}{(-8)} & 69.10\, \textcolor{blue}{(+1)} & 68.06\, \textcolor{blue}{(+1)} & 51.52\, \textcolor{red}{(-9)} & 48.43\, \textcolor{red}{(-7)} & 70.86\, \textcolor{blue}{(+2)} & 83.98\, \textcolor{blue}{(+4)} \\
MiniMax-M1(IR) & 69.75 & 53.15\, \textcolor{blue}{(+2)} & 62.75\, \textcolor{blue}{(+2)} & 73.58\, \textcolor{red}{(-3)} & 61.22\, \textcolor{blue}{(+1)} & 68.33\, \textcolor{red}{(-4)} & 67.56\, \textcolor{red}{(-4)} & 54.96\, \textcolor{blue}{(+3)} & 54.37\, \textcolor{blue}{(+4)} & 69.68\, \textcolor{red}{(-5)} & 80.05\, \textcolor{red}{(-5)} \\
glm-4.5(CoT) & 69.28 & 48.05\, \textcolor{red}{(-8)} & 57.90\, \textcolor{red}{(-5)} & 74.37 & 59.89\, \textcolor{red}{(-3)} & 69.03\, \textcolor{blue}{(+2)} & 67.99\, \textcolor{blue}{(+2)} & 52.28\, \textcolor{red}{(-5)} & 49.20\, \textcolor{red}{(-4)} & 70.82\, \textcolor{blue}{(+3)} & 83.37\, \textcolor{blue}{(+3)} \\
claude-sonnet-4-20250514 & 68.87 & 51.15\, \textcolor{blue}{(+1)} & 60.13\, \textcolor{blue}{(+1)} & 73.20\, \textcolor{red}{(-2)} & 60.82 & 68.79 & 67.70 & 54.15\, \textcolor{blue}{(+2)} & 51.40 & 70.23 & 81.99\, \textcolor{blue}{(+1)} \\
claude-sonnet-4-20250514(CoT) & 68.86 & 50.29\, \textcolor{blue}{(+1)} & 59.01\, \textcolor{blue}{(+1)} & 73.86\, \textcolor{blue}{(+1)} & 57.91\, \textcolor{red}{(-6)} & 68.89\, \textcolor{blue}{(+3)} & 67.67 & 53.58\, \textcolor{blue}{(+1)} & 45.59\, \textcolor{red}{(-7)} & 70.31\, \textcolor{blue}{(+2)} & 82.82\, \textcolor{blue}{(+4)} \\
doubao-seed-1-6-250615(CoT) & 68.23 & 50.08\, \textcolor{blue}{(+1)} & 58.74\, \textcolor{blue}{(+1)} & 72.01\, \textcolor{red}{(-1)} & 59.73\, \textcolor{red}{(-1)} & 67.30\, \textcolor{red}{(-1)} & 67.22\, \textcolor{red}{(-1)} & 52.68\, \textcolor{blue}{(+1)} & 50.43\, \textcolor{blue}{(+1)} & 69.03\, \textcolor{red}{(-2)} & 79.56\, \textcolor{red}{(-2)} \\
gpt-4o-2024-11-20(CoT) & 68.18 & 44.81\, \textcolor{red}{(-6)} & 54.56\, \textcolor{red}{(-4)} & 74.93\, \textcolor{blue}{(+7)} & 56.57\, \textcolor{red}{(-6)} & 68.60\, \textcolor{blue}{(+2)} & 69.07\, \textcolor{blue}{(+11)} & 50.04\, \textcolor{red}{(-5)} & 43.99\, \textcolor{red}{(-10)} & 74.81\, \textcolor{blue}{(+14)} & 83.86\, \textcolor{blue}{(+9)} \\
qwen3-235b-a22b(IR) & 68.07 & 53.46\, \textcolor{blue}{(+9)} & 60.72\, \textcolor{blue}{(+6)} & 70.19\, \textcolor{red}{(-2)} & 62.11\, \textcolor{blue}{(+10)} & 66.45\, \textcolor{red}{(-1)} & 67.02 & 54.61\, \textcolor{blue}{(+7)} & 55.59\, \textcolor{blue}{(+13)} & 68.00\, \textcolor{red}{(-2)} & 74.59\, \textcolor{red}{(-4)} \\
gpt-4.1-mini-2025-04-14(CoT) & 67.65 & 48.29\, \textcolor{red}{(-1)} & 55.28\, \textcolor{red}{(-1)} & 70.94 & 59.72\, \textcolor{blue}{(+1)} & 66.73\, \textcolor{blue}{(+1)} & 67.83\, \textcolor{blue}{(+6)} & 51.95 & 47.38\, \textcolor{red}{(-1)} & 69.02 & 76.99 \\
o3-mini-2025-01-31(IR) & 67.39 & 49.95\, \textcolor{blue}{(+4)} & 56.67\, \textcolor{blue}{(+1)} & 69.83\, \textcolor{red}{(-1)} & 60.86\, \textcolor{blue}{(+7)} & 65.62\, \textcolor{red}{(-1)} & 67.00\, \textcolor{blue}{(+1)} & 52.33\, \textcolor{blue}{(+3)} & 51.41\, \textcolor{blue}{(+7)} & 67.91\, \textcolor{red}{(-1)} & 73.96\, \textcolor{red}{(-3)} \\
gpt-5-nano-2025-08-07(IR) & 66.81 & 51.42\, \textcolor{blue}{(+9)} & 58.43\, \textcolor{blue}{(+5)} & 68.60\, \textcolor{red}{(-1)} & 61.61\, \textcolor{blue}{(+12)} & 64.59\, \textcolor{red}{(-1)} & 66.50\, \textcolor{red}{(-1)} & 52.59\, \textcolor{blue}{(+5)} & 55.37\, \textcolor{blue}{(+15)} & 67.17\, \textcolor{red}{(-3)} & 71.80\, \textcolor{red}{(-4)} \\
claude-3-5-sonnet-20241022(CoT) & 66.08 & 43.66\, \textcolor{red}{(-3)} & 53.21 & 71.73\, \textcolor{blue}{(+4)} & 54.42\, \textcolor{red}{(-5)} & 65.86\, \textcolor{blue}{(+2)} & 66.67\, \textcolor{blue}{(+1)} & 47.88\, \textcolor{red}{(-2)} & 42.70\, \textcolor{red}{(-8)} & 70.54\, \textcolor{blue}{(+10)} & 80.82\, \textcolor{blue}{(+7)} \\
doubao-seed-1-6-250615 & 65.01 & 44.42\, \textcolor{red}{(-1)} & 50.47\, \textcolor{red}{(-3)} & 68.17 & 55.86\, \textcolor{red}{(-2)} & 63.69 & 66.17 & 46.83\, \textcolor{red}{(-2)} & 42.80\, \textcolor{red}{(-6)} & 67.72\, \textcolor{blue}{(+1)} & 75.19\, \textcolor{blue}{(+2)} \\
doubao-seed-1-6-flash-250715(IR) & 64.45 & 48.32\, \textcolor{blue}{(+5)} & 51.62\, \textcolor{blue}{(+1)} & 65.43\, \textcolor{red}{(-4)} & 58.13\, \textcolor{blue}{(+4)} & 62.09 & 65.77\, \textcolor{red}{(-2)} & 50.34\, \textcolor{blue}{(+3)} & 47.76\, \textcolor{blue}{(+5)} & 65.98\, \textcolor{red}{(-3)} & 67.09\, \textcolor{red}{(-6)} \\
glm-4.5-air(CoT) & 64.38 & 42.64\, \textcolor{red}{(-3)} & 49.46\, \textcolor{red}{(-3)} & 67.40\, \textcolor{blue}{(+1)} & 56.37\, \textcolor{blue}{(+1)} & 61.86\, \textcolor{red}{(-1)} & 66.07\, \textcolor{blue}{(+1)} & 45.14\, \textcolor{red}{(-2)} & 44.18\, \textcolor{red}{(-1)} & 67.52\, \textcolor{blue}{(+2)} & 73.05\, \textcolor{blue}{(+1)} \\
spark-X1(IR) & 63.79 & 47.06\, \textcolor{blue}{(+4)} & 50.74\, \textcolor{blue}{(+1)} & 65.20\, \textcolor{red}{(-3)} & 57.00\, \textcolor{blue}{(+4)} & 61.68\, \textcolor{red}{(-1)} & 65.49\, \textcolor{red}{(-1)} & 49.59\, \textcolor{blue}{(+3)} & 45.31\, \textcolor{blue}{(+3)} & 66.07 & 67.84\, \textcolor{red}{(-2)} \\
glm-4.5-air(IR) & 63.71 & 43.18 & 51.46\, \textcolor{blue}{(+3)} & 66.57\, \textcolor{blue}{(+1)} & 53.81\, \textcolor{red}{(-1)} & 59.46\, \textcolor{red}{(-2)} & 64.14\, \textcolor{red}{(-1)} & 44.46\, \textcolor{red}{(-1)} & 46.40\, \textcolor{blue}{(+6)} & 65.51\, \textcolor{red}{(-2)} & 69.48 \\
qwen3-235b-a22b(CoT) & 63.59 & 43.65\, \textcolor{blue}{(+2)} & 48.58\, \textcolor{red}{(-1)} & 66.02\, \textcolor{blue}{(+1)} & 55.25\, \textcolor{blue}{(+2)} & 62.03\, \textcolor{blue}{(+3)} & 66.00\, \textcolor{blue}{(+3)} & 46.77\, \textcolor{blue}{(+2)} & 42.40\, \textcolor{red}{(-3)} & 66.98\, \textcolor{blue}{(+3)} & 71.31\, \textcolor{blue}{(+2)} \\
MiniMax-Text-01(CoT) & 63.00 & 40.82 & 50.21\, \textcolor{blue}{(+2)} & 67.02\, \textcolor{blue}{(+4)} & 53.65\, \textcolor{red}{(-1)} & 61.60\, \textcolor{blue}{(+1)} & 63.95\, \textcolor{red}{(-2)} & 43.73 & 43.25\, \textcolor{blue}{(+1)} & 65.84\, \textcolor{blue}{(+1)} & 75.54\, \textcolor{blue}{(+9)} \\
doubao-seed-1-6-flash-250715(CoT) & 60.88 & 40.49 & 43.07\, \textcolor{red}{(-5)} & 62.05 & 52.51\, \textcolor{red}{(-3)} & 57.71 & 64.06 & 42.41 & 40.02\, \textcolor{red}{(-8)} & 64.45 & 64.21\, \textcolor{red}{(-2)} \\
qwen3-30b-a3b(CoT) & 60.15 & 39.18 & 41.11\, \textcolor{red}{(-8)} & 61.07\, \textcolor{red}{(-2)} & 53.60 & 57.01 & 64.12\, \textcolor{blue}{(+2)} & 41.64 & 40.76\, \textcolor{red}{(-6)} & 64.20 & 62.88\, \textcolor{red}{(-4)} \\
qwen3-32b(CoT) & 59.16 & 37.84 & 39.23\, \textcolor{red}{(-9)} & 60.40\, \textcolor{red}{(-2)} & 50.07\, \textcolor{red}{(-4)} & 56.91 & 63.83 & 41.12 & 35.05\, \textcolor{red}{(-14)} & 64.13 & 63.19\, \textcolor{red}{(-1)} \\
gemini-2.5-flash & 58.84 & 36.68 & 48.44\, \textcolor{blue}{(+3)} & 61.56\, \textcolor{blue}{(+2)} & 53.14\, \textcolor{blue}{(+1)} & 54.60\, \textcolor{red}{(-2)} & 57.30\, \textcolor{red}{(-1)} & 38.06\, \textcolor{red}{(-1)} & 44.65\, \textcolor{blue}{(+8)} & 58.49\, \textcolor{red}{(-1)} & 66.01\, \textcolor{blue}{(+2)} \\
deepseek-V3 & 58.82 & 34.78\, \textcolor{red}{(-1)} & 47.66\, \textcolor{blue}{(+3)} & 61.52\, \textcolor{blue}{(+2)} & 53.77\, \textcolor{blue}{(+5)} & 54.75 & 56.80\, \textcolor{red}{(-2)} & 36.14\, \textcolor{red}{(-1)} & 44.65\, \textcolor{blue}{(+10)} & 57.36\, \textcolor{red}{(-1)} & 67.14\, \textcolor{blue}{(+5)} \\
gpt-4.1-nano-2025-04-14(CoT) & 58.22 & 35.77\, \textcolor{blue}{(+1)} & 39.52\, \textcolor{red}{(-5)} & 59.86 & 50.87 & 55.24\, \textcolor{blue}{(+2)} & 63.06\, \textcolor{blue}{(+2)} & 39.66\, \textcolor{blue}{(+2)} & 39.78\, \textcolor{red}{(-4)} & 63.74\, \textcolor{blue}{(+2)} & 63.07\, \textcolor{blue}{(+1)} \\
gpt-4.1-2025-04-14 & 54.85 & 30.21\, \textcolor{red}{(-1)} & 46.10\, \textcolor{blue}{(+4)} & 58.49 & 48.52\, \textcolor{red}{(-4)} & 50.43 & 52.13\, \textcolor{red}{(-1)} & 32.93 & 38.88\, \textcolor{red}{(-6)} & 55.21\, \textcolor{red}{(-1)} & 62.07 \\
gpt-4o-2024-11-20 & 53.97 & 28.78\, \textcolor{red}{(-2)} & 45.76\, \textcolor{blue}{(+4)} & 57.44 & 48.84\, \textcolor{red}{(-1)} & 49.50\, \textcolor{red}{(-1)} & 50.34\, \textcolor{red}{(-4)} & 31.35\, \textcolor{red}{(-2)} & 41.31\, \textcolor{blue}{(+3)} & 52.96\, \textcolor{red}{(-1)} & 61.98 \\
doubao-seed-1-6-flash-250715 & 53.85 & 31.20\, \textcolor{blue}{(+2)} & 34.29\, \textcolor{red}{(-6)} & 54.13\, \textcolor{red}{(-2)} & 51.18\, \textcolor{blue}{(+4)} & 47.58\, \textcolor{red}{(-3)} & 56.98\, \textcolor{blue}{(+3)} & 32.78\, \textcolor{blue}{(+1)} & 41.05\, \textcolor{blue}{(+3)} & 56.76\, \textcolor{blue}{(+2)} & 52.51\, \textcolor{red}{(-3)} \\
glm-4.5 & 53.77 & 27.18\, \textcolor{red}{(-2)} & 43.00\, \textcolor{blue}{(+3)} & 55.51 & 49.24\, \textcolor{blue}{(+2)} & 48.03 & 51.11 & 28.10\, \textcolor{red}{(-2)} & 41.68\, \textcolor{blue}{(+6)} & 50.75\, \textcolor{red}{(-1)} & 60.09 \\
claude-3-5-sonnet-20241022 & 53.63 & 29.78\, \textcolor{blue}{(+2)} & 44.99\, \textcolor{blue}{(+6)} & 56.47\, \textcolor{blue}{(+2)} & 47.69\, \textcolor{red}{(-3)} & 49.51\, \textcolor{blue}{(+3)} & 50.72 & 32.06\, \textcolor{blue}{(+2)} & 38.87\, \textcolor{red}{(-3)} & 51.92\, \textcolor{blue}{(+1)} & 61.24\, \textcolor{blue}{(+2)} \\
grok-3 & 52.19 & 28.12\, \textcolor{blue}{(+1)} & 42.96\, \textcolor{blue}{(+4)} & 53.52 & 48.72\, \textcolor{blue}{(+2)} & 47.62\, \textcolor{blue}{(+1)} & 50.01\, \textcolor{red}{(-1)} & 29.94\, \textcolor{blue}{(+1)} & 41.95\, \textcolor{blue}{(+9)} & 50.25 & 57.72\, \textcolor{blue}{(+1)} \\
qwen3-235b-a22b & 50.24 & 26.76 & 36.62\, \textcolor{red}{(-1)} & 49.83\, \textcolor{red}{(-1)} & 47.94 & 45.18 & 51.42\, \textcolor{blue}{(+4)} & 27.74 & 39.32\, \textcolor{blue}{(+2)} & 50.06 & 52.49 \\
gpt-4.1-mini-2025-04-14 & 50.03 & 25.90\, \textcolor{red}{(-1)} & 37.56\, \textcolor{blue}{(+1)} & 50.07\, \textcolor{blue}{(+1)} & 48.44\, \textcolor{blue}{(+2)} & 43.07 & 48.93 & 26.09\, \textcolor{red}{(-1)} & 39.17\, \textcolor{blue}{(+2)} & 47.49 & 51.17\, \textcolor{red}{(-1)} \\
MiniMax-Text-01 & 48.85 & 25.98\, \textcolor{blue}{(+1)} & 41.34\, \textcolor{blue}{(+6)} & 49.40 & 47.23 & 42.09 & 45.72\, \textcolor{red}{(-1)} & 26.11\, \textcolor{blue}{(+1)} & 40.77\, \textcolor{blue}{(+8)} & 44.52\, \textcolor{red}{(-1)} & 51.87\, \textcolor{blue}{(+1)} \\
qwen3-30b-a3b & 46.62 & 23.44 & 29.14\, \textcolor{red}{(-1)} & 45.40 & 45.40 & 38.53 & 48.58\, \textcolor{blue}{(+1)} & 23.42 & 35.19 & 46.76\, \textcolor{blue}{(+1)} & 42.06 \\
qwen3-32b & 43.80 & 20.61 & 28.78\, \textcolor{red}{(-1)} & 41.59\, \textcolor{red}{(-1)} & 42.97\, \textcolor{red}{(-1)} & 35.80 & 44.66 & 20.32 & 33.34\, \textcolor{red}{(-2)} & 41.66 & 38.85\, \textcolor{red}{(-1)} \\
glm-4.5-air & 43.64 & 18.30 & 31.57\, \textcolor{blue}{(+2)} & 42.37\, \textcolor{blue}{(+1)} & 43.60\, \textcolor{blue}{(+1)} & 34.25 & 42.77 & 17.82 & 36.61\, \textcolor{blue}{(+3)} & 40.36 & 40.94\, \textcolor{blue}{(+1)} \\
gpt-4.1-nano-2025-04-14 & 38.79 & 17.25 & 26.77 & 36.67 & 39.48 & 29.95 & 39.21 & 17.57 & 33.67\, \textcolor{blue}{(+1)} & 36.62 & 32.45 \\
\bottomrule
\end{tabular}}
\label{tab:meme_scores_summary_full_curated}
\end{table}

\begin{table}[H]
\centering
\caption{\textbf{Meme Scores (Curated Population; MATH-500).} The table reports meme scores across models, including property-derived 1D meme scores, predefined 2D meme scores (Mastery, Ingenuity, and Robustness), and a predefined 3D meme score (Caution). Models are sorted by Accuracy. {\color{blue}Blue} indicates rank improvement compared with Accuracy rank, and {\color{red}red} indicates rank degradation.}
\resizebox{\textwidth}{!}{
\begin{tabular}{cccccccccccc}
\toprule
\multirow{2}{*}{\textbf{Model}} & \multirow{2}{*}{\textbf{Accuracy}} &
\multicolumn{6}{c}{\textbf{Property-derived 1D meme scores}} &
\multicolumn{4}{c}{\textbf{Predefined 2D/3D meme scores}} \\
\cmidrule(lr){3-8}\cmidrule(lr){9-12}
 &  & \textbf{Difficulty} & \textbf{Uniqueness} & \textbf{Risk} & \textbf{Surprise} & \textbf{Typicality} & \textbf{Bridge} & \textbf{Mastery} & \textbf{Ingenuity} & \textbf{Robustness} & \textbf{Caution} \\
\midrule
grok-4-0709(IR) & 98.80 & 96.52 & 93.67 & 99.23\, \textcolor{red}{(-2)} & 97.10\, \textcolor{red}{(-2)} & 99.05 & 99.88\, \textcolor{red}{(-5)} & 96.83 & 93.57 & 99.89\, \textcolor{red}{(-6)} & 99.79\, \textcolor{red}{(-3)} \\
gpt-5-2025-08-07(IR) & 98.80 & 95.60 & 91.43 & 99.38\, \textcolor{blue}{(+1)} & 97.71\, \textcolor{blue}{(+1)} & 98.83 & 100.00\, \textcolor{blue}{(+1)} & 95.34\, \textcolor{red}{(-1)} & 91.87 & 100.00\, \textcolor{blue}{(+1)} & 99.91\, \textcolor{blue}{(+1)} \\
doubao-seed-1-6-250615(IR) & 98.60 & 95.38 & 91.09\, \textcolor{red}{(-1)} & 99.26\, \textcolor{blue}{(+1)} & 96.34\, \textcolor{red}{(-3)} & 98.73 & 100.00\, \textcolor{blue}{(+1)} & 95.24\, \textcolor{red}{(-1)} & 89.17\, \textcolor{red}{(-3)} & 100.00\, \textcolor{blue}{(+1)} & 99.86\, \textcolor{blue}{(+1)} \\
o3-2025-04-16(IR) & 98.20 & 94.95 & 91.13\, \textcolor{blue}{(+1)} & 98.74\, \textcolor{red}{(-2)} & 96.35\, \textcolor{red}{(-1)} & 98.52 & 99.80\, \textcolor{red}{(-4)} & 95.02\, \textcolor{red}{(-1)} & 90.60 & 99.78\, \textcolor{red}{(-4)} & 99.64\, \textcolor{red}{(-2)} \\
qwen3-235b-a22b(IR) & 98.20 & 93.50\, \textcolor{red}{(-3)} & 87.98\, \textcolor{red}{(-3)} & 98.88 & 97.37\, \textcolor{blue}{(+3)} & 98.52\, \textcolor{blue}{(+1)} & 100.00\, \textcolor{blue}{(+2)} & 94.29\, \textcolor{red}{(-2)} & 90.46 & 100.00\, \textcolor{blue}{(+2)} & 99.81\, \textcolor{blue}{(+2)} \\
deepseek-R1(IR) & 98.20 & 93.90 & 88.33\, \textcolor{red}{(-1)} & 98.99\, \textcolor{blue}{(+2)} & 95.76\, \textcolor{red}{(-1)} & 98.45 & 99.80\, \textcolor{red}{(-1)} & 94.47 & 88.08\, \textcolor{red}{(-1)} & 99.89 & 99.77\, \textcolor{blue}{(+1)} \\
gemini-2.5-flash(IR) & 97.80 & 94.23\, \textcolor{blue}{(+2)} & 89.46\, \textcolor{blue}{(+2)} & 98.56 & 93.40\, \textcolor{red}{(-3)} & 98.34 & 99.43\, \textcolor{red}{(-9)} & 95.85\, \textcolor{blue}{(+5)} & 85.99\, \textcolor{red}{(-2)} & 99.56\, \textcolor{red}{(-6)} & 99.27\, \textcolor{red}{(-3)} \\
gpt-5-nano-2025-08-07(IR) & 97.80 & 93.56\, \textcolor{blue}{(+1)} & 88.84\, \textcolor{blue}{(+2)} & 98.38 & 96.55\, \textcolor{blue}{(+4)} & 97.98 & 99.60\, \textcolor{red}{(-3)} & 93.50\, \textcolor{red}{(-1)} & 91.09\, \textcolor{blue}{(+5)} & 99.58\, \textcolor{red}{(-3)} & 99.38\, \textcolor{red}{(-1)} \\
claude-sonnet-4-20250514(IR) & 97.20 & 91.09\, \textcolor{red}{(-2)} & 84.84\, \textcolor{red}{(-3)} & 97.95\, \textcolor{red}{(-1)} & 95.69\, \textcolor{blue}{(+1)} & 97.78\, \textcolor{red}{(-1)} & 100.00\, \textcolor{blue}{(+5)} & 92.47\, \textcolor{red}{(-3)} & 87.26\, \textcolor{blue}{(+1)} & 100.00\, \textcolor{blue}{(+5)} & 99.44\, \textcolor{blue}{(+2)} \\
MiniMax-M1(IR) & 97.20 & 91.58\, \textcolor{blue}{(+1)} & 85.33\, \textcolor{blue}{(+1)} & 98.06\, \textcolor{blue}{(+1)} & 94.11\, \textcolor{blue}{(+1)} & 97.66\, \textcolor{red}{(-1)} & 99.66 & 92.93\, \textcolor{red}{(-1)} & 82.88\, \textcolor{red}{(-1)} & 99.62 & 99.17\, \textcolor{red}{(-1)} \\
gpt-5-mini-2025-08-07(IR) & 96.80 & 91.26\, \textcolor{blue}{(+1)} & 84.76\, \textcolor{red}{(-2)} & 97.80 & 91.68\, \textcolor{red}{(-1)} & 97.80\, \textcolor{blue}{(+2)} & 99.56\, \textcolor{red}{(-2)} & 93.16\, \textcolor{blue}{(+1)} & 78.04\, \textcolor{red}{(-4)} & 99.58\, \textcolor{red}{(-1)} & 99.38\, \textcolor{blue}{(+3)} \\
doubao-seed-1-6-flash-250715(IR) & 96.40 & 90.91 & 85.28\, \textcolor{blue}{(+2)} & 97.18 & 91.65\, \textcolor{red}{(-1)} & 97.59 & 99.40\, \textcolor{red}{(-5)} & 94.26\, \textcolor{blue}{(+4)} & 83.68\, \textcolor{blue}{(+2)} & 99.53\, \textcolor{red}{(-3)} & 98.85\, \textcolor{red}{(-1)} \\
grok-3(CoT) & 95.40 & 86.97\, \textcolor{red}{(-4)} & 78.68\, \textcolor{red}{(-5)} & 96.49 & 91.76\, \textcolor{blue}{(+2)} & 96.85\, \textcolor{red}{(-1)} & 100.00\, \textcolor{blue}{(+8)} & 90.40\, \textcolor{red}{(-3)} & 80.65 & 100.00\, \textcolor{blue}{(+8)} & 99.01\, \textcolor{blue}{(+1)} \\
glm-4.5(IR) & 95.40 & 90.30\, \textcolor{blue}{(+1)} & 85.24\, \textcolor{blue}{(+3)} & 96.19\, \textcolor{red}{(-1)} & 89.49\, \textcolor{red}{(-2)} & 96.01\, \textcolor{red}{(-3)} & 98.45\, \textcolor{red}{(-7)} & 91.15\, \textcolor{blue}{(+1)} & 80.42 & 98.37\, \textcolor{red}{(-6)} & 97.77\, \textcolor{red}{(-5)} \\
o3-mini-2025-01-31(IR) & 95.40 & 87.44 & 79.51 & 96.35\, \textcolor{blue}{(+1)} & 91.49\, \textcolor{blue}{(+1)} & 96.97\, \textcolor{blue}{(+2)} & 99.47\, \textcolor{blue}{(+1)} & 91.15\, \textcolor{blue}{(+1)} & 77.94\, \textcolor{red}{(-1)} & 99.42\, \textcolor{red}{(-2)} & 98.84\, \textcolor{blue}{(+1)} \\
gemini-2.5-pro(IR) & 95.20 & 89.21\, \textcolor{blue}{(+2)} & 83.19\, \textcolor{blue}{(+2)} & 95.95\, \textcolor{red}{(-2)} & 91.15\, \textcolor{blue}{(+1)} & 95.45\, \textcolor{red}{(-2)} & 97.23\, \textcolor{red}{(-13)} & 89.50\, \textcolor{red}{(-1)} & 81.76\, \textcolor{blue}{(+4)} & 97.00\, \textcolor{red}{(-14)} & 97.35\, \textcolor{red}{(-5)} \\
spark-X1(IR) & 94.80 & 87.18\, \textcolor{blue}{(+1)} & 79.31\, \textcolor{blue}{(+1)} & 96.03\, \textcolor{blue}{(+1)} & 87.59\, \textcolor{red}{(-2)} & 96.77\, \textcolor{blue}{(+2)} & 99.58\, \textcolor{blue}{(+5)} & 91.12\, \textcolor{blue}{(+2)} & 71.11\, \textcolor{red}{(-3)} & 99.54\, \textcolor{blue}{(+3)} & 98.79\, \textcolor{blue}{(+2)} \\
kimi-k2-0711-preview(CoT) & 94.80 & 85.64 & 76.46\, \textcolor{red}{(-1)} & 96.03\, \textcolor{blue}{(+1)} & 89.16\, \textcolor{blue}{(+1)} & 96.23\, \textcolor{blue}{(+2)} & 99.44\, \textcolor{blue}{(+3)} & 88.88 & 72.20\, \textcolor{red}{(-1)} & 99.51\, \textcolor{blue}{(+2)} & 98.68\, \textcolor{blue}{(+2)} \\
doubao-seed-1-6-250615(CoT) & 93.40 & 82.46\, \textcolor{red}{(-2)} & 73.00\, \textcolor{red}{(-3)} & 94.50 & 88.50\, \textcolor{blue}{(+1)} & 95.23\, \textcolor{red}{(-1)} & 99.27\, \textcolor{blue}{(+1)} & 86.86\, \textcolor{red}{(-2)} & 72.68\, \textcolor{blue}{(+1)} & 99.22\, \textcolor{blue}{(+1)} & 97.94\, \textcolor{blue}{(+1)} \\
gemini-2.5-flash(CoT) & 93.00 & 83.24 & 74.16 & 94.16\, \textcolor{red}{(-1)} & 86.19 & 94.91\, \textcolor{red}{(-1)} & 98.36\, \textcolor{red}{(-2)} & 87.47 & 69.74\, \textcolor{red}{(-2)} & 98.31\, \textcolor{red}{(-2)} & 97.39 \\
glm-4.5-air(IR) & 92.80 & 85.14\, \textcolor{blue}{(+2)} & 78.95\, \textcolor{blue}{(+4)} & 93.57\, \textcolor{red}{(-1)} & 83.22\, \textcolor{red}{(-3)} & 94.00\, \textcolor{red}{(-1)} & 97.22\, \textcolor{red}{(-9)} & 86.79\, \textcolor{red}{(-1)} & 74.11\, \textcolor{blue}{(+4)} & 97.26\, \textcolor{red}{(-8)} & 96.16\, \textcolor{red}{(-4)} \\
gpt-4.1-mini-2025-04-14(CoT) & 92.80 & 81.32 & 70.55\, \textcolor{red}{(-1)} & 94.38\, \textcolor{blue}{(+2)} & 85.08 & 95.41\, \textcolor{blue}{(+3)} & 99.71\, \textcolor{blue}{(+13)} & 87.53\, \textcolor{blue}{(+3)} & 63.94\, \textcolor{red}{(-2)} & 99.70\, \textcolor{blue}{(+13)} & 98.18\, \textcolor{blue}{(+5)} \\
claude-sonnet-4-20250514 & 91.00 & 77.35 & 66.81\, \textcolor{red}{(-1)} & 92.10\, \textcolor{red}{(-1)} & 85.69\, \textcolor{blue}{(+2)} & 93.18 & 98.45\, \textcolor{blue}{(+3)} & 82.59\, \textcolor{red}{(-1)} & 67.15 & 98.36\, \textcolor{blue}{(+2)} & 96.57 \\
claude-sonnet-4-20250514(CoT) & 90.60 & 76.81\, \textcolor{red}{(-1)} & 65.26\, \textcolor{red}{(-1)} & 92.12\, \textcolor{blue}{(+1)} & 82.01\, \textcolor{red}{(-1)} & 93.13 & 98.65\, \textcolor{blue}{(+5)} & 82.62\, \textcolor{blue}{(+1)} & 57.52\, \textcolor{red}{(-7)} & 98.57\, \textcolor{blue}{(+5)} & 96.71\, \textcolor{blue}{(+2)} \\
gpt-4.1-2025-04-14(CoT) & 89.80 & 75.92\, \textcolor{red}{(-1)} & 63.66\, \textcolor{red}{(-1)} & 91.51 & 79.96\, \textcolor{red}{(-3)} & 92.56 & 97.88\, \textcolor{blue}{(+2)} & 81.97 & 54.85\, \textcolor{red}{(-9)} & 97.70 & 96.24\, \textcolor{blue}{(+1)} \\
glm-4.5(CoT) & 89.80 & 74.37\, \textcolor{red}{(-2)} & 63.28\, \textcolor{red}{(-3)} & 90.77\, \textcolor{red}{(-1)} & 84.66\, \textcolor{blue}{(+3)} & 92.35 & 97.85\, \textcolor{blue}{(+1)} & 80.78\, \textcolor{red}{(-1)} & 63.03\, \textcolor{blue}{(+1)} & 97.59\, \textcolor{red}{(-1)} & 95.93 \\
deepseek-V3(CoT) & 89.60 & 73.95\, \textcolor{red}{(-3)} & 62.18\, \textcolor{red}{(-3)} & 90.93\, \textcolor{blue}{(+1)} & 81.97\, \textcolor{blue}{(+1)} & 91.09 & 97.74\, \textcolor{blue}{(+1)} & 77.44\, \textcolor{red}{(-3)} & 58.99\, \textcolor{red}{(-2)} & 97.60\, \textcolor{blue}{(+1)} & 95.55 \\
glm-4.5-air(CoT) & 89.20 & 74.91\, \textcolor{blue}{(+1)} & 63.58\, \textcolor{blue}{(+1)} & 90.68 & 81.30\, \textcolor{blue}{(+1)} & 90.59\, \textcolor{red}{(-1)} & 97.87\, \textcolor{blue}{(+4)} & 78.58\, \textcolor{red}{(-1)} & 59.69\, \textcolor{blue}{(+2)} & 97.95\, \textcolor{blue}{(+5)} & 95.01 \\
doubao-seed-1-6-250615 & 88.60 & 72.59\, \textcolor{red}{(-3)} & 59.92\, \textcolor{red}{(-3)} & 90.09 & 79.50 & 90.22\, \textcolor{red}{(-1)} & 97.36\, \textcolor{blue}{(+1)} & 76.40\, \textcolor{red}{(-3)} & 56.26\, \textcolor{red}{(-3)} & 97.36\, \textcolor{blue}{(+1)} & 94.71 \\
doubao-seed-1-6-flash-250715(CoT) & 88.20 & 74.13\, \textcolor{blue}{(+1)} & 63.54\, \textcolor{blue}{(+2)} & 89.31 & 78.44\, \textcolor{red}{(-2)} & 88.99\, \textcolor{red}{(-1)} & 95.95\, \textcolor{red}{(-2)} & 76.81\, \textcolor{red}{(-1)} & 59.37\, \textcolor{blue}{(+2)} & 95.84\, \textcolor{red}{(-1)} & 92.81\, \textcolor{red}{(-1)} \\
qwen3-235b-a22b(CoT) & 87.60 & 72.76 & 61.00 & 89.04 & 77.27\, \textcolor{red}{(-2)} & 90.63\, \textcolor{blue}{(+3)} & 97.65\, \textcolor{blue}{(+4)} & 78.62\, \textcolor{blue}{(+3)} & 56.07\, \textcolor{red}{(-2)} & 97.77\, \textcolor{blue}{(+7)} & 94.66\, \textcolor{blue}{(+1)} \\
kimi-k2-0711-preview & 86.80 & 77.12\, \textcolor{blue}{(+8)} & 73.03\, \textcolor{blue}{(+11)} & 86.54\, \textcolor{red}{(-1)} & 79.11\, \textcolor{blue}{(+1)} & 88.84 & 90.76\, \textcolor{red}{(-3)} & 81.74\, \textcolor{blue}{(+6)} & 70.27\, \textcolor{blue}{(+11)} & 89.59\, \textcolor{red}{(-3)} & 90.33\, \textcolor{red}{(-2)} \\
qwen3-30b-a3b(CoT) & 86.60 & 69.53 & 59.53 & 87.34\, \textcolor{blue}{(+1)} & 79.16\, \textcolor{blue}{(+3)} & 87.98 & 96.02\, \textcolor{blue}{(+2)} & 74.03 & 59.39\, \textcolor{blue}{(+6)} & 95.56\, \textcolor{blue}{(+1)} & 92.23\, \textcolor{blue}{(+1)} \\
qwen3-32b(CoT) & 83.40 & 66.35 & 53.88 & 85.00 & 69.14\, \textcolor{red}{(-2)} & 87.27 & 95.16\, \textcolor{blue}{(+1)} & 73.75 & 43.82\, \textcolor{red}{(-6)} & 94.93\, \textcolor{blue}{(+1)} & 91.81\, \textcolor{blue}{(+1)} \\
gpt-4.1-nano-2025-04-14(CoT) & 80.60 & 61.40 & 48.53\, \textcolor{red}{(-1)} & 82.14 & 68.36\, \textcolor{red}{(-3)} & 84.20 & 94.51\, \textcolor{blue}{(+1)} & 69.06 & 42.59\, \textcolor{red}{(-7)} & 94.37\, \textcolor{blue}{(+1)} & 89.77 \\
doubao-seed-1-6-flash-250715 & 75.40 & 57.73 & 50.78\, \textcolor{blue}{(+1)} & 74.70\, \textcolor{red}{(-2)} & 74.67\, \textcolor{blue}{(+2)} & 76.67 & 84.04\, \textcolor{red}{(-3)} & 63.00 & 57.57\, \textcolor{blue}{(+6)} & 82.57\, \textcolor{red}{(-3)} & 80.78\, \textcolor{red}{(-3)} \\
MiniMax-Text-01(CoT) & 75.20 & 53.15\, \textcolor{red}{(-1)} & 44.36\, \textcolor{red}{(-1)} & 74.95 & 69.32\, \textcolor{blue}{(+2)} & 75.56\, \textcolor{red}{(-2)} & 87.90\, \textcolor{red}{(-1)} & 57.51\, \textcolor{red}{(-2)} & 48.36 & 86.36\, \textcolor{red}{(-1)} & 80.97\, \textcolor{red}{(-1)} \\
claude-3-5-sonnet-20241022(CoT) & 75.00 & 52.64\, \textcolor{red}{(-1)} & 43.42\, \textcolor{red}{(-2)} & 74.66\, \textcolor{red}{(-1)} & 68.44\, \textcolor{blue}{(+1)} & 76.34\, \textcolor{blue}{(+1)} & 88.16\, \textcolor{blue}{(+1)} & 58.28 & 43.96\, \textcolor{red}{(-1)} & 86.49\, \textcolor{blue}{(+1)} & 81.28\, \textcolor{blue}{(+1)} \\
gpt-4o-2024-11-20(CoT) & 75.00 & 53.82\, \textcolor{blue}{(+2)} & 43.96 & 75.19\, \textcolor{blue}{(+3)} & 67.68 & 76.22\, \textcolor{blue}{(+1)} & 88.79\, \textcolor{blue}{(+3)} & 58.99\, \textcolor{blue}{(+2)} & 43.40\, \textcolor{red}{(-2)} & 87.56\, \textcolor{blue}{(+3)} & 81.83\, \textcolor{blue}{(+3)} \\
gemini-2.5-flash & 65.00 & 46.83 & 47.95\, \textcolor{blue}{(+3)} & 61.28 & 67.02 & 61.02 & 69.62 & 47.55 & 54.14\, \textcolor{blue}{(+5)} & 64.39 & 62.28 \\
deepseek-V3 & 63.60 & 42.02 & 42.87 & 59.42 & 66.19 & 59.92 & 68.02 & 42.16 & 50.25\, \textcolor{blue}{(+5)} & 62.02 & 61.87 \\
qwen3-235b-a22b & 53.60 & 32.46\, \textcolor{red}{(-1)} & 36.98 & 48.00 & 56.61\, \textcolor{red}{(-1)} & 50.97 & 59.57 & 34.44 & 46.99\, \textcolor{blue}{(+4)} & 52.69 & 51.73 \\
qwen3-30b-a3b & 52.80 & 32.64\, \textcolor{blue}{(+1)} & 35.12 & 47.84 & 56.73\, \textcolor{blue}{(+1)} & 47.99 & 58.50 & 33.19 & 42.24 & 52.31 & 48.80 \\
glm-4.5 & 49.20 & 25.47\, \textcolor{red}{(-1)} & 33.65 & 41.85\, \textcolor{red}{(-1)} & 53.40\, \textcolor{red}{(-1)} & 42.50\, \textcolor{red}{(-1)} & 52.34 & 24.65\, \textcolor{red}{(-1)} & 42.21 & 43.27\, \textcolor{red}{(-1)} & 41.91\, \textcolor{red}{(-1)} \\
gpt-4.1-mini-2025-04-14 & 49.00 & 27.66\, \textcolor{blue}{(+1)} & 33.36 & 42.26\, \textcolor{blue}{(+1)} & 54.38\, \textcolor{blue}{(+1)} & 42.58\, \textcolor{blue}{(+1)} & 51.85 & 27.63\, \textcolor{blue}{(+1)} & 41.25 & 43.68\, \textcolor{blue}{(+1)} & 42.30\, \textcolor{blue}{(+1)} \\
grok-3 & 41.00 & 17.89\, \textcolor{red}{(-1)} & 29.26 & 31.96\, \textcolor{red}{(-1)} & 48.62 & 34.94 & 43.77\, \textcolor{red}{(-1)} & 18.10\, \textcolor{red}{(-1)} & 41.19 & 33.45\, \textcolor{red}{(-2)} & 33.08 \\
qwen3-32b & 40.60 & 18.57\, \textcolor{blue}{(+1)} & 27.46\, \textcolor{red}{(-1)} & 32.57\, \textcolor{blue}{(+1)} & 45.23\, \textcolor{red}{(-2)} & 33.39\, \textcolor{red}{(-1)} & 43.81\, \textcolor{blue}{(+1)} & 18.42\, \textcolor{blue}{(+1)} & 34.24\, \textcolor{red}{(-2)} & 34.15\, \textcolor{blue}{(+1)} & 32.00\, \textcolor{red}{(-1)} \\
claude-3-5-sonnet-20241022 & 39.60 & 16.27\, \textcolor{red}{(-1)} & 24.80\, \textcolor{red}{(-3)} & 31.44 & 43.97\, \textcolor{red}{(-2)} & 33.65\, \textcolor{blue}{(+1)} & 43.61 & 17.02 & 29.48\, \textcolor{red}{(-3)} & 33.65\, \textcolor{blue}{(+1)} & 32.17\, \textcolor{blue}{(+1)} \\
MiniMax-Text-01 & 38.60 & 16.85\, \textcolor{blue}{(+1)} & 27.46\, \textcolor{blue}{(+2)} & 29.73 & 46.17\, \textcolor{blue}{(+2)} & 30.59\, \textcolor{red}{(-1)} & 40.22\, \textcolor{red}{(-1)} & 16.28\, \textcolor{red}{(-1)} & 35.72\, \textcolor{blue}{(+1)} & 29.91\, \textcolor{red}{(-1)} & 28.04\, \textcolor{red}{(-1)} \\
gpt-4o-2024-11-20 & 38.40 & 15.24 & 26.49\, \textcolor{blue}{(+1)} & 29.31 & 45.61\, \textcolor{blue}{(+2)} & 33.28\, \textcolor{blue}{(+1)} & 41.02\, \textcolor{blue}{(+1)} & 16.72\, \textcolor{blue}{(+1)} & 35.90\, \textcolor{blue}{(+3)} & 30.59\, \textcolor{blue}{(+1)} & 31.52\, \textcolor{blue}{(+1)} \\
gpt-4.1-2025-04-14 & 36.60 & 13.05\, \textcolor{red}{(-2)} & 23.68\, \textcolor{red}{(-1)} & 27.41 & 41.73\, \textcolor{red}{(-1)} & 29.62 & 40.07 & 13.62\, \textcolor{red}{(-2)} & 27.15\, \textcolor{red}{(-1)} & 29.06 & 27.02 \\
glm-4.5-air & 36.40 & 14.74\, \textcolor{blue}{(+1)} & 26.43\, \textcolor{blue}{(+2)} & 26.94 & 43.67\, \textcolor{blue}{(+1)} & 27.38 & 37.52 & 13.80 & 34.22\, \textcolor{blue}{(+2)} & 26.58 & 24.31 \\
gpt-4.1-nano-2025-04-14 & 31.40 & 14.48\, \textcolor{blue}{(+1)} & 21.74 & 23.61 & 34.29 & 25.28 & 33.50 & 14.56\, \textcolor{blue}{(+2)} & 23.95 & 24.49 & 22.44 \\
\bottomrule
\end{tabular}}
\label{tab:meme_scores_summary_full_curated-MATH-500}
\end{table}

\begin{table}[H]
\centering
\caption{\textbf{Meme Scores (Curated Population; MMLU-Redux).} The table reports meme scores across models, including property-derived 1D meme scores, predefined 2D meme scores (Mastery, Ingenuity, and Robustness), and a predefined 3D meme score (Caution). Models are sorted by Accuracy. {\color{blue}Blue} indicates rank improvement compared with Accuracy rank, and {\color{red}red} indicates rank degradation.}
\resizebox{\textwidth}{!}{
\begin{tabular}{cccccccccccc}
\toprule
\multirow{2}{*}{\textbf{Model}} & \multirow{2}{*}{\textbf{Accuracy}} &
\multicolumn{6}{c}{\textbf{Property-derived 1D meme scores}} &
\multicolumn{4}{c}{\textbf{Predefined 2D/3D meme scores}} \\
\cmidrule(lr){3-8}\cmidrule(lr){9-12}
 &  & \textbf{Difficulty} & \textbf{Uniqueness} & \textbf{Risk} & \textbf{Surprise} & \textbf{Typicality} & \textbf{Bridge} & \textbf{Mastery} & \textbf{Ingenuity} & \textbf{Robustness} & \textbf{Caution} \\
\midrule
gpt-5-2025-08-07(IR) & 95.28 & 68.60 & 67.17 & 96.02 & 75.32\, \textcolor{red}{(-1)} & 92.06 & 97.67 & 70.93 & 42.14\, \textcolor{red}{(-1)} & 97.71 & 97.88 \\
gemini-2.5-pro(IR) & 95.00 & 66.76 & 65.32 & 95.71 & 75.79\, \textcolor{blue}{(+1)} & 91.54 & 97.64 & 68.87 & 40.17\, \textcolor{red}{(-5)} & 97.68 & 97.66 \\
o3-2025-04-16(IR) & 94.72 & 66.73 & 65.25 & 95.59 & 73.44\, \textcolor{red}{(-3)} & 90.82\, \textcolor{red}{(-1)} & 97.43\, \textcolor{red}{(-2)} & 68.78 & 40.79 & 97.51\, \textcolor{red}{(-2)} & 97.10\, \textcolor{red}{(-2)} \\
grok-4-0709(IR) & 94.72 & 66.16 & 64.72 & 95.46 & 74.95\, \textcolor{blue}{(+1)} & 91.13\, \textcolor{blue}{(+1)} & 97.48 & 68.25 & 40.74 & 97.53\, \textcolor{blue}{(+1)} & 97.32\, \textcolor{blue}{(+1)} \\
claude-sonnet-4-20250514(IR) & 94.24 & 62.12\, \textcolor{red}{(-1)} & 60.47\, \textcolor{red}{(-1)} & 95.10 & 73.49 & 90.04 & 97.52\, \textcolor{blue}{(+2)} & 63.98\, \textcolor{red}{(-1)} & 35.75\, \textcolor{red}{(-11)} & 97.53\, \textcolor{blue}{(+1)} & 97.15\, \textcolor{blue}{(+1)} \\
claude-sonnet-4-20250514 & 94.00 & 62.59\, \textcolor{blue}{(+1)} & 61.18\, \textcolor{blue}{(+1)} & 94.64 & 74.95\, \textcolor{blue}{(+2)} & 89.90 & 97.39 & 65.06\, \textcolor{blue}{(+1)} & 40.55\, \textcolor{blue}{(+1)} & 97.44\, \textcolor{red}{(-1)} & 96.41 \\
deepseek-R1(IR) & 93.61 & 60.70 & 59.06\, \textcolor{red}{(-1)} & 94.54 & 70.47\, \textcolor{red}{(-11)} & 88.80\, \textcolor{red}{(-3)} & 97.22\, \textcolor{red}{(-5)} & 62.55\, \textcolor{red}{(-2)} & 33.88\, \textcolor{red}{(-26)} & 97.33\, \textcolor{red}{(-5)} & 96.08\, \textcolor{red}{(-2)} \\
claude-sonnet-4-20250514(CoT) & 93.54 & 59.98\, \textcolor{red}{(-2)} & 58.36\, \textcolor{red}{(-2)} & 94.43 & 71.35\, \textcolor{red}{(-3)} & 88.92\, \textcolor{red}{(-1)} & 97.35 & 62.12\, \textcolor{red}{(-3)} & 34.63\, \textcolor{red}{(-17)} & 97.41 & 96.17\, \textcolor{blue}{(+1)} \\
gpt-5-mini-2025-08-07(IR) & 93.50 & 60.52 & 58.96 & 94.36\, \textcolor{red}{(-1)} & 71.30\, \textcolor{red}{(-3)} & 88.96\, \textcolor{blue}{(+2)} & 97.30\, \textcolor{red}{(-1)} & 62.74\, \textcolor{blue}{(+1)} & 35.70\, \textcolor{red}{(-8)} & 97.39 & 96.07\, \textcolor{red}{(-1)} \\
gemini-2.5-flash(IR) & 93.48 & 60.67\, \textcolor{blue}{(+2)} & 59.10\, \textcolor{blue}{(+3)} & 94.40\, \textcolor{blue}{(+1)} & 70.76\, \textcolor{red}{(-6)} & 88.94\, \textcolor{blue}{(+2)} & 97.35\, \textcolor{blue}{(+3)} & 63.23\, \textcolor{blue}{(+3)} & 34.96\, \textcolor{red}{(-10)} & 97.46\, \textcolor{blue}{(+4)} & 96.05\, \textcolor{red}{(-1)} \\
doubao-seed-1-6-250615(IR) & 93.44 & 59.36\, \textcolor{red}{(-1)} & 57.70\, \textcolor{red}{(-1)} & 94.36 & 70.82\, \textcolor{red}{(-4)} & 88.65 & 97.17\, \textcolor{red}{(-2)} & 61.91\, \textcolor{red}{(-1)} & 33.52\, \textcolor{red}{(-23)} & 97.30\, \textcolor{red}{(-2)} & 96.09\, \textcolor{blue}{(+3)} \\
grok-3(CoT) & 93.22 & 59.93\, \textcolor{blue}{(+1)} & 58.35\, \textcolor{blue}{(+1)} & 94.15 & 69.57\, \textcolor{red}{(-12)} & 88.39 & 96.94\, \textcolor{red}{(-5)} & 62.19\, \textcolor{blue}{(+2)} & 35.87\, \textcolor{red}{(-3)} & 97.13\, \textcolor{red}{(-3)} & 95.64 \\
gpt-4.1-2025-04-14(CoT) & 92.91 & 56.72\, \textcolor{red}{(-2)} & 55.17\, \textcolor{red}{(-2)} & 93.66\, \textcolor{red}{(-1)} & 72.19\, \textcolor{blue}{(+4)} & 87.84\, \textcolor{red}{(-1)} & 97.30\, \textcolor{blue}{(+2)} & 59.68\, \textcolor{red}{(-2)} & 34.58\, \textcolor{red}{(-16)} & 97.36\, \textcolor{blue}{(+2)} & 95.29\, \textcolor{red}{(-1)} \\
kimi-k2-0711-preview(CoT) & 92.91 & 56.87 & 55.33 & 93.67\, \textcolor{blue}{(+1)} & 72.53\, \textcolor{blue}{(+6)} & 88.10\, \textcolor{blue}{(+1)} & 97.35\, \textcolor{blue}{(+5)} & 59.70 & 34.62\, \textcolor{red}{(-13)} & 97.39\, \textcolor{blue}{(+4)} & 95.50\, \textcolor{blue}{(+1)} \\
qwen3-235b-a22b(IR) & 92.74 & 59.15\, \textcolor{blue}{(+2)} & 57.64\, \textcolor{blue}{(+2)} & 93.62 & 68.56\, \textcolor{red}{(-20)} & 87.34 & 96.56\, \textcolor{red}{(-5)} & 61.62\, \textcolor{blue}{(+2)} & 35.21\, \textcolor{red}{(-3)} & 96.79\, \textcolor{red}{(-5)} & 94.37\, \textcolor{red}{(-2)} \\
doubao-seed-1-6-250615(CoT) & 92.52 & 56.55 & 54.89 & 93.54 & 69.13\, \textcolor{red}{(-14)} & 86.89\, \textcolor{red}{(-1)} & 96.95 & 59.25 & 32.10\, \textcolor{red}{(-27)} & 97.11 & 94.67 \\
gemini-2.5-flash(CoT) & 92.48 & 55.80 & 54.13 & 93.51 & 69.52\, \textcolor{red}{(-8)} & 87.05\, \textcolor{blue}{(+1)} & 96.97\, \textcolor{blue}{(+3)} & 58.72 & 31.81\, \textcolor{red}{(-28)} & 97.16\, \textcolor{blue}{(+3)} & 94.90\, \textcolor{blue}{(+2)} \\
claude-3-5-sonnet-20241022(CoT) & 91.48 & 55.38 & 53.98 & 92.46\, \textcolor{red}{(-1)} & 69.40\, \textcolor{red}{(-8)} & 84.56\, \textcolor{red}{(-5)} & 96.11\, \textcolor{red}{(-7)} & 57.31\, \textcolor{red}{(-1)} & 34.75\, \textcolor{red}{(-6)} & 96.37\, \textcolor{red}{(-8)} & 92.07\, \textcolor{red}{(-5)} \\
MiniMax-M1(IR) & 91.46 & 54.78 & 53.20\, \textcolor{red}{(-1)} & 92.55\, \textcolor{blue}{(+1)} & 66.41\, \textcolor{red}{(-30)} & 85.08 & 96.27\, \textcolor{red}{(-4)} & 57.41\, \textcolor{blue}{(+1)} & 33.03\, \textcolor{red}{(-19)} & 96.61\, \textcolor{red}{(-3)} & 92.78 \\
glm-4.5(CoT) & 91.37 & 52.08\, \textcolor{red}{(-4)} & 50.60\, \textcolor{red}{(-4)} & 92.16\, \textcolor{red}{(-2)} & 70.12\, \textcolor{red}{(-1)} & 85.04 & 96.90\, \textcolor{blue}{(+2)} & 55.36\, \textcolor{red}{(-2)} & 32.55\, \textcolor{red}{(-19)} & 97.02\, \textcolor{blue}{(+2)} & 92.71\, \textcolor{red}{(-1)} \\
deepseek-V3(CoT) & 91.35 & 52.10\, \textcolor{red}{(-2)} & 50.60\, \textcolor{red}{(-2)} & 92.20 & 70.03\, \textcolor{red}{(-1)} & 85.26\, \textcolor{blue}{(+3)} & 96.85\, \textcolor{blue}{(+2)} & 55.27\, \textcolor{red}{(-3)} & 33.24\, \textcolor{red}{(-15)} & 96.94\, \textcolor{blue}{(+2)} & 93.13\, \textcolor{blue}{(+3)} \\
doubao-seed-1-6-250615 & 91.28 & 51.98\, \textcolor{red}{(-3)} & 50.45\, \textcolor{red}{(-3)} & 92.20\, \textcolor{blue}{(+2)} & 69.16\, \textcolor{red}{(-6)} & 84.79\, \textcolor{blue}{(+1)} & 96.52\, \textcolor{blue}{(+1)} & 54.85\, \textcolor{red}{(-3)} & 31.52\, \textcolor{red}{(-24)} & 96.65\, \textcolor{blue}{(+1)} & 92.71\, \textcolor{blue}{(+2)} \\
MiniMax-Text-01(CoT) & 91.05 & 54.59\, \textcolor{blue}{(+3)} & 53.26\, \textcolor{blue}{(+4)} & 91.95 & 68.90\, \textcolor{red}{(-8)} & 84.51\, \textcolor{red}{(-1)} & 96.09\, \textcolor{red}{(-4)} & 57.22\, \textcolor{blue}{(+2)} & 34.88\, \textcolor{blue}{(+2)} & 96.36\, \textcolor{red}{(-4)} & 91.71\, \textcolor{red}{(-1)} \\
gpt-4o-2024-11-20(CoT) & 90.89 & 51.35\, \textcolor{red}{(-2)} & 49.98\, \textcolor{red}{(-2)} & 91.69\, \textcolor{red}{(-3)} & 70.20\, \textcolor{blue}{(+4)} & 83.76\, \textcolor{red}{(-2)} & 96.36\, \textcolor{blue}{(+2)} & 53.94\, \textcolor{red}{(-4)} & 32.43\, \textcolor{red}{(-16)} & 96.51\, \textcolor{blue}{(+1)} & 91.42\, \textcolor{red}{(-1)} \\
gpt-4.1-mini-2025-04-14(CoT) & 90.85 & 50.88\, \textcolor{red}{(-2)} & 49.33\, \textcolor{red}{(-2)} & 91.85\, \textcolor{blue}{(+1)} & 68.48\, \textcolor{red}{(-11)} & 84.60\, \textcolor{blue}{(+3)} & 96.96\, \textcolor{blue}{(+10)} & 54.62\, \textcolor{red}{(-1)} & 31.02\, \textcolor{red}{(-23)} & 97.08\, \textcolor{blue}{(+8)} & 92.56\, \textcolor{blue}{(+3)} \\
o3-mini-2025-01-31(IR) & 90.81 & 52.24\, \textcolor{blue}{(+4)} & 50.71\, \textcolor{blue}{(+4)} & 91.81\, \textcolor{blue}{(+1)} & 67.22\, \textcolor{red}{(-19)} & 83.75\, \textcolor{red}{(-1)} & 96.06\, \textcolor{red}{(-2)} & 55.30\, \textcolor{blue}{(+3)} & 31.22\, \textcolor{red}{(-21)} & 96.34\, \textcolor{red}{(-2)} & 91.42 \\
gpt-5-nano-2025-08-07(IR) & 90.76 & 53.97\, \textcolor{blue}{(+6)} & 52.56\, \textcolor{blue}{(+6)} & 91.77\, \textcolor{blue}{(+1)} & 68.47\, \textcolor{red}{(-10)} & 83.81\, \textcolor{blue}{(+2)} & 96.12\, \textcolor{blue}{(+3)} & 57.30\, \textcolor{blue}{(+7)} & 34.76\, \textcolor{blue}{(+4)} & 96.41\, \textcolor{blue}{(+3)} & 90.96 \\
qwen3-235b-a22b(CoT) & 90.02 & 50.44 & 49.06 & 91.03 & 68.06\, \textcolor{red}{(-11)} & 82.32 & 96.10\, \textcolor{blue}{(+2)} & 53.97\, \textcolor{blue}{(+1)} & 33.16\, \textcolor{red}{(-9)} & 96.39\, \textcolor{blue}{(+3)} & 89.99\, \textcolor{red}{(-1)} \\
glm-4.5(IR) & 89.87 & 48.58\, \textcolor{red}{(-2)} & 47.05\, \textcolor{red}{(-4)} & 90.93 & 67.72\, \textcolor{red}{(-13)} & 82.07 & 95.60\, \textcolor{red}{(-1)} & 51.57\, \textcolor{red}{(-2)} & 28.64\, \textcolor{red}{(-23)} & 95.85\, \textcolor{red}{(-1)} & 90.11\, \textcolor{blue}{(+1)} \\
doubao-seed-1-6-flash-250715(IR) & 89.79 & 49.99\, \textcolor{blue}{(+1)} & 48.64\, \textcolor{blue}{(+1)} & 90.71 & 68.06\, \textcolor{red}{(-10)} & 81.74 & 95.61\, \textcolor{blue}{(+1)} & 52.77 & 32.18\, \textcolor{red}{(-12)} & 95.92\, \textcolor{blue}{(+1)} & 89.30 \\
kimi-k2-0711-preview & 89.20 & 48.44\, \textcolor{red}{(-1)} & 47.41 & 89.98 & 71.69\, \textcolor{blue}{(+21)} & 79.50\, \textcolor{red}{(-1)} & 93.42\, \textcolor{red}{(-6)} & 49.93\, \textcolor{red}{(-1)} & 34.62\, \textcolor{blue}{(+5)} & 93.69\, \textcolor{red}{(-6)} & 86.62\, \textcolor{red}{(-1)} \\
spark-X1(IR) & 88.52 & 49.66\, \textcolor{blue}{(+2)} & 48.39\, \textcolor{blue}{(+2)} & 89.47 & 69.15\, \textcolor{blue}{(+3)} & 80.19\, \textcolor{blue}{(+1)} & 94.01\, \textcolor{red}{(-1)} & 53.45\, \textcolor{blue}{(+3)} & 34.48\, \textcolor{blue}{(+1)} & 94.52\, \textcolor{red}{(-1)} & 87.07\, \textcolor{blue}{(+1)} \\
deepseek-V3 & 88.22 & 45.81\, \textcolor{red}{(-4)} & 44.84\, \textcolor{red}{(-5)} & 89.03\, \textcolor{red}{(-1)} & 70.63\, \textcolor{blue}{(+16)} & 77.50\, \textcolor{red}{(-1)} & 93.51\, \textcolor{red}{(-3)} & 47.54\, \textcolor{red}{(-2)} & 34.80\, \textcolor{blue}{(+11)} & 93.81\, \textcolor{red}{(-3)} & 84.66\, \textcolor{red}{(-1)} \\
glm-4.5-air(CoT) & 88.05 & 43.65\, \textcolor{red}{(-6)} & 42.34\, \textcolor{red}{(-7)} & 89.04\, \textcolor{blue}{(+1)} & 66.55\, \textcolor{red}{(-14)} & 78.63\, \textcolor{blue}{(+1)} & 95.44\, \textcolor{blue}{(+3)} & 47.24\, \textcolor{red}{(-4)} & 29.12\, \textcolor{red}{(-17)} & 95.75\, \textcolor{blue}{(+3)} & 86.62\, \textcolor{blue}{(+1)} \\
claude-3-5-sonnet-20241022 & 87.92 & 48.19\, \textcolor{blue}{(+2)} & 47.31\, \textcolor{blue}{(+3)} & 88.91 & 70.34\, \textcolor{blue}{(+16)} & 75.97\, \textcolor{red}{(-5)} & 91.61\, \textcolor{red}{(-6)} & 48.28\, \textcolor{blue}{(+2)} & 36.17\, \textcolor{blue}{(+22)} & 92.14\, \textcolor{red}{(-5)} & 82.85\, \textcolor{red}{(-5)} \\
gemini-2.5-flash & 87.89 & 46.96\, \textcolor{blue}{(+1)} & 46.02\, \textcolor{blue}{(+1)} & 88.84 & 69.29\, \textcolor{blue}{(+9)} & 76.34\, \textcolor{red}{(-3)} & 92.27\, \textcolor{red}{(-2)} & 47.71\, \textcolor{blue}{(+2)} & 34.58\, \textcolor{blue}{(+8)} & 92.72\, \textcolor{red}{(-2)} & 83.34\, \textcolor{red}{(-2)} \\
grok-3 & 87.79 & 46.05\, \textcolor{blue}{(+1)} & 45.21\, \textcolor{blue}{(+1)} & 88.57 & 71.26\, \textcolor{blue}{(+24)} & 76.37\, \textcolor{red}{(-1)} & 92.10\, \textcolor{red}{(-2)} & 47.30 & 36.51\, \textcolor{blue}{(+25)} & 92.51\, \textcolor{red}{(-2)} & 83.21\, \textcolor{red}{(-2)} \\
qwen3-30b-a3b(CoT) & 87.29 & 44.53\, \textcolor{red}{(-1)} & 43.44\, \textcolor{red}{(-1)} & 88.16 & 68.18 & 76.63\, \textcolor{blue}{(+1)} & 94.29\, \textcolor{blue}{(+6)} & 47.53\, \textcolor{blue}{(+2)} & 34.51\, \textcolor{blue}{(+8)} & 94.66\, \textcolor{blue}{(+6)} & 83.71\, \textcolor{blue}{(+1)} \\
qwen3-32b(CoT) & 87.20 & 43.54\, \textcolor{red}{(-2)} & 42.38\, \textcolor{red}{(-1)} & 88.15 & 67.85\, \textcolor{red}{(-2)} & 76.64\, \textcolor{blue}{(+3)} & 93.92\, \textcolor{blue}{(+4)} & 46.03\, \textcolor{red}{(-2)} & 32.09\, \textcolor{red}{(-5)} & 94.28\, \textcolor{blue}{(+4)} & 84.14\, \textcolor{blue}{(+3)} \\
gpt-4.1-2025-04-14 & 87.09 & 45.60\, \textcolor{blue}{(+2)} & 44.85\, \textcolor{blue}{(+3)} & 87.93\, \textcolor{red}{(-1)} & 71.24\, \textcolor{blue}{(+26)} & 73.59\, \textcolor{red}{(-2)} & 89.94\, \textcolor{red}{(-5)} & 45.11\, \textcolor{red}{(-2)} & 35.14\, \textcolor{blue}{(+21)} & 90.42\, \textcolor{red}{(-5)} & 80.21\, \textcolor{red}{(-2)} \\
doubao-seed-1-6-flash-250715(CoT) & 86.96 & 43.44\, \textcolor{red}{(-1)} & 42.22\, \textcolor{red}{(-1)} & 87.99\, \textcolor{blue}{(+1)} & 65.27\, \textcolor{red}{(-11)} & 76.68\, \textcolor{blue}{(+6)} & 93.99\, \textcolor{blue}{(+7)} & 46.67\, \textcolor{blue}{(+1)} & 29.50\, \textcolor{red}{(-9)} & 94.43\, \textcolor{blue}{(+7)} & 84.23\, \textcolor{blue}{(+6)} \\
MiniMax-Text-01 & 86.83 & 47.40\, \textcolor{blue}{(+8)} & 46.89\, \textcolor{blue}{(+8)} & 87.43 & 72.86\, \textcolor{blue}{(+35)} & 72.81\, \textcolor{red}{(-2)} & 89.65\, \textcolor{red}{(-4)} & 46.85\, \textcolor{blue}{(+3)} & 40.27\, \textcolor{blue}{(+36)} & 90.13\, \textcolor{red}{(-4)} & 78.30\, \textcolor{red}{(-2)} \\
glm-4.5 & 86.35 & 38.71\, \textcolor{red}{(-5)} & 37.86\, \textcolor{red}{(-5)} & 87.05 & 68.74\, \textcolor{blue}{(+11)} & 73.59\, \textcolor{blue}{(+2)} & 91.77\, \textcolor{blue}{(+3)} & 40.19\, \textcolor{red}{(-3)} & 30.89\, \textcolor{red}{(-6)} & 92.04\, \textcolor{blue}{(+2)} & 81.01\, \textcolor{blue}{(+2)} \\
gpt-4o-2024-11-20 & 85.96 & 42.85\, \textcolor{blue}{(+1)} & 42.21\, \textcolor{blue}{(+1)} & 86.75 & 69.94\, \textcolor{blue}{(+21)} & 71.11\, \textcolor{red}{(-2)} & 88.72\, \textcolor{red}{(-4)} & 41.97\, \textcolor{red}{(-1)} & 33.40\, \textcolor{blue}{(+9)} & 89.19\, \textcolor{red}{(-4)} & 77.51\, \textcolor{red}{(-1)} \\
qwen3-235b-a22b & 85.92 & 41.49\, \textcolor{blue}{(+1)} & 40.76\, \textcolor{blue}{(+1)} & 86.69 & 68.57\, \textcolor{blue}{(+11)} & 72.86\, \textcolor{blue}{(+2)} & 90.90\, \textcolor{blue}{(+2)} & 42.20\, \textcolor{blue}{(+1)} & 33.94\, \textcolor{blue}{(+13)} & 91.24\, \textcolor{blue}{(+1)} & 79.66\, \textcolor{blue}{(+2)} \\
gpt-4.1-mini-2025-04-14 & 84.57 & 39.40 & 38.78 & 85.36 & 67.47\, \textcolor{blue}{(+3)} & 69.29\, \textcolor{red}{(-2)} & 88.97\, \textcolor{red}{(-1)} & 39.31\, \textcolor{red}{(-1)} & 32.22\, \textcolor{blue}{(+5)} & 89.35\, \textcolor{red}{(-1)} & 75.54\, \textcolor{red}{(-2)} \\
gpt-4.1-nano-2025-04-14(CoT) & 84.02 & 40.47\, \textcolor{blue}{(+2)} & 39.63\, \textcolor{blue}{(+2)} & 84.78\, \textcolor{red}{(-1)} & 66.41\, \textcolor{red}{(-3)} & 71.57\, \textcolor{blue}{(+2)} & 91.37\, \textcolor{blue}{(+5)} & 44.67\, \textcolor{blue}{(+4)} & 38.91\, \textcolor{blue}{(+39)} & 91.79\, \textcolor{blue}{(+5)} & 77.37\, \textcolor{blue}{(+1)} \\
qwen3-32b & 83.83 & 39.16\, \textcolor{blue}{(+1)} & 38.71\, \textcolor{blue}{(+1)} & 84.46\, \textcolor{red}{(-1)} & 68.68\, \textcolor{blue}{(+15)} & 67.24\, \textcolor{red}{(-1)} & 87.93\, \textcolor{red}{(-1)} & 38.48 & 36.03\, \textcolor{blue}{(+34)} & 88.39\, \textcolor{red}{(-1)} & 72.86\, \textcolor{red}{(-1)} \\
glm-4.5-air(IR) & 83.46 & 35.87 & 34.78 & 84.79\, \textcolor{blue}{(+2)} & 58.89\, \textcolor{red}{(-4)} & 69.32\, \textcolor{blue}{(+2)} & 90.87\, \textcolor{blue}{(+5)} & 38.03 & 23.39\, \textcolor{red}{(-4)} & 91.51\, \textcolor{blue}{(+6)} & 77.00\, \textcolor{blue}{(+2)} \\
glm-4.5-air & 81.33 & 32.46\, \textcolor{red}{(-2)} & 32.25\, \textcolor{red}{(-3)} & 81.96 & 67.41\, \textcolor{blue}{(+6)} & 61.71\, \textcolor{red}{(-1)} & 86.19 & 31.87\, \textcolor{red}{(-3)} & 36.51\, \textcolor{blue}{(+39)} & 86.63 & 66.96 \\
qwen3-30b-a3b & 81.09 & 34.38\, \textcolor{blue}{(+1)} & 34.21\, \textcolor{blue}{(+1)} & 81.64 & 66.93\, \textcolor{blue}{(+5)} & 61.87\, \textcolor{blue}{(+1)} & 85.39 & 34.00\, \textcolor{blue}{(+1)} & 37.63\, \textcolor{blue}{(+41)} & 85.83 & 66.57 \\
doubao-seed-1-6-flash-250715 & 80.14 & 32.82\, \textcolor{blue}{(+1)} & 32.66 & 80.78 & 66.55\, \textcolor{blue}{(+5)} & 60.14 & 84.96 & 32.42 & 38.36\, \textcolor{blue}{(+43)} & 85.39 & 64.70 \\
gpt-4.1-nano-2025-04-14 & 76.29 & 32.40 & 32.71\, \textcolor{blue}{(+2)} & 76.13 & 65.27\, \textcolor{blue}{(+2)} & 55.64 & 80.71 & 33.04\, \textcolor{blue}{(+2)} & 42.20\, \textcolor{blue}{(+52)} & 80.70 & 56.94 \\
\bottomrule
\end{tabular}}
\label{tab:meme_scores_summary_full_curated-MMLU-Redux}
\end{table}

\begin{table}[H]
\centering
\caption{\textbf{Meme Scores (Curated Population; SimpleQA).} The table reports meme scores across models, including property-derived 1D meme scores, predefined 2D meme scores (Mastery, Ingenuity, and Robustness), and a predefined 3D meme score (Caution). Models are sorted by Accuracy. {\color{blue}Blue} indicates rank improvement compared with Accuracy rank, and {\color{red}red} indicates rank degradation.}
\resizebox{\textwidth}{!}{
\begin{tabular}{cccccccccccc}
\toprule
\multirow{2}{*}{\textbf{Model}} & \multirow{2}{*}{\textbf{Accuracy}} &
\multicolumn{6}{c}{\textbf{Property-derived 1D meme scores}} &
\multicolumn{4}{c}{\textbf{Predefined 2D/3D meme scores}} \\
\cmidrule(lr){3-8}\cmidrule(lr){9-12}
 &  & \textbf{Difficulty} & \textbf{Uniqueness} & \textbf{Risk} & \textbf{Surprise} & \textbf{Typicality} & \textbf{Bridge} & \textbf{Mastery} & \textbf{Ingenuity} & \textbf{Robustness} & \textbf{Caution} \\
\midrule
gemini-2.5-pro(IR) & 57.51 & 49.56 & 81.55 & 78.56 & 46.76 & 66.12 & 42.37 & 59.87 & 65.60\, \textcolor{red}{(-1)} & 66.06 & 91.02 \\
grok-4-0709(IR) & 52.24 & 43.42 & 80.04 & 73.19 & 42.81 & 59.98 & 35.28 & 52.54 & 66.35\, \textcolor{blue}{(+1)} & 57.27 & 88.48\, \textcolor{red}{(-1)} \\
o3-2025-04-16(IR) & 50.51 & 41.41 & 78.83\, \textcolor{red}{(-1)} & 72.11 & 41.95\, \textcolor{red}{(-1)} & 59.76 & 33.96 & 52.28 & 62.99\, \textcolor{red}{(-1)} & 57.07 & 88.73\, \textcolor{blue}{(+1)} \\
gpt-5-2025-08-07(IR) & 50.44 & 41.35 & 78.96\, \textcolor{blue}{(+1)} & 71.77 & 42.34\, \textcolor{blue}{(+1)} & 58.77 & 33.46 & 51.10 & 64.83\, \textcolor{blue}{(+1)} & 55.76 & 88.17 \\
gpt-4.1-2025-04-14(CoT) & 45.08 & 35.81 & 74.87 & 65.55 & 36.54 & 52.82 & 28.88 & 44.61 & 60.10\, \textcolor{red}{(-1)} & 49.79 & 84.07 \\
grok-3(CoT) & 41.45 & 31.98 & 73.13 & 60.69 & 35.21 & 48.76 & 23.80\, \textcolor{red}{(-1)} & 40.33 & 61.22\, \textcolor{blue}{(+1)} & 41.71\, \textcolor{red}{(-1)} & 80.30 \\
gpt-4.1-2025-04-14 & 40.87 & 31.97 & 69.79 & 60.14 & 32.61 & 48.10 & 26.38\, \textcolor{blue}{(+1)} & 40.06 & 54.36\, \textcolor{red}{(-4)} & 46.14\, \textcolor{blue}{(+1)} & 78.98 \\
gpt-4o-2024-11-20(CoT) & 38.65 & 29.26 & 69.73 & 57.89 & 31.82 & 45.83 & 22.06 & 37.20 & 56.15\, \textcolor{blue}{(+1)} & 40.37 & 78.33 \\
gpt-4o-2024-11-20 & 37.54 & 28.24 & 68.58 & 56.25 & 30.97 & 44.10 & 21.27 & 35.38 & 54.63\, \textcolor{red}{(-1)} & 39.10 & 76.92 \\
kimi-k2-0711-preview(CoT) & 35.71 & 26.40 & 68.04 & 52.71 & 30.81 & 41.11 & 17.12 & 32.30 & 55.87\, \textcolor{blue}{(+2)} & 30.45 & 73.40 \\
claude-3-5-sonnet-20241022 & 33.38 & 24.87 & 62.85\, \textcolor{red}{(-3)} & 49.05\, \textcolor{red}{(-1)} & 28.76\, \textcolor{red}{(-2)} & 38.92 & 16.93 & 30.87 & 50.97\, \textcolor{red}{(-8)} & 29.96 & 68.68\, \textcolor{red}{(-2)} \\
kimi-k2-0711-preview & 33.33 & 24.25 & 65.42\, \textcolor{blue}{(+1)} & 49.19\, \textcolor{blue}{(+1)} & 29.54\, \textcolor{blue}{(+1)} & 38.13 & 15.98 & 29.43 & 55.30\, \textcolor{blue}{(+3)} & 28.69\, \textcolor{red}{(-1)} & 69.82\, \textcolor{blue}{(+1)} \\
claude-3-5-sonnet-20241022(CoT) & 31.76 & 22.96 & 62.23\, \textcolor{red}{(-3)} & 48.06 & 25.41\, \textcolor{red}{(-9)} & 36.68 & 15.74 & 28.03 & 49.39\, \textcolor{red}{(-9)} & 28.77\, \textcolor{blue}{(+1)} & 69.11\, \textcolor{blue}{(+1)} \\
gemini-2.5-flash(IR) & 31.02 & 21.93 & 63.47\, \textcolor{blue}{(+2)} & 46.57 & 28.24 & 35.11 & 13.41\, \textcolor{red}{(-2)} & 26.36 & 53.62\, \textcolor{blue}{(+1)} & 24.89 & 66.75 \\
gemini-2.5-flash(CoT) & 30.74 & 21.88 & 62.67 & 45.57 & 28.86\, \textcolor{blue}{(+3)} & 34.67 & 13.61 & 26.15 & 53.71\, \textcolor{blue}{(+3)} & 24.51\, \textcolor{red}{(-1)} & 65.30 \\
deepseek-R1(IR) & 29.03 & 19.62\, \textcolor{red}{(-1)} & 63.20\, \textcolor{blue}{(+3)} & 44.14 & 26.28 & 31.60 & 10.30\, \textcolor{red}{(-1)} & 22.28\, \textcolor{red}{(-1)} & 53.04\, \textcolor{blue}{(+1)} & 20.03\, \textcolor{red}{(-1)} & 65.12 \\
deepseek-V3(CoT) & 28.32 & 19.12\, \textcolor{red}{(-1)} & 62.12 & 42.36\, \textcolor{red}{(-1)} & 26.53\, \textcolor{blue}{(+2)} & 30.95\, \textcolor{red}{(-1)} & 9.60\, \textcolor{red}{(-3)} & 21.85\, \textcolor{red}{(-1)} & 53.05\, \textcolor{blue}{(+3)} & 18.03\, \textcolor{red}{(-3)} & 63.27 \\
glm-4.5(IR) & 28.16 & 19.02\, \textcolor{red}{(-1)} & 61.55 & 42.38\, \textcolor{blue}{(+1)} & 25.68\, \textcolor{red}{(-1)} & 30.85\, \textcolor{red}{(-1)} & 10.19 & 21.80\, \textcolor{red}{(-1)} & 52.11\, \textcolor{blue}{(+2)} & 19.16 & 62.91 \\
grok-3 & 27.79 & 20.42\, \textcolor{blue}{(+3)} & 54.42\, \textcolor{red}{(-4)} & 40.05\, \textcolor{red}{(-1)} & 26.27\, \textcolor{blue}{(+2)} & 31.55\, \textcolor{blue}{(+2)} & 14.15\, \textcolor{blue}{(+5)} & 24.43\, \textcolor{blue}{(+3)} & 48.15\, \textcolor{red}{(-5)} & 24.79\, \textcolor{blue}{(+4)} & 56.88\, \textcolor{red}{(-3)} \\
glm-4.5(CoT) & 26.68 & 17.69 & 59.83\, \textcolor{blue}{(+1)} & 40.19\, \textcolor{blue}{(+1)} & 24.89\, \textcolor{red}{(-3)} & 29.71 & 9.20\, \textcolor{red}{(-2)} & 20.70 & 52.02\, \textcolor{blue}{(+3)} & 17.86\, \textcolor{red}{(-1)} & 61.47\, \textcolor{blue}{(+1)} \\
glm-4.5 & 25.77 & 17.35 & 57.48\, \textcolor{blue}{(+1)} & 37.64 & 25.57 & 27.99 & 9.23 & 19.47 & 51.92\, \textcolor{blue}{(+3)} & 16.94\, \textcolor{red}{(-1)} & 57.35 \\
deepseek-V3 & 24.64 & 16.51 & 55.27 & 36.13\, \textcolor{red}{(-1)} & 24.50\, \textcolor{red}{(-2)} & 26.83 & 8.87\, \textcolor{red}{(-1)} & 18.71\, \textcolor{red}{(-1)} & 48.90\, \textcolor{red}{(-1)} & 16.27\, \textcolor{red}{(-1)} & 54.89\, \textcolor{red}{(-2)} \\
claude-sonnet-4-20250514(IR) & 23.74 & 15.15\, \textcolor{red}{(-1)} & 55.51\, \textcolor{blue}{(+2)} & 36.82\, \textcolor{blue}{(+1)} & 21.49\, \textcolor{red}{(-11)} & 26.17\, \textcolor{red}{(-1)} & 7.92\, \textcolor{red}{(-1)} & 17.30\, \textcolor{red}{(-1)} & 46.48\, \textcolor{red}{(-7)} & 16.10\, \textcolor{red}{(-1)} & 58.09\, \textcolor{blue}{(+3)} \\
gemini-2.5-flash & 23.65 & 16.24\, \textcolor{blue}{(+1)} & 51.36\, \textcolor{red}{(-5)} & 34.57\, \textcolor{red}{(-1)} & 23.11\, \textcolor{red}{(-5)} & 26.44\, \textcolor{blue}{(+1)} & 10.00\, \textcolor{blue}{(+5)} & 18.93\, \textcolor{blue}{(+2)} & 45.23\, \textcolor{red}{(-8)} & 18.37\, \textcolor{blue}{(+5)} & 52.43\, \textcolor{red}{(-3)} \\
MiniMax-Text-01(CoT) & 22.75 & 14.72\, \textcolor{red}{(-1)} & 53.01 & 34.17\, \textcolor{red}{(-1)} & 22.73\, \textcolor{red}{(-5)} & 24.72 & 7.85 & 16.46 & 46.52\, \textcolor{red}{(-2)} & 14.79\, \textcolor{red}{(-2)} & 53.93 \\
gpt-5-mini-2025-08-07(IR) & 22.54 & 14.97\, \textcolor{blue}{(+1)} & 52.20\, \textcolor{red}{(-1)} & 31.89\, \textcolor{red}{(-2)} & 25.86\, \textcolor{blue}{(+8)} & 23.70\, \textcolor{red}{(-1)} & 7.50 & 16.39 & 50.51\, \textcolor{blue}{(+6)} & 12.62\, \textcolor{red}{(-4)} & 46.91\, \textcolor{red}{(-5)} \\
claude-sonnet-4-20250514(CoT) & 22.45 & 14.08 & 53.42\, \textcolor{blue}{(+3)} & 35.03\, \textcolor{blue}{(+3)} & 20.37\, \textcolor{red}{(-11)} & 24.62\, \textcolor{blue}{(+1)} & 7.02\, \textcolor{red}{(-2)} & 15.99 & 44.61\, \textcolor{red}{(-7)} & 14.95\, \textcolor{blue}{(+2)} & 55.58\, \textcolor{blue}{(+4)} \\
claude-sonnet-4-20250514 & 21.61 & 13.50\, \textcolor{red}{(-1)} & 52.40\, \textcolor{blue}{(+2)} & 32.85\, \textcolor{blue}{(+1)} & 21.84\, \textcolor{red}{(-4)} & 23.28 & 7.28 & 14.80\, \textcolor{red}{(-1)} & 46.50\, \textcolor{red}{(-1)} & 14.90\, \textcolor{blue}{(+2)} & 52.99\, \textcolor{blue}{(+2)} \\
MiniMax-Text-01 & 21.13 & 13.69\, \textcolor{blue}{(+1)} & 49.68\, \textcolor{red}{(-2)} & 31.04 & 22.65\, \textcolor{red}{(-2)} & 22.89 & 7.29\, \textcolor{blue}{(+2)} & 15.21\, \textcolor{blue}{(+1)} & 46.34\, \textcolor{red}{(-2)} & 13.53\, \textcolor{blue}{(+1)} & 49.28\, \textcolor{blue}{(+1)} \\
doubao-seed-1-6-250615(IR) & 21.10 & 13.23 & 51.85\, \textcolor{blue}{(+2)} & 30.89 & 24.40\, \textcolor{blue}{(+5)} & 22.22\, \textcolor{red}{(-1)} & 6.57\, \textcolor{red}{(-2)} & 14.20\, \textcolor{red}{(-1)} & 49.79\, \textcolor{blue}{(+9)} & 12.53\, \textcolor{red}{(-1)} & 48.92\, \textcolor{blue}{(+1)} \\
MiniMax-M1(IR) & 20.60 & 13.10 & 49.73\, \textcolor{blue}{(+1)} & 30.14 & 23.15\, \textcolor{blue}{(+3)} & 22.24\, \textcolor{blue}{(+1)} & 6.76 & 14.53\, \textcolor{blue}{(+1)} & 47.20\, \textcolor{blue}{(+6)} & 12.81\, \textcolor{blue}{(+2)} & 48.19\, \textcolor{blue}{(+1)} \\
gpt-4.1-mini-2025-04-14(CoT) & 19.30 & 12.68 & 45.96\, \textcolor{red}{(-1)} & 26.58\, \textcolor{red}{(-1)} & 25.60\, \textcolor{blue}{(+12)} & 20.18 & 6.81\, \textcolor{blue}{(+2)} & 13.70 & 47.19\, \textcolor{blue}{(+6)} & 10.29\, \textcolor{red}{(-1)} & 40.24\, \textcolor{red}{(-1)} \\
doubao-seed-1-6-250615(CoT) & 18.77 & 11.22 & 48.32\, \textcolor{blue}{(+1)} & 27.98\, \textcolor{blue}{(+1)} & 21.55 & 19.79 & 5.44\, \textcolor{red}{(-2)} & 11.91 & 46.51\, \textcolor{blue}{(+5)} & 10.76\, \textcolor{blue}{(+1)} & 46.06\, \textcolor{blue}{(+1)} \\
gpt-4.1-mini-2025-04-14 & 16.53 & 10.63 & 40.54\, \textcolor{red}{(-3)} & 22.59 & 23.48\, \textcolor{blue}{(+7)} & 17.33 & 5.98\, \textcolor{blue}{(+1)} & 11.32 & 44.05\, \textcolor{red}{(-1)} & 9.43 & 35.69\, \textcolor{red}{(-2)} \\
o3-mini-2025-01-31(IR) & 15.95 & 10.18 & 39.79\, \textcolor{red}{(-3)} & 21.32\, \textcolor{red}{(-3)} & 23.88\, \textcolor{blue}{(+9)} & 16.12\, \textcolor{red}{(-1)} & 5.48\, \textcolor{blue}{(+1)} & 10.54 & 45.09\, \textcolor{blue}{(+2)} & 7.97\, \textcolor{red}{(-2)} & 31.63\, \textcolor{red}{(-3)} \\
glm-4.5-air(CoT) & 15.88 & 9.36 & 42.46\, \textcolor{blue}{(+2)} & 22.48\, \textcolor{blue}{(+1)} & 21.25\, \textcolor{blue}{(+1)} & 16.37\, \textcolor{blue}{(+1)} & 4.89 & 9.59 & 43.74 & 8.87 & 37.54\, \textcolor{blue}{(+1)} \\
doubao-seed-1-6-250615 & 15.14 & 8.69 & 41.03\, \textcolor{blue}{(+2)} & 22.22\, \textcolor{blue}{(+1)} & 18.93\, \textcolor{red}{(-5)} & 16.06 & 4.63 & 9.23 & 40.62\, \textcolor{red}{(-2)} & 9.14\, \textcolor{blue}{(+2)} & 38.15\, \textcolor{blue}{(+3)} \\
glm-4.5-air(IR) & 14.86 & 8.52 & 40.65\, \textcolor{blue}{(+2)} & 21.35\, \textcolor{blue}{(+1)} & 19.31\, \textcolor{red}{(-3)} & 15.05 & 4.34\, \textcolor{red}{(-2)} & 8.55 & 41.70\, \textcolor{blue}{(+1)} & 7.77\, \textcolor{red}{(-1)} & 35.26\, \textcolor{blue}{(+1)} \\
qwen3-235b-a22b(IR) & 13.27 & 7.72\, \textcolor{red}{(-1)} & 36.54 & 18.06\, \textcolor{red}{(-1)} & 20.39\, \textcolor{blue}{(+2)} & 13.48\, \textcolor{red}{(-1)} & 4.50 & 7.93 & 41.11\, \textcolor{blue}{(+1)} & 7.21\, \textcolor{red}{(-1)} & 29.58\, \textcolor{red}{(-1)} \\
glm-4.5-air & 13.20 & 7.70\, \textcolor{red}{(-1)} & 36.03 & 18.20\, \textcolor{blue}{(+1)} & 19.73 & 13.66\, \textcolor{blue}{(+1)} & 4.59\, \textcolor{blue}{(+2)} & 7.79 & 39.11\, \textcolor{red}{(-1)} & 7.89\, \textcolor{blue}{(+2)} & 31.56\, \textcolor{blue}{(+1)} \\
qwen3-235b-a22b(CoT) & 13.15 & 7.76\, \textcolor{blue}{(+2)} & 35.69 & 17.98 & 20.43\, \textcolor{blue}{(+5)} & 13.12 & 4.24 & 7.71 & 37.96\, \textcolor{red}{(-1)} & 6.78 & 29.27 \\
gpt-5-nano-2025-08-07(IR) & 11.88 & 6.73 & 33.89 & 15.63 & 19.81\, \textcolor{blue}{(+3)} & 11.96 & 3.77\, \textcolor{red}{(-1)} & 6.97 & 40.27\, \textcolor{blue}{(+2)} & 5.52\, \textcolor{red}{(-1)} & 25.06\, \textcolor{red}{(-1)} \\
qwen3-235b-a22b & 11.21 & 6.32 & 32.13 & 14.79 & 18.63\, \textcolor{red}{(-1)} & 11.69 & 3.79\, \textcolor{blue}{(+1)} & 6.58 & 37.03\, \textcolor{red}{(-1)} & 6.26\, \textcolor{blue}{(+1)} & 26.07\, \textcolor{blue}{(+1)} \\
gpt-4.1-nano-2025-04-14(CoT) & 10.06 & 5.42 & 30.39 & 12.67 & 17.85\, \textcolor{red}{(-1)} & 9.95 & 3.30\, \textcolor{red}{(-1)} & 5.25 & 37.83\, \textcolor{blue}{(+1)} & 5.06 & 22.09 \\
gpt-4.1-nano-2025-04-14 & 8.69 & 4.87 & 25.86 & 10.27 & 18.87\, \textcolor{blue}{(+2)} & 8.94 & 3.42\, \textcolor{blue}{(+1)} & 5.11 & 34.85 & 4.68 & 17.97 \\
spark-X1(IR) & 8.07 & 4.32 & 24.51 & 10.09 & 14.24\, \textcolor{red}{(-2)} & 8.08 & 2.89 & 4.19 & 30.35\, \textcolor{red}{(-1)} & 4.16 & 17.66 \\
doubao-seed-1-6-flash-250715(CoT) & 7.47 & 3.90\, \textcolor{red}{(-2)} & 23.46 & 8.85 & 13.83\, \textcolor{red}{(-2)} & 7.45 & 2.25\, \textcolor{red}{(-3)} & 3.74\, \textcolor{red}{(-2)} & 31.20\, \textcolor{blue}{(+1)} & 3.07\, \textcolor{red}{(-1)} & 15.58 \\
doubao-seed-1-6-flash-250715(IR) & 7.14 & 4.06 & 20.94\, \textcolor{red}{(-1)} & 8.40 & 14.69\, \textcolor{blue}{(+1)} & 6.93 & 2.31 & 3.97 & 27.41\, \textcolor{red}{(-3)} & 2.49\, \textcolor{red}{(-1)} & 13.11\, \textcolor{red}{(-1)} \\
qwen3-32b & 6.98 & 4.09\, \textcolor{blue}{(+2)} & 20.18\, \textcolor{red}{(-2)} & 7.72\, \textcolor{red}{(-1)} & 15.00\, \textcolor{blue}{(+3)} & 6.76\, \textcolor{red}{(-1)} & 2.25 & 4.06\, \textcolor{blue}{(+2)} & 29.73\, \textcolor{blue}{(+1)} & 2.44\, \textcolor{red}{(-1)} & 11.68\, \textcolor{red}{(-3)} \\
qwen3-32b(CoT) & 6.87 & 3.62 & 21.44\, \textcolor{blue}{(+2)} & 8.05\, \textcolor{blue}{(+1)} & 13.21\, \textcolor{red}{(-1)} & 6.81\, \textcolor{blue}{(+1)} & 2.41\, \textcolor{blue}{(+3)} & 3.56 & 29.23\, \textcolor{blue}{(+1)} & 3.17\, \textcolor{blue}{(+3)} & 13.62\, \textcolor{blue}{(+2)} \\
qwen3-30b-a3b(CoT) & 6.56 & 3.49 & 20.36\, \textcolor{blue}{(+1)} & 7.71 & 13.44\, \textcolor{blue}{(+1)} & 6.43 & 2.05 & 3.36 & 28.37\, \textcolor{blue}{(+1)} & 2.38 & 12.71\, \textcolor{blue}{(+1)} \\
doubao-seed-1-6-flash-250715 & 6.01 & 3.05\, \textcolor{red}{(-1)} & 19.41 & 6.91 & 12.33\, \textcolor{red}{(-1)} & 5.94 & 1.95 & 2.92\, \textcolor{red}{(-1)} & 27.21 & 2.33 & 12.06\, \textcolor{blue}{(+1)} \\
qwen3-30b-a3b & 5.96 & 3.29\, \textcolor{blue}{(+1)} & 18.10 & 6.72 & 12.54\, \textcolor{blue}{(+1)} & 5.73 & 1.84 & 3.07\, \textcolor{blue}{(+1)} & 25.69 & 2.13 & 10.80 \\
\bottomrule
\end{tabular}}
\label{tab:meme_scores_summary_full_curated-SimpleQA}
\end{table}

\subsection{Meme Score Results on the Open LLM Population}
\label{app:MemeScore_Leaderboard_openllm}
Table~\ref{tab:meme_scores_summary_OpenLLM} reports Meme Scores for the top-50 models in the Open LLM population, ranked by overall accuracy; Tables~\ref{tab:meme_scores_summary_OpenLLM-BBH}, \ref{tab:meme_scores_summary_OpenLLM-GPQA-Diamond}, \ref{tab:meme_scores_summary_OpenLLM-IFEval}, \ref{tab:meme_scores_summary_OpenLLM-MATH}, \ref{tab:meme_scores_summary_OpenLLM-MMLU-Pro}, and \ref{tab:meme_scores_summary_OpenLLM-MUSR} report the corresponding top-50 results computed on BBH, GPQA-Diamond, IFEval, MATH, MMLU-Pro, and MUSR, respectively.

As shown in Table~\ref{tab:meme_scores_summary_OpenLLM-GPQA-Diamond}, Meme Scores induced by the cluster-based MPPs (\emph{typicality} and \emph{bridge}), exhibit a large number of zero values for several MSs, including \emph{Typicality}, \emph{Bridge}, \emph{Mastery}, \emph{Robustness}, and \emph{Caution}.
This phenomenon arises from a structural mismatch between the model population and the dataset: while the Open LLM population contains 4,479 models, GPQA-Diamond consists of only 198 items. As a result, the probe–probe similarity matrix $S$ becomes sparse after threshold $\tau$ pruning, leading to only a small number of clusters when grouping items by perception spans. Models that do not perform well on these few items exhibiting specific behavioral patterns thus receive zero Meme Scores.
This observation suggests that, when applying the Probing Memes paradigm, the choice of Meme Scores should take dataset characteristics into account, and cluster-based Meme Scores may be relatively less informative for small or sparse datasets.

\begin{table}[H]
\centering
\caption{\textbf{Meme Scores (Open LLM Population; Averaged over BBH, GPQA-Diamond, IFEVal, MATH, MMLU-Pro, and MUSR).} The table reports meme scores across models, including property-derived 1D meme scores, predefined 2D meme scores (Mastery, Ingenuity, and Robustness), and a predefined 3D meme score (Caution). Models are sorted by Accuracy, and only the \textbf{top-50 models} are shown. {\color{blue}Blue} indicates rank improvement compared with Accuracy rank, and {\color{red}red} indicates rank degradation.}
\resizebox{\textwidth}{!}{
\begin{tabular}{cccccccccccc}
\toprule
\multirow{2}{*}{\textbf{Model}} & \multirow{2}{*}{\textbf{Accuracy}} &
\multicolumn{6}{c}{\textbf{Property-derived 1D meme scores}} &
\multicolumn{4}{c}{\textbf{Predefined 2D/3D meme scores}} \\
\cmidrule(lr){3-8}\cmidrule(lr){9-12}
 &  & \textbf{Difficulty} & \textbf{Uniqueness} & \textbf{Risk} & \textbf{Surprise} & \textbf{Typicality} & \textbf{Bridge} & \textbf{Mastery} & \textbf{Ingenuity} & \textbf{Robustness} & \textbf{Caution} \\
\midrule
calme-3.2-instruct-78b & 60.08 & 50.48 & 71.01\, \textcolor{red}{(-1)} & 73.95\, \textcolor{red}{(-1)} & 39.04 & 56.83\, \textcolor{red}{(-4)} & 36.36\, \textcolor{red}{(-1)} & 48.00\, \textcolor{red}{(-1)} & 53.18 & 54.15\, \textcolor{red}{(-4)} & 73.61\, \textcolor{red}{(-28)} \\
CalmeRys-78B-Orpo-v0.1 & 59.77 & 50.16 & 70.49\, \textcolor{red}{(-3)} & 73.42\, \textcolor{red}{(-2)} & 38.51 & 56.50\, \textcolor{red}{(-6)} & 34.64\, \textcolor{red}{(-4)} & 47.89\, \textcolor{red}{(-1)} & 52.74 & 53.44\, \textcolor{red}{(-11)} & 73.02\, \textcolor{red}{(-48)} \\
calme-3.1-instruct-78b & 59.30 & 49.63 & 70.47\, \textcolor{red}{(-3)} & 73.36\, \textcolor{red}{(-3)} & 38.34 & 56.58\, \textcolor{red}{(-3)} & 36.10 & 47.58\, \textcolor{red}{(-1)} & 52.65 & 53.88\, \textcolor{red}{(-5)} & 73.90\, \textcolor{red}{(-23)} \\
calme-2.4-rys-78b & 59.30 & 49.42 & 70.30\, \textcolor{red}{(-4)} & 73.09\, \textcolor{red}{(-3)} & 37.79 & 55.73\, \textcolor{red}{(-8)} & 34.25\, \textcolor{red}{(-4)} & 46.83\, \textcolor{red}{(-1)} & 52.46 & 52.68\, \textcolor{red}{(-16)} & 72.64\, \textcolor{red}{(-66)} \\
FluentlyLM-Prinum & 57.88 & 46.45 & 70.59\, \textcolor{blue}{(+1)} & 73.70\, \textcolor{blue}{(+2)} & 34.30\, \textcolor{red}{(-3)} & 59.05\, \textcolor{blue}{(+4)} & 34.02\, \textcolor{red}{(-4)} & 48.36\, \textcolor{blue}{(+4)} & 50.57\, \textcolor{red}{(-2)} & 54.59\, \textcolor{blue}{(+1)} & 85.21\, \textcolor{blue}{(+4)} \\
Homer-v1.0-Qwen2.5-72B & 57.63 & 46.23 & 71.04\, \textcolor{blue}{(+5)} & 74.22\, \textcolor{blue}{(+5)} & 34.38\, \textcolor{red}{(-1)} & 54.19\, \textcolor{red}{(-12)} & 29.07\, \textcolor{red}{(-95)} & 43.72\, \textcolor{red}{(-11)} & 50.56\, \textcolor{red}{(-2)} & 45.37\, \textcolor{red}{(-214)} & 74.48\, \textcolor{red}{(-16)} \\
ultiima-72B & 56.81 & 45.26\, \textcolor{red}{(-3)} & 70.43 & 73.41\, \textcolor{blue}{(+2)} & 33.50\, \textcolor{red}{(-6)} & 53.54\, \textcolor{red}{(-13)} & 30.17\, \textcolor{red}{(-41)} & 43.01\, \textcolor{red}{(-13)} & 49.92\, \textcolor{red}{(-2)} & 47.32\, \textcolor{red}{(-119)} & 73.80\, \textcolor{red}{(-20)} \\
Gilgamesh-72B & 56.71 & 46.13\, \textcolor{blue}{(+1)} & 69.14\, \textcolor{red}{(-2)} & 71.37\, \textcolor{red}{(-14)} & 35.00\, \textcolor{blue}{(+2)} & 54.25\, \textcolor{red}{(-9)} & 30.55\, \textcolor{red}{(-32)} & 43.93\, \textcolor{red}{(-8)} & 50.74\, \textcolor{blue}{(+3)} & 48.13\, \textcolor{red}{(-74)} & 74.00\, \textcolor{red}{(-16)} \\
shuttle-3 & 56.36 & 44.43\, \textcolor{red}{(-3)} & 70.77\, \textcolor{blue}{(+6)} & 72.49\, \textcolor{blue}{(+1)} & 32.77\, \textcolor{red}{(-8)} & 53.08\, \textcolor{red}{(-12)} & 28.57\, \textcolor{red}{(-135)} & 42.33\, \textcolor{red}{(-14)} & 50.71\, \textcolor{blue}{(+3)} & 46.56\, \textcolor{red}{(-159)} & 73.56\, \textcolor{red}{(-22)} \\
T3Q-Qwen2.5-14B-Instruct-1M-e3 & 56.03 & 45.89\, \textcolor{blue}{(+2)} & 67.19\, \textcolor{red}{(-21)} & 71.60\, \textcolor{red}{(-7)} & 33.65\, \textcolor{red}{(-1)} & 52.05\, \textcolor{red}{(-32)} & 28.69\, \textcolor{red}{(-124)} & 42.77\, \textcolor{red}{(-11)} & 47.22\, \textcolor{red}{(-35)} & 44.63\, \textcolor{red}{(-263)} & 71.58\, \textcolor{red}{(-129)} \\
T3Q-qwen2.5-14b-v1.0-e3 & 56.03 & 45.89\, \textcolor{blue}{(+3)} & 67.19\, \textcolor{red}{(-20)} & 71.60\, \textcolor{red}{(-6)} & 33.65 & 52.05\, \textcolor{red}{(-31)} & 28.69\, \textcolor{red}{(-123)} & 42.77\, \textcolor{red}{(-10)} & 47.22\, \textcolor{red}{(-34)} & 44.63\, \textcolor{red}{(-262)} & 71.58\, \textcolor{red}{(-128)} \\
test-2.5-72B & 55.82 & 45.10\, \textcolor{blue}{(+1)} & 67.85\, \textcolor{red}{(-1)} & 70.81\, \textcolor{red}{(-28)} & 33.74\, \textcolor{blue}{(+2)} & 51.46\, \textcolor{red}{(-41)} & 28.25\, \textcolor{red}{(-161)} & 41.56\, \textcolor{red}{(-21)} & 48.77\, \textcolor{red}{(-3)} & 44.81\, \textcolor{red}{(-245)} & 70.91\, \textcolor{red}{(-181)} \\
Qwen2.5-72B-Instruct-abliterated & 55.10 & 43.65\, \textcolor{red}{(-1)} & 68.81\, \textcolor{blue}{(+2)} & 71.22\, \textcolor{red}{(-11)} & 32.86\, \textcolor{red}{(-3)} & 51.70\, \textcolor{red}{(-36)} & 28.74\, \textcolor{red}{(-117)} & 40.85\, \textcolor{red}{(-35)} & 49.11\, \textcolor{blue}{(+1)} & 44.62\, \textcolor{red}{(-262)} & 72.70\, \textcolor{red}{(-55)} \\
RYS-XLarge & 55.00 & 43.62\, \textcolor{red}{(-1)} & 67.58\, \textcolor{red}{(-7)} & 70.88\, \textcolor{red}{(-25)} & 31.89\, \textcolor{red}{(-9)} & 52.47\, \textcolor{red}{(-19)} & 28.83\, \textcolor{red}{(-108)} & 42.04\, \textcolor{red}{(-11)} & 47.20\, \textcolor{red}{(-33)} & 47.00\, \textcolor{red}{(-131)} & 73.00\, \textcolor{red}{(-37)} \\
calme-2.1-rys-78b & 54.96 & 43.53\, \textcolor{red}{(-1)} & 67.98\, \textcolor{blue}{(+3)} & 70.11\, \textcolor{red}{(-46)} & 33.23\, \textcolor{blue}{(+1)} & 54.00\, \textcolor{red}{(-4)} & 32.60\, \textcolor{blue}{(+2)} & 43.06\, \textcolor{red}{(-4)} & 48.85\, \textcolor{blue}{(+2)} & 50.11\, \textcolor{red}{(-27)} & 74.27\, \textcolor{red}{(-8)} \\
sky-t1-coder-32b-flash & 54.95 & 43.93\, \textcolor{blue}{(+3)} & 67.63\, \textcolor{red}{(-4)} & 69.47\, \textcolor{red}{(-68)} & 33.74\, \textcolor{blue}{(+7)} & 55.14\, \textcolor{blue}{(+2)} & 34.83\, \textcolor{blue}{(+11)} & 44.53\, \textcolor{blue}{(+3)} & 49.69\, \textcolor{blue}{(+6)} & 53.37\, \textcolor{blue}{(+2)} & 72.62\, \textcolor{red}{(-57)} \\
tempmotacilla-cinerea-0308 & 54.86 & 43.04\, \textcolor{red}{(-1)} & 67.16\, \textcolor{red}{(-16)} & 72.24\, \textcolor{blue}{(+8)} & 29.90\, \textcolor{red}{(-69)} & 52.76\, \textcolor{red}{(-6)} & 30.19\, \textcolor{red}{(-30)} & 42.04\, \textcolor{red}{(-9)} & 45.29\, \textcolor{red}{(-126)} & 50.72\, \textcolor{red}{(-16)} & 72.90\, \textcolor{red}{(-38)} \\
ultiima-72B-v1.5 & 54.62 & 42.90\, \textcolor{red}{(-3)} & 69.53\, \textcolor{blue}{(+9)} & 71.13\, \textcolor{red}{(-14)} & 32.19\, \textcolor{red}{(-3)} & 51.37\, \textcolor{red}{(-37)} & 28.63\, \textcolor{red}{(-122)} & 40.52\, \textcolor{red}{(-33)} & 49.54\, \textcolor{blue}{(+7)} & 44.90\, \textcolor{red}{(-236)} & 72.52\, \textcolor{red}{(-65)} \\
Rombos-LLM-V2.5-Qwen-72b & 54.59 & 43.29\, \textcolor{blue}{(+2)} & 67.69\, \textcolor{blue}{(+2)} & 70.38\, \textcolor{red}{(-33)} & 32.68\, \textcolor{blue}{(+1)} & 52.30\, \textcolor{red}{(-19)} & 29.32\, \textcolor{red}{(-66)} & 41.85\, \textcolor{red}{(-10)} & 48.19\, \textcolor{red}{(-4)} & 45.90\, \textcolor{red}{(-176)} & 72.54\, \textcolor{red}{(-61)} \\
li-14b-v0.4 & 54.40 & 42.46\, \textcolor{red}{(-8)} & 67.37\, \textcolor{red}{(-8)} & 70.42\, \textcolor{red}{(-30)} & 30.46\, \textcolor{red}{(-34)} & 57.55\, \textcolor{blue}{(+17)} & 32.00\, \textcolor{blue}{(+3)} & 46.28\, \textcolor{blue}{(+13)} & 46.71\, \textcolor{red}{(-54)} & 53.79\, \textcolor{blue}{(+11)} & 84.47\, \textcolor{blue}{(+18)} \\
calme-2.2-rys-78b & 54.38 & 42.95\, \textcolor{blue}{(+1)} & 67.41\, \textcolor{red}{(-5)} & 69.71\, \textcolor{red}{(-54)} & 32.67\, \textcolor{blue}{(+2)} & 50.61\, \textcolor{red}{(-58)} & 29.66\, \textcolor{red}{(-50)} & 39.57\, \textcolor{red}{(-63)} & 48.43\, \textcolor{blue}{(+2)} & 46.85\, \textcolor{red}{(-133)} & 71.16\, \textcolor{red}{(-158)} \\
Cheng-2 & 54.37 & 42.29\, \textcolor{red}{(-14)} & 67.77\, \textcolor{blue}{(+7)} & 70.46\, \textcolor{red}{(-26)} & 30.66\, \textcolor{red}{(-22)} & 52.41\, \textcolor{red}{(-12)} & 29.87\, \textcolor{red}{(-37)} & 41.21\, \textcolor{red}{(-19)} & 47.40\, \textcolor{red}{(-17)} & 50.13\, \textcolor{red}{(-18)} & 72.69\, \textcolor{red}{(-47)} \\
calme-2.2-qwen2-72b & 54.31 & 42.53\, \textcolor{red}{(-4)} & 67.40\, \textcolor{red}{(-4)} & 70.18\, \textcolor{red}{(-35)} & 31.24\, \textcolor{red}{(-6)} & 51.52\, \textcolor{red}{(-27)} & 29.87\, \textcolor{red}{(-37)} & 40.48\, \textcolor{red}{(-29)} & 46.81\, \textcolor{red}{(-44)} & 47.97\, \textcolor{red}{(-68)} & 72.37\, \textcolor{red}{(-66)} \\
Qwentile2.5-32B-Instruct & 54.25 & 42.86\, \textcolor{blue}{(+1)} & 67.11\, \textcolor{red}{(-13)} & 69.49\, \textcolor{red}{(-59)} & 31.80\, \textcolor{red}{(-1)} & 54.99\, \textcolor{blue}{(+9)} & 33.73\, \textcolor{blue}{(+14)} & 44.24\, \textcolor{blue}{(+9)} & 47.75\, \textcolor{red}{(-6)} & 55.74\, \textcolor{blue}{(+22)} & 82.04\, \textcolor{blue}{(+14)} \\
calme-2.1-qwen2-72b & 54.22 & 42.63 & 66.98\, \textcolor{red}{(-16)} & 70.38\, \textcolor{red}{(-26)} & 31.07\, \textcolor{red}{(-10)} & 51.02\, \textcolor{red}{(-39)} & 28.93\, \textcolor{red}{(-88)} & 40.32\, \textcolor{red}{(-33)} & 46.04\, \textcolor{red}{(-84)} & 46.44\, \textcolor{red}{(-151)} & 72.07\, \textcolor{red}{(-77)} \\
calme-2.3-rys-78b & 54.22 & 42.84\, \textcolor{blue}{(+2)} & 67.34\, \textcolor{red}{(-3)} & 69.29\, \textcolor{red}{(-67)} & 33.22\, \textcolor{blue}{(+11)} & 52.52\, \textcolor{red}{(-5)} & 34.41\, \textcolor{blue}{(+19)} & 41.74\, \textcolor{red}{(-4)} & 48.47\, \textcolor{blue}{(+8)} & 50.20\, \textcolor{red}{(-12)} & 71.36\, \textcolor{red}{(-136)} \\
ZYH-LLM-Qwen2.5-14B-V4 & 54.07 & 42.34\, \textcolor{red}{(-5)} & 67.13\, \textcolor{red}{(-9)} & 69.47\, \textcolor{red}{(-58)} & 30.48\, \textcolor{red}{(-26)} & 57.78\, \textcolor{blue}{(+25)} & 31.77\, \textcolor{blue}{(+5)} & 46.67\, \textcolor{blue}{(+21)} & 47.03\, \textcolor{red}{(-31)} & 53.98\, \textcolor{blue}{(+21)} & 84.42\, \textcolor{blue}{(+24)} \\
EVA-Qwen2.5-72B-v0.2 & 54.03 & 42.33\, \textcolor{red}{(-5)} & 67.68\, \textcolor{blue}{(+10)} & 69.38\, \textcolor{red}{(-59)} & 32.23\, \textcolor{blue}{(+8)} & 50.39\, \textcolor{red}{(-65)} & 28.23\, \textcolor{red}{(-147)} & 39.60\, \textcolor{red}{(-53)} & 48.75\, \textcolor{blue}{(+12)} & 45.50\, \textcolor{red}{(-185)} & 71.07\, \textcolor{red}{(-155)} \\
Cheng-2-v1.1 & 54.02 & 41.91\, \textcolor{red}{(-11)} & 67.45\, \textcolor{blue}{(+4)} & 70.16\, \textcolor{red}{(-30)} & 30.56\, \textcolor{red}{(-19)} & 52.30\, \textcolor{red}{(-10)} & 29.38\, \textcolor{red}{(-52)} & 41.09\, \textcolor{red}{(-13)} & 47.17\, \textcolor{red}{(-21)} & 48.75\, \textcolor{red}{(-40)} & 72.55\, \textcolor{red}{(-48)} \\
Qwen2.5-14B-1M-YOYO-V3 & 54.00 & 41.90\, \textcolor{red}{(-11)} & 67.65\, \textcolor{blue}{(+11)} & 69.86\, \textcolor{red}{(-41)} & 30.18\, \textcolor{red}{(-39)} & 52.75\, \textcolor{blue}{(+5)} & 30.97\, \textcolor{blue}{(+1)} & 41.54\, \textcolor{red}{(-4)} & 47.51\, \textcolor{red}{(-4)} & 51.26\, \textcolor{blue}{(+4)} & 73.23\, \textcolor{red}{(-11)} \\
virtuoso-small-v2-tensopolis-v1 & 53.97 & 42.33\, \textcolor{red}{(-3)} & 66.20\, \textcolor{red}{(-43)} & 71.37\, \textcolor{blue}{(+8)} & 29.26\, \textcolor{red}{(-96)} & 52.56\, \textcolor{blue}{(+2)} & 27.97\, \textcolor{red}{(-169)} & 41.38\, \textcolor{red}{(-7)} & 43.68\, \textcolor{red}{(-214)} & 47.04\, \textcolor{red}{(-113)} & 74.60\, \textcolor{blue}{(+10)} \\
miscii-14b-0218 & 53.93 & 41.77\, \textcolor{red}{(-15)} & 66.45\, \textcolor{red}{(-29)} & 71.86\, \textcolor{blue}{(+22)} & 29.10\, \textcolor{red}{(-108)} & 51.86\, \textcolor{red}{(-13)} & 30.88 & 40.65\, \textcolor{red}{(-17)} & 44.17\, \textcolor{red}{(-185)} & 53.47\, \textcolor{blue}{(+20)} & 73.28\, \textcolor{red}{(-7)} \\
Lix-14B-v0.1 & 53.92 & 41.86\, \textcolor{red}{(-10)} & 66.51\, \textcolor{red}{(-23)} & 71.20\, \textcolor{blue}{(+6)} & 28.77\, \textcolor{red}{(-127)} & 52.65\, \textcolor{blue}{(+6)} & 29.95\, \textcolor{red}{(-23)} & 41.57\, \textcolor{blue}{(+1)} & 44.67\, \textcolor{red}{(-144)} & 49.79\, \textcolor{red}{(-12)} & 73.99\, \textcolor{blue}{(+8)} \\
QwentileSwap & 53.92 & 42.55\, \textcolor{blue}{(+8)} & 66.71\, \textcolor{red}{(-11)} & 70.16\, \textcolor{red}{(-26)} & 30.94\, \textcolor{red}{(-4)} & 51.82\, \textcolor{red}{(-12)} & 29.30\, \textcolor{red}{(-54)} & 41.09\, \textcolor{red}{(-9)} & 46.17\, \textcolor{red}{(-72)} & 46.70\, \textcolor{red}{(-127)} & 73.20\, \textcolor{red}{(-8)} \\
Qwen2.5-14B-YOYO-V4-p1 & 53.91 & 41.60\, \textcolor{red}{(-14)} & 67.74\, \textcolor{blue}{(+19)} & 69.77\, \textcolor{red}{(-38)} & 30.73\, \textcolor{red}{(-6)} & 56.40\, \textcolor{blue}{(+26)} & 30.99\, \textcolor{blue}{(+7)} & 44.95\, \textcolor{blue}{(+24)} & 47.67\, \textcolor{blue}{(+4)} & 52.70\, \textcolor{blue}{(+16)} & 83.23\, \textcolor{blue}{(+30)} \\
NQLSG-Qwen2.5-14B-MegaFusion-v9.1 & 53.88 & 41.95\, \textcolor{red}{(-3)} & 66.59\, \textcolor{red}{(-15)} & 70.28\, \textcolor{red}{(-19)} & 29.72\, \textcolor{red}{(-58)} & 53.07\, \textcolor{blue}{(+14)} & 30.70\, \textcolor{red}{(-2)} & 42.14\, \textcolor{blue}{(+12)} & 45.71\, \textcolor{red}{(-87)} & 51.08\, \textcolor{blue}{(+6)} & 73.11\, \textcolor{red}{(-9)} \\
Set-70b & 53.86 & 42.98\, \textcolor{blue}{(+18)} & 67.49\, \textcolor{blue}{(+13)} & 69.10\, \textcolor{red}{(-63)} & 32.02\, \textcolor{blue}{(+15)} & 49.84\, \textcolor{red}{(-82)} & 28.18\, \textcolor{red}{(-141)} & 39.40\, \textcolor{red}{(-53)} & 47.76\, \textcolor{blue}{(+8)} & 43.15\, \textcolor{red}{(-355)} & 70.34\, \textcolor{red}{(-201)} \\
Rombos-LLM-V2.5-Qwen-32b & 53.85 & 42.89\, \textcolor{blue}{(+16)} & 66.22\, \textcolor{red}{(-32)} & 69.51\, \textcolor{red}{(-43)} & 31.83\, \textcolor{blue}{(+14)} & 50.50\, \textcolor{red}{(-52)} & 28.53\, \textcolor{red}{(-110)} & 40.36\, \textcolor{red}{(-16)} & 46.42\, \textcolor{red}{(-56)} & 44.04\, \textcolor{red}{(-272)} & 70.87\, \textcolor{red}{(-157)} \\
Arcee-Nova & 53.85 & 42.10\, \textcolor{blue}{(+1)} & 67.25\, \textcolor{blue}{(+9)} & 69.55\, \textcolor{red}{(-40)} & 31.21\, \textcolor{blue}{(+8)} & 50.98\, \textcolor{red}{(-29)} & 29.31\, \textcolor{red}{(-47)} & 39.96\, \textcolor{red}{(-29)} & 47.19\, \textcolor{red}{(-9)} & 47.28\, \textcolor{red}{(-88)} & 71.75\, \textcolor{red}{(-87)} \\
li-14b-v0.4-slerp0.1 & 53.85 & 41.87\, \textcolor{red}{(-2)} & 66.53\, \textcolor{red}{(-14)} & 70.73\, \textcolor{red}{(-1)} & 29.12\, \textcolor{red}{(-98)} & 52.75\, \textcolor{blue}{(+16)} & 30.82\, \textcolor{blue}{(+7)} & 41.67\, \textcolor{blue}{(+9)} & 45.30\, \textcolor{red}{(-102)} & 51.17\, \textcolor{blue}{(+13)} & 73.74\, \textcolor{blue}{(+12)} \\
Qwen2-72B-Orpo-v0.1 & 53.84 & 42.39\, \textcolor{blue}{(+11)} & 66.38\, \textcolor{red}{(-23)} & 69.93\, \textcolor{red}{(-28)} & 30.94\, \textcolor{blue}{(+4)} & 50.70\, \textcolor{red}{(-35)} & 28.63\, \textcolor{red}{(-98)} & 40.00\, \textcolor{red}{(-26)} & 45.90\, \textcolor{red}{(-73)} & 46.24\, \textcolor{red}{(-139)} & 71.88\, \textcolor{red}{(-74)} \\
NQLSG-Qwen2.5-14B-MegaFusion-v8.7 & 53.78 & 41.80\, \textcolor{red}{(-4)} & 66.47\, \textcolor{red}{(-17)} & 70.07\, \textcolor{red}{(-21)} & 29.76\, \textcolor{red}{(-49)} & 52.35\, \textcolor{blue}{(+7)} & 29.42\, \textcolor{red}{(-37)} & 41.29\, \textcolor{blue}{(+2)} & 45.76\, \textcolor{red}{(-78)} & 49.10\, \textcolor{red}{(-17)} & 72.52\, \textcolor{red}{(-40)} \\
MG-FinalMix-72B & 53.73 & 42.14\, \textcolor{blue}{(+6)} & 65.86\, \textcolor{red}{(-51)} & 69.75\, \textcolor{red}{(-31)} & 30.30\, \textcolor{red}{(-20)} & 50.69\, \textcolor{red}{(-34)} & 28.37\, \textcolor{red}{(-120)} & 40.11\, \textcolor{red}{(-21)} & 44.71\, \textcolor{red}{(-131)} & 46.05\, \textcolor{red}{(-146)} & 71.33\, \textcolor{red}{(-124)} \\
ultiima-32B & 53.68 & 42.39\, \textcolor{blue}{(+15)} & 67.14\, \textcolor{blue}{(+10)} & 70.51\, \textcolor{red}{(-2)} & 31.12\, \textcolor{blue}{(+11)} & 51.39\, \textcolor{red}{(-10)} & 29.03\, \textcolor{red}{(-60)} & 41.05 & 46.31\, \textcolor{red}{(-56)} & 44.73\, \textcolor{red}{(-220)} & 72.36\, \textcolor{red}{(-46)} \\
NQLSG-Qwen2.5-14B-MegaFusion-v9.2 & 53.67 & 41.83\, \textcolor{blue}{(+1)} & 66.22\, \textcolor{red}{(-26)} & 70.24\, \textcolor{red}{(-11)} & 29.12\, \textcolor{red}{(-94)} & 52.65\, \textcolor{blue}{(+17)} & 29.65\, \textcolor{red}{(-27)} & 41.99\, \textcolor{blue}{(+18)} & 45.28\, \textcolor{red}{(-99)} & 49.87\, \textcolor{blue}{(+1)} & 72.39\, \textcolor{red}{(-43)} \\
Galactic-Qwen-14B-Exp2 & 53.60 & 41.82\, \textcolor{blue}{(+1)} & 66.96\, \textcolor{blue}{(+4)} & 71.20\, \textcolor{blue}{(+20)} & 30.09\, \textcolor{red}{(-30)} & 50.53\, \textcolor{red}{(-41)} & 26.64\, \textcolor{red}{(-331)} & 39.88\, \textcolor{red}{(-24)} & 45.46\, \textcolor{red}{(-87)} & 42.42\, \textcolor{red}{(-419)} & 72.25\, \textcolor{red}{(-50)} \\
RYS-XLarge-base & 53.57 & 42.36\, \textcolor{blue}{(+16)} & 65.95\, \textcolor{red}{(-39)} & 69.08\, \textcolor{red}{(-54)} & 31.49\, \textcolor{blue}{(+19)} & 50.90\, \textcolor{red}{(-26)} & 29.81\, \textcolor{red}{(-17)} & 40.56\, \textcolor{red}{(-3)} & 46.54\, \textcolor{red}{(-43)} & 47.15\, \textcolor{red}{(-89)} & 71.64\, \textcolor{red}{(-88)} \\
Qwen2.5-14B-YOYO-V4 & 53.56 & 41.53\, \textcolor{red}{(-3)} & 67.13\, \textcolor{blue}{(+13)} & 69.01\, \textcolor{red}{(-55)} & 30.43\, \textcolor{red}{(-10)} & 57.01\, \textcolor{blue}{(+44)} & 33.04\, \textcolor{blue}{(+36)} & 45.78\, \textcolor{blue}{(+40)} & 47.23\, \textcolor{blue}{(+5)} & 54.78\, \textcolor{blue}{(+45)} & 83.47\, \textcolor{blue}{(+44)} \\
EVA-abliterated-TIES-Qwen2.5-14B & 53.54 & 41.63\, \textcolor{blue}{(+1)} & 66.48\, \textcolor{red}{(-8)} & 70.07\, \textcolor{red}{(-13)} & 29.57\, \textcolor{red}{(-53)} & 51.32\, \textcolor{red}{(-8)} & 29.78\, \textcolor{red}{(-17)} & 40.35\, \textcolor{red}{(-6)} & 45.65\, \textcolor{red}{(-77)} & 49.48\, \textcolor{red}{(-3)} & 71.62\, \textcolor{red}{(-88)} \\
tempesthenno-sft-0309-ckpt10 & 53.49 & 41.33\, \textcolor{red}{(-9)} & 66.51\, \textcolor{red}{(-5)} & 71.75\, \textcolor{blue}{(+36)} & 28.79\, \textcolor{red}{(-108)} & 51.24\, \textcolor{red}{(-9)} & 30.04\, \textcolor{red}{(-1)} & 40.09\, \textcolor{red}{(-16)} & 43.95\, \textcolor{red}{(-180)} & 52.30\, \textcolor{blue}{(+26)} & 73.06\, \textcolor{blue}{(+3)} \\
\bottomrule
\end{tabular}}
\label{tab:meme_scores_summary_OpenLLM}
\end{table}

\begin{table}[H]
\centering
\caption{\textbf{Meme Scores (Open LLM Population; BBH).} The table reports meme scores across models, including property-derived 1D meme scores, predefined 2D meme scores (Mastery, Ingenuity, and Robustness), and a predefined 3D meme score (Caution). Models are sorted by Accuracy, and only the \textbf{top-50 models} are shown. {\color{blue}Blue} indicates rank improvement compared with Accuracy rank, and {\color{red}red} indicates rank degradation.}
\resizebox{\textwidth}{!}{
\begin{tabular}{cccccccccccc}
\toprule
\multirow{2}{*}{\textbf{Model}} & \multirow{2}{*}{\textbf{Accuracy}} &
\multicolumn{6}{c}{\textbf{Property-derived 1D meme scores}} &
\multicolumn{4}{c}{\textbf{Predefined 2D/3D meme scores}} \\
\cmidrule(lr){3-8}\cmidrule(lr){9-12}
 &  & \textbf{Difficulty} & \textbf{Uniqueness} & \textbf{Risk} & \textbf{Surprise} & \textbf{Typicality} & \textbf{Bridge} & \textbf{Mastery} & \textbf{Ingenuity} & \textbf{Robustness} & \textbf{Caution} \\
\midrule
Benchmaxx-Llama-3.2-1B-Instruct & 82.69 & 79.45 & 84.38 & 83.00\, \textcolor{red}{(-3)} & 80.73 & 82.80 & 82.08 & 78.96 & 81.59 & 84.02 & 86.41\, \textcolor{red}{(-758)} \\
T3Q-qwen2.5-14b-v1.0-e3 & 75.76 & 65.47 & 66.64\, \textcolor{red}{(-3)} & 84.25\, \textcolor{blue}{(+1)} & 51.65\, \textcolor{red}{(-4)} & 75.10 & 67.23\, \textcolor{red}{(-1)} & 62.46 & 45.25\, \textcolor{red}{(-5)} & 81.66\, \textcolor{red}{(-6)} & 92.61\, \textcolor{red}{(-3)} \\
T3Q-Qwen2.5-14B-Instruct-1M-e3 & 75.76 & 65.47\, \textcolor{blue}{(+1)} & 66.64\, \textcolor{red}{(-2)} & 84.25\, \textcolor{blue}{(+2)} & 51.65\, \textcolor{red}{(-3)} & 75.10\, \textcolor{blue}{(+1)} & 67.23 & 62.46\, \textcolor{blue}{(+1)} & 45.25\, \textcolor{red}{(-4)} & 81.66\, \textcolor{red}{(-5)} & 92.61\, \textcolor{red}{(-2)} \\
internlm2\_5-20b-llamafied & 74.82 & 64.73 & 67.02 & 82.71\, \textcolor{red}{(-2)} & 54.83\, \textcolor{blue}{(+1)} & 73.88 & 67.04\, \textcolor{red}{(-3)} & 61.23\, \textcolor{red}{(-1)} & 49.70\, \textcolor{red}{(-1)} & 80.81\, \textcolor{red}{(-26)} & 91.18\, \textcolor{red}{(-43)} \\
internlm2\_5-20b-chat & 74.65 & 63.94 & 68.36\, \textcolor{blue}{(+2)} & 81.81\, \textcolor{red}{(-4)} & 54.60\, \textcolor{blue}{(+1)} & 73.07 & 65.10\, \textcolor{red}{(-28)} & 59.65\, \textcolor{red}{(-2)} & 50.95\, \textcolor{blue}{(+2)} & 79.41\, \textcolor{red}{(-213)} & 90.15\, \textcolor{red}{(-173)} \\
shuttle-3 & 74.02 & 61.51\, \textcolor{red}{(-3)} & 65.34\, \textcolor{red}{(-2)} & 82.93\, \textcolor{blue}{(+1)} & 48.18\, \textcolor{red}{(-3)} & 72.76\, \textcolor{red}{(-1)} & 64.00\, \textcolor{red}{(-71)} & 56.99\, \textcolor{red}{(-7)} & 42.19\, \textcolor{red}{(-8)} & 80.43\, \textcolor{red}{(-57)} & 93.10\, \textcolor{blue}{(+4)} \\
ultiima-72B-v1.5 & 73.73 & 61.09\, \textcolor{red}{(-4)} & 64.14\, \textcolor{red}{(-7)} & 83.13\, \textcolor{blue}{(+4)} & 46.58\, \textcolor{red}{(-14)} & 72.59\, \textcolor{red}{(-2)} & 63.49\, \textcolor{red}{(-126)} & 56.66\, \textcolor{red}{(-7)} & 39.92\, \textcolor{red}{(-29)} & 80.20\, \textcolor{red}{(-92)} & 93.36\, \textcolor{blue}{(+6)} \\
calme-3.2-instruct-78b & 72.96 & 61.74 & 64.49\, \textcolor{red}{(-1)} & 81.47\, \textcolor{red}{(-8)} & 48.26 & 72.80\, \textcolor{blue}{(+2)} & 66.42\, \textcolor{red}{(-6)} & 59.01 & 42.42\, \textcolor{red}{(-3)} & 80.46\, \textcolor{red}{(-50)} & 91.17\, \textcolor{red}{(-40)} \\
Homer-v1.0-Qwen2.5-72B & 72.91 & 60.43\, \textcolor{red}{(-5)} & 63.30\, \textcolor{red}{(-11)} & 82.39\, \textcolor{blue}{(+1)} & 46.13\, \textcolor{red}{(-16)} & 72.01\, \textcolor{red}{(-3)} & 64.15\, \textcolor{red}{(-59)} & 56.50\, \textcolor{red}{(-6)} & 39.49\, \textcolor{red}{(-38)} & 80.45\, \textcolor{red}{(-52)} & 92.66\, \textcolor{blue}{(+6)} \\
calme-3.1-instruct-78b & 72.83 & 61.49 & 64.32\, \textcolor{red}{(-1)} & 81.41\, \textcolor{red}{(-8)} & 48.07 & 72.70\, \textcolor{blue}{(+2)} & 66.84 & 58.70\, \textcolor{blue}{(+1)} & 42.23\, \textcolor{red}{(-3)} & 80.90\, \textcolor{red}{(-16)} & 91.37\, \textcolor{red}{(-27)} \\
calme-2.2-qwen2.5-72b & 72.60 & 60.00\, \textcolor{red}{(-5)} & 63.87\, \textcolor{red}{(-7)} & 81.71 & 47.11\, \textcolor{red}{(-6)} & 71.42\, \textcolor{red}{(-6)} & 63.76\, \textcolor{red}{(-94)} & 55.82\, \textcolor{red}{(-10)} & 41.10\, \textcolor{red}{(-13)} & 79.88\, \textcolor{red}{(-136)} & 91.96\, \textcolor{red}{(-1)} \\
calme-2.4-rys-78b & 72.58 & 61.09 & 64.28 & 81.11\, \textcolor{red}{(-7)} & 47.95\, \textcolor{blue}{(+1)} & 72.45\, \textcolor{blue}{(+2)} & 66.60\, \textcolor{blue}{(+1)} & 58.25\, \textcolor{blue}{(+1)} & 42.33 & 80.72\, \textcolor{red}{(-28)} & 91.21\, \textcolor{red}{(-33)} \\
Qwen2.5-72B-2x-Instruct-TIES-v1.0 & 72.56 & 59.94\, \textcolor{red}{(-4)} & 63.87\, \textcolor{red}{(-4)} & 81.66\, \textcolor{blue}{(+1)} & 47.07\, \textcolor{red}{(-5)} & 71.34\, \textcolor{red}{(-5)} & 63.85\, \textcolor{red}{(-79)} & 55.67\, \textcolor{red}{(-11)} & 41.11\, \textcolor{red}{(-10)} & 79.96\, \textcolor{red}{(-121)} & 91.93 \\
Qwen2.5-72B-Instruct & 72.55 & 59.92\, \textcolor{red}{(-4)} & 63.94\, \textcolor{red}{(-1)} & 81.60\, \textcolor{blue}{(+1)} & 47.15\, \textcolor{red}{(-2)} & 71.25\, \textcolor{red}{(-7)} & 63.80\, \textcolor{red}{(-83)} & 55.55\, \textcolor{red}{(-12)} & 41.23\, \textcolor{red}{(-8)} & 79.92\, \textcolor{red}{(-125)} & 91.86\, \textcolor{red}{(-1)} \\
test-2.5-72B & 72.44 & 60.10 & 64.48\, \textcolor{blue}{(+5)} & 81.05\, \textcolor{red}{(-6)} & 47.66\, \textcolor{blue}{(+3)} & 71.62\, \textcolor{red}{(-1)} & 63.48\, \textcolor{red}{(-120)} & 56.07\, \textcolor{red}{(-3)} & 42.62\, \textcolor{blue}{(+5)} & 79.61\, \textcolor{red}{(-172)} & 91.76\, \textcolor{red}{(-8)} \\
calme-2.1-qwen2.5-72b & 72.44 & 59.86\, \textcolor{red}{(-4)} & 63.77\, \textcolor{red}{(-3)} & 81.53\, \textcolor{blue}{(+1)} & 47.04\, \textcolor{red}{(-3)} & 71.27\, \textcolor{red}{(-3)} & 63.82\, \textcolor{red}{(-78)} & 55.67\, \textcolor{red}{(-7)} & 41.09\, \textcolor{red}{(-9)} & 79.90\, \textcolor{red}{(-128)} & 91.80\, \textcolor{red}{(-5)} \\
CalmeRys-78B-Orpo-v0.1 & 72.41 & 60.93\, \textcolor{blue}{(+4)} & 63.92\, \textcolor{blue}{(+1)} & 81.05\, \textcolor{red}{(-3)} & 47.59\, \textcolor{blue}{(+4)} & 72.26\, \textcolor{blue}{(+6)} & 66.31\, \textcolor{blue}{(+2)} & 58.09\, \textcolor{blue}{(+5)} & 41.80\, \textcolor{red}{(-1)} & 80.53\, \textcolor{red}{(-36)} & 91.15\, \textcolor{red}{(-33)} \\
LeTriomphant2.2\_ECE\_iLAB & 72.37 & 59.62\, \textcolor{red}{(-3)} & 63.05\, \textcolor{red}{(-3)} & 81.76\, \textcolor{blue}{(+8)} & 45.72\, \textcolor{red}{(-9)} & 72.00\, \textcolor{blue}{(+5)} & 63.53\, \textcolor{red}{(-106)} & 56.35\, \textcolor{blue}{(+2)} & 39.46\, \textcolor{red}{(-31)} & 80.19\, \textcolor{red}{(-83)} & 92.64\, \textcolor{blue}{(+14)} \\
Gilgamesh-72B & 72.30 & 59.91 & 64.21\, \textcolor{blue}{(+6)} & 81.00\, \textcolor{red}{(-3)} & 47.29\, \textcolor{blue}{(+5)} & 71.68\, \textcolor{blue}{(+4)} & 63.68\, \textcolor{red}{(-93)} & 56.16\, \textcolor{blue}{(+2)} & 42.15\, \textcolor{blue}{(+4)} & 79.83\, \textcolor{red}{(-133)} & 91.85\, \textcolor{blue}{(+2)} \\
Rombos-LLM-V2.5-Qwen-72b & 72.10 & 59.39\, \textcolor{red}{(-3)} & 62.70\, \textcolor{red}{(-3)} & 81.57\, \textcolor{blue}{(+6)} & 45.52\, \textcolor{red}{(-9)} & 71.23\, \textcolor{red}{(-2)} & 63.86\, \textcolor{red}{(-70)} & 55.46\, \textcolor{red}{(-8)} & 39.12\, \textcolor{red}{(-37)} & 80.18\, \textcolor{red}{(-85)} & 92.17\, \textcolor{blue}{(+12)} \\
ultiima-72B & 71.97 & 59.16\, \textcolor{red}{(-4)} & 62.61\, \textcolor{red}{(-3)} & 81.46\, \textcolor{blue}{(+4)} & 45.36\, \textcolor{red}{(-9)} & 70.99\, \textcolor{red}{(-6)} & 63.50\, \textcolor{red}{(-109)} & 55.01\, \textcolor{red}{(-15)} & 38.99\, \textcolor{red}{(-40)} & 79.91\, \textcolor{red}{(-120)} & 92.12\, \textcolor{blue}{(+12)} \\
Galactic-Qwen-14B-Exp2 & 71.92 & 59.48 & 60.54\, \textcolor{red}{(-17)} & 82.43\, \textcolor{blue}{(+15)} & 43.16\, \textcolor{red}{(-57)} & 71.27\, \textcolor{blue}{(+2)} & 63.34\, \textcolor{red}{(-125)} & 56.03\, \textcolor{blue}{(+2)} & 35.07\, \textcolor{red}{(-515)} & 80.18\, \textcolor{red}{(-82)} & 92.56\, \textcolor{blue}{(+15)} \\
Qwen2.5-72B-Instruct-abliterated & 71.64 & 58.80\, \textcolor{red}{(-4)} & 62.79\, \textcolor{blue}{(+1)} & 80.92\, \textcolor{red}{(-1)} & 45.95\, \textcolor{red}{(-3)} & 70.76\, \textcolor{red}{(-6)} & 64.01\, \textcolor{red}{(-53)} & 54.60\, \textcolor{red}{(-18)} & 40.13\, \textcolor{red}{(-10)} & 80.10\, \textcolor{red}{(-89)} & 91.83\, \textcolor{blue}{(+5)} \\
Qwen2.5-72B-Instruct-abliterated & 71.35 & 58.49\, \textcolor{red}{(-5)} & 61.84\, \textcolor{red}{(-2)} & 80.96\, \textcolor{blue}{(+1)} & 44.73\, \textcolor{red}{(-14)} & 70.16\, \textcolor{red}{(-19)} & 63.39\, \textcolor{red}{(-120)} & 53.93\, \textcolor{red}{(-32)} & 38.31\, \textcolor{red}{(-65)} & 79.66\, \textcolor{red}{(-153)} & 91.52\, \textcolor{red}{(-8)} \\
internlm2\_5-7b-chat & 71.21 & 61.92\, \textcolor{blue}{(+18)} & 65.61\, \textcolor{blue}{(+18)} & 77.64\, \textcolor{red}{(-181)} & 53.21\, \textcolor{blue}{(+20)} & 71.09\, \textcolor{blue}{(+1)} & 63.95\, \textcolor{red}{(-57)} & 60.48\, \textcolor{blue}{(+19)} & 50.37\, \textcolor{blue}{(+21)} & 76.62\, \textcolor{red}{(-1011)} & 85.49\, \textcolor{red}{(-796)} \\
FluentlyLM-Prinum & 71.09 & 58.83 & 61.62\, \textcolor{red}{(-2)} & 80.69\, \textcolor{blue}{(+1)} & 46.18\, \textcolor{blue}{(+2)} & 71.10\, \textcolor{blue}{(+3)} & 67.16\, \textcolor{blue}{(+21)} & 56.05\, \textcolor{blue}{(+7)} & 39.57\, \textcolor{red}{(-19)} & 82.33\, \textcolor{blue}{(+23)} & 91.82\, \textcolor{blue}{(+6)} \\
calme-2.3-rys-78b & 70.97 & 58.54\, \textcolor{red}{(-1)} & 61.52\, \textcolor{red}{(-3)} & 80.48\, \textcolor{blue}{(+1)} & 45.07\, \textcolor{red}{(-5)} & 71.07\, \textcolor{blue}{(+2)} & 65.21\, \textcolor{red}{(-4)} & 55.81\, \textcolor{blue}{(+5)} & 38.59\, \textcolor{red}{(-46)} & 80.57\, \textcolor{red}{(-22)} & 91.52\, \textcolor{red}{(-6)} \\
BigQwen2.5-52B-Instruct & 70.88 & 59.37\, \textcolor{blue}{(+4)} & 62.60\, \textcolor{blue}{(+3)} & 79.55\, \textcolor{red}{(-38)} & 47.20\, \textcolor{blue}{(+13)} & 71.93\, \textcolor{blue}{(+14)} & 67.52\, \textcolor{blue}{(+26)} & 58.36\, \textcolor{blue}{(+18)} & 41.63\, \textcolor{blue}{(+8)} & 81.72\, \textcolor{blue}{(+21)} & 90.64\, \textcolor{red}{(-89)} \\
calme-2.1-rys-78b & 70.85 & 57.79\, \textcolor{red}{(-6)} & 61.71\, \textcolor{blue}{(+2)} & 80.30\, \textcolor{blue}{(+2)} & 44.76\, \textcolor{red}{(-7)} & 70.73\, \textcolor{red}{(-1)} & 64.29\, \textcolor{red}{(-25)} & 54.44\, \textcolor{red}{(-17)} & 38.65\, \textcolor{red}{(-41)} & 80.32\, \textcolor{red}{(-49)} & 91.92\, \textcolor{blue}{(+15)} \\
calme-2.2-rys-78b & 70.71 & 57.77\, \textcolor{red}{(-9)} & 61.39\, \textcolor{red}{(-1)} & 80.29\, \textcolor{blue}{(+1)} & 44.56\, \textcolor{red}{(-12)} & 70.54\, \textcolor{red}{(-1)} & 64.52\, \textcolor{red}{(-14)} & 54.38\, \textcolor{red}{(-17)} & 38.18\, \textcolor{red}{(-64)} & 80.41\, \textcolor{red}{(-37)} & 91.74\, \textcolor{blue}{(+6)} \\
EVA-Qwen2.5-72B-v0.2 & 70.53 & 57.66\, \textcolor{red}{(-10)} & 61.55\, \textcolor{blue}{(+2)} & 79.93\, \textcolor{red}{(-3)} & 44.25\, \textcolor{red}{(-21)} & 71.04\, \textcolor{blue}{(+5)} & 63.92\, \textcolor{red}{(-54)} & 55.30\, \textcolor{red}{(-1)} & 38.37\, \textcolor{red}{(-54)} & 80.24\, \textcolor{red}{(-62)} & 91.52 \\
qwen2.5-test-32b-it & 70.43 & 57.93\, \textcolor{blue}{(+2)} & 60.63\, \textcolor{red}{(-5)} & 80.29\, \textcolor{blue}{(+4)} & 44.73\, \textcolor{red}{(-5)} & 70.79\, \textcolor{blue}{(+4)} & 67.10\, \textcolor{blue}{(+26)} & 55.58\, \textcolor{blue}{(+7)} & 37.96\, \textcolor{red}{(-76)} & 82.41\, \textcolor{blue}{(+30)} & 91.80\, \textcolor{blue}{(+10)} \\
RYS-XLarge & 70.41 & 57.01\, \textcolor{red}{(-19)} & 60.93 & 80.18\, \textcolor{blue}{(+3)} & 43.47\, \textcolor{red}{(-38)} & 70.28\, \textcolor{red}{(-3)} & 63.90\, \textcolor{red}{(-54)} & 53.80\, \textcolor{red}{(-35)} & 37.16\, \textcolor{red}{(-152)} & 80.29\, \textcolor{red}{(-56)} & 91.85\, \textcolor{blue}{(+17)} \\
RYS-XLarge-base & 70.39 & 57.05\, \textcolor{red}{(-17)} & 60.88 & 80.17\, \textcolor{blue}{(+3)} & 43.44\, \textcolor{red}{(-39)} & 70.26\, \textcolor{red}{(-3)} & 63.96\, \textcolor{red}{(-47)} & 53.79\, \textcolor{red}{(-35)} & 37.11\, \textcolor{red}{(-158)} & 80.32\, \textcolor{red}{(-43)} & 91.83\, \textcolor{blue}{(+15)} \\
Apollo\_v2-32B & 70.36 & 57.91\, \textcolor{blue}{(+3)} & 60.59\, \textcolor{red}{(-3)} & 80.16\, \textcolor{blue}{(+3)} & 44.44\, \textcolor{red}{(-13)} & 70.50\, \textcolor{blue}{(+3)} & 66.47\, \textcolor{blue}{(+22)} & 55.26\, \textcolor{blue}{(+1)} & 37.70\, \textcolor{red}{(-99)} & 82.03\, \textcolor{blue}{(+29)} & 91.54\, \textcolor{blue}{(+6)} \\
70B-L3.3-Cirrus-x1 & 70.17 & 57.77\, \textcolor{red}{(-2)} & 60.35\, \textcolor{red}{(-6)} & 79.91 & 44.03\, \textcolor{red}{(-20)} & 69.34\, \textcolor{red}{(-30)} & 64.43\, \textcolor{red}{(-11)} & 53.79\, \textcolor{red}{(-34)} & 37.47\, \textcolor{red}{(-120)} & 80.02\, \textcolor{red}{(-85)} & 90.41\, \textcolor{red}{(-106)} \\
Rombos-LLM-V2.5-Qwen-32b & 70.13 & 57.85\, \textcolor{blue}{(+4)} & 60.30\, \textcolor{red}{(-8)} & 79.97\, \textcolor{blue}{(+4)} & 44.70\, \textcolor{red}{(-2)} & 70.37\, \textcolor{blue}{(+3)} & 66.23\, \textcolor{blue}{(+20)} & 55.47\, \textcolor{blue}{(+10)} & 38.01\, \textcolor{red}{(-66)} & 81.53\, \textcolor{blue}{(+26)} & 91.12\, \textcolor{red}{(-16)} \\
BunderMaxx-1010 & 70.12 & 62.78\, \textcolor{blue}{(+32)} & 70.62\, \textcolor{blue}{(+36)} & 72.74\, \textcolor{red}{(-585)} & 62.03\, \textcolor{blue}{(+36)} & 70.31\, \textcolor{blue}{(+3)} & 66.51\, \textcolor{blue}{(+26)} & 61.53\, \textcolor{blue}{(+34)} & 63.11\, \textcolor{blue}{(+36)} & 73.49\, \textcolor{red}{(-2151)} & 80.11\, \textcolor{red}{(-1845)} \\
novablast-preview & 70.10 & 57.93\, \textcolor{blue}{(+8)} & 60.36\, \textcolor{red}{(-2)} & 79.86 & 44.87\, \textcolor{blue}{(+6)} & 70.26\, \textcolor{blue}{(+1)} & 66.24\, \textcolor{blue}{(+23)} & 55.43\, \textcolor{blue}{(+10)} & 38.27\, \textcolor{red}{(-52)} & 81.45\, \textcolor{blue}{(+27)} & 90.96\, \textcolor{red}{(-29)} \\
Qwen2.5-95B-Instruct & 70.05 & 56.60\, \textcolor{red}{(-35)} & 61.01\, \textcolor{blue}{(+8)} & 79.75\, \textcolor{red}{(-5)} & 43.61\, \textcolor{red}{(-23)} & 69.45\, \textcolor{red}{(-20)} & 61.78\, \textcolor{red}{(-256)} & 52.55\, \textcolor{red}{(-111)} & 37.50\, \textcolor{red}{(-115)} & 78.65\, \textcolor{red}{(-365)} & 91.68\, \textcolor{blue}{(+13)} \\
ultiima-32B & 70.03 & 57.82\, \textcolor{blue}{(+7)} & 60.27\, \textcolor{red}{(-5)} & 79.82 & 44.60\, \textcolor{blue}{(+1)} & 70.20\, \textcolor{blue}{(+1)} & 65.98\, \textcolor{blue}{(+21)} & 55.33\, \textcolor{blue}{(+11)} & 37.98\, \textcolor{red}{(-65)} & 81.32\, \textcolor{blue}{(+24)} & 90.95\, \textcolor{red}{(-31)} \\
RombosBeagle-v2beta-MGS-32B & 70.03 & 57.71\, \textcolor{blue}{(+2)} & 60.16\, \textcolor{red}{(-13)} & 79.89\, \textcolor{blue}{(+5)} & 44.44\, \textcolor{red}{(-5)} & 70.19\, \textcolor{blue}{(+1)} & 65.92\, \textcolor{blue}{(+19)} & 55.24\, \textcolor{blue}{(+7)} & 37.75\, \textcolor{red}{(-83)} & 81.33\, \textcolor{blue}{(+26)} & 91.06\, \textcolor{red}{(-16)} \\
TheBeagle-v2beta-32B-MGS & 70.01 & 57.78\, \textcolor{blue}{(+6)} & 60.16\, \textcolor{red}{(-11)} & 79.84\, \textcolor{blue}{(+3)} & 44.49\, \textcolor{red}{(-3)} & 70.10\, \textcolor{red}{(-2)} & 65.95\, \textcolor{blue}{(+21)} & 55.27\, \textcolor{blue}{(+10)} & 37.84\, \textcolor{red}{(-75)} & 81.29\, \textcolor{blue}{(+24)} & 90.86\, \textcolor{red}{(-42)} \\
Set-70b & 69.99 & 57.65\, \textcolor{blue}{(+2)} & 59.98\, \textcolor{red}{(-21)} & 79.87\, \textcolor{blue}{(+6)} & 43.70\, \textcolor{red}{(-16)} & 68.98\, \textcolor{red}{(-55)} & 64.86\, \textcolor{blue}{(+7)} & 53.43\, \textcolor{red}{(-54)} & 36.75\, \textcolor{red}{(-187)} & 80.29\, \textcolor{red}{(-44)} & 90.28\, \textcolor{red}{(-115)} \\
Oxyge1-33B & 69.99 & 57.79\, \textcolor{blue}{(+9)} & 60.22\, \textcolor{red}{(-7)} & 79.79\, \textcolor{blue}{(+1)} & 44.56\, \textcolor{blue}{(+4)} & 70.18\, \textcolor{blue}{(+3)} & 65.98\, \textcolor{blue}{(+24)} & 55.31\, \textcolor{blue}{(+14)} & 37.95\, \textcolor{red}{(-65)} & 81.32\, \textcolor{blue}{(+27)} & 90.93\, \textcolor{red}{(-30)} \\
Linkbricks-Horizon-AI-Avengers-V2-32B & 69.94 & 57.49\, \textcolor{blue}{(+1)} & 60.23\, \textcolor{red}{(-4)} & 79.80\, \textcolor{blue}{(+4)} & 44.51\, \textcolor{blue}{(+1)} & 70.21\, \textcolor{blue}{(+7)} & 65.90\, \textcolor{blue}{(+22)} & 54.97\, \textcolor{blue}{(+9)} & 38.02\, \textcolor{red}{(-56)} & 81.41\, \textcolor{blue}{(+32)} & 91.25\, \textcolor{blue}{(+5)} \\
Llama-3.1-SauerkrautLM-70b-Instruct & 69.93 & 57.62\, \textcolor{blue}{(+4)} & 60.24\, \textcolor{red}{(-2)} & 79.59\, \textcolor{red}{(-14)} & 43.50\, \textcolor{red}{(-22)} & 69.76\, \textcolor{red}{(-3)} & 61.95\, \textcolor{red}{(-229)} & 54.22\, \textcolor{red}{(-3)} & 36.97\, \textcolor{red}{(-158)} & 78.16\, \textcolor{red}{(-488)} & 90.69\, \textcolor{red}{(-63)} \\
L3.3-MS-Nevoria-70b & 69.89 & 57.59\, \textcolor{blue}{(+4)} & 60.07\, \textcolor{red}{(-14)} & 79.59\, \textcolor{red}{(-14)} & 43.39\, \textcolor{red}{(-26)} & 69.51\, \textcolor{red}{(-9)} & 63.94\, \textcolor{red}{(-35)} & 54.07\, \textcolor{red}{(-5)} & 36.79\, \textcolor{red}{(-180)} & 79.84\, \textcolor{red}{(-103)} & 90.41\, \textcolor{red}{(-93)} \\
magnum-v2-72b & 69.84 & 56.14\, \textcolor{red}{(-64)} & 60.07\, \textcolor{red}{(-12)} & 79.93\, \textcolor{blue}{(+14)} & 42.39\, \textcolor{red}{(-52)} & 70.42\, \textcolor{blue}{(+16)} & 63.10\, \textcolor{red}{(-122)} & 53.89\, \textcolor{red}{(-11)} & 35.86\, \textcolor{red}{(-337)} & 79.89\, \textcolor{red}{(-96)} & 91.99\, \textcolor{blue}{(+38)} \\
QwentileSwap & 69.79 & 57.09\, \textcolor{blue}{(+1)} & 60.34\, \textcolor{blue}{(+7)} & 79.55\, \textcolor{red}{(-15)} & 44.18\, \textcolor{red}{(-3)} & 70.14\, \textcolor{blue}{(+6)} & 63.97\, \textcolor{red}{(-29)} & 54.52\, \textcolor{blue}{(+6)} & 37.84\, \textcolor{red}{(-66)} & 80.30\, \textcolor{red}{(-35)} & 91.24\, \textcolor{blue}{(+7)} \\
\bottomrule
\end{tabular}}
\label{tab:meme_scores_summary_OpenLLM-BBH}
\end{table}

\begin{table}[H]
\centering
\caption{\textbf{Meme Scores (Open LLM Population; GPQA-Diamond).} The table reports meme scores across models, including property-derived 1D meme scores, predefined 2D meme scores (Mastery, Ingenuity, and Robustness), and a predefined 3D meme score (Caution). Models are sorted by Accuracy, and only the \textbf{top-50 models} are shown. {\color{blue}Blue} indicates rank improvement compared with Accuracy rank, and {\color{red}red} indicates rank degradation.}
\resizebox{\textwidth}{!}{
\begin{tabular}{cccccccccccc}
\toprule
\multirow{2}{*}{\textbf{Model}} & \multirow{2}{*}{\textbf{Accuracy}} &
\multicolumn{6}{c}{\textbf{Property-derived 1D meme scores}} &
\multicolumn{4}{c}{\textbf{Predefined 2D/3D meme scores}} \\
\cmidrule(lr){3-8}\cmidrule(lr){9-12}
 &  & \textbf{Difficulty} & \textbf{Uniqueness} & \textbf{Risk} & \textbf{Surprise} & \textbf{Typicality} & \textbf{Bridge} & \textbf{Mastery} & \textbf{Ingenuity} & \textbf{Robustness} & \textbf{Caution} \\
\midrule
EVA-Qwen2.5-72B-v0.2 & 45.45 & 36.85\, \textcolor{red}{(-4)} & 57.17\, \textcolor{red}{(-1)} & 59.88\, \textcolor{red}{(-80)} & 29.87\, \textcolor{red}{(-61)} & 0.00\, \textcolor{red}{(-923)} & 0.00\, \textcolor{red}{(-969)} & 0.00\, \textcolor{red}{(-926)} & 44.26\, \textcolor{red}{(-1)} & 0.00\, \textcolor{red}{(-973)} & 0.00\, \textcolor{red}{(-973)} \\
LeTriomphant2.2\_ECE\_iLAB & 44.95 & 36.30\, \textcolor{red}{(-4)} & 56.27\, \textcolor{red}{(-1)} & 61.93\, \textcolor{red}{(-21)} & 28.24\, \textcolor{red}{(-151)} & 0.00\, \textcolor{red}{(-923)} & 0.00\, \textcolor{red}{(-969)} & 0.00\, \textcolor{red}{(-926)} & 40.56\, \textcolor{red}{(-86)} & 0.00\, \textcolor{red}{(-915)} & 0.00\, \textcolor{red}{(-879)} \\
T3Q-qwen2.5-14b-v1.0-e3 & 44.44 & 37.32\, \textcolor{blue}{(+1)} & 53.04\, \textcolor{red}{(-52)} & 61.45\, \textcolor{red}{(-34)} & 28.52\, \textcolor{red}{(-127)} & 0.00\, \textcolor{red}{(-923)} & 0.00\, \textcolor{red}{(-969)} & 0.00\, \textcolor{red}{(-921)} & 36.70\, \textcolor{red}{(-318)} & 0.00\, \textcolor{red}{(-928)} & 0.00\, \textcolor{red}{(-941)} \\
T3Q-Qwen2.5-14B-Instruct-1M-e3 & 44.44 & 37.32\, \textcolor{blue}{(+2)} & 53.04\, \textcolor{red}{(-51)} & 61.45\, \textcolor{red}{(-33)} & 28.52\, \textcolor{red}{(-126)} & 0.00\, \textcolor{red}{(-922)} & 0.00\, \textcolor{red}{(-968)} & 0.00\, \textcolor{red}{(-920)} & 36.70\, \textcolor{red}{(-317)} & 0.00\, \textcolor{red}{(-927)} & 0.00\, \textcolor{red}{(-940)} \\
ultiima-72B & 44.44 & 35.99\, \textcolor{red}{(-3)} & 55.31\, \textcolor{red}{(-4)} & 62.36\, \textcolor{red}{(-10)} & 28.26\, \textcolor{red}{(-147)} & 0.00\, \textcolor{red}{(-921)} & 0.00\, \textcolor{red}{(-967)} & 0.00\, \textcolor{red}{(-925)} & 38.91\, \textcolor{red}{(-133)} & 0.00\, \textcolor{red}{(-904)} & 0.00\, \textcolor{red}{(-865)} \\
Llama-3.1-Nemotron-lorablated-70B & 43.94 & 37.27\, \textcolor{blue}{(+2)} & 52.28\, \textcolor{red}{(-80)} & 58.45\, \textcolor{red}{(-125)} & 27.82\, \textcolor{red}{(-187)} & 0.00\, \textcolor{red}{(-923)} & 0.00\, \textcolor{red}{(-971)} & 0.00\, \textcolor{red}{(-920)} & 37.96\, \textcolor{red}{(-180)} & 0.00\, \textcolor{red}{(-1013)} & 0.00\, \textcolor{red}{(-1176)} \\
Barcenas-14b-phi-4-v2 & 43.94 & 39.00\, \textcolor{blue}{(+6)} & 49.72\, \textcolor{red}{(-223)} & 56.23\, \textcolor{red}{(-214)} & 30.58\, \textcolor{red}{(-20)} & 25.36\, \textcolor{red}{(-303)} & 10.81\, \textcolor{red}{(-806)} & 24.46\, \textcolor{red}{(-303)} & 37.38\, \textcolor{red}{(-233)} & 18.97\, \textcolor{red}{(-414)} & 64.75\, \textcolor{red}{(-140)} \\
test-2.5-72B & 43.94 & 35.43\, \textcolor{red}{(-6)} & 55.25\, \textcolor{red}{(-3)} & 59.87\, \textcolor{red}{(-75)} & 28.91\, \textcolor{red}{(-100)} & 0.00\, \textcolor{red}{(-921)} & 0.00\, \textcolor{red}{(-967)} & 0.00\, \textcolor{red}{(-925)} & 40.88\, \textcolor{red}{(-64)} & 0.00\, \textcolor{red}{(-968)} & 0.00\, \textcolor{red}{(-938)} \\
Homer-v1.0-Qwen2.5-72B & 43.94 & 35.52\, \textcolor{red}{(-4)} & 54.69\, \textcolor{red}{(-13)} & 61.78\, \textcolor{red}{(-17)} & 27.51\, \textcolor{red}{(-218)} & 0.00\, \textcolor{red}{(-920)} & 0.00\, \textcolor{red}{(-966)} & 0.00\, \textcolor{red}{(-923)} & 37.94\, \textcolor{red}{(-180)} & 0.00\, \textcolor{red}{(-911)} & 0.00\, \textcolor{red}{(-867)} \\
Qwen2-VL-72B-Instruct & 43.43 & 35.60\, \textcolor{blue}{(+1)} & 53.96\, \textcolor{red}{(-22)} & 56.86\, \textcolor{red}{(-184)} & 29.17\, \textcolor{red}{(-81)} & 0.00\, \textcolor{red}{(-922)} & 0.00\, \textcolor{red}{(-971)} & 0.00\, \textcolor{red}{(-921)} & 41.01\, \textcolor{red}{(-57)} & 0.00\, \textcolor{red}{(-1067)} & 0.00\, \textcolor{red}{(-1146)} \\
magnum-v2-72b & 43.43 & 35.25\, \textcolor{red}{(-5)} & 54.52\, \textcolor{red}{(-12)} & 57.23\, \textcolor{red}{(-163)} & 29.27\, \textcolor{red}{(-78)} & 0.00\, \textcolor{red}{(-921)} & 0.00\, \textcolor{red}{(-967)} & 0.00\, \textcolor{red}{(-924)} & 41.29\, \textcolor{red}{(-47)} & 0.00\, \textcolor{red}{(-1048)} & 0.00\, \textcolor{red}{(-1037)} \\
Galactic-Qwen-14B-Exp2 & 43.43 & 33.64\, \textcolor{red}{(-22)} & 56.09\, \textcolor{blue}{(+7)} & 63.84\, \textcolor{blue}{(+11)} & 25.06\, \textcolor{red}{(-646)} & 0.00\, \textcolor{red}{(-920)} & 0.00\, \textcolor{red}{(-966)} & 0.00\, \textcolor{red}{(-940)} & 37.74\, \textcolor{red}{(-199)} & 0.00\, \textcolor{red}{(-883)} & 0.00\, \textcolor{red}{(-842)} \\
calme-2.1-rys-78b & 43.43 & 35.59\, \textcolor{blue}{(+3)} & 53.80\, \textcolor{red}{(-24)} & 58.11\, \textcolor{red}{(-126)} & 29.62\, \textcolor{red}{(-61)} & 17.14\, \textcolor{red}{(-501)} & 16.08\, \textcolor{red}{(-494)} & 16.93\, \textcolor{red}{(-501)} & 39.81\, \textcolor{red}{(-97)} & 16.29\, \textcolor{red}{(-761)} & 16.39\, \textcolor{red}{(-400)} \\
shuttle-3 & 43.43 & 32.56\, \textcolor{red}{(-46)} & 58.67\, \textcolor{blue}{(+13)} & 60.19\, \textcolor{red}{(-57)} & 26.67\, \textcolor{red}{(-308)} & 0.00\, \textcolor{red}{(-918)} & 0.00\, \textcolor{red}{(-964)} & 0.00\, \textcolor{red}{(-960)} & 44.36\, \textcolor{blue}{(+13)} & 0.00\, \textcolor{red}{(-950)} & 0.00\, \textcolor{red}{(-872)} \\
Phi-4-RRStock & 42.93 & 35.43 & 53.12\, \textcolor{red}{(-36)} & 55.34\, \textcolor{red}{(-255)} & 29.50\, \textcolor{red}{(-64)} & 0.00\, \textcolor{red}{(-921)} & 0.00\, \textcolor{red}{(-968)} & 0.00\, \textcolor{red}{(-919)} & 41.58\, \textcolor{red}{(-24)} & 0.00\, \textcolor{red}{(-1126)} & 0.00\, \textcolor{red}{(-1282)} \\
MiniusLight-24B-v1d-test & 42.93 & 33.69\, \textcolor{red}{(-17)} & 55.10\, \textcolor{blue}{(+3)} & 60.88\, \textcolor{red}{(-39)} & 25.64\, \textcolor{red}{(-476)} & 0.00\, \textcolor{red}{(-920)} & 0.00\, \textcolor{red}{(-966)} & 0.00\, \textcolor{red}{(-935)} & 38.29\, \textcolor{red}{(-152)} & 0.00\, \textcolor{red}{(-932)} & 0.00\, \textcolor{red}{(-884)} \\
ultiima-72B-v1.5 & 42.93 & 34.10\, \textcolor{red}{(-9)} & 54.28\, \textcolor{red}{(-11)} & 60.55\, \textcolor{red}{(-46)} & 26.00\, \textcolor{red}{(-403)} & 0.00\, \textcolor{red}{(-919)} & 0.00\, \textcolor{red}{(-968)} & 0.00\, \textcolor{red}{(-927)} & 37.03\, \textcolor{red}{(-261)} & 0.00\, \textcolor{red}{(-939)} & 0.00\, \textcolor{red}{(-880)} \\
Qwentile2.5-32B-Instruct & 42.93 & 34.69\, \textcolor{red}{(-2)} & 54.44\, \textcolor{red}{(-7)} & 56.57\, \textcolor{red}{(-192)} & 29.65\, \textcolor{red}{(-53)} & 25.36\, \textcolor{red}{(-293)} & 23.59\, \textcolor{red}{(-231)} & 24.46\, \textcolor{red}{(-295)} & 42.32 & 49.09\, \textcolor{red}{(-56)} & 64.75\, \textcolor{red}{(-126)} \\
Set-70b & 42.93 & 36.09\, \textcolor{blue}{(+12)} & 51.47\, \textcolor{red}{(-101)} & 58.58\, \textcolor{red}{(-106)} & 26.92\, \textcolor{red}{(-268)} & 0.00\, \textcolor{red}{(-917)} & 0.00\, \textcolor{red}{(-964)} & 0.00\, \textcolor{red}{(-910)} & 36.57\, \textcolor{red}{(-323)} & 0.00\, \textcolor{red}{(-996)} & 0.00\, \textcolor{red}{(-1111)} \\
tempesthenno-kto-0205-ckpt80 & 42.42 & 34.40\, \textcolor{red}{(-3)} & 52.11\, \textcolor{red}{(-74)} & 63.13\, \textcolor{blue}{(+15)} & 24.41\, \textcolor{red}{(-922)} & 0.00\, \textcolor{red}{(-922)} & 0.00\, \textcolor{red}{(-969)} & 0.00\, \textcolor{red}{(-921)} & 33.48\, \textcolor{red}{(-1097)} & 0.00\, \textcolor{red}{(-879)} & 0.00\, \textcolor{red}{(-860)} \\
calme-2.2-rys-78b & 42.42 & 34.72\, \textcolor{blue}{(+2)} & 52.58\, \textcolor{red}{(-58)} & 57.75\, \textcolor{red}{(-126)} & 28.51\, \textcolor{red}{(-112)} & 0.00\, \textcolor{red}{(-921)} & 0.00\, \textcolor{red}{(-966)} & 0.00\, \textcolor{red}{(-917)} & 38.71\, \textcolor{red}{(-124)} & 0.00\, \textcolor{red}{(-1013)} & 0.00\, \textcolor{red}{(-1008)} \\
phi4-qwq-sky-t1 & 42.42 & 32.67\, \textcolor{red}{(-31)} & 56.15\, \textcolor{blue}{(+18)} & 56.60\, \textcolor{red}{(-185)} & 26.40\, \textcolor{red}{(-342)} & 0.00\, \textcolor{red}{(-918)} & 0.00\, \textcolor{red}{(-964)} & 0.00\, \textcolor{red}{(-945)} & 43.06\, \textcolor{blue}{(+14)} & 0.00\, \textcolor{red}{(-1065)} & 0.00\, \textcolor{red}{(-1055)} \\
Phi-4-AbliteratedRP & 42.42 & 33.91\, \textcolor{red}{(-7)} & 54.03\, \textcolor{red}{(-7)} & 56.09\, \textcolor{red}{(-209)} & 27.42\, \textcolor{red}{(-212)} & 0.00\, \textcolor{red}{(-919)} & 0.00\, \textcolor{red}{(-964)} & 0.00\, \textcolor{red}{(-925)} & 40.67\, \textcolor{red}{(-57)} & 0.00\, \textcolor{red}{(-1086)} & 0.00\, \textcolor{red}{(-1173)} \\
Blaze-14B-xElite & 42.42 & 33.54\, \textcolor{red}{(-13)} & 54.87\, \textcolor{blue}{(+5)} & 55.20\, \textcolor{red}{(-256)} & 27.34\, \textcolor{red}{(-214)} & 0.00\, \textcolor{red}{(-916)} & 0.00\, \textcolor{red}{(-965)} & 0.00\, \textcolor{red}{(-931)} & 42.61\, \textcolor{blue}{(+9)} & 0.00\, \textcolor{red}{(-1125)} & 0.00\, \textcolor{red}{(-1210)} \\
PathFinderAi3.0 & 42.42 & 33.03\, \textcolor{red}{(-21)} & 55.66\, \textcolor{blue}{(+18)} & 56.65\, \textcolor{red}{(-180)} & 28.09\, \textcolor{red}{(-142)} & 0.00\, \textcolor{red}{(-917)} & 12.78\, \textcolor{red}{(-601)} & 0.00\, \textcolor{red}{(-936)} & 43.68\, \textcolor{blue}{(+20)} & 30.12\, \textcolor{red}{(-210)} & 0.00\, \textcolor{red}{(-1042)} \\
openbuddy-llama3.3-70b-v24.1-131k & 41.92 & 33.45\, \textcolor{red}{(-14)} & 52.95\, \textcolor{red}{(-33)} & 58.67\, \textcolor{red}{(-95)} & 25.15\, \textcolor{red}{(-609)} & 0.00\, \textcolor{red}{(-920)} & 0.00\, \textcolor{red}{(-967)} & 0.00\, \textcolor{red}{(-932)} & 37.39\, \textcolor{red}{(-212)} & 0.00\, \textcolor{red}{(-985)} & 0.00\, \textcolor{red}{(-980)} \\
FluentlyLM-Prinum & 41.92 & 33.40\, \textcolor{red}{(-14)} & 53.66\, \textcolor{red}{(-11)} & 56.35\, \textcolor{red}{(-188)} & 27.61\, \textcolor{red}{(-185)} & 25.36\, \textcolor{red}{(-285)} & 10.81\, \textcolor{red}{(-787)} & 24.46\, \textcolor{red}{(-287)} & 40.59\, \textcolor{red}{(-59)} & 18.97\, \textcolor{red}{(-393)} & 64.75\, \textcolor{red}{(-118)} \\
Yi-1.5-9B & 41.92 & 34.08\, \textcolor{blue}{(+1)} & 52.64\, \textcolor{red}{(-47)} & 55.31\, \textcolor{red}{(-245)} & 28.68\, \textcolor{red}{(-92)} & 0.00\, \textcolor{red}{(-918)} & 0.00\, \textcolor{red}{(-963)} & 0.00\, \textcolor{red}{(-917)} & 40.86\, \textcolor{red}{(-47)} & 0.00\, \textcolor{red}{(-1116)} & 0.00\, \textcolor{red}{(-1205)} \\
Apollo-70B & 41.92 & 34.46\, \textcolor{blue}{(+7)} & 51.19\, \textcolor{red}{(-108)} & 58.93\, \textcolor{red}{(-85)} & 25.35\, \textcolor{red}{(-535)} & 0.00\, \textcolor{red}{(-917)} & 0.00\, \textcolor{red}{(-964)} & 0.00\, \textcolor{red}{(-911)} & 35.42\, \textcolor{red}{(-522)} & 0.00\, \textcolor{red}{(-975)} & 0.00\, \textcolor{red}{(-1034)} \\
phi-4-14b & 41.92 & 32.04\, \textcolor{red}{(-53)} & 55.48\, \textcolor{blue}{(+22)} & 57.28\, \textcolor{red}{(-142)} & 24.96\, \textcolor{red}{(-721)} & 0.00\, \textcolor{red}{(-916)} & 0.00\, \textcolor{red}{(-963)} & 0.00\, \textcolor{red}{(-965)} & 40.36\, \textcolor{red}{(-65)} & 0.00\, \textcolor{red}{(-1027)} & 0.00\, \textcolor{red}{(-993)} \\
phi-4-300steps & 41.92 & 31.75\, \textcolor{red}{(-76)} & 55.83\, \textcolor{blue}{(+25)} & 58.11\, \textcolor{red}{(-107)} & 24.61\, \textcolor{red}{(-831)} & 0.00\, \textcolor{red}{(-915)} & 0.00\, \textcolor{red}{(-961)} & 0.00\, \textcolor{red}{(-986)} & 40.04\, \textcolor{red}{(-73)} & 0.00\, \textcolor{red}{(-995)} & 0.00\, \textcolor{red}{(-946)} \\
EVA-Qwen2.5-14B-v0.2 & 41.92 & 32.96\, \textcolor{red}{(-16)} & 53.55\, \textcolor{red}{(-11)} & 60.14\, \textcolor{red}{(-41)} & 24.24\, \textcolor{red}{(-964)} & 0.00\, \textcolor{red}{(-914)} & 0.00\, \textcolor{red}{(-961)} & 0.00\, \textcolor{red}{(-931)} & 36.53\, \textcolor{red}{(-318)} & 0.00\, \textcolor{red}{(-934)} & 0.00\, \textcolor{red}{(-921)} \\
tempesthenno-0120 & 41.92 & 32.89\, \textcolor{red}{(-16)} & 53.25\, \textcolor{red}{(-16)} & 62.59\, \textcolor{blue}{(+21)} & 23.59\, \textcolor{red}{(-1177)} & 0.00\, \textcolor{red}{(-913)} & 0.00\, \textcolor{red}{(-960)} & 0.00\, \textcolor{red}{(-931)} & 34.41\, \textcolor{red}{(-803)} & 0.00\, \textcolor{red}{(-873)} & 0.00\, \textcolor{red}{(-833)} \\
Progenitor-V1.1-LLaMa-70B & 41.92 & 34.85\, \textcolor{blue}{(+16)} & 50.59\, \textcolor{red}{(-126)} & 58.74\, \textcolor{red}{(-84)} & 24.83\, \textcolor{red}{(-757)} & 0.00\, \textcolor{red}{(-912)} & 0.00\, \textcolor{red}{(-959)} & 0.00\, \textcolor{red}{(-903)} & 34.28\, \textcolor{red}{(-837)} & 0.00\, \textcolor{red}{(-974)} & 0.00\, \textcolor{red}{(-1042)} \\
SuperThoughts-CoT-14B-16k-o1-QwQ & 41.92 & 33.06\, \textcolor{red}{(-10)} & 54.36\, \textcolor{blue}{(+8)} & 54.89\, \textcolor{red}{(-274)} & 27.24\, \textcolor{red}{(-213)} & 0.00\, \textcolor{red}{(-911)} & 0.00\, \textcolor{red}{(-958)} & 0.00\, \textcolor{red}{(-925)} & 42.21\, \textcolor{blue}{(+10)} & 0.00\, \textcolor{red}{(-1140)} & 0.00\, \textcolor{red}{(-1209)} \\
Dracarys-72B-Instruct & 41.92 & 32.43\, \textcolor{red}{(-30)} & 54.50\, \textcolor{blue}{(+12)} & 58.39\, \textcolor{red}{(-97)} & 25.20\, \textcolor{red}{(-579)} & 0.00\, \textcolor{red}{(-910)} & 0.00\, \textcolor{red}{(-957)} & 0.00\, \textcolor{red}{(-943)} & 37.61\, \textcolor{red}{(-183)} & 0.00\, \textcolor{red}{(-985)} & 0.00\, \textcolor{red}{(-905)} \\
ultiima-14B-v0.4 & 41.41 & 31.26\, \textcolor{red}{(-113)} & 54.87\, \textcolor{blue}{(+19)} & 61.24\, \textcolor{red}{(-4)} & 23.53\, \textcolor{red}{(-1190)} & 0.00\, \textcolor{red}{(-920)} & 0.00\, \textcolor{red}{(-964)} & 0.00\, \textcolor{red}{(-1016)} & 37.18\, \textcolor{red}{(-228)} & 0.00\, \textcolor{red}{(-898)} & 0.00\, \textcolor{red}{(-835)} \\
L3.3-Nevoria-R1-70b & 41.41 & 34.18\, \textcolor{blue}{(+13)} & 50.30\, \textcolor{red}{(-151)} & 58.17\, \textcolor{red}{(-97)} & 24.46\, \textcolor{red}{(-886)} & 0.00\, \textcolor{red}{(-919)} & 0.00\, \textcolor{red}{(-963)} & 0.00\, \textcolor{red}{(-905)} & 33.60\, \textcolor{red}{(-1051)} & 0.00\, \textcolor{red}{(-985)} & 0.00\, \textcolor{red}{(-1064)} \\
Rombos-LLM-V2.5-Qwen-32b & 41.41 & 34.06\, \textcolor{blue}{(+10)} & 51.04\, \textcolor{red}{(-106)} & 55.59\, \textcolor{red}{(-216)} & 28.08\, \textcolor{red}{(-129)} & 0.00\, \textcolor{red}{(-930)} & 0.00\, \textcolor{red}{(-962)} & 0.00\, \textcolor{red}{(-908)} & 37.97\, \textcolor{red}{(-146)} & 0.00\, \textcolor{red}{(-1089)} & 0.00\, \textcolor{red}{(-1192)} \\
orca\_mini\_phi-4 & 41.41 & 35.56\, \textcolor{blue}{(+29)} & 48.94\, \textcolor{red}{(-256)} & 53.96\, \textcolor{red}{(-338)} & 29.46\, \textcolor{red}{(-41)} & 25.36\, \textcolor{red}{(-273)} & 10.81\, \textcolor{red}{(-775)} & 24.46\, \textcolor{red}{(-271)} & 37.86\, \textcolor{red}{(-159)} & 18.97\, \textcolor{red}{(-382)} & 64.75\, \textcolor{red}{(-108)} \\
merge\_model\_20250308\_4 & 41.41 & 32.83\, \textcolor{red}{(-10)} & 52.13\, \textcolor{red}{(-51)} & 61.45\, \textcolor{blue}{(+2)} & 24.09\, \textcolor{red}{(-1009)} & 0.00\, \textcolor{red}{(-916)} & 0.00\, \textcolor{red}{(-960)} & 0.00\, \textcolor{red}{(-924)} & 34.58\, \textcolor{red}{(-748)} & 0.00\, \textcolor{red}{(-892)} & 0.00\, \textcolor{red}{(-854)} \\
Meta-Llama-3-70B & 41.41 & 33.56\, \textcolor{blue}{(+6)} & 51.95\, \textcolor{red}{(-58)} & 55.26\, \textcolor{red}{(-234)} & 26.02\, \textcolor{red}{(-375)} & 0.00\, \textcolor{red}{(-914)} & 0.00\, \textcolor{red}{(-959)} & 0.00\, \textcolor{red}{(-912)} & 39.40\, \textcolor{red}{(-83)} & 0.00\, \textcolor{red}{(-1105)} & 0.00\, \textcolor{red}{(-1299)} \\
Qwen2.5-14B-ReasoningMerge & 41.41 & 32.13\, \textcolor{red}{(-33)} & 53.10\, \textcolor{red}{(-9)} & 62.15\, \textcolor{blue}{(+25)} & 22.99\, \textcolor{red}{(-1376)} & 0.00\, \textcolor{red}{(-914)} & 0.00\, \textcolor{red}{(-958)} & 0.00\, \textcolor{red}{(-946)} & 34.28\, \textcolor{red}{(-827)} & 0.00\, \textcolor{red}{(-869)} & 0.00\, \textcolor{red}{(-821)} \\
YiSM-34B-0rn & 41.41 & 35.25\, \textcolor{blue}{(+27)} & 49.66\, \textcolor{red}{(-197)} & 52.59\, \textcolor{red}{(-434)} & 29.52\, \textcolor{red}{(-34)} & 0.00\, \textcolor{red}{(-913)} & 0.00\, \textcolor{red}{(-957)} & 0.00\, \textcolor{red}{(-892)} & 39.51\, \textcolor{red}{(-72)} & 0.00\, \textcolor{red}{(-1287)} & 0.00\, \textcolor{red}{(-1466)} \\
phi-4-abliterated & 41.41 & 31.60\, \textcolor{red}{(-83)} & 54.86\, \textcolor{blue}{(+25)} & 56.99\, \textcolor{red}{(-139)} & 25.09\, \textcolor{red}{(-605)} & 0.00\, \textcolor{red}{(-912)} & 0.00\, \textcolor{red}{(-956)} & 0.00\, \textcolor{red}{(-988)} & 39.89\, \textcolor{red}{(-63)} & 0.00\, \textcolor{red}{(-1023)} & 0.00\, \textcolor{red}{(-992)} \\
Epimetheus-14B-Axo & 41.41 & 32.51\, \textcolor{red}{(-17)} & 52.59\, \textcolor{red}{(-32)} & 61.90\, \textcolor{blue}{(+22)} & 23.91\, \textcolor{red}{(-1054)} & 0.00\, \textcolor{red}{(-923)} & 0.00\, \textcolor{red}{(-955)} & 0.00\, \textcolor{red}{(-931)} & 34.62\, \textcolor{red}{(-736)} & 0.00\, \textcolor{red}{(-872)} & 0.00\, \textcolor{red}{(-827)} \\
orca\_mini\_v9\_2\_14B & 41.41 & 35.56\, \textcolor{blue}{(+36)} & 48.94\, \textcolor{red}{(-249)} & 53.96\, \textcolor{red}{(-331)} & 29.46\, \textcolor{red}{(-34)} & 25.36\, \textcolor{red}{(-266)} & 10.81\, \textcolor{red}{(-768)} & 24.46\, \textcolor{red}{(-264)} & 37.86\, \textcolor{red}{(-152)} & 18.97\, \textcolor{red}{(-375)} & 64.75\, \textcolor{red}{(-101)} \\
Qwen2.5-14B-Vimarckoso & 41.41 & 31.84\, \textcolor{red}{(-51)} & 53.55\, \textcolor{blue}{(+6)} & 62.89\, \textcolor{blue}{(+42)} & 22.37\, \textcolor{red}{(-1627)} & 0.00\, \textcolor{red}{(-909)} & 0.00\, \textcolor{red}{(-953)} & 0.00\, \textcolor{red}{(-962)} & 34.26\, \textcolor{red}{(-830)} & 0.00\, \textcolor{red}{(-852)} & 0.00\, \textcolor{red}{(-812)} \\
Qwentessential-14B-v1 & 41.41 & 32.08\, \textcolor{red}{(-31)} & 53.02\, \textcolor{red}{(-8)} & 63.49\, \textcolor{blue}{(+47)} & 22.00\, \textcolor{red}{(-1778)} & 0.00\, \textcolor{red}{(-908)} & 0.00\, \textcolor{red}{(-952)} & 0.00\, \textcolor{red}{(-944)} & 33.21\, \textcolor{red}{(-1197)} & 0.00\, \textcolor{red}{(-847)} & 0.00\, \textcolor{red}{(-807)} \\
Mistral-Small-24B-Instruct-2501-bf16 & 41.41 & 33.50\, \textcolor{blue}{(+12)} & 51.38\, \textcolor{red}{(-78)} & 59.81\, \textcolor{red}{(-35)} & 24.77\, \textcolor{red}{(-764)} & 0.00\, \textcolor{red}{(-907)} & 0.00\, \textcolor{red}{(-951)} & 0.00\, \textcolor{red}{(-906)} & 34.96\, \textcolor{red}{(-617)} & 0.00\, \textcolor{red}{(-928)} & 0.00\, \textcolor{red}{(-920)} \\
\bottomrule
\end{tabular}}
\label{tab:meme_scores_summary_OpenLLM-GPQA-Diamond}
\end{table}

\begin{table}[H]
\centering
\caption{\textbf{Meme Scores (Open LLM Population; IFEval).} The table reports meme scores across models, including property-derived 1D meme scores, predefined 2D meme scores (Mastery, Ingenuity, and Robustness), and a predefined 3D meme score (Caution). Models are sorted by Accuracy, and only the \textbf{top-50 models} are shown. {\color{blue}Blue} indicates rank improvement compared with Accuracy rank, and {\color{red}red} indicates rank degradation.}
\resizebox{\textwidth}{!}{
\begin{tabular}{cccccccccccc}
\toprule
\multirow{2}{*}{\textbf{Model}} & \multirow{2}{*}{\textbf{Accuracy}} &
\multicolumn{6}{c}{\textbf{Property-derived 1D meme scores}} &
\multicolumn{4}{c}{\textbf{Predefined 2D/3D meme scores}} \\
\cmidrule(lr){3-8}\cmidrule(lr){9-12}
 &  & \textbf{Difficulty} & \textbf{Uniqueness} & \textbf{Risk} & \textbf{Surprise} & \textbf{Typicality} & \textbf{Bridge} & \textbf{Mastery} & \textbf{Ingenuity} & \textbf{Robustness} & \textbf{Caution} \\
\midrule
Llama-3.3-70B-Instruct & 87.99 & 81.24 & 87.13 & 95.20 & 59.54 & 84.76 & 36.65 & 75.60 & 57.81\, \textcolor{red}{(-1)} & 55.35 & 98.26 \\
Llama-3.1-70B-Instruct & 84.29 & 75.86\, \textcolor{red}{(-3)} & 82.31\, \textcolor{red}{(-1)} & 93.79 & 54.15\, \textcolor{red}{(-7)} & 80.94\, \textcolor{red}{(-3)} & 14.08\, \textcolor{red}{(-153)} & 70.01\, \textcolor{red}{(-3)} & 51.77\, \textcolor{red}{(-12)} & 34.44\, \textcolor{red}{(-125)} & 97.55\, \textcolor{red}{(-1)} \\
Qwen2.5-72B-Instruct & 83.92 & 76.17\, \textcolor{red}{(-1)} & 81.06\, \textcolor{red}{(-7)} & 93.14\, \textcolor{red}{(-2)} & 55.84\, \textcolor{red}{(-2)} & 81.54 & 23.90\, \textcolor{red}{(-3)} & 71.80 & 52.97\, \textcolor{red}{(-5)} & 42.15\, \textcolor{red}{(-5)} & 96.60\, \textcolor{red}{(-7)} \\
Llama-3.1-SauerkrautLM-70b-Instruct & 83.92 & 76.44\, \textcolor{blue}{(+2)} & 80.41\, \textcolor{red}{(-11)} & 93.37 & 53.05\, \textcolor{red}{(-9)} & 82.19\, \textcolor{blue}{(+2)} & 19.55\, \textcolor{red}{(-20)} & 73.12\, \textcolor{blue}{(+2)} & 49.41\, \textcolor{red}{(-41)} & 40.17\, \textcolor{red}{(-13)} & 96.74\, \textcolor{red}{(-4)} \\
calme-2.1-qwen2.5-72b & 83.92 & 76.21\, \textcolor{blue}{(+2)} & 81.56\, \textcolor{red}{(-1)} & 93.14\, \textcolor{red}{(-1)} & 56.57\, \textcolor{blue}{(+1)} & 81.23\, \textcolor{blue}{(+1)} & 26.24\, \textcolor{blue}{(+2)} & 71.31\, \textcolor{blue}{(+1)} & 54.40 & 45.16\, \textcolor{blue}{(+2)} & 96.86 \\
ZYH-LLM-Qwen2.5-14B-V3 & 83.55 & 75.18 & 82.99\, \textcolor{blue}{(+4)} & 92.48\, \textcolor{red}{(-4)} & 56.84\, \textcolor{blue}{(+3)} & 79.57\, \textcolor{red}{(-4)} & 20.63\, \textcolor{red}{(-8)} & 68.17\, \textcolor{red}{(-8)} & 55.44\, \textcolor{blue}{(+2)} & 39.98\, \textcolor{red}{(-15)} & 96.78\, \textcolor{red}{(-1)} \\
calme-2.3-llama3.1-70b & 83.36 & 74.90\, \textcolor{red}{(-1)} & 81.61\, \textcolor{blue}{(+2)} & 92.72\, \textcolor{red}{(-1)} & 52.44\, \textcolor{red}{(-13)} & 79.05\, \textcolor{red}{(-7)} & 14.68\, \textcolor{red}{(-119)} & 68.66\, \textcolor{red}{(-6)} & 50.28\, \textcolor{red}{(-19)} & 26.72\, \textcolor{red}{(-1143)} & 94.11\, \textcolor{red}{(-119)} \\
Qwen2.5-72B-Instruct-abliterated & 83.36 & 74.55\, \textcolor{red}{(-2)} & 81.22 & 93.55\, \textcolor{blue}{(+5)} & 53.47\, \textcolor{red}{(-2)} & 80.63\, \textcolor{blue}{(+2)} & 23.61\, \textcolor{blue}{(+1)} & 69.66\, \textcolor{blue}{(+1)} & 50.88\, \textcolor{red}{(-12)} & 43.20\, \textcolor{blue}{(+2)} & 97.70\, \textcolor{blue}{(+6)} \\
calme-2.2-llama3.1-70b & 83.36 & 74.95\, \textcolor{blue}{(+2)} & 81.99\, \textcolor{blue}{(+5)} & 92.68 & 52.09\, \textcolor{red}{(-18)} & 80.23\, \textcolor{blue}{(+2)} & 18.29\, \textcolor{red}{(-24)} & 69.37 & 49.99\, \textcolor{red}{(-22)} & 39.01\, \textcolor{red}{(-20)} & 96.78\, \textcolor{blue}{(+3)} \\
orca\_mini\_v8\_1\_70b & 83.18 & 74.70\, \textcolor{blue}{(+1)} & 81.12\, \textcolor{blue}{(+1)} & 92.84\, \textcolor{blue}{(+3)} & 52.41\, \textcolor{red}{(-12)} & 80.17\, \textcolor{blue}{(+2)} & 20.51\, \textcolor{red}{(-8)} & 69.30\, \textcolor{red}{(-1)} & 49.62\, \textcolor{red}{(-28)} & 40.70\, \textcolor{red}{(-3)} & 96.86\, \textcolor{blue}{(+6)} \\
Gilgamesh-72B & 82.07 & 73.58\, \textcolor{red}{(-1)} & 81.26\, \textcolor{blue}{(+4)} & 91.43\, \textcolor{red}{(-11)} & 57.58\, \textcolor{blue}{(+9)} & 79.99\, \textcolor{blue}{(+2)} & 20.61\, \textcolor{red}{(-4)} & 69.68\, \textcolor{blue}{(+5)} & 57.13\, \textcolor{blue}{(+8)} & 39.29\, \textcolor{red}{(-13)} & 96.01\, \textcolor{red}{(-15)} \\
RYS-Llama3.1-Large & 82.07 & 73.29\, \textcolor{red}{(-1)} & 80.52\, \textcolor{red}{(-2)} & 91.81\, \textcolor{red}{(-2)} & 51.78\, \textcolor{red}{(-18)} & 79.57\, \textcolor{blue}{(+1)} & 17.81\, \textcolor{red}{(-30)} & 68.69 & 49.52\, \textcolor{red}{(-28)} & 38.35\, \textcolor{red}{(-28)} & 96.17\, \textcolor{red}{(-8)} \\
calme-2.2-qwen2.5-72b & 81.89 & 73.83\, \textcolor{blue}{(+2)} & 79.54\, \textcolor{red}{(-14)} & 91.70\, \textcolor{red}{(-4)} & 55.69\, \textcolor{blue}{(+6)} & 79.52\, \textcolor{blue}{(+1)} & 28.34\, \textcolor{blue}{(+11)} & 69.36\, \textcolor{blue}{(+3)} & 53.13\, \textcolor{blue}{(+6)} & 45.45\, \textcolor{blue}{(+11)} & 96.05\, \textcolor{red}{(-11)} \\
Qwen2.5-14B-Instruct-1M & 81.70 & 73.16 & 80.81\, \textcolor{blue}{(+3)} & 90.91\, \textcolor{red}{(-19)} & 52.47\, \textcolor{red}{(-4)} & 78.19\, \textcolor{red}{(-8)} & 20.03\, \textcolor{red}{(-6)} & 66.84\, \textcolor{red}{(-9)} & 51.38\, \textcolor{red}{(-2)} & 41.17\, \textcolor{blue}{(+5)} & 95.31\, \textcolor{red}{(-31)} \\
Qwen2.5-14B-Instruct-1M-GRPO-Reasoning & 81.70 & 73.16\, \textcolor{blue}{(+1)} & 80.81\, \textcolor{blue}{(+4)} & 90.91\, \textcolor{red}{(-18)} & 52.47\, \textcolor{red}{(-3)} & 78.19\, \textcolor{red}{(-7)} & 20.03\, \textcolor{red}{(-5)} & 66.84\, \textcolor{red}{(-8)} & 51.38\, \textcolor{red}{(-1)} & 41.17\, \textcolor{blue}{(+6)} & 95.31\, \textcolor{red}{(-30)} \\
Rombos-LLM-V2.6-Qwen-14b & 81.70 & 73.00 & 79.56\, \textcolor{red}{(-9)} & 91.80\, \textcolor{blue}{(+1)} & 52.19\, \textcolor{red}{(-9)} & 77.55\, \textcolor{red}{(-12)} & 12.36\, \textcolor{red}{(-282)} & 65.69\, \textcolor{red}{(-23)} & 49.50\, \textcolor{red}{(-25)} & 33.19\, \textcolor{red}{(-185)} & 95.99\, \textcolor{red}{(-11)} \\
calme-2.1-llama3.1-70b & 81.52 & 72.57\, \textcolor{red}{(-2)} & 80.30\, \textcolor{blue}{(+1)} & 91.47\, \textcolor{red}{(-4)} & 51.34\, \textcolor{red}{(-19)} & 78.49\, \textcolor{red}{(-2)} & 19.97\, \textcolor{red}{(-5)} & 67.09\, \textcolor{red}{(-4)} & 49.19\, \textcolor{red}{(-31)} & 40.91\, \textcolor{blue}{(+5)} & 96.18\, \textcolor{red}{(-2)} \\
Qwen2.5-14B-1M-YOYO-V3 & 81.52 & 72.64\, \textcolor{blue}{(+1)} & 80.04\, \textcolor{blue}{(+1)} & 91.55\, \textcolor{red}{(-1)} & 52.18\, \textcolor{red}{(-8)} & 77.97\, \textcolor{red}{(-6)} & 21.22\, \textcolor{blue}{(+6)} & 66.26\, \textcolor{red}{(-14)} & 50.36\, \textcolor{red}{(-7)} & 40.57\, \textcolor{blue}{(+3)} & 96.31\, \textcolor{blue}{(+3)} \\
test-2.5-72B & 81.33 & 72.35\, \textcolor{red}{(-4)} & 79.58\, \textcolor{red}{(-5)} & 91.59\, \textcolor{blue}{(+1)} & 55.73\, \textcolor{blue}{(+13)} & 79.03\, \textcolor{blue}{(+4)} & 17.98\, \textcolor{red}{(-20)} & 68.13\, \textcolor{blue}{(+4)} & 54.07\, \textcolor{blue}{(+13)} & 36.14\, \textcolor{red}{(-59)} & 96.16\, \textcolor{red}{(-3)} \\
virtuoso-small-v2-tensopolis-v1 & 81.33 & 72.41\, \textcolor{red}{(-1)} & 78.50\, \textcolor{red}{(-20)} & 91.92\, \textcolor{blue}{(+8)} & 50.78\, \textcolor{red}{(-28)} & 78.77\, \textcolor{blue}{(+3)} & 13.13\, \textcolor{red}{(-195)} & 67.68\, \textcolor{blue}{(+3)} & 47.84\, \textcolor{red}{(-63)} & 33.53\, \textcolor{red}{(-162)} & 96.23\, \textcolor{blue}{(+3)} \\
Qwen2.5-95B-Instruct & 81.33 & 72.02\, \textcolor{red}{(-4)} & 79.30\, \textcolor{red}{(-11)} & 92.00\, \textcolor{blue}{(+10)} & 51.32\, \textcolor{red}{(-16)} & 78.27 & 13.46\, \textcolor{red}{(-170)} & 66.51\, \textcolor{red}{(-6)} & 49.03\, \textcolor{red}{(-31)} & 34.44\, \textcolor{red}{(-107)} & 96.60\, \textcolor{blue}{(+10)} \\
EXAONE-3.5-32B-Instruct & 81.15 & 72.45\, \textcolor{blue}{(+2)} & 78.06\, \textcolor{red}{(-26)} & 91.81\, \textcolor{blue}{(+9)} & 50.72\, \textcolor{red}{(-27)} & 77.84\, \textcolor{red}{(-5)} & 16.52\, \textcolor{red}{(-38)} & 66.42\, \textcolor{red}{(-7)} & 47.08\, \textcolor{red}{(-84)} & 37.34\, \textcolor{red}{(-27)} & 96.12\, \textcolor{red}{(-1)} \\
solar-pro-preview-instruct & 81.15 & 71.96\, \textcolor{red}{(-3)} & 79.79\, \textcolor{blue}{(+3)} & 91.49\, \textcolor{blue}{(+3)} & 51.83\, \textcolor{red}{(-6)} & 78.34\, \textcolor{blue}{(+3)} & 17.12\, \textcolor{red}{(-32)} & 66.85\, \textcolor{blue}{(+1)} & 48.60\, \textcolor{red}{(-40)} & 38.13\, \textcolor{red}{(-18)} & 96.39\, \textcolor{blue}{(+9)} \\
Qwen2.5-14B-YOYO-V4 & 81.15 & 72.27 & 80.75\, \textcolor{blue}{(+11)} & 90.64\, \textcolor{red}{(-18)} & 52.62\, \textcolor{blue}{(+9)} & 78.88\, \textcolor{blue}{(+8)} & 21.29\, \textcolor{blue}{(+13)} & 68.02\, \textcolor{blue}{(+8)} & 51.60\, \textcolor{blue}{(+9)} & 40.66\, \textcolor{blue}{(+10)} & 95.63\, \textcolor{red}{(-15)} \\
orca\_mini\_v9\_2\_70b & 80.96 & 72.60\, \textcolor{blue}{(+7)} & 79.54\, \textcolor{red}{(-3)} & 90.36\, \textcolor{red}{(-19)} & 52.47\, \textcolor{blue}{(+8)} & 77.49\, \textcolor{red}{(-8)} & 20.41\, \textcolor{blue}{(+6)} & 67.15\, \textcolor{blue}{(+5)} & 50.46\, \textcolor{blue}{(+2)} & 40.16\, \textcolor{blue}{(+7)} & 93.48\, \textcolor{red}{(-172)} \\
Mistral-Large-Instruct-2411 & 80.96 & 72.36\, \textcolor{blue}{(+4)} & 78.48\, \textcolor{red}{(-15)} & 91.24\, \textcolor{blue}{(+3)} & 52.22\, \textcolor{blue}{(+2)} & 79.46\, \textcolor{blue}{(+13)} & 13.13\, \textcolor{red}{(-188)} & 69.37\, \textcolor{blue}{(+18)} & 48.83\, \textcolor{red}{(-32)} & 33.54\, \textcolor{red}{(-155)} & 95.83\, \textcolor{red}{(-6)} \\
ZYH-LLM-Qwen2.5-14B-V4 & 80.96 & 71.95 & 79.80\, \textcolor{blue}{(+8)} & 90.96\, \textcolor{red}{(-3)} & 51.27\, \textcolor{red}{(-11)} & 77.50\, \textcolor{red}{(-5)} & 15.38\, \textcolor{red}{(-73)} & 65.71\, \textcolor{red}{(-11)} & 49.88\, \textcolor{red}{(-7)} & 35.96\, \textcolor{red}{(-55)} & 95.75\, \textcolor{red}{(-9)} \\
Cheng-2 & 80.78 & 71.35\, \textcolor{red}{(-2)} & 80.03\, \textcolor{blue}{(+10)} & 90.85\, \textcolor{red}{(-7)} & 53.42\, \textcolor{blue}{(+17)} & 77.53\, \textcolor{red}{(-2)} & 13.06\, \textcolor{red}{(-189)} & 65.50\, \textcolor{red}{(-13)} & 52.35\, \textcolor{blue}{(+19)} & 32.42\, \textcolor{red}{(-232)} & 95.72\, \textcolor{red}{(-9)} \\
Llama-3.1-Tulu-3-70B & 80.78 & 71.59 & 79.36\, \textcolor{red}{(-1)} & 91.14\, \textcolor{blue}{(+4)} & 50.39\, \textcolor{red}{(-24)} & 78.66\, \textcolor{blue}{(+11)} & 15.95\, \textcolor{red}{(-47)} & 67.59\, \textcolor{blue}{(+11)} & 47.79\, \textcolor{red}{(-57)} & 36.86\, \textcolor{red}{(-30)} & 96.21\, \textcolor{blue}{(+11)} \\
zetasepic-abliteratedV2-Qwen2.5-32B-Inst-BaseMerge-TIES & 80.59 & 71.73\, \textcolor{blue}{(+2)} & 78.52\, \textcolor{red}{(-9)} & 90.96\, \textcolor{blue}{(+2)} & 51.36\, \textcolor{red}{(-4)} & 77.55\, \textcolor{blue}{(+1)} & 15.74\, \textcolor{red}{(-57)} & 66.16\, \textcolor{red}{(-3)} & 48.59\, \textcolor{red}{(-34)} & 36.87\, \textcolor{red}{(-28)} & 95.58\, \textcolor{red}{(-11)} \\
Qwen2.5-32B-Instruct & 80.59 & 70.91\, \textcolor{red}{(-4)} & 78.62\, \textcolor{red}{(-7)} & 91.77\, \textcolor{blue}{(+15)} & 51.53\, \textcolor{red}{(-1)} & 77.97\, \textcolor{blue}{(+6)} & 11.22\, \textcolor{red}{(-406)} & 66.15\, \textcolor{red}{(-3)} & 48.76\, \textcolor{red}{(-28)} & 31.24\, \textcolor{red}{(-352)} & 96.69\, \textcolor{blue}{(+22)} \\
Qwen2.5-32B-Instruct-abliterated-v2 & 80.59 & 71.26 & 79.27\, \textcolor{red}{(-1)} & 91.01\, \textcolor{blue}{(+6)} & 53.36\, \textcolor{blue}{(+20)} & 77.88\, \textcolor{blue}{(+6)} & 13.21\, \textcolor{red}{(-174)} & 66.31\, \textcolor{blue}{(+2)} & 52.09\, \textcolor{blue}{(+22)} & 33.70\, \textcolor{red}{(-135)} & 95.88\, \textcolor{blue}{(+2)} \\
Q2.5-Veltha-14B & 80.22 & 71.30\, \textcolor{blue}{(+2)} & 77.42\, \textcolor{red}{(-28)} & 90.95\, \textcolor{blue}{(+2)} & 49.25\, \textcolor{red}{(-43)} & 76.99\, \textcolor{red}{(-5)} & 11.64\, \textcolor{red}{(-342)} & 65.38\, \textcolor{red}{(-9)} & 45.61\, \textcolor{red}{(-145)} & 31.94\, \textcolor{red}{(-278)} & 95.53\, \textcolor{red}{(-9)} \\
Josiefied-Qwen2.5-14B-Instruct-abliterated-v4 & 80.22 & 70.60\, \textcolor{red}{(-4)} & 79.33\, \textcolor{blue}{(+3)} & 90.81\, \textcolor{red}{(-2)} & 52.74\, \textcolor{blue}{(+20)} & 76.58\, \textcolor{red}{(-8)} & 15.49\, \textcolor{red}{(-62)} & 64.06\, \textcolor{red}{(-34)} & 50.88\, \textcolor{blue}{(+15)} & 35.54\, \textcolor{red}{(-56)} & 95.94\, \textcolor{blue}{(+5)} \\
Qwen2.5-14B-Instruct-abliterated-v2 & 80.22 & 70.41\, \textcolor{red}{(-4)} & 79.74\, \textcolor{blue}{(+14)} & 90.69\, \textcolor{red}{(-4)} & 51.77\, \textcolor{blue}{(+4)} & 76.23\, \textcolor{red}{(-14)} & 13.06\, \textcolor{red}{(-183)} & 63.39\, \textcolor{red}{(-58)} & 50.52\, \textcolor{blue}{(+13)} & 33.88\, \textcolor{red}{(-125)} & 95.81\, \textcolor{blue}{(+2)} \\
Configurable-Llama-3.1-8B-Instruct & 80.04 & 70.37\, \textcolor{red}{(-4)} & 78.67\, \textcolor{red}{(-1)} & 90.80\, \textcolor{red}{(-1)} & 50.20\, \textcolor{red}{(-21)} & 75.21\, \textcolor{red}{(-43)} & 18.57\, \textcolor{blue}{(+5)} & 61.88\, \textcolor{red}{(-106)} & 47.80\, \textcolor{red}{(-49)} & 38.44\, \textcolor{red}{(-3)} & 95.80\, \textcolor{blue}{(+1)} \\
Virtuoso-Small-v2 & 79.85 & 69.79\, \textcolor{red}{(-9)} & 78.27\, \textcolor{red}{(-9)} & 91.18\, \textcolor{blue}{(+13)} & 51.19\, \textcolor{red}{(-3)} & 77.05 & 14.88\, \textcolor{red}{(-84)} & 64.67\, \textcolor{red}{(-18)} & 48.72\, \textcolor{red}{(-23)} & 34.07\, \textcolor{red}{(-112)} & 96.48\, \textcolor{blue}{(+24)} \\
lambda-qwen2.5-14b-dpo-test & 79.85 & 70.68\, \textcolor{blue}{(+1)} & 78.95\, \textcolor{blue}{(+4)} & 90.00\, \textcolor{red}{(-11)} & 51.47\, \textcolor{blue}{(+5)} & 75.83\, \textcolor{red}{(-23)} & 15.25\, \textcolor{red}{(-67)} & 63.49\, \textcolor{red}{(-50)} & 50.10\, \textcolor{blue}{(+9)} & 35.47\, \textcolor{red}{(-55)} & 95.08\, \textcolor{red}{(-17)} \\
Josiefied-abliteratedV4-Qwen2.5-14B-Inst-BaseMerge-TIES & 79.67 & 70.07\, \textcolor{red}{(-3)} & 78.35\, \textcolor{red}{(-5)} & 90.51\, \textcolor{red}{(-4)} & 49.61\, \textcolor{red}{(-28)} & 75.53\, \textcolor{red}{(-30)} & 18.20\, \textcolor{blue}{(+4)} & 62.72\, \textcolor{red}{(-74)} & 47.01\, \textcolor{red}{(-71)} & 38.54\, \textcolor{blue}{(+1)} & 95.65\, \textcolor{blue}{(+1)} \\
Llama-3.1-Tulu-3-70B-DPO & 79.67 & 70.93\, \textcolor{blue}{(+7)} & 77.40\, \textcolor{red}{(-22)} & 90.05\, \textcolor{red}{(-8)} & 48.75\, \textcolor{red}{(-56)} & 77.52\, \textcolor{blue}{(+9)} & 16.81\, \textcolor{red}{(-16)} & 66.75\, \textcolor{blue}{(+15)} & 45.50\, \textcolor{red}{(-148)} & 38.12\, \textcolor{red}{(-3)} & 95.01\, \textcolor{red}{(-20)} \\
Cheng-2-v1.1 & 79.67 & 69.84\, \textcolor{red}{(-4)} & 79.63\, \textcolor{blue}{(+19)} & 89.95\, \textcolor{red}{(-9)} & 52.55\, \textcolor{blue}{(+25)} & 75.97\, \textcolor{red}{(-15)} & 10.93\, \textcolor{red}{(-444)} & 63.58\, \textcolor{red}{(-39)} & 52.02\, \textcolor{blue}{(+30)} & 25.15\, \textcolor{red}{(-1466)} & 94.91\, \textcolor{red}{(-25)} \\
Dinobot-Opus-14B-Exp & 79.67 & 69.98\, \textcolor{red}{(-1)} & 78.27\, \textcolor{red}{(-3)} & 90.65\, \textcolor{blue}{(+1)} & 49.58\, \textcolor{red}{(-26)} & 75.53\, \textcolor{red}{(-27)} & 18.20\, \textcolor{blue}{(+7)} & 62.64\, \textcolor{red}{(-73)} & 46.96\, \textcolor{red}{(-70)} & 38.54\, \textcolor{blue}{(+5)} & 95.81\, \textcolor{blue}{(+8)} \\
Linkbricks-Horizon-AI-Avengers-V3-32B & 79.48 & 69.40\, \textcolor{red}{(-10)} & 77.81\, \textcolor{red}{(-7)} & 90.97\, \textcolor{blue}{(+16)} & 50.23\, \textcolor{red}{(-12)} & 76.74\, \textcolor{blue}{(+3)} & 13.27\, \textcolor{red}{(-161)} & 64.39\, \textcolor{red}{(-16)} & 47.56\, \textcolor{red}{(-47)} & 33.73\, \textcolor{red}{(-122)} & 96.28\, \textcolor{blue}{(+27)} \\
Awqward2.5-32B-Instruct & 79.48 & 69.55\, \textcolor{red}{(-5)} & 77.58\, \textcolor{red}{(-11)} & 90.96\, \textcolor{blue}{(+15)} & 51.25\, \textcolor{blue}{(+5)} & 76.99\, \textcolor{blue}{(+5)} & 11.22\, \textcolor{red}{(-394)} & 64.95\, \textcolor{red}{(-2)} & 48.86\, \textcolor{red}{(-12)} & 31.24\, \textcolor{red}{(-340)} & 96.17\, \textcolor{blue}{(+23)} \\
miscii-14b-1028 & 79.48 & 68.84\, \textcolor{red}{(-16)} & 79.44\, \textcolor{blue}{(+16)} & 90.73\, \textcolor{blue}{(+7)} & 51.85\, \textcolor{blue}{(+17)} & 77.07\, \textcolor{blue}{(+9)} & 12.95\, \textcolor{red}{(-179)} & 64.50\, \textcolor{red}{(-13)} & 51.19\, \textcolor{blue}{(+27)} & 32.94\, \textcolor{red}{(-166)} & 96.53\, \textcolor{blue}{(+33)} \\
Llama-3.1-Tulu-3-8B & 79.48 & 69.94\, \textcolor{blue}{(+2)} & 79.54\, \textcolor{blue}{(+20)} & 89.53\, \textcolor{red}{(-16)} & 50.83 & 76.60\, \textcolor{blue}{(+5)} & 17.77\, \textcolor{blue}{(+3)} & 64.66\, \textcolor{red}{(-10)} & 49.07\, \textcolor{red}{(-5)} & 38.60\, \textcolor{blue}{(+10)} & 95.06\, \textcolor{red}{(-10)} \\
Dearly\_Beloved-8B-TIES & 79.48 & 70.33\, \textcolor{blue}{(+6)} & 78.92\, \textcolor{blue}{(+12)} & 89.53\, \textcolor{red}{(-16)} & 50.22\, \textcolor{red}{(-9)} & 76.48\, \textcolor{blue}{(+2)} & 18.26\, \textcolor{blue}{(+13)} & 64.80\, \textcolor{red}{(-3)} & 48.21\, \textcolor{red}{(-27)} & 38.89\, \textcolor{blue}{(+17)} & 94.89\, \textcolor{red}{(-21)} \\
Qwen2.5-14B-YOYO-V4-p1 & 79.30 & 69.64 & 78.26\, \textcolor{blue}{(+1)} & 89.92\, \textcolor{red}{(-3)} & 50.45\, \textcolor{red}{(-4)} & 76.28 & 10.29\, \textcolor{red}{(-528)} & 64.11\, \textcolor{red}{(-17)} & 48.68\, \textcolor{red}{(-13)} & 31.04\, \textcolor{red}{(-363)} & 95.24 \\
Qwen2.5-14B-Gutenberg-Instruct-Slerpeno & 79.30 & 69.31\, \textcolor{red}{(-8)} & 79.63\, \textcolor{blue}{(+26)} & 89.67\, \textcolor{red}{(-8)} & 52.41\, \textcolor{blue}{(+28)} & 75.07\, \textcolor{red}{(-37)} & 11.67\, \textcolor{red}{(-322)} & 62.21\, \textcolor{red}{(-82)} & 51.93\, \textcolor{blue}{(+36)} & 25.95\, \textcolor{red}{(-1222)} & 94.83\, \textcolor{red}{(-21)} \\
Qwen2.5-14B-it-restore & 79.30 & 69.32\, \textcolor{red}{(-6)} & 78.70\, \textcolor{blue}{(+14)} & 90.12\, \textcolor{blue}{(+4)} & 50.84\, \textcolor{blue}{(+6)} & 74.75\, \textcolor{red}{(-51)} & 12.82\, \textcolor{red}{(-189)} & 61.69\, \textcolor{red}{(-99)} & 50.02\, \textcolor{blue}{(+20)} & 27.79\, \textcolor{red}{(-871)} & 95.03\, \textcolor{red}{(-9)} \\
\bottomrule
\end{tabular}}
\label{tab:meme_scores_summary_OpenLLM-IFEval}
\end{table}

\begin{table}[H]
\centering
\caption{\textbf{Meme Scores (Open LLM Population; MATH).} The table reports meme scores across models, including property-derived 1D meme scores, predefined 2D meme scores (Mastery, Ingenuity, and Robustness), and a predefined 3D meme score (Caution). Models are sorted by Accuracy, and only the \textbf{top-50 models} are shown. {\color{blue}Blue} indicates rank improvement compared with Accuracy rank, and {\color{red}red} indicates rank degradation.}
\resizebox{\textwidth}{!}{
\begin{tabular}{cccccccccccc}
\toprule
\multirow{2}{*}{\textbf{Model}} & \multirow{2}{*}{\textbf{Accuracy}} &
\multicolumn{6}{c}{\textbf{Property-derived 1D meme scores}} &
\multicolumn{4}{c}{\textbf{Predefined 2D/3D meme scores}} \\
\cmidrule(lr){3-8}\cmidrule(lr){9-12}
 &  & \textbf{Difficulty} & \textbf{Uniqueness} & \textbf{Risk} & \textbf{Surprise} & \textbf{Typicality} & \textbf{Bridge} & \textbf{Mastery} & \textbf{Ingenuity} & \textbf{Robustness} & \textbf{Caution} \\
\midrule
Gauss-Opus-14B-R999 & 57.55 & 51.04 & 86.38\, \textcolor{red}{(-6)} & 84.85\, \textcolor{red}{(-2)} & 36.48\, \textcolor{red}{(-4)} & 76.48\, \textcolor{red}{(-3)} & 37.16\, \textcolor{red}{(-1)} & 72.31\, \textcolor{red}{(-3)} & 74.21\, \textcolor{red}{(-7)} & 66.17 & 93.69\, \textcolor{red}{(-10)} \\
Sombrero-Opus-14B-Sm1 & 56.65 & 49.91 & 86.69\, \textcolor{red}{(-3)} & 84.44\, \textcolor{red}{(-6)} & 37.17 & 77.33\, \textcolor{blue}{(+1)} & 34.87\, \textcolor{red}{(-2)} & 73.25\, \textcolor{blue}{(+1)} & 75.85 & 64.95\, \textcolor{red}{(-1)} & 93.78\, \textcolor{red}{(-8)} \\
Viper-Coder-v1.6-r999 & 56.57 & 49.73 & 86.89 & 85.10\, \textcolor{blue}{(+2)} & 36.35\, \textcolor{red}{(-4)} & 76.99\, \textcolor{blue}{(+1)} & 35.10 & 72.68\, \textcolor{blue}{(+1)} & 75.02\, \textcolor{red}{(-2)} & 65.06\, \textcolor{blue}{(+1)} & 94.64 \\
o-distil-qwen & 56.50 & 49.71 & 86.72 & 84.53\, \textcolor{red}{(-2)} & 36.86\, \textcolor{blue}{(+1)} & 75.92\, \textcolor{red}{(-4)} & 34.46\, \textcolor{red}{(-2)} & 71.56\, \textcolor{red}{(-6)} & 74.89\, \textcolor{red}{(-2)} & 63.91 & 93.63\, \textcolor{red}{(-8)} \\
Apollo-70B & 56.12 & 49.21\, \textcolor{red}{(-1)} & 87.03\, \textcolor{blue}{(+4)} & 84.31\, \textcolor{red}{(-4)} & 36.27\, \textcolor{red}{(-4)} & 76.86\, \textcolor{blue}{(+2)} & 33.45\, \textcolor{red}{(-4)} & 72.60\, \textcolor{blue}{(+2)} & 75.82\, \textcolor{blue}{(+2)} & 62.31\, \textcolor{red}{(-7)} & 94.12\, \textcolor{red}{(-2)} \\
li-14b-v0.4 & 55.74 & 48.70\, \textcolor{red}{(-2)} & 87.00\, \textcolor{blue}{(+4)} & 85.06\, \textcolor{blue}{(+4)} & 36.13\, \textcolor{red}{(-4)} & 76.44\, \textcolor{blue}{(+1)} & 32.96\, \textcolor{red}{(-5)} & 71.99\, \textcolor{blue}{(+1)} & 75.13\, \textcolor{blue}{(+2)} & 63.25\, \textcolor{red}{(-4)} & 94.65\, \textcolor{blue}{(+4)} \\
Qwen2.5-Math-7B-CFT & 55.74 & 49.28\, \textcolor{blue}{(+2)} & 84.09\, \textcolor{red}{(-27)} & 83.12\, \textcolor{red}{(-15)} & 35.40\, \textcolor{red}{(-4)} & 75.17\, \textcolor{red}{(-9)} & 34.77\, \textcolor{blue}{(+2)} & 71.28\, \textcolor{red}{(-6)} & 71.55\, \textcolor{red}{(-30)} & 63.78\, \textcolor{blue}{(+1)} & 91.76\, \textcolor{red}{(-34)} \\
NQLSG-Qwen2.5-14B-MegaFusion-v8 & 55.59 & 48.69\, \textcolor{red}{(-1)} & 85.99\, \textcolor{red}{(-3)} & 84.70\, \textcolor{blue}{(+4)} & 34.37\, \textcolor{red}{(-10)} & 75.71\, \textcolor{red}{(-2)} & 33.56 & 71.32\, \textcolor{red}{(-4)} & 72.68\, \textcolor{red}{(-13)} & 63.46 & 94.18\, \textcolor{blue}{(+3)} \\
tempmotacilla-cinerea-0308 & 55.51 & 48.56\, \textcolor{red}{(-1)} & 86.30 & 84.61\, \textcolor{blue}{(+4)} & 35.08\, \textcolor{red}{(-3)} & 75.62\, \textcolor{red}{(-3)} & 33.35\, \textcolor{red}{(-1)} & 71.18\, \textcolor{red}{(-5)} & 73.90 & 63.40 & 94.15\, \textcolor{blue}{(+3)} \\
DeepSeek-R1-Distill-Qwen-25.5B-Brainstorm & 55.36 & 48.49\, \textcolor{red}{(-1)} & 86.45\, \textcolor{blue}{(+4)} & 83.08\, \textcolor{red}{(-13)} & 36.77\, \textcolor{blue}{(+6)} & 75.23\, \textcolor{red}{(-4)} & 32.69\, \textcolor{red}{(-5)} & 70.72\, \textcolor{red}{(-6)} & 75.90\, \textcolor{blue}{(+9)} & 59.09\, \textcolor{red}{(-23)} & 93.44\, \textcolor{red}{(-6)} \\
Qwen2.5-14B-Instruct & 55.29 & 48.71\, \textcolor{blue}{(+4)} & 84.12\, \textcolor{red}{(-22)} & 83.29\, \textcolor{red}{(-6)} & 36.40\, \textcolor{blue}{(+5)} & 75.66 & 33.59\, \textcolor{blue}{(+4)} & 71.90\, \textcolor{blue}{(+5)} & 72.49\, \textcolor{red}{(-11)} & 63.66\, \textcolor{blue}{(+4)} & 91.59\, \textcolor{red}{(-34)} \\
Coma-II-14B & 55.14 & 48.41 & 85.45\, \textcolor{red}{(-3)} & 82.53\, \textcolor{red}{(-19)} & 36.30\, \textcolor{blue}{(+4)} & 75.77\, \textcolor{blue}{(+3)} & 32.82 & 71.64\, \textcolor{blue}{(+4)} & 74.26\, \textcolor{blue}{(+5)} & 57.52\, \textcolor{red}{(-28)} & 93.14\, \textcolor{red}{(-11)} \\
NQLSG-Qwen2.5-14B-MegaFusion-v9.1 & 54.68 & 47.66\, \textcolor{red}{(-1)} & 85.77\, \textcolor{blue}{(+1)} & 83.95\, \textcolor{blue}{(+3)} & 34.49\, \textcolor{red}{(-4)} & 76.05\, \textcolor{blue}{(+6)} & 31.41\, \textcolor{red}{(-8)} & 71.84\, \textcolor{blue}{(+6)} & 73.84\, \textcolor{blue}{(+3)} & 61.60\, \textcolor{red}{(-3)} & 93.47\, \textcolor{red}{(-1)} \\
Viper-Coder-v1.1 & 54.61 & 47.83\, \textcolor{blue}{(+1)} & 84.85\, \textcolor{red}{(-9)} & 82.60\, \textcolor{red}{(-16)} & 34.69\, \textcolor{red}{(-1)} & 74.82\, \textcolor{red}{(-3)} & 32.74 & 70.54\, \textcolor{red}{(-3)} & 72.76\, \textcolor{red}{(-4)} & 59.63\, \textcolor{red}{(-17)} & 92.66\, \textcolor{red}{(-15)} \\
Cheng-2 & 54.38 & 47.35 & 85.70\, \textcolor{blue}{(+2)} & 83.52\, \textcolor{blue}{(+3)} & 33.77\, \textcolor{red}{(-9)} & 73.55\, \textcolor{red}{(-14)} & 31.96\, \textcolor{red}{(-2)} & 68.72\, \textcolor{red}{(-15)} & 73.13\, \textcolor{blue}{(+3)} & 61.38\, \textcolor{red}{(-3)} & 93.24\, \textcolor{red}{(-5)} \\
sky-t1-coder-32b-flash & 54.23 & 47.22\, \textcolor{red}{(-2)} & 85.40 & 83.36\, \textcolor{blue}{(+1)} & 34.86\, \textcolor{blue}{(+2)} & 74.05\, \textcolor{red}{(-9)} & 31.23\, \textcolor{red}{(-7)} & 69.48\, \textcolor{red}{(-8)} & 72.86\, \textcolor{red}{(-1)} & 60.11\, \textcolor{red}{(-9)} & 93.29\, \textcolor{red}{(-2)} \\
NQLSG-Qwen2.5-14B-MegaFusion-v8.7 & 54.08 & 47.25 & 84.13\, \textcolor{red}{(-15)} & 82.98\, \textcolor{red}{(-8)} & 33.95\, \textcolor{red}{(-4)} & 74.37\, \textcolor{red}{(-4)} & 32.38\, \textcolor{blue}{(+1)} & 70.10\, \textcolor{red}{(-2)} & 71.80\, \textcolor{red}{(-15)} & 63.81\, \textcolor{blue}{(+12)} & 92.03\, \textcolor{red}{(-18)} \\
FluentlyLM-Prinum & 54.00 & 47.31\, \textcolor{blue}{(+2)} & 83.99\, \textcolor{red}{(-18)} & 81.44\, \textcolor{red}{(-19)} & 34.96\, \textcolor{blue}{(+5)} & 72.52\, \textcolor{red}{(-18)} & 32.80\, \textcolor{blue}{(+5)} & 68.01\, \textcolor{red}{(-15)} & 72.70\, \textcolor{red}{(-2)} & 60.18\, \textcolor{red}{(-6)} & 91.16\, \textcolor{red}{(-42)} \\
Dyson-14b & 53.93 & 46.94 & 84.74\, \textcolor{red}{(-5)} & 83.36\, \textcolor{blue}{(+3)} & 33.62\, \textcolor{red}{(-6)} & 74.02\, \textcolor{red}{(-7)} & 31.94\, \textcolor{blue}{(+1)} & 69.44\, \textcolor{red}{(-6)} & 71.74\, \textcolor{red}{(-15)} & 62.41\, \textcolor{blue}{(+8)} & 93.20\, \textcolor{red}{(-2)} \\
ZYH-LLM-Qwen2.5-14B-V4 & 53.93 & 46.64\, \textcolor{red}{(-4)} & 86.16\, \textcolor{blue}{(+10)} & 84.47\, \textcolor{blue}{(+13)} & 33.23\, \textcolor{red}{(-10)} & 76.09\, \textcolor{blue}{(+14)} & 30.54\, \textcolor{red}{(-10)} & 71.59\, \textcolor{blue}{(+11)} & 72.44\, \textcolor{red}{(-3)} & 61.66\, \textcolor{blue}{(+5)} & 94.74\, \textcolor{blue}{(+19)} \\
Cheng-2-v1.1 & 53.93 & 46.89\, \textcolor{blue}{(+1)} & 85.20\, \textcolor{blue}{(+2)} & 83.14 & 34.54\, \textcolor{blue}{(+5)} & 74.15\, \textcolor{red}{(-2)} & 30.72\, \textcolor{red}{(-7)} & 69.62\, \textcolor{red}{(-2)} & 72.95\, \textcolor{blue}{(+6)} & 59.99\, \textcolor{red}{(-5)} & 92.83\, \textcolor{red}{(-5)} \\
NQLSG-Qwen2.5-14B-MegaFusion-v8.9 & 53.70 & 46.73\, \textcolor{red}{(-1)} & 84.49\, \textcolor{red}{(-6)} & 82.98\, \textcolor{red}{(-4)} & 33.45\, \textcolor{red}{(-4)} & 74.54\, \textcolor{blue}{(+2)} & 31.50\, \textcolor{blue}{(+3)} & 70.16\, \textcolor{blue}{(+4)} & 72.09\, \textcolor{red}{(-7)} & 61.89\, \textcolor{blue}{(+9)} & 92.77\, \textcolor{red}{(-5)} \\
Qwen2.5-14B-it-restore & 53.70 & 46.82\, \textcolor{blue}{(+1)} & 83.89\, \textcolor{red}{(-14)} & 82.94\, \textcolor{red}{(-4)} & 33.31\, \textcolor{red}{(-5)} & 74.24\, \textcolor{blue}{(+1)} & 30.87\, \textcolor{red}{(-3)} & 69.97\, \textcolor{blue}{(+2)} & 70.26\, \textcolor{red}{(-32)} & 60.61\, \textcolor{blue}{(+1)} & 92.28\, \textcolor{red}{(-9)} \\
Qwen2.5-14B-1M-YOYO-V3 & 53.55 & 46.21\, \textcolor{red}{(-4)} & 86.32\, \textcolor{blue}{(+16)} & 83.75\, \textcolor{blue}{(+13)} & 32.46\, \textcolor{red}{(-22)} & 74.14 & 30.20\, \textcolor{red}{(-10)} & 69.20\, \textcolor{red}{(-3)} & 73.05\, \textcolor{blue}{(+11)} & 59.81\, \textcolor{red}{(-4)} & 94.44\, \textcolor{blue}{(+20)} \\
Sombrero-Opus-14B-Elite5 & 53.55 & 46.45 & 85.08\, \textcolor{blue}{(+3)} & 83.18\, \textcolor{blue}{(+5)} & 33.89\, \textcolor{blue}{(+2)} & 73.84\, \textcolor{red}{(-3)} & 30.36\, \textcolor{red}{(-6)} & 69.18\, \textcolor{red}{(-3)} & 72.93\, \textcolor{blue}{(+9)} & 59.95\, \textcolor{red}{(-2)} & 93.14\, \textcolor{blue}{(+1)} \\
Qwen2.5-14B-YOYO-V4 & 53.47 & 46.43 & 84.65\, \textcolor{red}{(-1)} & 83.07\, \textcolor{blue}{(+2)} & 33.18\, \textcolor{red}{(-6)} & 73.91\, \textcolor{red}{(-1)} & 31.20\, \textcolor{blue}{(+2)} & 69.33 & 71.55\, \textcolor{red}{(-13)} & 61.53\, \textcolor{blue}{(+9)} & 93.17\, \textcolor{blue}{(+4)} \\
Qwen2.5-14B-YOYO-V4-p1 & 53.32 & 46.27 & 84.71\, \textcolor{blue}{(+2)} & 82.63\, \textcolor{red}{(-2)} & 34.03\, \textcolor{blue}{(+7)} & 73.40\, \textcolor{red}{(-3)} & 30.21\, \textcolor{red}{(-6)} & 68.76\, \textcolor{red}{(-2)} & 72.42\, \textcolor{blue}{(+3)} & 59.13\, \textcolor{red}{(-5)} & 92.75\, \textcolor{red}{(-1)} \\
li-14b-v0.4-slerp0.1 & 53.32 & 46.13\, \textcolor{red}{(-3)} & 85.18\, \textcolor{blue}{(+8)} & 83.44\, \textcolor{blue}{(+14)} & 32.28\, \textcolor{red}{(-23)} & 74.56\, \textcolor{blue}{(+10)} & 29.87\, \textcolor{red}{(-9)} & 70.02\, \textcolor{blue}{(+8)} & 71.69\, \textcolor{red}{(-7)} & 59.67\, \textcolor{red}{(-2)} & 93.80\, \textcolor{blue}{(+19)} \\
NQLSG-Qwen2.5-14B-MegaFusion-v9.2 & 53.32 & 46.15\, \textcolor{red}{(-1)} & 85.34\, \textcolor{blue}{(+11)} & 82.80\, \textcolor{blue}{(+1)} & 32.79\, \textcolor{red}{(-9)} & 75.19\, \textcolor{blue}{(+14)} & 29.63\, \textcolor{red}{(-10)} & 70.80\, \textcolor{blue}{(+14)} & 73.30\, \textcolor{blue}{(+18)} & 57.95\, \textcolor{red}{(-7)} & 93.25\, \textcolor{blue}{(+10)} \\
Imbue-14b & 53.17 & 45.98\, \textcolor{red}{(-3)} & 85.09\, \textcolor{blue}{(+9)} & 83.21\, \textcolor{blue}{(+11)} & 33.26\, \textcolor{blue}{(+1)} & 73.12\, \textcolor{red}{(-1)} & 30.86\, \textcolor{blue}{(+3)} & 68.21\, \textcolor{red}{(-2)} & 72.31\, \textcolor{blue}{(+4)} & 61.78\, \textcolor{blue}{(+16)} & 93.52\, \textcolor{blue}{(+17)} \\
Dinobot-Opus-14B-Exp & 53.17 & 46.02\, \textcolor{red}{(-1)} & 84.67\, \textcolor{blue}{(+5)} & 83.44\, \textcolor{blue}{(+18)} & 32.30\, \textcolor{red}{(-19)} & 75.58\, \textcolor{blue}{(+18)} & 29.91\, \textcolor{red}{(-5)} & 71.38\, \textcolor{blue}{(+20)} & 70.74\, \textcolor{red}{(-18)} & 61.13\, \textcolor{blue}{(+11)} & 93.45\, \textcolor{blue}{(+16)} \\
qwen2.5-14b-tensopolis-v1 & 52.95 & 46.19\, \textcolor{blue}{(+3)} & 83.03\, \textcolor{red}{(-17)} & 81.02\, \textcolor{red}{(-8)} & 33.93\, \textcolor{blue}{(+10)} & 72.62\, \textcolor{red}{(-3)} & 31.41\, \textcolor{blue}{(+12)} & 68.34\, \textcolor{blue}{(+1)} & 71.08\, \textcolor{red}{(-14)} & 60.21\, \textcolor{blue}{(+9)} & 90.72\, \textcolor{red}{(-43)} \\
Lix-14B-v0.1 & 52.95 & 45.71\, \textcolor{red}{(-3)} & 85.40\, \textcolor{blue}{(+16)} & 82.49\, \textcolor{blue}{(+1)} & 32.57\, \textcolor{red}{(-8)} & 72.87\, \textcolor{blue}{(+1)} & 29.43\, \textcolor{red}{(-8)} & 67.87\, \textcolor{red}{(-3)} & 72.75\, \textcolor{blue}{(+14)} & 56.27\, \textcolor{red}{(-13)} & 93.38\, \textcolor{blue}{(+16)} \\
Qwen2.5-14B-HyperMarck-dl & 52.87 & 45.82\, \textcolor{red}{(-1)} & 84.41\, \textcolor{blue}{(+5)} & 81.90\, \textcolor{blue}{(+1)} & 33.01 & 72.51\, \textcolor{red}{(-3)} & 30.32\, \textcolor{blue}{(+2)} & 67.64\, \textcolor{red}{(-4)} & 71.55\, \textcolor{red}{(-4)} & 57.82\, \textcolor{red}{(-4)} & 92.86\, \textcolor{blue}{(+9)} \\
ZYH-LLM-Qwen2.5-14B-V3 & 52.72 & 45.35\, \textcolor{red}{(-5)} & 85.51\, \textcolor{blue}{(+21)} & 83.22\, \textcolor{blue}{(+17)} & 32.38\, \textcolor{red}{(-13)} & 74.56\, \textcolor{blue}{(+16)} & 28.99\, \textcolor{red}{(-8)} & 69.90\, \textcolor{blue}{(+13)} & 72.09\, \textcolor{blue}{(+7)} & 59.69\, \textcolor{blue}{(+6)} & 94.01\, \textcolor{blue}{(+27)} \\
MFDOOM-14B & 52.64 & 45.85\, \textcolor{blue}{(+2)} & 82.76\, \textcolor{red}{(-19)} & 81.04\, \textcolor{red}{(-3)} & 32.49\, \textcolor{red}{(-8)} & 71.30\, \textcolor{red}{(-7)} & 31.25\, \textcolor{blue}{(+14)} & 66.64\, \textcolor{red}{(-7)} & 69.97\, \textcolor{red}{(-25)} & 58.24\, \textcolor{blue}{(+1)} & 91.23\, \textcolor{red}{(-23)} \\
mocha-14B & 52.64 & 45.65 & 83.68\, \textcolor{red}{(-4)} & 81.88\, \textcolor{blue}{(+3)} & 33.20\, \textcolor{blue}{(+6)} & 72.42\, \textcolor{red}{(-1)} & 30.65\, \textcolor{blue}{(+8)} & 67.75 & 71.75\, \textcolor{blue}{(+4)} & 61.19\, \textcolor{blue}{(+18)} & 91.98\, \textcolor{red}{(-1)} \\
ultiima-72B & 52.42 & 45.39\, \textcolor{red}{(-1)} & 84.31\, \textcolor{blue}{(+8)} & 80.42\, \textcolor{red}{(-9)} & 32.91\, \textcolor{blue}{(+3)} & 71.25\, \textcolor{red}{(-6)} & 29.77 & 66.29\, \textcolor{red}{(-8)} & 73.03\, \textcolor{blue}{(+24)} & 56.13\, \textcolor{red}{(-9)} & 91.23\, \textcolor{red}{(-20)} \\
Condor-Opus-14B-Exp & 52.27 & 45.49\, \textcolor{blue}{(+1)} & 82.91\, \textcolor{red}{(-11)} & 79.67\, \textcolor{red}{(-20)} & 33.15\, \textcolor{blue}{(+6)} & 71.48\, \textcolor{red}{(-3)} & 30.06\, \textcolor{blue}{(+4)} & 66.95\, \textcolor{red}{(-3)} & 71.10\, \textcolor{red}{(-6)} & 53.08\, \textcolor{red}{(-41)} & 90.47\, \textcolor{red}{(-38)} \\
Qwen2.5-14B-ReasoningMerge & 52.04 & 45.00\, \textcolor{red}{(-2)} & 83.75\, \textcolor{blue}{(+2)} & 80.68\, \textcolor{red}{(-2)} & 33.38\, \textcolor{blue}{(+13)} & 72.13\, \textcolor{blue}{(+1)} & 29.50 & 67.34\, \textcolor{blue}{(+1)} & 71.36\, \textcolor{red}{(-1)} & 55.46\, \textcolor{red}{(-13)} & 92.12\, \textcolor{blue}{(+7)} \\
Lo-Phi-14b & 51.96 & 45.04 & 82.87\, \textcolor{red}{(-10)} & 80.64\, \textcolor{red}{(-2)} & 32.37\, \textcolor{red}{(-8)} & 71.24\, \textcolor{red}{(-4)} & 30.90\, \textcolor{blue}{(+16)} & 66.50\, \textcolor{red}{(-3)} & 70.35\, \textcolor{red}{(-12)} & 60.97\, \textcolor{blue}{(+20)} & 91.01\, \textcolor{red}{(-25)} \\
lamarckvergence-14b-tensopolis-v1 & 51.66 & 44.55\, \textcolor{red}{(-1)} & 83.08\, \textcolor{red}{(-5)} & 81.62\, \textcolor{blue}{(+6)} & 32.78\, \textcolor{blue}{(+3)} & 71.88\, \textcolor{blue}{(+2)} & 28.83\, \textcolor{red}{(-2)} & 67.20\, \textcolor{blue}{(+2)} & 69.04\, \textcolor{red}{(-38)} & 57.78\, \textcolor{blue}{(+3)} & 92.01\, \textcolor{blue}{(+5)} \\
Qwen2.5-14B-YOYO-V4-p2 & 51.66 & 44.45\, \textcolor{red}{(-1)} & 84.00\, \textcolor{blue}{(+8)} & 81.12\, \textcolor{blue}{(+5)} & 32.61\, \textcolor{blue}{(+3)} & 72.66\, \textcolor{blue}{(+10)} & 27.28\, \textcolor{red}{(-15)} & 67.94\, \textcolor{blue}{(+9)} & 72.21\, \textcolor{blue}{(+16)} & 55.21\, \textcolor{red}{(-11)} & 91.87\, \textcolor{blue}{(+4)} \\
Qwen2.5-14B-YOYO-latest-V2 & 51.59 & 44.28\, \textcolor{red}{(-2)} & 84.27\, \textcolor{blue}{(+13)} & 81.66\, \textcolor{blue}{(+9)} & 32.39\, \textcolor{red}{(-3)} & 72.63\, \textcolor{blue}{(+10)} & 28.28\, \textcolor{red}{(-2)} & 67.87\, \textcolor{blue}{(+9)} & 72.01\, \textcolor{blue}{(+13)} & 58.35\, \textcolor{blue}{(+10)} & 92.56\, \textcolor{blue}{(+14)} \\
miscii-14b-0218 & 51.44 & 44.36 & 83.26\, \textcolor{blue}{(+1)} & 80.29\, \textcolor{red}{(-3)} & 32.87\, \textcolor{blue}{(+9)} & 71.71\, \textcolor{blue}{(+4)} & 28.38 & 66.98\, \textcolor{blue}{(+4)} & 71.63\, \textcolor{blue}{(+9)} & 54.23\, \textcolor{red}{(-20)} & 91.64\, \textcolor{blue}{(+1)} \\
absolute-o1-7b & 50.83 & 43.53\, \textcolor{red}{(-2)} & 83.71\, \textcolor{blue}{(+7)} & 80.51\, \textcolor{blue}{(+1)} & 31.23\, \textcolor{red}{(-17)} & 71.15\, \textcolor{red}{(-1)} & 28.28\, \textcolor{red}{(-1)} & 66.14\, \textcolor{red}{(-1)} & 71.27\, \textcolor{blue}{(+3)} & 56.33\, \textcolor{blue}{(+1)} & 92.01\, \textcolor{blue}{(+10)} \\
Qwen2.5-THREADRIPPER-Small-AnniversaryEdition & 50.76 & 43.43\, \textcolor{red}{(-4)} & 83.71\, \textcolor{blue}{(+7)} & 80.59\, \textcolor{blue}{(+3)} & 32.05\, \textcolor{red}{(-8)} & 70.59\, \textcolor{red}{(-4)} & 27.97\, \textcolor{red}{(-3)} & 65.42\, \textcolor{red}{(-6)} & 71.18\, \textcolor{blue}{(+3)} & 57.11\, \textcolor{blue}{(+6)} & 92.09\, \textcolor{blue}{(+13)} \\
pre-cursa-o1-v1.3 & 50.76 & 43.50\, \textcolor{red}{(-2)} & 83.45\, \textcolor{blue}{(+5)} & 80.29\, \textcolor{red}{(-1)} & 32.48\, \textcolor{blue}{(+3)} & 70.36\, \textcolor{red}{(-6)} & 27.85\, \textcolor{red}{(-5)} & 65.30\, \textcolor{red}{(-8)} & 70.95\, \textcolor{blue}{(+1)} & 56.01\, \textcolor{red}{(-1)} & 91.69\, \textcolor{blue}{(+5)} \\
pre-cursa-o1-v1.2 & 50.68 & 43.52 & 82.66\, \textcolor{red}{(-7)} & 80.17\, \textcolor{red}{(-1)} & 32.53\, \textcolor{blue}{(+6)} & 71.17\, \textcolor{blue}{(+3)} & 27.95\, \textcolor{red}{(-3)} & 66.44\, \textcolor{blue}{(+4)} & 70.28\, \textcolor{red}{(-5)} & 56.05\, \textcolor{blue}{(+1)} & 91.34\, \textcolor{red}{(-3)} \\
MFGRIMM-14B & 50.60 & 43.54\, \textcolor{blue}{(+3)} & 81.97\, \textcolor{red}{(-27)} & 80.00\, \textcolor{red}{(-1)} & 31.03\, \textcolor{red}{(-21)} & 70.21\, \textcolor{red}{(-7)} & 27.96\, \textcolor{red}{(-1)} & 65.44\, \textcolor{red}{(-2)} & 68.94\, \textcolor{red}{(-32)} & 56.86\, \textcolor{blue}{(+7)} & 90.27\, \textcolor{red}{(-30)} \\
\bottomrule
\end{tabular}}
\label{tab:meme_scores_summary_OpenLLM-MATH}
\end{table}

\begin{table}[H]
\centering
\caption{\textbf{Meme Scores (Open LLM Population; MMLU-Pro).} The table reports meme scores across models, including property-derived 1D meme scores, predefined 2D meme scores (Mastery, Ingenuity, and Robustness), and a predefined 3D meme score (Caution). Models are sorted by Accuracy, and only the \textbf{top-50 models} are shown. {\color{blue}Blue} indicates rank improvement compared with Accuracy rank, and {\color{red}red} indicates rank degradation.}
\resizebox{\textwidth}{!}{
\begin{tabular}{cccccccccccc}
\toprule
\multirow{2}{*}{\textbf{Model}} & \multirow{2}{*}{\textbf{Accuracy}} &
\multicolumn{6}{c}{\textbf{Property-derived 1D meme scores}} &
\multicolumn{4}{c}{\textbf{Predefined 2D/3D meme scores}} \\
\cmidrule(lr){3-8}\cmidrule(lr){9-12}
 &  & \textbf{Difficulty} & \textbf{Uniqueness} & \textbf{Risk} & \textbf{Surprise} & \textbf{Typicality} & \textbf{Bridge} & \textbf{Mastery} & \textbf{Ingenuity} & \textbf{Robustness} & \textbf{Caution} \\
\midrule
calme-3.2-instruct-78b & 73.03 & 64.53 & 86.45 & 87.05 & 44.61 & 76.87 & 55.27 & 69.47 & 62.54 & 81.34 & 95.01 \\
calme-3.1-instruct-78b & 71.85 & 62.94 & 85.91 & 86.52 & 42.42 & 75.79 & 53.19 & 68.03 & 60.96 & 80.36 & 94.85 \\
CalmeRys-78B-Orpo-v0.1 & 70.12 & 60.80 & 84.93\, \textcolor{red}{(-1)} & 85.41 & 39.86 & 74.04 & 50.85 & 65.87 & 59.19 & 79.16 & 94.12\, \textcolor{red}{(-1)} \\
calme-2.4-rys-78b & 70.02 & 60.63 & 84.94\, \textcolor{blue}{(+1)} & 85.40 & 39.69 & 73.98 & 50.55 & 65.73 & 59.08 & 79.03 & 94.25\, \textcolor{blue}{(+1)} \\
Reflection-Llama-3.1-70B & 63.41 & 52.37 & 81.56 & 81.03\, \textcolor{red}{(-3)} & 34.39 & 67.21 & 43.20 & 56.94 & 53.30 & 74.84 & 92.57\, \textcolor{red}{(-10)} \\
Arcee-Blitz & 61.54 & 50.12 & 79.66\, \textcolor{red}{(-2)} & 80.47\, \textcolor{red}{(-13)} & 27.92 & 65.67 & 41.05 & 55.02 & 47.99 & 74.72 & 92.22\, \textcolor{red}{(-29)} \\
Homer-v1.0-Qwen2.5-72B & 61.45 & 49.14 & 80.70\, \textcolor{blue}{(+1)} & 81.93\, \textcolor{blue}{(+2)} & 24.95 & 65.49 & 37.03\, \textcolor{red}{(-3)} & 54.05 & 46.86 & 73.10\, \textcolor{red}{(-4)} & 93.58\, \textcolor{blue}{(+2)} \\
ultiima-72B-v1.5 & 60.54 & 47.81 & 80.63\, \textcolor{blue}{(+1)} & 81.46\, \textcolor{blue}{(+2)} & 23.84\, \textcolor{red}{(-1)} & 64.39 & 35.16\, \textcolor{red}{(-14)} & 52.47 & 46.35 & 72.02\, \textcolor{red}{(-57)} & 93.42\, \textcolor{blue}{(+2)} \\
Qwen2.5-72B & 59.68 & 47.23 & 79.44\, \textcolor{red}{(-2)} & 80.24\, \textcolor{red}{(-14)} & 24.52\, \textcolor{blue}{(+1)} & 63.64 & 36.37\, \textcolor{red}{(-4)} & 51.95 & 46.15 & 72.57\, \textcolor{red}{(-8)} & 92.42\, \textcolor{red}{(-15)} \\
QwentileSwap & 59.46 & 46.95 & 78.43\, \textcolor{red}{(-16)} & 81.11\, \textcolor{blue}{(+3)} & 21.70\, \textcolor{red}{(-11)} & 63.39 & 35.85\, \textcolor{red}{(-4)} & 51.74\, \textcolor{red}{(-2)} & 42.06\, \textcolor{red}{(-39)} & 73.01\, \textcolor{red}{(-2)} & 92.79\, \textcolor{blue}{(+1)} \\
Rombos-LLM-V2.5-Qwen-72b & 59.35 & 46.65 & 79.49\, \textcolor{blue}{(+1)} & 80.28\, \textcolor{red}{(-11)} & 22.95 & 63.18 & 34.75\, \textcolor{red}{(-13)} & 51.25\, \textcolor{red}{(-7)} & 45.37 & 71.74\, \textcolor{red}{(-101)} & 92.50\, \textcolor{red}{(-11)} \\
Rombos-LLM-V2.5-Qwen-32b & 59.16 & 46.60\, \textcolor{red}{(-2)} & 78.30\, \textcolor{red}{(-17)} & 80.76\, \textcolor{blue}{(+3)} & 21.37\, \textcolor{red}{(-17)} & 63.08\, \textcolor{red}{(-5)} & 35.35\, \textcolor{red}{(-9)} & 51.33\, \textcolor{red}{(-4)} & 41.90\, \textcolor{red}{(-42)} & 72.53\, \textcolor{red}{(-7)} & 92.54\, \textcolor{red}{(-5)} \\
novablast-preview & 59.15 & 46.64\, \textcolor{blue}{(+1)} & 78.14\, \textcolor{red}{(-22)} & 80.76\, \textcolor{blue}{(+3)} & 21.31\, \textcolor{red}{(-17)} & 63.15\, \textcolor{red}{(-1)} & 35.59\, \textcolor{red}{(-3)} & 51.47 & 41.51\, \textcolor{red}{(-49)} & 72.68\, \textcolor{red}{(-2)} & 92.54\, \textcolor{red}{(-6)} \\
TheBeagle-v2beta-32B-MGS & 59.15 & 46.61\, \textcolor{blue}{(+1)} & 78.24\, \textcolor{red}{(-16)} & 80.74\, \textcolor{blue}{(+1)} & 21.42\, \textcolor{red}{(-12)} & 63.03\, \textcolor{red}{(-4)} & 35.42\, \textcolor{red}{(-5)} & 51.29\, \textcolor{red}{(-3)} & 41.90\, \textcolor{red}{(-39)} & 72.50\, \textcolor{red}{(-6)} & 92.51\, \textcolor{red}{(-6)} \\
ultiima-32B & 59.10 & 46.56 & 78.15\, \textcolor{red}{(-18)} & 80.75\, \textcolor{blue}{(+4)} & 21.26\, \textcolor{red}{(-16)} & 63.12 & 35.56\, \textcolor{red}{(-3)} & 51.39 & 41.51\, \textcolor{red}{(-46)} & 72.66\, \textcolor{red}{(-1)} & 92.65\, \textcolor{blue}{(+3)} \\
Oxyge1-33B & 59.09 & 46.55 & 78.15\, \textcolor{red}{(-18)} & 80.74\, \textcolor{blue}{(+4)} & 21.25\, \textcolor{red}{(-16)} & 63.12 & 35.58\, \textcolor{red}{(-1)} & 51.40\, \textcolor{blue}{(+2)} & 41.51\, \textcolor{red}{(-44)} & 72.69\, \textcolor{blue}{(+2)} & 92.64\, \textcolor{blue}{(+3)} \\
RombosBeagle-v2beta-MGS-32B & 59.08 & 46.51\, \textcolor{red}{(-2)} & 78.23\, \textcolor{red}{(-14)} & 80.71\, \textcolor{blue}{(+3)} & 21.38\, \textcolor{red}{(-10)} & 63.01\, \textcolor{red}{(-2)} & 35.39\, \textcolor{red}{(-3)} & 51.25\, \textcolor{red}{(-2)} & 41.84\, \textcolor{red}{(-40)} & 72.57\, \textcolor{red}{(-1)} & 92.58\, \textcolor{blue}{(+3)} \\
ultiima-72B & 59.06 & 46.24\, \textcolor{red}{(-2)} & 79.37\, \textcolor{blue}{(+6)} & 80.19\, \textcolor{red}{(-6)} & 22.56\, \textcolor{blue}{(+4)} & 62.90\, \textcolor{red}{(-2)} & 34.29\, \textcolor{red}{(-14)} & 50.84\, \textcolor{red}{(-2)} & 44.97\, \textcolor{blue}{(+2)} & 71.52\, \textcolor{red}{(-142)} & 92.54 \\
T3Q-qwen2.5-14b-v1.0-e3 & 58.84 & 46.55\, \textcolor{blue}{(+2)} & 77.18\, \textcolor{red}{(-36)} & 80.62\, \textcolor{blue}{(+4)} & 21.81 & 63.17\, \textcolor{blue}{(+7)} & 36.37\, \textcolor{blue}{(+8)} & 51.86\, \textcolor{blue}{(+9)} & 40.75\, \textcolor{red}{(-61)} & 73.29\, \textcolor{blue}{(+10)} & 92.36\, \textcolor{red}{(-10)} \\
T3Q-Qwen2.5-14B-Instruct-1M-e3 & 58.84 & 46.55\, \textcolor{blue}{(+3)} & 77.18\, \textcolor{red}{(-35)} & 80.62\, \textcolor{blue}{(+5)} & 21.81\, \textcolor{blue}{(+1)} & 63.17\, \textcolor{blue}{(+8)} & 36.37\, \textcolor{blue}{(+9)} & 51.86\, \textcolor{blue}{(+10)} & 40.75\, \textcolor{red}{(-60)} & 73.29\, \textcolor{blue}{(+11)} & 92.36\, \textcolor{red}{(-9)} \\
Qwentile2.5-32B-Instruct & 58.79 & 45.87\, \textcolor{red}{(-1)} & 78.88\, \textcolor{blue}{(+1)} & 80.54\, \textcolor{blue}{(+4)} & 21.12\, \textcolor{red}{(-12)} & 62.84 & 33.95\, \textcolor{red}{(-21)} & 50.72 & 43.04\, \textcolor{red}{(-5)} & 71.89\, \textcolor{red}{(-65)} & 92.80\, \textcolor{blue}{(+13)} \\
Qwen2.5-72B-Instruct-abliterated & 58.72 & 45.86\, \textcolor{red}{(-1)} & 79.27\, \textcolor{blue}{(+9)} & 79.75\, \textcolor{red}{(-5)} & 22.52\, \textcolor{blue}{(+7)} & 62.46\, \textcolor{red}{(-1)} & 33.78\, \textcolor{red}{(-23)} & 50.33\, \textcolor{red}{(-1)} & 45.27\, \textcolor{blue}{(+10)} & 71.08\, \textcolor{red}{(-182)} & 92.21\, \textcolor{red}{(-14)} \\
Apollo\_v2-32B & 58.69 & 45.97\, \textcolor{blue}{(+2)} & 78.19\, \textcolor{red}{(-9)} & 80.49\, \textcolor{blue}{(+5)} & 21.37\, \textcolor{red}{(-5)} & 62.62\, \textcolor{blue}{(+1)} & 34.88 & 50.69\, \textcolor{blue}{(+1)} & 42.01\, \textcolor{red}{(-28)} & 72.28\, \textcolor{red}{(-6)} & 92.56\, \textcolor{blue}{(+7)} \\
LeTriomphant2.2\_ECE\_iLAB & 58.51 & 45.59\, \textcolor{red}{(-2)} & 79.20\, \textcolor{blue}{(+10)} & 79.60\, \textcolor{red}{(-6)} & 22.42\, \textcolor{blue}{(+7)} & 62.27 & 33.92\, \textcolor{red}{(-20)} & 50.09\, \textcolor{red}{(-1)} & 45.09\, \textcolor{blue}{(+11)} & 71.21\, \textcolor{red}{(-172)} & 92.18\, \textcolor{red}{(-14)} \\
test-2.5-72B & 58.37 & 45.59 & 78.98\, \textcolor{blue}{(+8)} & 79.15\, \textcolor{red}{(-27)} & 22.78\, \textcolor{blue}{(+12)} & 62.21 & 34.35\, \textcolor{red}{(-6)} & 50.11\, \textcolor{blue}{(+1)} & 45.47\, \textcolor{blue}{(+15)} & 71.32\, \textcolor{red}{(-164)} & 91.95\, \textcolor{red}{(-51)} \\
EVA-Qwen2.5-72B-v0.2 & 58.13 & 45.59\, \textcolor{blue}{(+2)} & 77.93\, \textcolor{red}{(-14)} & 79.07\, \textcolor{red}{(-42)} & 22.44\, \textcolor{blue}{(+10)} & 62.00\, \textcolor{red}{(-1)} & 34.73\, \textcolor{blue}{(+1)} & 50.07 & 43.68\, \textcolor{blue}{(+8)} & 71.67\, \textcolor{red}{(-99)} & 91.65\, \textcolor{red}{(-114)} \\
FluentlyLM-Prinum & 58.08 & 44.71\, \textcolor{red}{(-5)} & 78.96\, \textcolor{blue}{(+9)} & 80.38\, \textcolor{blue}{(+7)} & 19.99\, \textcolor{red}{(-33)} & 62.10\, \textcolor{blue}{(+1)} & 32.60\, \textcolor{red}{(-91)} & 49.54\, \textcolor{red}{(-3)} & 42.28\, \textcolor{red}{(-14)} & 71.12\, \textcolor{red}{(-173)} & 93.03\, \textcolor{blue}{(+20)} \\
Qwen2.5-32B & 58.05 & 45.36\, \textcolor{blue}{(+1)} & 77.92\, \textcolor{red}{(-14)} & 79.38\, \textcolor{red}{(-7)} & 21.60\, \textcolor{blue}{(+4)} & 61.82\, \textcolor{red}{(-2)} & 34.61\, \textcolor{blue}{(+2)} & 49.91 & 42.89\, \textcolor{red}{(-3)} & 71.88\, \textcolor{red}{(-59)} & 91.51\, \textcolor{red}{(-170)} \\
Gilgamesh-72B & 58.02 & 44.98 & 79.07\, \textcolor{blue}{(+13)} & 79.15\, \textcolor{red}{(-25)} & 22.10\, \textcolor{blue}{(+11)} & 61.85 & 33.68\, \textcolor{red}{(-20)} & 49.53\, \textcolor{red}{(-2)} & 45.01\, \textcolor{blue}{(+14)} & 71.05\, \textcolor{red}{(-179)} & 92.10\, \textcolor{red}{(-20)} \\
Linkbricks-Horizon-AI-Avengers-V1-32B & 57.93 & 44.60\, \textcolor{red}{(-3)} & 78.90\, \textcolor{blue}{(+11)} & 80.04\, \textcolor{blue}{(+5)} & 20.06\, \textcolor{red}{(-28)} & 61.93\, \textcolor{blue}{(+2)} & 32.40\, \textcolor{red}{(-108)} & 49.37\, \textcolor{red}{(-3)} & 42.78\, \textcolor{red}{(-2)} & 71.00\, \textcolor{red}{(-184)} & 92.76\, \textcolor{blue}{(+20)} \\
RomboUltima-32B & 57.89 & 44.83 & 78.41\, \textcolor{blue}{(+4)} & 79.64\, \textcolor{blue}{(+2)} & 20.79\, \textcolor{red}{(-8)} & 61.77 & 33.04\, \textcolor{red}{(-42)} & 49.42\, \textcolor{red}{(-1)} & 42.95\, \textcolor{blue}{(+2)} & 71.02\, \textcolor{red}{(-181)} & 92.23\, \textcolor{red}{(-3)} \\
sky-t1-coder-32b-flash & 57.82 & 44.53\, \textcolor{red}{(-2)} & 79.08\, \textcolor{blue}{(+17)} & 79.53\, \textcolor{red}{(-1)} & 20.92\, \textcolor{red}{(-5)} & 61.75\, \textcolor{red}{(-1)} & 32.56\, \textcolor{red}{(-90)} & 49.20\, \textcolor{red}{(-4)} & 44.30\, \textcolor{blue}{(+15)} & 70.98\, \textcolor{red}{(-188)} & 92.44\, \textcolor{blue}{(+9)} \\
qwen2.5-test-32b-it & 57.65 & 44.31\, \textcolor{red}{(-4)} & 78.50\, \textcolor{blue}{(+8)} & 79.98\, \textcolor{blue}{(+7)} & 19.82\, \textcolor{red}{(-28)} & 61.76\, \textcolor{blue}{(+1)} & 32.44\, \textcolor{red}{(-102)} & 49.23\, \textcolor{red}{(-1)} & 41.89\, \textcolor{red}{(-22)} & 71.03\, \textcolor{red}{(-177)} & 92.70\, \textcolor{blue}{(+22)} \\
Linkbricks-Horizon-AI-Avengers-V5-32B & 57.62 & 44.34\, \textcolor{red}{(-1)} & 78.68\, \textcolor{blue}{(+12)} & 79.53\, \textcolor{blue}{(+2)} & 20.57\, \textcolor{red}{(-10)} & 61.53\, \textcolor{red}{(-1)} & 32.88\, \textcolor{red}{(-54)} & 48.98\, \textcolor{red}{(-4)} & 43.17\, \textcolor{blue}{(+10)} & 71.07\, \textcolor{red}{(-172)} & 92.39\, \textcolor{blue}{(+7)} \\
PathFinderAi3.0 & 57.57 & 44.89\, \textcolor{blue}{(+5)} & 77.55\, \textcolor{red}{(-11)} & 78.76\, \textcolor{red}{(-87)} & 21.65\, \textcolor{blue}{(+12)} & 61.14\, \textcolor{red}{(-4)} & 34.23\, \textcolor{blue}{(+1)} & 49.08\, \textcolor{red}{(-2)} & 42.73\, \textcolor{blue}{(+1)} & 71.46\, \textcolor{red}{(-135)} & 91.08\, \textcolor{red}{(-231)} \\
Linkbricks-Horizon-AI-Avengers-V4-32B & 57.52 & 44.25\, \textcolor{red}{(-3)} & 78.59\, \textcolor{blue}{(+13)} & 79.41\, \textcolor{blue}{(+2)} & 20.49\, \textcolor{red}{(-10)} & 61.31 & 32.81\, \textcolor{red}{(-58)} & 48.79\, \textcolor{red}{(-3)} & 43.06\, \textcolor{blue}{(+11)} & 71.03\, \textcolor{red}{(-175)} & 92.10\, \textcolor{red}{(-11)} \\
Qwen2-72B & 57.31 & 45.20\, \textcolor{blue}{(+9)} & 75.70\, \textcolor{red}{(-115)} & 78.53\, \textcolor{red}{(-120)} & 21.54\, \textcolor{blue}{(+12)} & 61.04\, \textcolor{red}{(-5)} & 37.73\, \textcolor{blue}{(+28)} & 49.68\, \textcolor{blue}{(+8)} & 39.22\, \textcolor{red}{(-68)} & 73.31\, \textcolor{blue}{(+29)} & 90.55\, \textcolor{red}{(-311)} \\
Awqward2.5-32B-Instruct & 57.23 & 43.95\, \textcolor{red}{(-4)} & 77.84\, \textcolor{red}{(-5)} & 79.65\, \textcolor{blue}{(+10)} & 19.45\, \textcolor{red}{(-38)} & 61.07\, \textcolor{red}{(-3)} & 32.49\, \textcolor{red}{(-92)} & 48.42\, \textcolor{red}{(-4)} & 40.78\, \textcolor{red}{(-41)} & 70.99\, \textcolor{red}{(-180)} & 92.50\, \textcolor{blue}{(+17)} \\
huihui-ai-abliterated-Qwen2.5-32B-Inst-BaseMerge-TIES & 57.21 & 44.32\, \textcolor{blue}{(+3)} & 77.23\, \textcolor{red}{(-15)} & 79.09\, \textcolor{red}{(-26)} & 20.30\, \textcolor{red}{(-13)} & 61.31\, \textcolor{blue}{(+2)} & 33.78\, \textcolor{red}{(-7)} & 49.20\, \textcolor{blue}{(+4)} & 41.22\, \textcolor{red}{(-33)} & 71.59\, \textcolor{red}{(-103)} & 91.67\, \textcolor{red}{(-98)} \\
Linkbricks-Horizon-AI-Avengers-V2-32B & 57.20 & 43.89\, \textcolor{red}{(-3)} & 77.92\, \textcolor{red}{(-1)} & 79.58\, \textcolor{blue}{(+9)} & 19.56\, \textcolor{red}{(-30)} & 61.11 & 32.50\, \textcolor{red}{(-86)} & 48.54 & 40.86\, \textcolor{red}{(-38)} & 70.96\, \textcolor{red}{(-184)} & 92.37\, \textcolor{blue}{(+12)} \\
Qwen2-VL-72B-Instruct & 57.17 & 43.85\, \textcolor{red}{(-3)} & 78.51\, \textcolor{blue}{(+17)} & 78.93\, \textcolor{red}{(-48)} & 20.51\, \textcolor{red}{(-4)} & 61.16\, \textcolor{blue}{(+3)} & 32.27\, \textcolor{red}{(-106)} & 48.51 & 43.27\, \textcolor{blue}{(+20)} & 70.54\, \textcolor{red}{(-221)} & 92.13\, \textcolor{red}{(-3)} \\
shuttle-3 & 57.16 & 43.48\, \textcolor{red}{(-5)} & 79.53\, \textcolor{blue}{(+33)} & 78.89\, \textcolor{red}{(-60)} & 20.78\, \textcolor{blue}{(+2)} & 60.93\, \textcolor{red}{(-1)} & 30.95\, \textcolor{red}{(-285)} & 47.85\, \textcolor{red}{(-8)} & 45.05\, \textcolor{blue}{(+28)} & 69.65\, \textcolor{red}{(-345)} & 92.41\, \textcolor{blue}{(+17)} \\
Galactic-Qwen-14B-Exp2 & 56.91 & 44.12\, \textcolor{blue}{(+2)} & 75.36\, \textcolor{red}{(-129)} & 80.36\, \textcolor{blue}{(+22)} & 18.23\, \textcolor{red}{(-72)} & 61.66\, \textcolor{blue}{(+9)} & 34.28\, \textcolor{blue}{(+10)} & 49.99\, \textcolor{blue}{(+16)} & 36.16\, \textcolor{red}{(-243)} & 72.75\, \textcolor{blue}{(+30)} & 92.40\, \textcolor{blue}{(+17)} \\
zetasepic-abliteratedV2-Qwen2.5-32B-Inst-BaseMerge-TIES & 56.85 & 43.51\, \textcolor{red}{(-2)} & 77.73 & 79.20\, \textcolor{red}{(-1)} & 19.26\, \textcolor{red}{(-36)} & 60.58\, \textcolor{red}{(-1)} & 32.11\, \textcolor{red}{(-133)} & 47.99\, \textcolor{red}{(-2)} & 41.01\, \textcolor{red}{(-33)} & 70.65\, \textcolor{red}{(-211)} & 91.88\, \textcolor{red}{(-46)} \\
QwQ-32B-Preview & 56.78 & 44.14\, \textcolor{blue}{(+5)} & 76.86\, \textcolor{red}{(-17)} & 77.93\, \textcolor{red}{(-146)} & 21.67\, \textcolor{blue}{(+23)} & 60.43\, \textcolor{red}{(-4)} & 34.56\, \textcolor{blue}{(+18)} & 48.37\, \textcolor{blue}{(+2)} & 42.38\, \textcolor{blue}{(+6)} & 71.60\, \textcolor{red}{(-94)} & 90.69\, \textcolor{red}{(-280)} \\
Linkbricks-Horizon-AI-Avengers-V6-32B & 56.72 & 43.40\, \textcolor{red}{(-2)} & 77.39\, \textcolor{red}{(-3)} & 79.26\, \textcolor{blue}{(+7)} & 19.21\, \textcolor{red}{(-36)} & 60.62\, \textcolor{blue}{(+2)} & 32.45\, \textcolor{red}{(-87)} & 47.95\, \textcolor{red}{(-1)} & 40.22\, \textcolor{red}{(-43)} & 70.93\, \textcolor{red}{(-182)} & 92.26\, \textcolor{blue}{(+14)} \\
Qwen2.5-32B-Instruct & 56.67 & 43.32\, \textcolor{red}{(-2)} & 77.36\, \textcolor{red}{(-3)} & 79.25\, \textcolor{blue}{(+7)} & 19.09\, \textcolor{red}{(-42)} & 60.56 & 32.26\, \textcolor{red}{(-103)} & 47.85\, \textcolor{red}{(-2)} & 40.08\, \textcolor{red}{(-45)} & 70.84\, \textcolor{red}{(-190)} & 92.29\, \textcolor{blue}{(+16)} \\
Linkbricks-Horizon-AI-Avengers-V3-32B & 56.64 & 43.31\, \textcolor{red}{(-2)} & 77.30\, \textcolor{red}{(-3)} & 79.22\, \textcolor{blue}{(+4)} & 19.09\, \textcolor{red}{(-42)} & 60.54 & 32.35\, \textcolor{red}{(-92)} & 47.86 & 40.01\, \textcolor{red}{(-45)} & 70.89\, \textcolor{red}{(-187)} & 92.23\, \textcolor{blue}{(+15)} \\
lambda-qwen2.5-32b-dpo-test & 56.57 & 43.61\, \textcolor{blue}{(+4)} & 76.39\, \textcolor{red}{(-26)} & 78.90\, \textcolor{red}{(-50)} & 19.15\, \textcolor{red}{(-37)} & 60.56\, \textcolor{blue}{(+3)} & 33.48\, \textcolor{red}{(-5)} & 48.28\, \textcolor{blue}{(+5)} & 38.90\, \textcolor{red}{(-69)} & 71.53\, \textcolor{red}{(-108)} & 91.70\, \textcolor{red}{(-79)} \\
Qwen2.5-72B-2x-Instruct-TIES-v1.0 & 56.28 & 42.83\, \textcolor{red}{(-5)} & 78.09\, \textcolor{blue}{(+14)} & 78.04\, \textcolor{red}{(-131)} & 20.25\, \textcolor{red}{(-4)} & 60.09\, \textcolor{red}{(-1)} & 31.27\, \textcolor{red}{(-245)} & 47.30\, \textcolor{red}{(-6)} & 43.27\, \textcolor{blue}{(+28)} & 69.75\, \textcolor{red}{(-308)} & 91.43\, \textcolor{red}{(-173)} \\
\bottomrule
\end{tabular}}
\label{tab:meme_scores_summary_OpenLLM-MMLU-Pro}
\end{table}

\begin{table}[H]
\centering
\caption{\textbf{Meme Scores (Open LLM Population; MUSR).} The table reports meme scores across models, including property-derived 1D meme scores, predefined 2D meme scores (Mastery, Ingenuity, and Robustness), and a predefined 3D meme score (Caution). Models are sorted by Accuracy, and only the \textbf{top-50 models} are shown. {\color{blue}Blue} indicates rank improvement compared with Accuracy rank, and {\color{red}red} indicates rank degradation.}
\resizebox{\textwidth}{!}{
\begin{tabular}{cccccccccccc}
\toprule
\multirow{2}{*}{\textbf{Model}} & \multirow{2}{*}{\textbf{Accuracy}} &
\multicolumn{6}{c}{\textbf{Property-derived 1D meme scores}} &
\multicolumn{4}{c}{\textbf{Predefined 2D/3D meme scores}} \\
\cmidrule(lr){3-8}\cmidrule(lr){9-12}
 &  & \textbf{Difficulty} & \textbf{Uniqueness} & \textbf{Risk} & \textbf{Surprise} & \textbf{Typicality} & \textbf{Bridge} & \textbf{Mastery} & \textbf{Ingenuity} & \textbf{Robustness} & \textbf{Caution} \\
\midrule
calme-3.2-instruct-78b & 60.05 & 47.72 & 77.52\, \textcolor{red}{(-2)} & 66.51\, \textcolor{red}{(-5)} & 44.02 & 62.80\, \textcolor{red}{(-3)} & 47.97\, \textcolor{red}{(-59)} & 50.20\, \textcolor{red}{(-3)} & 72.14\, \textcolor{red}{(-1)} & 61.28\, \textcolor{red}{(-132)} & 81.70\, \textcolor{red}{(-319)} \\
T3Q-Qwen2.5-14B-Instruct-1M-e3 & 58.99 & 45.06\, \textcolor{red}{(-2)} & 77.99\, \textcolor{blue}{(+1)} & 69.44\, \textcolor{blue}{(+1)} & 39.55\, \textcolor{red}{(-6)} & 65.48\, \textcolor{blue}{(+1)} & 48.49\, \textcolor{red}{(-39)} & 52.70\, \textcolor{blue}{(+1)} & 70.03\, \textcolor{red}{(-5)} & 62.25\, \textcolor{red}{(-63)} & 87.27\, \textcolor{red}{(-2)} \\
T3Q-qwen2.5-14b-v1.0-e3 & 58.99 & 45.06\, \textcolor{red}{(-1)} & 77.99\, \textcolor{blue}{(+2)} & 69.44\, \textcolor{blue}{(+2)} & 39.55\, \textcolor{red}{(-5)} & 65.48\, \textcolor{blue}{(+2)} & 48.49\, \textcolor{red}{(-38)} & 52.70\, \textcolor{blue}{(+2)} & 70.03\, \textcolor{red}{(-4)} & 62.25\, \textcolor{red}{(-62)} & 87.27\, \textcolor{red}{(-1)} \\
CalmeRys-78B-Orpo-v0.1 & 58.86 & 46.29\, \textcolor{blue}{(+2)} & 76.23\, \textcolor{red}{(-1)} & 66.81\, \textcolor{red}{(-1)} & 40.56 & 63.35\, \textcolor{blue}{(+1)} & 46.56\, \textcolor{red}{(-145)} & 50.99\, \textcolor{blue}{(+1)} & 68.81\, \textcolor{red}{(-15)} & 60.84\, \textcolor{red}{(-192)} & 82.92\, \textcolor{red}{(-228)} \\
calme-3.1-instruct-78b & 58.73 & 45.81\, \textcolor{blue}{(+2)} & 76.85\, \textcolor{blue}{(+1)} & 65.88\, \textcolor{red}{(-4)} & 41.81\, \textcolor{blue}{(+3)} & 62.37 & 47.35\, \textcolor{red}{(-82)} & 49.34\, \textcolor{red}{(-1)} & 70.56 & 60.89\, \textcolor{red}{(-182)} & 82.22\, \textcolor{red}{(-271)} \\
calme-2.4-rys-78b & 57.54 & 44.52 & 75.71 & 65.11\, \textcolor{red}{(-15)} & 39.46\, \textcolor{red}{(-4)} & 61.44\, \textcolor{red}{(-1)} & 45.38\, \textcolor{red}{(-434)} & 48.27\, \textcolor{red}{(-1)} & 68.41\, \textcolor{red}{(-22)} & 59.65\, \textcolor{red}{(-698)} & 81.73\, \textcolor{red}{(-309)} \\
\_Spydaz\_Web\_AI\_AGI\_R1\_OmG\_Coder & 56.08 & 43.33 & 74.61\, \textcolor{red}{(-3)} & 62.63\, \textcolor{red}{(-99)} & 40.91\, \textcolor{blue}{(+4)} & 62.07\, \textcolor{blue}{(+1)} & 43.37\, \textcolor{red}{(-3307)} & 49.52\, \textcolor{blue}{(+2)} & 71.42\, \textcolor{blue}{(+4)} & 58.40\, \textcolor{red}{(-2863)} & 80.68\, \textcolor{red}{(-398)} \\
DS-R1-Distill-Q2.5-14B-Harmony\_V0.1 & 55.56 & 42.12 & 73.32\, \textcolor{red}{(-13)} & 67.38\, \textcolor{blue}{(+5)} & 35.10\, \textcolor{red}{(-106)} & 60.13\, \textcolor{red}{(-1)} & 48.80\, \textcolor{red}{(-25)} & 46.91 & 64.76\, \textcolor{red}{(-487)} & 62.79\, \textcolor{red}{(-34)} & 83.56\, \textcolor{red}{(-180)} \\
miscii-14b-1M-0128 & 54.23 & 40.07\, \textcolor{red}{(-5)} & 72.71\, \textcolor{red}{(-30)} & 67.03\, \textcolor{blue}{(+5)} & 32.76\, \textcolor{red}{(-348)} & 59.89\, \textcolor{red}{(-1)} & 47.07\, \textcolor{red}{(-103)} & 45.46\, \textcolor{red}{(-2)} & 60.83\, \textcolor{red}{(-2082)} & 61.65\, \textcolor{red}{(-86)} & 85.28\, \textcolor{red}{(-59)} \\
\_Spydaz\_Web\_AI\_AGI\_R1\_Top\_Student & 53.84 & 38.79\, \textcolor{red}{(-16)} & 75.70\, \textcolor{blue}{(+3)} & 60.66\, \textcolor{red}{(-230)} & 37.78\, \textcolor{red}{(-9)} & 58.16\, \textcolor{red}{(-20)} & 45.18\, \textcolor{red}{(-561)} & 42.22\, \textcolor{red}{(-54)} & 72.71\, \textcolor{blue}{(+9)} & 59.73\, \textcolor{red}{(-594)} & 79.50\, \textcolor{red}{(-599)} \\
DeepSeek-R1-Distill-Qwen-14B & 53.57 & 39.67\, \textcolor{red}{(-4)} & 72.14\, \textcolor{red}{(-46)} & 65.12\, \textcolor{red}{(-8)} & 33.46\, \textcolor{red}{(-229)} & 58.42\, \textcolor{red}{(-12)} & 47.74\, \textcolor{red}{(-63)} & 44.73\, \textcolor{red}{(-6)} & 64.06\, \textcolor{red}{(-727)} & 61.72\, \textcolor{red}{(-75)} & 82.07\, \textcolor{red}{(-278)} \\
Coma-II-14B & 53.44 & 39.29\, \textcolor{red}{(-6)} & 71.88\, \textcolor{red}{(-64)} & 66.14\, \textcolor{blue}{(+5)} & 31.99\, \textcolor{red}{(-526)} & 58.64\, \textcolor{red}{(-3)} & 51.12\, \textcolor{blue}{(+10)} & 44.04\, \textcolor{red}{(-16)} & 60.49\, \textcolor{red}{(-2250)} & 65.04\, \textcolor{blue}{(+11)} & 85.16\, \textcolor{red}{(-66)} \\
Galactic-Qwen-14B-Exp2 & 53.44 & 37.07\, \textcolor{red}{(-28)} & 75.58\, \textcolor{blue}{(+5)} & 65.56\, \textcolor{red}{(-2)} & 33.07\, \textcolor{red}{(-284)} & 61.39\, \textcolor{blue}{(+5)} & 44.92\, \textcolor{red}{(-819)} & 45.17 & 66.91\, \textcolor{red}{(-78)} & 59.52\, \textcolor{red}{(-834)} & 87.70\, \textcolor{blue}{(+12)} \\
o-distil-qwen & 53.31 & 39.28\, \textcolor{red}{(-5)} & 72.06\, \textcolor{red}{(-49)} & 64.94\, \textcolor{red}{(-9)} & 33.28\, \textcolor{red}{(-250)} & 58.29\, \textcolor{red}{(-13)} & 47.29\, \textcolor{red}{(-81)} & 44.54\, \textcolor{red}{(-6)} & 63.96\, \textcolor{red}{(-766)} & 61.28\, \textcolor{red}{(-118)} & 82.11\, \textcolor{red}{(-272)} \\
Gauss-Opus-14B-R999 & 53.31 & 38.87\, \textcolor{red}{(-9)} & 72.15\, \textcolor{red}{(-41)} & 66.13\, \textcolor{blue}{(+7)} & 31.71\, \textcolor{red}{(-604)} & 59.09\, \textcolor{blue}{(+3)} & 48.44\, \textcolor{red}{(-30)} & 44.29\, \textcolor{red}{(-9)} & 61.33\, \textcolor{red}{(-1861)} & 62.46\, \textcolor{red}{(-39)} & 85.07\, \textcolor{red}{(-69)} \\
Lion-Lamarck-v.1.1.0 & 53.17 & 40.30\, \textcolor{blue}{(+3)} & 69.99\, \textcolor{red}{(-329)} & 65.25\, \textcolor{red}{(-1)} & 32.98\, \textcolor{red}{(-302)} & 58.52\, \textcolor{red}{(-1)} & 46.99\, \textcolor{red}{(-107)} & 45.72\, \textcolor{blue}{(+6)} & 60.46\, \textcolor{red}{(-2255)} & 60.78\, \textcolor{red}{(-195)} & 82.16\, \textcolor{red}{(-267)} \\
Qwen2.5-14B-Vimarckoso-v3-IF-Variant & 53.04 & 40.55\, \textcolor{blue}{(+6)} & 69.39\, \textcolor{red}{(-474)} & 63.90\, \textcolor{red}{(-27)} & 33.11\, \textcolor{red}{(-271)} & 56.49\, \textcolor{red}{(-72)} & 45.57\, \textcolor{red}{(-345)} & 44.06\, \textcolor{red}{(-10)} & 59.12\, \textcolor{red}{(-2947)} & 59.17\, \textcolor{red}{(-2146)} & 78.37\, \textcolor{red}{(-798)} \\
Sombrero-Opus-14B-Sm1 & 52.91 & 38.05\, \textcolor{red}{(-14)} & 72.42\, \textcolor{red}{(-31)} & 65.84\, \textcolor{blue}{(+8)} & 30.79\, \textcolor{red}{(-881)} & 59.20\, \textcolor{blue}{(+7)} & 47.75\, \textcolor{red}{(-55)} & 44.36\, \textcolor{red}{(-4)} & 60.77\, \textcolor{red}{(-2109)} & 62.27\, \textcolor{red}{(-46)} & 85.70\, \textcolor{red}{(-31)} \\
Lion-Lamarck-v.1.0.9 & 52.91 & 38.89\, \textcolor{red}{(-4)} & 71.64\, \textcolor{red}{(-73)} & 64.85\, \textcolor{red}{(-6)} & 32.85\, \textcolor{red}{(-319)} & 58.51 & 47.21\, \textcolor{red}{(-80)} & 44.43\, \textcolor{red}{(-2)} & 63.26\, \textcolor{red}{(-1074)} & 61.22\, \textcolor{red}{(-120)} & 83.26\, \textcolor{red}{(-188)} \\
Llama-3-8B-Instruct\_dare\_ties-density-0.9 & 52.25 & 40.50\, \textcolor{blue}{(+8)} & 67.19\, \textcolor{red}{(-1231)} & 64.55\, \textcolor{red}{(-10)} & 31.73\, \textcolor{red}{(-596)} & 56.68\, \textcolor{red}{(-60)} & 49.05\, \textcolor{red}{(-2)} & 44.70\, \textcolor{blue}{(+2)} & 54.38\, \textcolor{red}{(-4133)} & 63.62\, \textcolor{blue}{(+3)} & 80.08\, \textcolor{red}{(-480)} \\
Viper-Coder-v1.1 & 52.12 & 36.29\, \textcolor{red}{(-35)} & 72.83\, \textcolor{red}{(-12)} & 65.72\, \textcolor{blue}{(+10)} & 29.48\, \textcolor{red}{(-1372)} & 58.46\, \textcolor{blue}{(+1)} & 46.01\, \textcolor{red}{(-208)} & 42.68\, \textcolor{red}{(-29)} & 60.43\, \textcolor{red}{(-2269)} & 61.01\, \textcolor{red}{(-144)} & 86.37\, \textcolor{blue}{(+5)} \\
Viper-Coder-v1.6-r999 & 52.12 & 36.29\, \textcolor{red}{(-34)} & 72.83\, \textcolor{red}{(-11)} & 65.72\, \textcolor{blue}{(+11)} & 29.48\, \textcolor{red}{(-1371)} & 58.46\, \textcolor{blue}{(+2)} & 46.01\, \textcolor{red}{(-207)} & 42.68\, \textcolor{red}{(-28)} & 60.43\, \textcolor{red}{(-2268)} & 61.01\, \textcolor{red}{(-143)} & 86.37\, \textcolor{blue}{(+6)} \\
\_Spydaz\_Web\_AI\_AGI\_R1\_Math\_Teacher & 52.12 & 39.30\, \textcolor{blue}{(+6)} & 70.39\, \textcolor{red}{(-247)} & 59.08\, \textcolor{red}{(-310)} & 36.52\, \textcolor{red}{(-24)} & 58.06\, \textcolor{red}{(-8)} & 45.47\, \textcolor{red}{(-372)} & 44.62\, \textcolor{blue}{(+4)} & 66.32\, \textcolor{red}{(-135)} & 59.77\, \textcolor{red}{(-551)} & 77.95\, \textcolor{red}{(-888)} \\
IF-reasoning-experiment-40 & 51.85 & 37.74\, \textcolor{red}{(-10)} & 69.98\, \textcolor{red}{(-322)} & 64.97\, \textcolor{blue}{(+2)} & 29.43\, \textcolor{red}{(-1407)} & 57.91\, \textcolor{red}{(-13)} & 45.28\, \textcolor{red}{(-473)} & 44.89\, \textcolor{blue}{(+9)} & 55.85\, \textcolor{red}{(-3924)} & 60.63\, \textcolor{red}{(-215)} & 82.80\, \textcolor{red}{(-219)} \\
Condor-Opus-14B-Exp & 51.85 & 37.33\, \textcolor{red}{(-11)} & 71.09\, \textcolor{red}{(-135)} & 64.07\, \textcolor{red}{(-11)} & 31.10\, \textcolor{red}{(-778)} & 58.28\, \textcolor{red}{(-3)} & 47.80\, \textcolor{red}{(-45)} & 43.44\, \textcolor{red}{(-12)} & 60.24\, \textcolor{red}{(-2367)} & 62.44\, \textcolor{red}{(-31)} & 84.60\, \textcolor{red}{(-100)} \\
Sombrero-Opus-14B-Sm4 & 51.85 & 35.87\, \textcolor{red}{(-35)} & 72.75\, \textcolor{red}{(-12)} & 65.63\, \textcolor{blue}{(+13)} & 29.54\, \textcolor{red}{(-1344)} & 58.33\, \textcolor{blue}{(+1)} & 49.00\, \textcolor{blue}{(+2)} & 41.75\, \textcolor{red}{(-60)} & 60.94\, \textcolor{red}{(-2016)} & 63.34\, \textcolor{blue}{(+3)} & 86.87\, \textcolor{blue}{(+18)} \\
\_Spydaz\_Web\_AI\_AGI\_R1\_Math\_AdvancedStudent & 51.85 & 39.33\, \textcolor{blue}{(+11)} & 69.55\, \textcolor{red}{(-417)} & 58.98\, \textcolor{red}{(-312)} & 35.93\, \textcolor{red}{(-46)} & 57.39\, \textcolor{red}{(-22)} & 43.53\, \textcolor{red}{(-3184)} & 44.13\, \textcolor{blue}{(+1)} & 64.94\, \textcolor{red}{(-400)} & 58.28\, \textcolor{red}{(-2916)} & 77.21\, \textcolor{red}{(-1046)} \\
Lamarck-14B-v0.6-002-model\_stock & 51.72 & 37.26\, \textcolor{red}{(-9)} & 70.26\, \textcolor{red}{(-265)} & 65.16\, \textcolor{blue}{(+10)} & 29.38\, \textcolor{red}{(-1426)} & 58.22\, \textcolor{red}{(-1)} & 45.19\, \textcolor{red}{(-539)} & 44.83\, \textcolor{blue}{(+12)} & 56.52\, \textcolor{red}{(-3773)} & 60.07\, \textcolor{red}{(-387)} & 83.28\, \textcolor{red}{(-176)} \\
Qwen2.5-14B-ReasoningMerge & 51.59 & 35.68\, \textcolor{red}{(-42)} & 72.32\, \textcolor{red}{(-25)} & 65.53\, \textcolor{blue}{(+13)} & 29.08\, \textcolor{red}{(-1554)} & 58.29\, \textcolor{blue}{(+3)} & 47.23\, \textcolor{red}{(-69)} & 42.05\, \textcolor{red}{(-41)} & 59.58\, \textcolor{red}{(-2728)} & 61.62\, \textcolor{red}{(-69)} & 87.14\, \textcolor{blue}{(+23)} \\
Llama-3-8B-Instruct\_breadcrumbs-density-0.7-gamma-0.01 & 51.59 & 39.12\, \textcolor{blue}{(+10)} & 67.26\, \textcolor{red}{(-1193)} & 64.93\, \textcolor{blue}{(+6)} & 29.59\, \textcolor{red}{(-1318)} & 55.98\, \textcolor{red}{(-104)} & 48.89\, \textcolor{blue}{(+4)} & 43.72\, \textcolor{red}{(-3)} & 52.05\, \textcolor{red}{(-4309)} & 64.31\, \textcolor{blue}{(+25)} & 81.35\, \textcolor{red}{(-311)} \\
DeepSeek-R1-Distill-Qwen-25.5B-Brainstorm & 51.46 & 36.69\, \textcolor{red}{(-16)} & 71.18\, \textcolor{red}{(-114)} & 63.46\, \textcolor{red}{(-29)} & 31.27\, \textcolor{red}{(-711)} & 56.86\, \textcolor{red}{(-42)} & 43.90\, \textcolor{red}{(-2971)} & 41.78\, \textcolor{red}{(-50)} & 62.59\, \textcolor{red}{(-1312)} & 57.90\, \textcolor{red}{(-3101)} & 82.73\, \textcolor{red}{(-218)} \\
Qwentinuum-14B-v013 & 51.46 & 37.87\, \textcolor{red}{(-1)} & 68.75\, \textcolor{red}{(-619)} & 64.65\, \textcolor{blue}{(+4)} & 29.08\, \textcolor{red}{(-1549)} & 57.03\, \textcolor{red}{(-31)} & 44.41\, \textcolor{red}{(-2253)} & 44.33\, \textcolor{blue}{(+9)} & 54.43\, \textcolor{red}{(-4116)} & 59.49\, \textcolor{red}{(-889)} & 81.58\, \textcolor{red}{(-299)} \\
Qwenvergence-14B-v12-Prose-DS & 51.46 & 34.95\, \textcolor{red}{(-61)} & 72.99\, \textcolor{blue}{(+7)} & 65.62\, \textcolor{blue}{(+19)} & 28.78\, \textcolor{red}{(-1687)} & 58.61\, \textcolor{blue}{(+17)} & 46.87\, \textcolor{red}{(-96)} & 41.77\, \textcolor{red}{(-51)} & 61.25\, \textcolor{red}{(-1882)} & 61.57\, \textcolor{red}{(-69)} & 87.38\, \textcolor{blue}{(+30)} \\
Llama-3-8B-Instruct\_breadcrumbs-density-0.9-gamma-0.01 & 51.46 & 38.90\, \textcolor{blue}{(+12)} & 67.27\, \textcolor{red}{(-1185)} & 64.80\, \textcolor{blue}{(+8)} & 29.43\, \textcolor{red}{(-1395)} & 55.87\, \textcolor{red}{(-113)} & 48.10\, \textcolor{red}{(-24)} & 43.52\, \textcolor{red}{(-2)} & 52.10\, \textcolor{red}{(-4301)} & 63.68\, \textcolor{blue}{(+20)} & 81.33\, \textcolor{red}{(-310)} \\
Qwenvergence-14B-v9 & 51.32 & 36.60\, \textcolor{red}{(-14)} & 70.21\, \textcolor{red}{(-267)} & 65.11\, \textcolor{blue}{(+15)} & 28.71\, \textcolor{red}{(-1713)} & 58.69\, \textcolor{blue}{(+21)} & 44.51\, \textcolor{red}{(-2157)} & 45.00\, \textcolor{blue}{(+21)} & 56.16\, \textcolor{red}{(-3841)} & 59.58\, \textcolor{red}{(-750)} & 84.96\, \textcolor{red}{(-60)} \\
Orca-2-13b & 51.19 & 37.11\, \textcolor{red}{(-4)} & 70.70\, \textcolor{red}{(-179)} & 60.94\, \textcolor{red}{(-183)} & 33.04\, \textcolor{red}{(-269)} & 56.03\, \textcolor{red}{(-92)} & 45.62\, \textcolor{red}{(-301)} & 42.58\, \textcolor{red}{(-17)} & 62.17\, \textcolor{red}{(-1479)} & 60.35\, \textcolor{red}{(-279)} & 78.44\, \textcolor{red}{(-767)} \\
NQLSG-Qwen2.5-14B-OriginalFusion & 51.19 & 36.86\, \textcolor{red}{(-7)} & 69.69\, \textcolor{red}{(-380)} & 64.58\, \textcolor{blue}{(+8)} & 29.08\, \textcolor{red}{(-1545)} & 58.86\, \textcolor{blue}{(+24)} & 46.23\, \textcolor{red}{(-153)} & 45.27\, \textcolor{blue}{(+25)} & 55.45\, \textcolor{red}{(-3976)} & 60.87\, \textcolor{red}{(-153)} & 84.51\, \textcolor{red}{(-94)} \\
Llama-3-8B-Instruct\_dare\_ties-density-0.7 & 51.06 & 38.73\, \textcolor{blue}{(+11)} & 66.93\, \textcolor{red}{(-1328)} & 63.37\, \textcolor{red}{(-23)} & 30.75\, \textcolor{red}{(-872)} & 55.74\, \textcolor{red}{(-120)} & 49.50\, \textcolor{blue}{(+26)} & 43.19\, \textcolor{red}{(-1)} & 54.11\, \textcolor{red}{(-4153)} & 64.18\, \textcolor{blue}{(+32)} & 79.80\, \textcolor{red}{(-514)} \\
Llama-3-8B-Instruct\_breadcrumbs-density-0.5-gamma-0.01 & 50.93 & 38.20\, \textcolor{blue}{(+9)} & 66.98\, \textcolor{red}{(-1304)} & 64.46\, \textcolor{blue}{(+8)} & 28.83\, \textcolor{red}{(-1662)} & 56.15\, \textcolor{red}{(-83)} & 47.80\, \textcolor{red}{(-32)} & 43.98\, \textcolor{blue}{(+9)} & 51.57\, \textcolor{red}{(-4333)} & 63.42\, \textcolor{blue}{(+17)} & 81.57\, \textcolor{red}{(-293)} \\
Barcenas-14b-phi-4 & 50.93 & 34.55\, \textcolor{red}{(-79)} & 72.77\, \textcolor{blue}{(+4)} & 63.95 & 29.56\, \textcolor{red}{(-1319)} & 58.38\, \textcolor{blue}{(+16)} & 48.61\, \textcolor{blue}{(+5)} & 41.76\, \textcolor{red}{(-45)} & 62.25\, \textcolor{red}{(-1443)} & 62.93\, \textcolor{blue}{(+3)} & 86.10\, \textcolor{blue}{(+16)} \\
Phi4-Slerp4-14B & 50.93 & 35.65\, \textcolor{red}{(-32)} & 71.12\, \textcolor{red}{(-116)} & 63.91\, \textcolor{red}{(-2)} & 29.72\, \textcolor{red}{(-1250)} & 58.43\, \textcolor{blue}{(+19)} & 49.27\, \textcolor{blue}{(+23)} & 42.86\, \textcolor{red}{(-4)} & 59.66\, \textcolor{red}{(-2667)} & 63.54\, \textcolor{blue}{(+21)} & 85.25\, \textcolor{red}{(-31)} \\
Phi-4-AbliteratedRP & 50.93 & 36.33\, \textcolor{red}{(-13)} & 69.97\, \textcolor{red}{(-305)} & 63.95\, \textcolor{blue}{(+3)} & 29.64\, \textcolor{red}{(-1277)} & 57.93\, \textcolor{blue}{(+9)} & 48.55\, \textcolor{blue}{(+3)} & 42.74\, \textcolor{red}{(-6)} & 58.16\, \textcolor{red}{(-3314)} & 62.95\, \textcolor{blue}{(+9)} & 84.54\, \textcolor{red}{(-86)} \\
Sombrero-Opus-14B-Sm2 & 50.79 & 33.82\, \textcolor{red}{(-112)} & 73.15\, \textcolor{blue}{(+18)} & 64.74\, \textcolor{blue}{(+16)} & 28.38\, \textcolor{red}{(-1860)} & 57.58\, \textcolor{blue}{(+2)} & 44.13\, \textcolor{red}{(-2655)} & 40.12\, \textcolor{red}{(-119)} & 60.87\, \textcolor{red}{(-2030)} & 59.29\, \textcolor{red}{(-2037)} & 87.43\, \textcolor{blue}{(+41)} \\
DRT-o1-7B & 50.79 & 36.20\, \textcolor{red}{(-14)} & 70.97\, \textcolor{red}{(-129)} & 60.95\, \textcolor{red}{(-173)} & 32.61\, \textcolor{red}{(-342)} & 58.52\, \textcolor{blue}{(+26)} & 45.38\, \textcolor{red}{(-397)} & 44.02\, \textcolor{blue}{(+15)} & 63.94\, \textcolor{red}{(-746)} & 59.03\, \textcolor{red}{(-2225)} & 81.34\, \textcolor{red}{(-298)} \\
\_Spydaz\_Web\_AI\_AGI\_R1\_Math\_Student & 50.79 & 37.19\, \textcolor{blue}{(+7)} & 70.83\, \textcolor{red}{(-142)} & 56.09\, \textcolor{red}{(-556)} & 37.08\, \textcolor{blue}{(+15)} & 55.62\, \textcolor{red}{(-132)} & 44.28\, \textcolor{red}{(-2374)} & 40.96\, \textcolor{red}{(-68)} & 69.13\, \textcolor{blue}{(+31)} & 58.04\, \textcolor{red}{(-3018)} & 75.47\, \textcolor{red}{(-1320)} \\
orca\_mini\_v7\_72b & 50.66 & 35.76\, \textcolor{red}{(-21)} & 71.21\, \textcolor{red}{(-95)} & 60.94\, \textcolor{red}{(-174)} & 32.70\, \textcolor{red}{(-324)} & 56.05\, \textcolor{red}{(-79)} & 48.84\, \textcolor{blue}{(+14)} & 40.28\, \textcolor{red}{(-106)} & 62.77\, \textcolor{red}{(-1224)} & 62.43\, \textcolor{red}{(-11)} & 81.77\, \textcolor{red}{(-266)} \\
Phi-4-Megatron-Empathetic & 50.66 & 34.12\, \textcolor{red}{(-94)} & 72.94\, \textcolor{blue}{(+20)} & 63.16\, \textcolor{red}{(-28)} & 29.81\, \textcolor{red}{(-1204)} & 56.95\, \textcolor{red}{(-20)} & 47.38\, \textcolor{red}{(-39)} & 39.45\, \textcolor{red}{(-182)} & 63.43\, \textcolor{red}{(-967)} & 61.69\, \textcolor{red}{(-42)} & 85.26\, \textcolor{red}{(-22)} \\
orca\_mini\_v3\_70b & 50.66 & 33.53\, \textcolor{red}{(-123)} & 74.97\, \textcolor{blue}{(+39)} & 60.03\, \textcolor{red}{(-233)} & 32.99\, \textcolor{red}{(-268)} & 55.88\, \textcolor{red}{(-98)} & 45.70\, \textcolor{red}{(-259)} & 38.05\, \textcolor{red}{(-315)} & 69.14\, \textcolor{blue}{(+35)} & 60.01\, \textcolor{red}{(-387)} & 81.63\, \textcolor{red}{(-281)} \\
Llama-3-8B-Instruct\_dare\_ties-density-0.3 & 50.66 & 38.87\, \textcolor{blue}{(+24)} & 65.67\, \textcolor{red}{(-1786)} & 63.20\, \textcolor{red}{(-20)} & 30.54\, \textcolor{red}{(-919)} & 55.06\, \textcolor{red}{(-199)} & 49.84\, \textcolor{blue}{(+40)} & 43.18\, \textcolor{blue}{(+9)} & 51.56\, \textcolor{red}{(-4324)} & 64.44\, \textcolor{blue}{(+45)} & 79.53\, \textcolor{red}{(-551)} \\
QwQ-Buddy-32B-Alpha & 50.53 & 35.66\, \textcolor{red}{(-22)} & 71.08\, \textcolor{red}{(-111)} & 60.82\, \textcolor{red}{(-176)} & 32.33\, \textcolor{red}{(-398)} & 56.19\, \textcolor{red}{(-69)} & 48.48\, \textcolor{blue}{(+7)} & 41.94\, \textcolor{red}{(-23)} & 63.77\, \textcolor{red}{(-813)} & 62.73\, \textcolor{blue}{(+6)} & 78.23\, \textcolor{red}{(-802)} \\
\bottomrule
\end{tabular}}
\label{tab:meme_scores_summary_OpenLLM-MUSR}
\end{table}

\section{Applications and Case Studies}
\label{app:Applications_case_studies}
\subsection{Characterizing Model Similarities and Differences Through Meme Scores}
\label{app:commondity_characterize_Meme_scores}
\begin{figure}[H]
  \centering
  \includegraphics[width=0.8\linewidth]{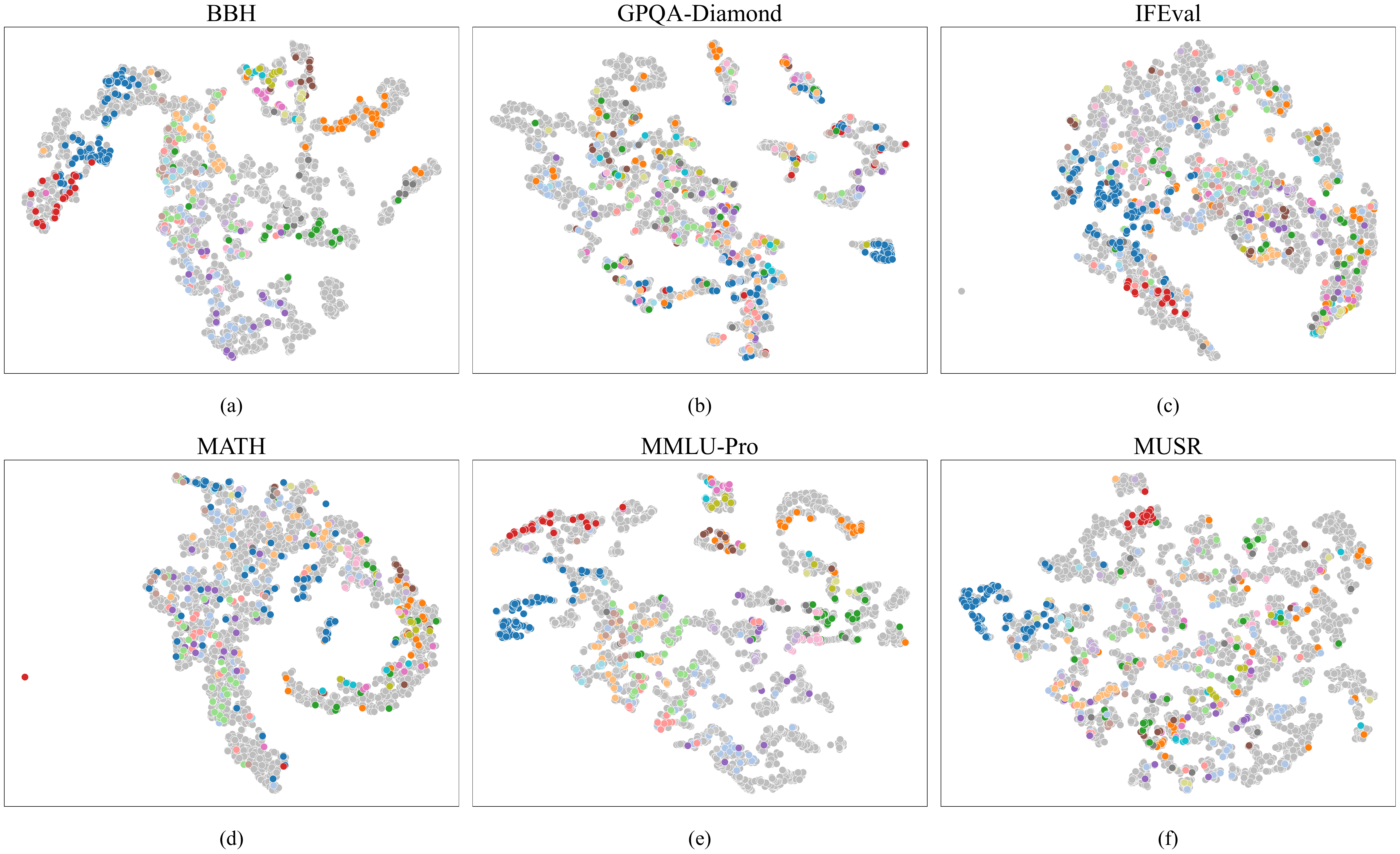}
  \caption{\textbf{Commonality and divergence among models revealed by Meme Scores (TSNE visualization).}}
  \label{fig:hf_combined_models_tsne}
\end{figure}

\begin{figure}[H]
  \centering
  \includegraphics[width=0.8\linewidth]{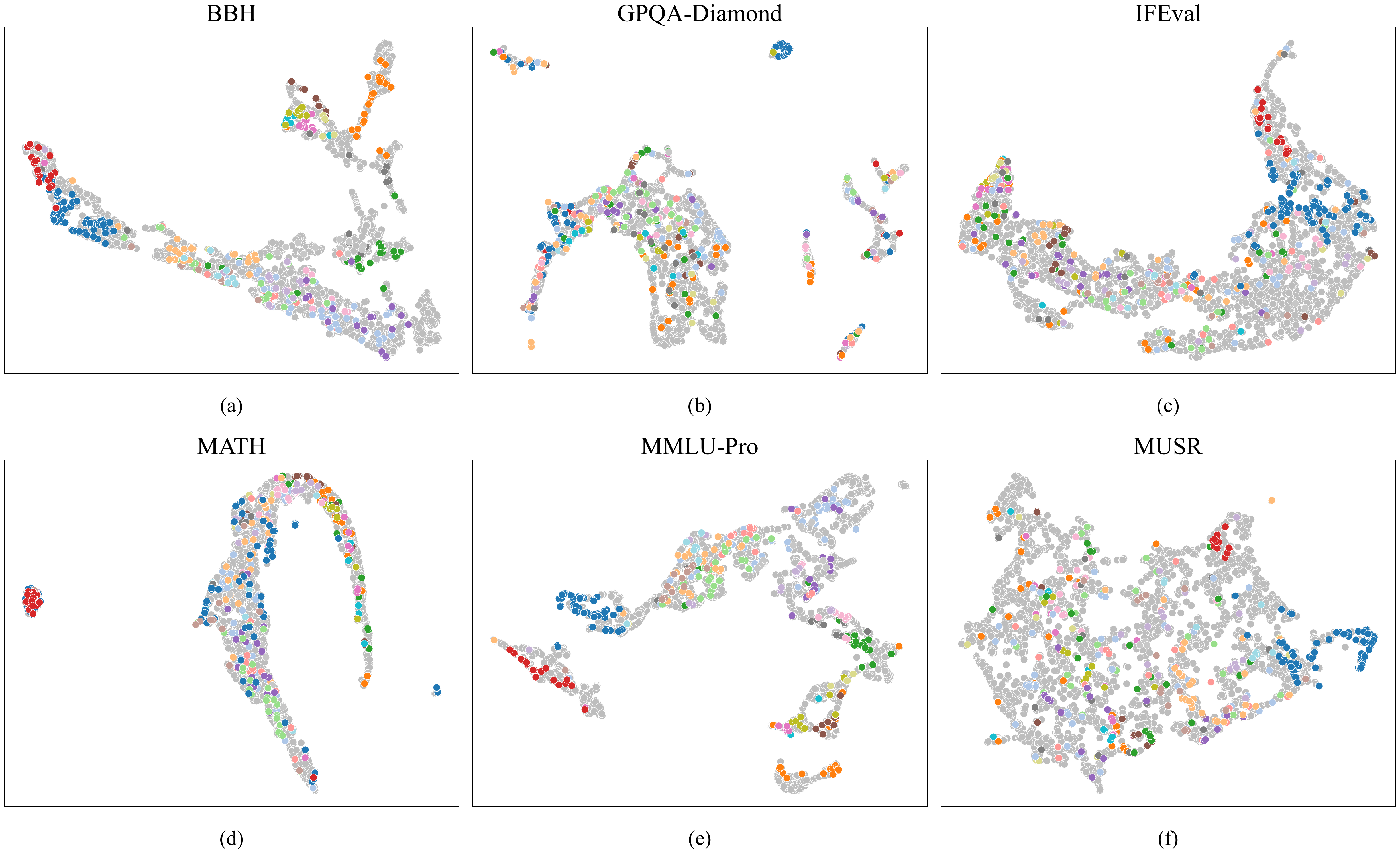}
  \caption{\textbf{Commonality and divergence among models revealed by Meme Scores (UMAP visualization).}}
  \label{fig:hf_combined_models_umap}
\end{figure}

Figure~\ref{fig:hf_combined_models_tsne} supplements the visualizations in Section~\ref{subsec:model_side_meme_profiling} by showing the t-SNE embeddings for all datasets in the Open LLM Leaderboard experiments. Models that share the same reported base model tend to lie close to one another in the embedding space, forming coherent local groups. This clustering pattern is consistent across datasets, suggesting that Meme Scores capture behavioral similarity. To further verify this structure, Figure~\ref{fig:hf_combined_models_umap} presents a UMAP projection based on the same 1D Meme Score representations. 

\begin{figure}[H]
  \centering
  \includegraphics[width=0.8\linewidth]{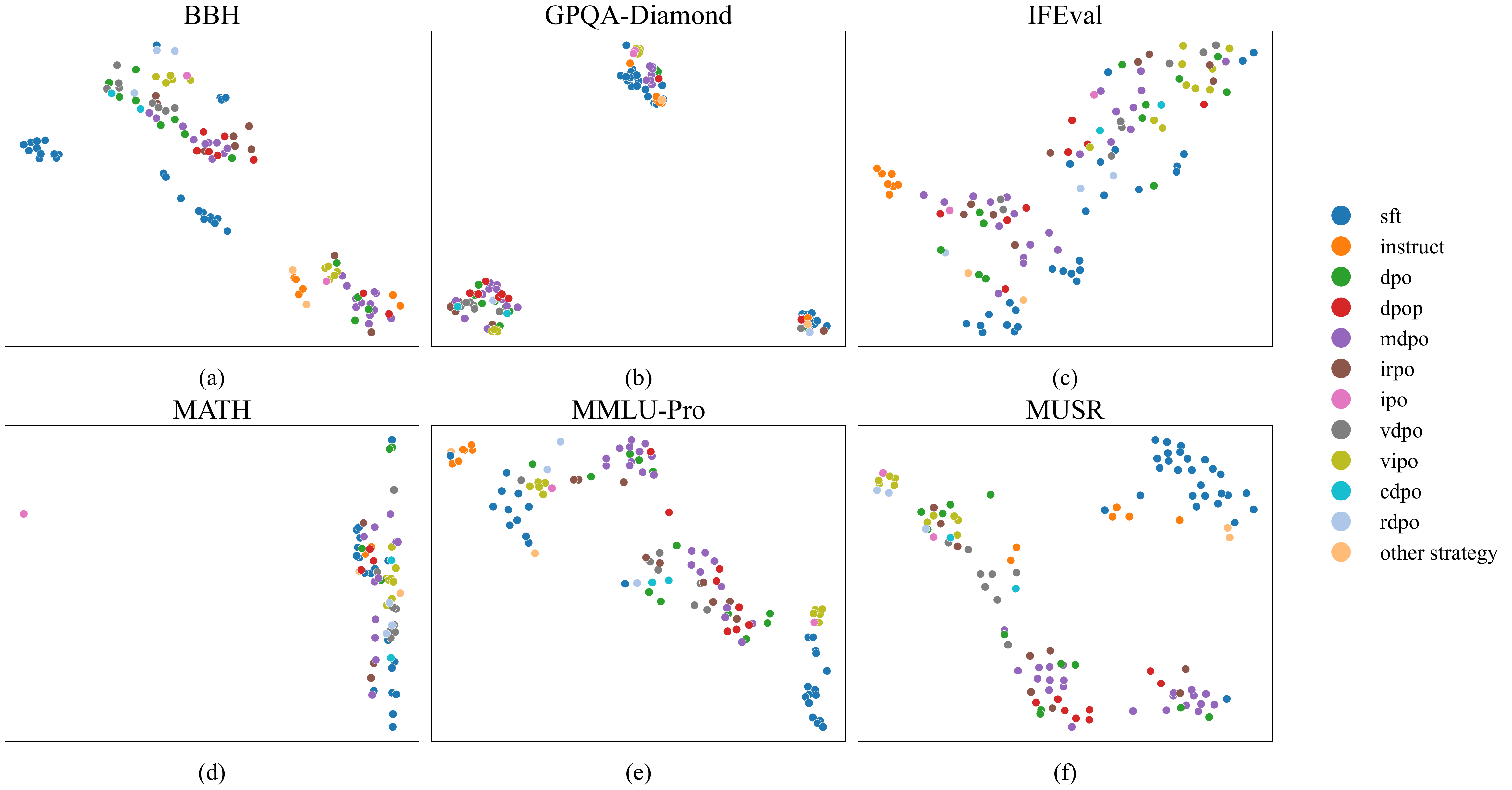}
  \caption{\textbf{Qwen2.5-0.5B models colored by training strategies (t-SNE visualization).}}
  \label{fig:qwen_training_strategies_tsne}
\end{figure}

\begin{figure}[H]
  \centering
  \includegraphics[width=0.8\linewidth]{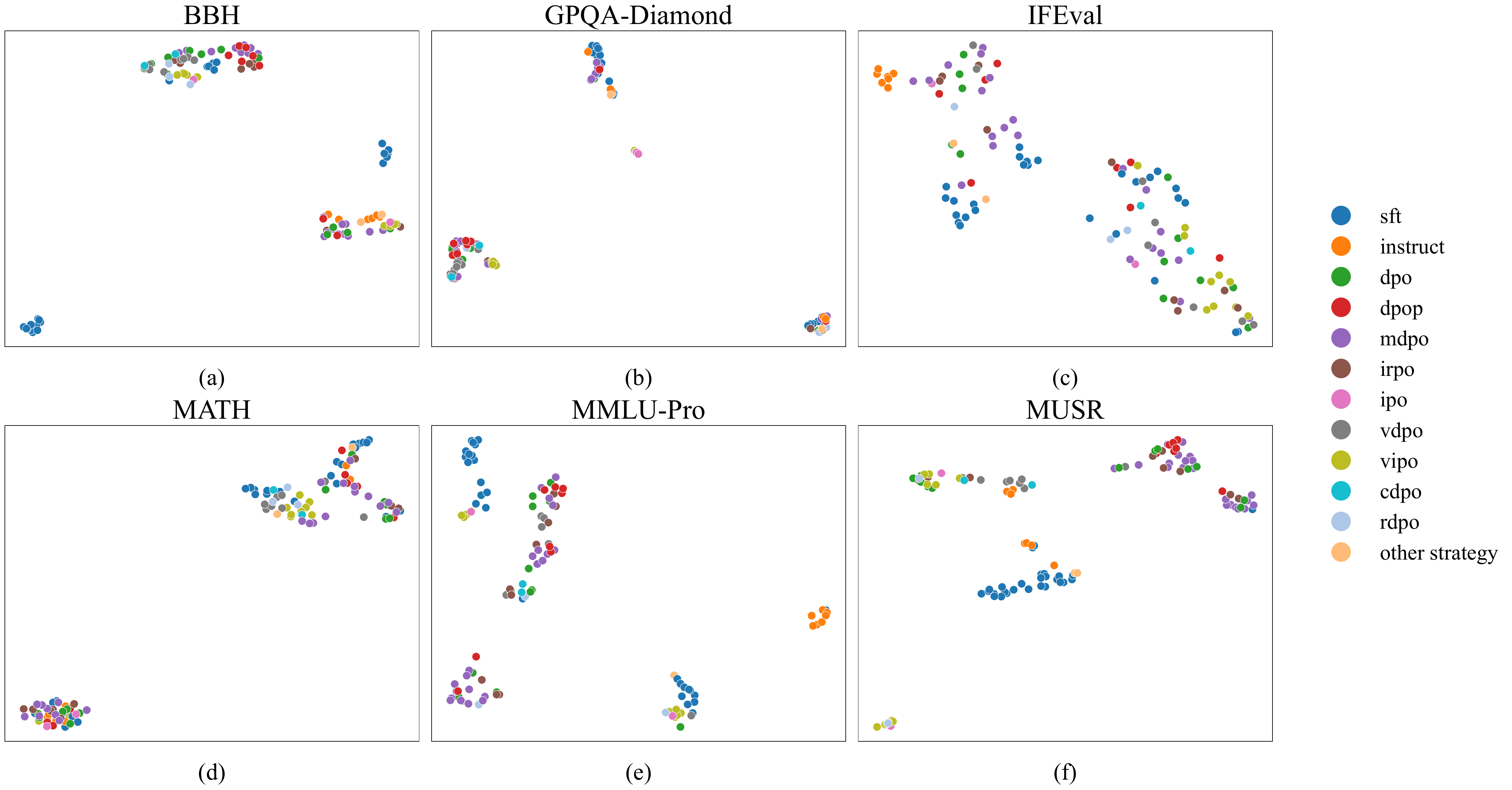}
  \caption{\textbf{Qwen2.5-0.5B models colored by training strategies (UMAP visualization).}}
  \label{fig:qwen_training_strategies_umap}
\end{figure}

This work also visualizes the Qwen2.5-0.5B family by coloring models according to their respective strategies. Figure~\ref{fig:qwen_training_strategies_tsne} (t-SNE) and Figure~\ref{fig:qwen_training_strategies_umap} (UMAP) show that models using the same strategy also appear close to one another in the visualization. Building on this characterization of models via the Probing Memes paradigm, researchers can focus on clusters of nearby models in this space to study how shared underlying design choices (e.g., base architectures, training strategies, or data sources) contribute to both common behaviors and systematic differences.

\subsection{Meme-Guided Model Selection Application on MATH}
\label{app:meme-guided-routing}
This section examines whether Meme Scores can support task-aware model selection on the MATH benchmark~\citep{hendrycksmath2021} by generalizing from Difficulty Meme Scores computed on MATH-500 to routing decisions on the remaining MATH items (excluding MATH-500). Based on Table~\ref{tab:memescores_definitions}, under comparable overall accuracy, a higher Difficulty meme score suggests stronger performance on harder items, whereas a lower Difficulty Meme Score suggests stronger performance on easier items. Models are paired to have similar overall accuracy on MATH-500, but different Difficulty Meme Scores, and MATH-FULL items are then routed between the paired models according to the dataset’s pre-annotated difficulty levels. Results indicate that, under a fixed total item budget, assigning harder items to high-Difficulty models and easier items to low-Difficulty models improves overall accuracy.

\noindent\textbf{Models.} Following Table~\ref{tab:model-selection-math500}, the study first computes each model’s \emph{Difficulty} Meme Score on MATH-500, and then selects two model pairs with similar overall accuracy but different \emph{Difficulty} scores.
For the stronger pair (Pair~1), doubao-seed-1-6-flash-250715 and MiniMax-Text-01(CoT) achieve comparable accuracy on MATH-500 (75.40 vs.\ 75.20), while the former has a higher \emph{Difficulty} score (57.73 vs.\ 53.15).
For the weaker pair (Pair~2), glm-4.5 and gpt-4.1-mini-2025-04-14 also have similar accuracy (49.20 vs.\ 49.00), with a noticeable difference in \emph{Difficulty} (25.47 vs.\ 27.66).

\begin{table}[H]
\centering
\small
\caption{\textbf{Difficulty-aware model selection.}
For each pair two models are selected with similar overall accuracy but different Difficulty meme score.
``Acc'' is the accuracy on MATH-500, ``Diff.'' is the Difficulty meme score on MATH-500.}
\label{tab:model-selection-math500}
\begin{tabular}{p{5.5cm}cc}
\toprule
Model & Acc & Diff. \\
\midrule
\multicolumn{3}{l}{\textit{Pair 1 (stronger models, hi-diff. vs.\ low-diff.)}} \\
doubao-seed-1-6-flash-250715 & 75.40 & 57.73 \\
MiniMax-Text-01(CoT)         & 75.20 & 53.15 \\
\midrule
\multicolumn{3}{l}{\textit{Pair 2 (weaker models, hi-diff. vs.\ low-diff.)}} \\
glm-4.5                      & 49.20 & 25.47 \\
gpt-4.1-mini-2025-04-14      & 49.00 & 27.66 \\
\bottomrule
\end{tabular}
\end{table}

\noindent\textbf{Meme Score based routing and baselines.}
The same pairs are then evaluated on the full test set of the MATH benchmark (MATH-FULL, excluding MATH-500). For each pair, the analysis considers the 4{,}500 items answered by both models, with 2{,}276 items labeled as Level~4 or 5 and 2{,}224 items labeled as Level~1, 2, or 3. The model with the higher \emph{Difficulty} Meme Score is treated as $M^{\mathrm{hi}}$ (high-\emph{Difficulty} model), and the other as $M^{\mathrm{lo}}$ (low-\emph{Difficulty} model). A simple \emph{Meme Score based routing} policy is then applied: items with level 4 or 5 are routed to $M^{\mathrm{hi}}$ (hard subset), and items with level 1, 2, or 3 are routed to $M^{\mathrm{lo}}$ (easy subset). To control for the effect of simply combining two models, a \emph{random per-level balanced routing} is used as a baseline: for each difficulty level, items are randomly split into two equal halves, one assigned to $M^{\mathrm{hi}}$ and the other to $M^{\mathrm{lo}}$, and this procedure is repeated 10 times with different random seeds (from 0 to 9). In addition, a \emph{best single model} baseline is included by evaluating each individual model in the pair on the full 4{,}500 items (without routing) and reporting the higher overall accuracy between the two models.

\noindent\textbf{Results.}
Table~\ref{tab:model-selection-mathfull} shows that \emph{Difficulty}-based routing consistently improves performance across both model pairs. 
For Pair~1, routing achieves 77.29\% overall accuracy, outperforming balanced routing by +3.15 points and the best single model by +1.20 points, with gains on both Hard (L4--5, +1.41) and Easy (L1--3, +4.91) subsets. 
For Pair~2, routing reaches 48.51\% overall accuracy, improving over balanced routing by +2.76 points and over the best single model by +1.69 points, again yielding clear gains on Hard (+3.61) and Easy (+1.89) subsets. 

\begin{table}[H]
\centering
\small
\caption{\textbf{Task-aware model selection on MATH-FULL (Difficulty-based routing).}
For each pair, a \emph{high-Difficulty} (hi-Diff) model and a \emph{low-Difficulty} (low-Diff) model are selected based on the \emph{Difficulty} meme score on MATH-500.
Routing assigns Level~4--5 items to the hi-Diff model and Level~1--3 items to the low-Diff model.
The balanced routing baseline uses a per-level 50/50 random split between the two models, repeated 10 times (seeds 0--9).
The table reports overall accuracy (mean $\pm$ std for random) and gains relative to baselines.}
\label{tab:model-selection-mathfull}
\begin{tabular}{p{5cm}p{2.5cm}p{2.5cm}p{5cm}}
\toprule
\multicolumn{4}{c}{\textit{Pair 1: doubao-seed-1-6-flash-250715 (hi-Diff) vs MiniMax-Text-01(CoT) (low-Diff)}} \\
 Method & Hard (L4--5) & Easy (L1--3) & Overall \\
\midrule
Difficulty Routing & \textbf{66.65} & \textbf{88.17} & \textbf{77.29} \\
Balanced Routing & 65.24 ($\pm$ 0.53) & 83.26 ($\pm$ 0.34) & 74.14 ($\pm$ 0.27) \\
Best Single Model (Without Routing) & -- & -- & 76.09 \\
Gains of Difficulty Routing
& \textbf{+1.41}
& \textbf{+4.91}
& \textbf{+3.15} vs Balanced,\ \textbf{+1.20} vs Single \\
\midrule
\multicolumn{4}{c}{\textit{Pair 2: gpt-4.1-mini-2025-04-14 (hi-Diff) vs glm-4.5 (low-Diff)}} \\
Method & Hard (L4--5) & Easy (L1--3) & Overall \\
\midrule
Difficulty Routing & \textbf{37.17} & \textbf{60.12} & \textbf{48.51} \\
Balanced Routing & 33.56 ($\pm$ 0.52) & 58.23 ($\pm$ 0.33) & 45.75 ($\pm$ 0.32) \\
Best Single Model (Without Routing) & -- & -- & 46.82 \\
Gains of Difficulty Routing
& \textbf{+3.61}
& \textbf{+1.89}
& \textbf{+2.76} vs Balanced,\ \textbf{+1.69} vs Single \\
\bottomrule
\end{tabular}
\end{table}

Overall, this generalization-based application suggests that routing items by difficulty to Difficulty-specialized models can produce a more accurate system than either random per-level splitting or using a single model alone. Moreover, under the Probing Memes Paradigm, Meme Score based model selection leverages interpretable behavioral traits and can improve system-level accuracy in tasks such as multi-agent pipelines, while keeping the total number of queried items fixed.

\subsection{Case Study: Diagnosing High-Surprise Items Beyond Random Guessing}
\label{app:high-surprise-case}
Motivated by the prevalence of high-surprise items in Figure~\ref{fig:datasets_landscape_3d}, particularly in SimpleQA, this subsection presents a small case study of the five highest-surprise items in each of the three datasets analyzed under the \emph{curated population}. To assess whether these high-surprise effects reflect genuine model capability rather than stochastic variability, each selected item is re-evaluated with repeated runs to measure the stability of the observed failures and successes. The source of surprise is dataset-dependent: on MATH-500, it is mainly driven by failures of stronger models, whereas on MMLU-Redux and SimpleQA, it is mainly driven by successes of weaker models.

For each selected item and each highlighted model, the evaluation issues 20 independent queries under three querying configurations: 
\begin{enumerate}
    \item \textbf{Deterministic setting ($T{=}0.0$).} Temperature is set to zero.
    \item \textbf{Higher-temperature sampling ($T{=}0.6$).} Temperature is set to $0.6$, which encourages more diverse model outputs.
    \item \textbf{Deterministic setting with hint ($T{=}0.0$ + hint).} Temperature is set to zero, and the prompt includes an explicit instruction:
    \emph{``If you do not know the answer or are not confident about which option is correct, do not guess. Instead, answer exactly `I don't know' or `I'm not sure'.''} (For MCQ datasets such as MMLU-Redux, ``do not guess'' is replaced by ``do not guess any option''.)
\end{enumerate}

\begin{table}[H]
\centering
\small
\setlength{\tabcolsep}{6pt}
\renewcommand{\arraystretch}{1.05}
\caption{\textbf{Repeat-evaluation outcomes for highlighted items across three datasets.}
Each cell reports the average correctness (\%) over 20 runs under three settings: $T{=}0.0$, $T{=}0.6$, and $T{=}0.0$ with an explicit hint. For models with internal reasoning (IR), the temperature is the model default (not forced to $T{=}0.0$).}
\label{tab:case-study-high-surprise}
\begin{tabular}{l p{5.5cm} c c c c}
\toprule
\textbf{Item} & \textbf{Model} & \textbf{Surprise} & \textbf{$T{=}0.0$ (IR: default)} & \textbf{$T{=}0.6$} & \textbf{$T{=}0.0$ (IR: default) + hint} \\
\midrule

\multicolumn{6}{l}{\textit{Dataset: MMLU-Redux (surprise: weaker-success)}}\\
\midrule
\multirow{1}{*}{Item~1} & doubao-seed-1-6-flash-250715 & 0.1530 & 75 & 60 & 25 \\
\addlinespace[2pt]
\multirow{1}{*}{Item~2} & doubao-seed-1-6-flash-250715 & 0.1530 & 100 & 95 & 100 \\
\addlinespace[2pt]
\multirow{2}{*}{Item~3} & gpt-4.1-nano-2025-04-14(CoT) & \multirow{2}{*}{0.1342} & 65 & 65 & 60 \\
                         & gpt-4.1-nano-2025-04-14 &                       & 10 & 40 & 0 \\
\addlinespace[2pt]
\multirow{1}{*}{Item~4} & glm-4.5-air & 0.1335 & 100 & 55 & 100 \\
\addlinespace[2pt]
\multirow{1}{*}{Item~5} & glm-4.5-air & 0.1335 & 15 & 15 & 0 \\
\addlinespace[2pt]

\midrule
\multicolumn{6}{l}{\textit{Dataset: MATH-500 (surprise: stronger-failure)}}\\
\midrule
\multirow{1}{*}{Item~1} & glm-4.5-air(IR) & 0.2257 & 95 & -- & 80 \\
\addlinespace[2pt]
\multirow{1}{*}{Item~2} & doubao-seed-1-6-flash-250715(CoT) & 0.1499 & 95 & 100 & 100 \\
\addlinespace[2pt]
\multirow{1}{*}{Item~3} & kimi-k2-0711-preview & 0.1269 & 0 & 25 & 15 \\
\addlinespace[2pt]
\multirow{1}{*}{Item~4} & kimi-k2-0711-preview & 0.1269 & 0 & 15 & 0 \\
\addlinespace[2pt]
\multirow{3}{*}{Item~5} & doubao-seed-1-6-flash-250715(IR) & \multirow{3}{*}{0.0957} & 40 & -- & 5 \\
                         & qwen3-32b(CoT) &                       & 100 & 100 & 100 \\
                         & gpt-4.1-nano-2025-04-14 &                       & 0 & 0 & 20 \\
\addlinespace[2pt]

\midrule
\multicolumn{6}{l}{\textit{Dataset: SimpleQA (surprise: weaker-success)}}\\
\midrule
\multirow{1}{*}{Item~1} & qwen3-30b-a3b & 0.2919 & 35 & 35 & 0 \\
\addlinespace[2pt]
\multirow{1}{*}{Item~2} & doubao-seed-1-6-flash-250715 & 0.2911 & 25 & 30 & 0 \\
\addlinespace[2pt]
\multirow{1}{*}{Item~3} & qwen3-30b-a3b(CoT) & 0.2820 & 10 & 15 & 0 \\
\addlinespace[2pt]
\multirow{1}{*}{Item~4} & qwen3-30b-a3b(CoT) & 0.2820 & 0 & 0 & 0 \\
\addlinespace[2pt]
\multirow{1}{*}{Item~5} & doubao-seed-1-6-flash-250715(IR) & 0.2724 & 20 & -- & 0 \\
\addlinespace[2pt]

\bottomrule

\end{tabular}
\end{table}

\noindent\textbf{Findings.}
Table~\ref{tab:case-study-high-surprise} reveals three empirical patterns:
\begin{enumerate}
  \item \textbf{High surprise mixes deterministic capability and stochasticity.}
  Roughly half of the high-surprise effects appear stable across repeated runs, consistent with deterministic knowledge and capability (MMLU-Redux Item~2, \texttt{doubao-seed-1-6-flash-250715}, accuracy: 100/95/100), while the rest are dominated by variability and collapse under re-evaluation (MMLU-Redux Item~5, \texttt{glm-4.5-air}, accuracy: 15/15/0).

  \item \textbf{Hints can help or hurt.}
  The ``do not guess'' hint can shift correctness in either direction, improving some cases while degrading others (MATH-500 Item~5, \texttt{gpt-4.1-nano-2025-04-14} improves with hint). On SimpleQA, many high-surprise ``successes'' behave like random guesses and vanish entirely once guessing is discouraged (all selected SimpleQA high-surprise items reach 0 with a hint).

  \item \textbf{Item-dependent stability within a model.}
  The same model can be stable on some high-surprise items but unstable on others (\texttt{glm-4.5-air} on Items 4 and 5 in MMLU-Redux), indicating that uncertainty is strongly probe-specific rather than a global model trait.
\end{enumerate}

Taken together, these results indicate that high-surprise effects reflect both genuine model behavior and randomness, and that repeated evaluation is necessary to interpret them reliably. The Probing Memes paradigm and the associated \emph{surprise} property \textbf{make such patterns visible at the probe level and support targeted case studies like this one.}

\section{Stability of Probe Properties and Meme Scores to Population Size}
\label{app:stability_population_size}
This experiment evaluates how stable and sensitive both Probe Properties and Meme Scores of models are to the size and composition of the model population used to estimate them. 

\noindent\textbf{Setup.}
The subsampling analyses use the same model population described in Section~\ref{subsec:experimental_setups}, considering a total of $|\mathcal{M}|=53$ models, and examine stability to population size by varying the subsample $size \in \{5,10,20,30,40,50\}$. For each $size$, the subsampling procedure is repeated $10$ times: in each repeat, a subset of $size$ models is drawn from $\mathcal{M}$ (The subsets are different across repeats). For each subsample, the complete pipeline is rerun using the $size$ selected models, producing a full set of Probe Properties and Meme Scores for all models in $\mathcal{M}$ derived from that subsample. Stability is then quantified by comparing all pairs of subsamples at the same $size$: at the data side, stability is measured by the Jensen--Shannon (JS) divergence (Gaussian KDE, with bandwidth $h$ chosen based on Silverman’s rule) between the resulting distributions of Probe Properties, and at the model side, stability is measured by the Spearman rank correlation between the corresponding Meme Scores.

\begin{table}[H]
\caption{\textbf{Properties' stability under random subsampling.}
$size$ (from 5 to 50) denotes the number of models used to estimate probe properties in each subsample.
Entries report the Jensen--Shannon (JS) divergence among subsample-based probe properties, averaged over datasets; lower is better. Bridge is undefined at $size{=}5$ because the perception spans becomes too sparse to form cross-cluster connections.}
\label{tab:probe-sampling-js-k}
\centering
\begin{tabular}{lcccccc}
\toprule
\multirow{2}{*}{Property} & \multicolumn{6}{c}{Jensen--Shannon (JS) divergence} \\
\cmidrule(lr){2-7}
 & $size=5$ & $size=10$ & $size=20$ & $size=30$ & $size=40$ & $size=50$ \\
\midrule
difficulty & 0.0329 & 0.0260 & 0.0153 & 0.0080 & 0.0053 & 0.0017 \\
risk & 0.1991 & 0.1077 & 0.0509 & 0.0425 & 0.0181 & 0.0029 \\
uniqueness & 0.5276 & 0.3239 & 0.1810 & 0.1036 & 0.0980 & 0.0289 \\
surprise & 0.1646 & 0.0957 & 0.0453 & 0.0270 & 0.0194 & 0.0123 \\
typicality & 0.4526 & 0.0756 & 0.0210 & 0.0164 & 0.0101 & 0.0103 \\
bridge & -- & 0.3459 & 0.1940 & 0.1839 & 0.1311 & 0.1123 \\
\bottomrule
\end{tabular}
\end{table}

\noindent\textbf{Stability of Probe Properties.}
As Table~\ref{tab:probe-sampling-js-k} shows, across all six MPPs, increasing the subsample size $size$ consistently reduces the JS divergence between subsample-based estimates, indicating improved stability on the data side. Empirically, by $size{=}20$, most MPPs already achieve low divergence (JS $<0.1$) except uniqueness and bridge, and by $size{=}40$, all properties satisfy JS $<0.1$ except bridge. This aligns with their definitions: uniqueness depends on the pairwise probe--probe relationships induced by the Perception Matrix, while bridge relies on the more complex cluster structure constructed from model behaviors and is easily affected by changes in cluster labels. As a result, both properties are more sensitive to the choice of $size$.

\begin{table}[H]
\caption{\textbf{Meme Scores' stability under random subsampling.} $size$ denotes the number of models used to estimate probe properties in each subsample. Entries report the mean Spearman rank correlation among subsample-based model scores across repeats; higher is better. Bridge-based Meme Scores (Bridge and Robustness) are undefined at $size{=}5$ because the perception spans becomes too sparse to form cross-cluster connections to calculate the bridge property.}
\label{tab:phemotype-spearman-k}
\centering
\begin{tabular}{lcccccc}
\toprule
\multirow{2}{*}{Meme Score} & \multicolumn{6}{c}{Spearman Rank Correlation} \\
\cmidrule(lr){2-7}
 & $size$ = 5 & $size$ = 10 & $size$ = 20 & $size$ = 30 & $size$ = 40 & $size$ = 50 \\
\midrule
Difficulty & 0.9577 & 0.9808 & 0.9925 & 0.9960 & 0.9980 & 0.9994 \\
Uniqueness & 0.9446 & 0.9718 & 0.9876 & 0.9928 & 0.9965 & 0.9989 \\
Risk & 0.9940 & 0.9954 & 0.9973 & 0.9983 & 0.9987 & 0.9996 \\
Surprise & 0.6954 & 0.7575 & 0.8476 & 0.9072 & 0.9539 & 0.9887 \\
Typicality & 0.9536 & 0.9817 & 0.9940 & 0.9961 & 0.9972 & 0.9984 \\
Bridge & -- & 0.9932 & 0.9932 & 0.9919 & 0.9929 & 0.9963 \\
Mastery & 0.9335 & 0.9687 & 0.9898 & 0.9943 & 0.9957 & 0.9977 \\
Ingenuity & 0.6131 & 0.6932 & 0.8196 & 0.8866 & 0.9295 & 0.9806 \\
Robustness & -- & 0.9892 & 0.9910 & 0.9902 & 0.9922 & 0.9958 \\
Caution & 0.9692 & 0.9839 & 0.9915 & 0.9932 & 0.9955 & 0.9978 \\
\bottomrule
\end{tabular}
\end{table}

\noindent\textbf{Stability of Meme Scores.}
Overall, models' Meme Score rankings are highly stable under random subsampling and improve as $size$ increases. By $size{=}30$, all Meme Scores except Ingenuity already achieve Spearman rank correlations above $0.9$, and by $size{=}40$, all Meme Scores exceed $0.9$.

\noindent\textbf{Summary.}
Both Probe Properties and Meme Scores become highly stable once the subsample $size$ reaches $30$--$40$ (i.e., more than half of $|\mathcal{M}|$). The few remaining instabilities are expected and interpretable given the definitions, since properties and scores that depend on global probe-to-probe relations or clustering structure and its induced labels are inherently more sensitive to population size.


\end{document}